\documentclass{article}

\PassOptionsToPackage{numbers, compress}{natbib}

    \usepackage[preprint]{neurips_2023}

\usepackage{algpseudocode}
\usepackage{algorithm}
\usepackage{setspace}   

\usepackage[utf8]{inputenc} 
\usepackage[T1]{fontenc}    
\usepackage{hyperref}       
\usepackage{url}            
\usepackage{booktabs}       
\usepackage{sidecap}
\usepackage{amsfonts}       
\usepackage{nicefrac}       
\usepackage{microtype}      
\usepackage{xcolor}         
\usepackage[title]{appendix}
\usepackage{amsmath,amssymb}
\usepackage{graphicx}
\usepackage{subfigure}
\usepackage{sidecap} 
\usepackage{tabularx} 
\usepackage{cleveref}  

\newtheorem{proposition}{Proposition}
\newtheorem{assumption}{Assumption}
\numberwithin{assumption}{section}
\numberwithin{proposition}{section}
\newtheorem{corollary}{Corollary}
\numberwithin{equation}{section}
\newtheorem{theorem}{Theorem}
\numberwithin{theorem}{section}

\numberwithin{lemma}{section}
\newenvironment{proof}{{\noindent\it Proof.}\ }{\hfill $\square$\par}

\usepackage{enumitem}
\usepackage{array}
\usepackage{multirow}
\usepackage{caption}
\usepackage{float}
\usepackage{setspace}
\usepackage{txfonts}

\title{
SciRE-Solver: Accelerating Diffusion Models Sampling by Score-integrand Solver with  Recursive Difference
}

\author{%
  Shigui Li\textsuperscript{1}\thanks{Code is available at \href{https://github.com/ShiguiLi/scire-solver}{https://github.com/ShiguiLi/scire-solver}
    }
    \\
  School of Mathematics\\
  South China University of Technology\\
  Guangzhou 510641, China\\
  \texttt{lishigui@mail.scut.edu.cn} \\
  \And
  Wei Chen\textsuperscript{1}\\
  School of Mathematics\\
  South China University of Technology\\
  Guangzhou 510641, China\\
  \texttt{maweichen@mail.scut.edu.cn} \\
  \AND
  Delu Zeng\textsuperscript{2}\thanks{Corresponding author
    }\\
  School of Electronic and Information Engineering\\
  South China University of Technology\\
  Guangzhou 510641, China\\
  \texttt{dlzeng@scut.edu.cn} \\
}

\begin{document}

\maketitle
\begin{abstract}
Diffusion models (DMs) have made significant progress in the fields of image, audio, and video generation. One downside of DMs is their slow iterative process. Recent algorithms for fast sampling are designed from the perspective of differential equations.
However, in higher-order algorithms based on Taylor expansion, estimating the derivative of the score function becomes intractable due to the complexity of large-scale, well-trained neural networks.
Driven by this motivation, in this work, we introduce the recursive difference (RD) method to calculate the derivative of the score function in the realm of DMs.
Based on the RD method and the truncated Taylor expansion of score-integrand, we propose \emph{SciRE-Solver} with the convergence order guarantee  for accelerating sampling  of DMs.
To further investigate the effectiveness of the RD method, we also propose a variant named \emph{SciREI-Solver} based on the RD method and exponential integrator.
Our proposed sampling algorithms with RD method attain state-of-the-art (SOTA) FIDs in comparison to existing training-free sampling algorithms, across both discrete-time and continuous-time pre-trained DMs, under various number of score function evaluations (NFE).
Remarkably, SciRE-Solver using a small NFEs demonstrates promising potential to surpass the FID achieved by some pre-trained models in their original papers using no fewer than $1000$ NFEs.
For example, we reach SOTA value of $2.40$ FID with $100$ NFE for continuous-time DM  and of $3.15$ FID with $84$ NFE for discrete-time DM on CIFAR-10, as well as of $2.17$ (2.02) FID with $18$ (50) NFE for discrete-time DM on CelebA 64$\times$64. 
\end{abstract}

\section{Introduction}
\begin{figure}[ht]
\vspace{-0.45cm}
\centering
\begin{tabular}{m{0.66cm}p{2.76cm}p{2.76cm}p{2.76cm}p{2.76cm}}
   ~~ &~~~~~~~~~~NFE=$10$& ~~~~~~~~~~NFE=$15$  &~~~~~~~~~~NFE=$20$ &~~~~~~~~~~NFE=$50$ \\
\multirow{-4.5}{*}{\parbox{0.8cm}{\centering DDIM}}
& \includegraphics[width=0.2226\textwidth]{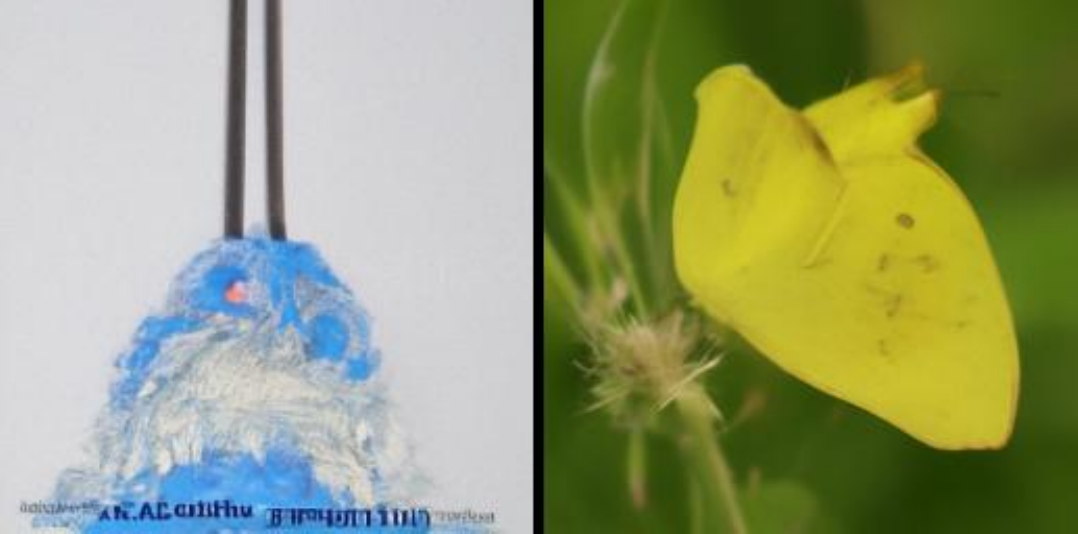} & \includegraphics[width=0.2226\textwidth]{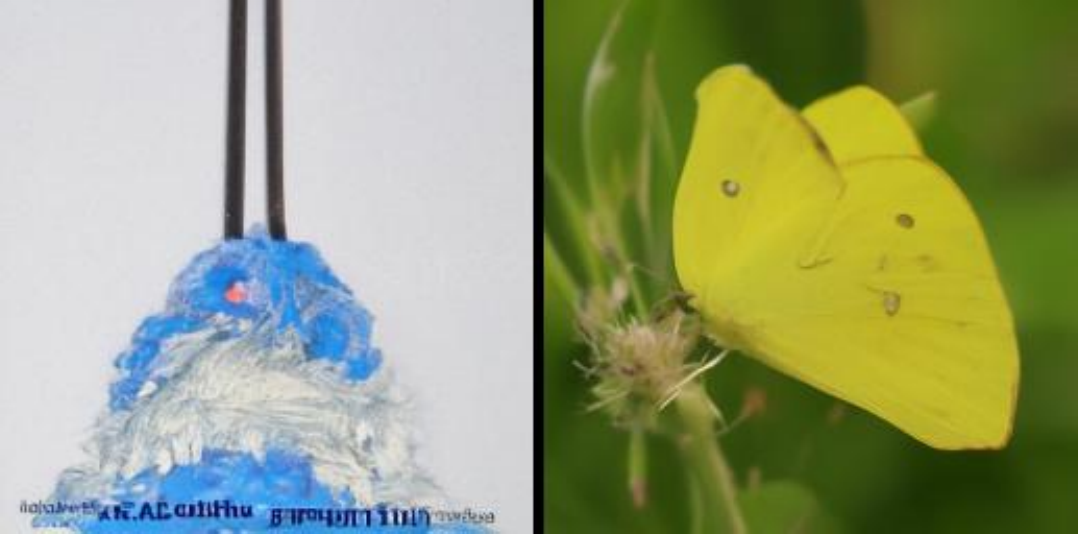} & \includegraphics[width=0.2226\textwidth]{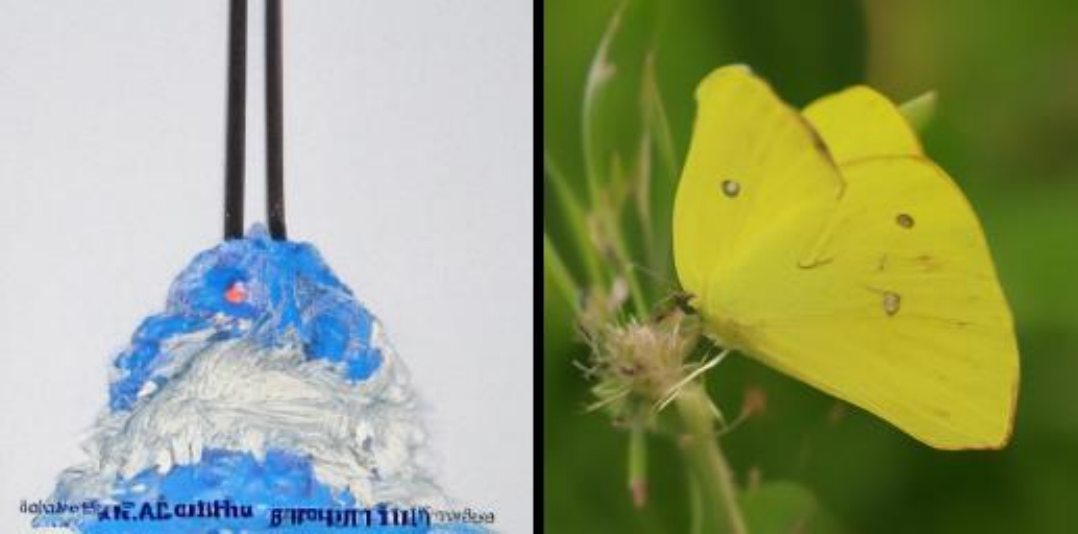} & \includegraphics[width=0.2226\textwidth]{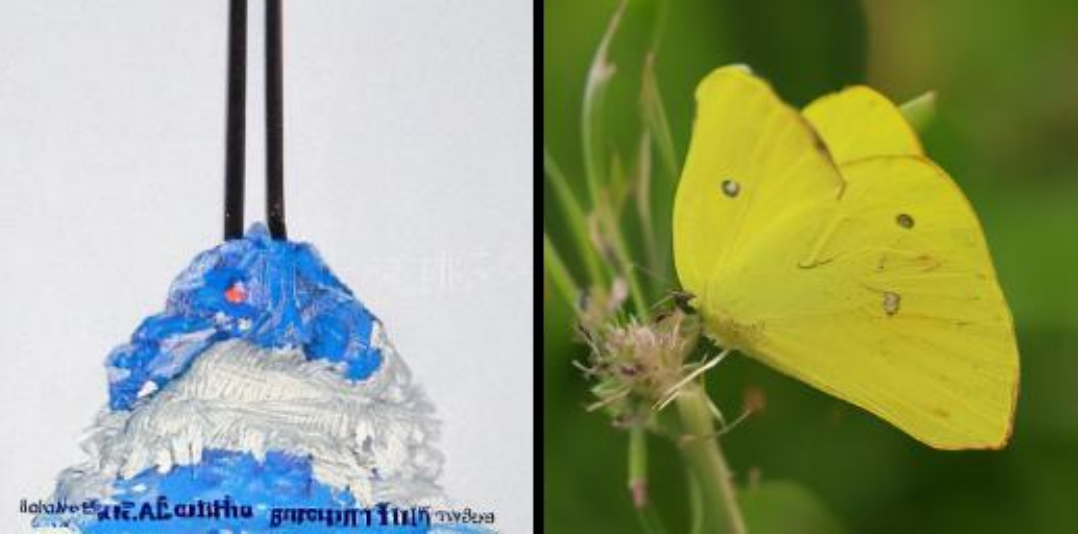}
\\
\multirow{-4.5}{*}{\parbox{0.8cm}{\centering DPM-Solver}}
& \includegraphics[width=0.2226\textwidth]{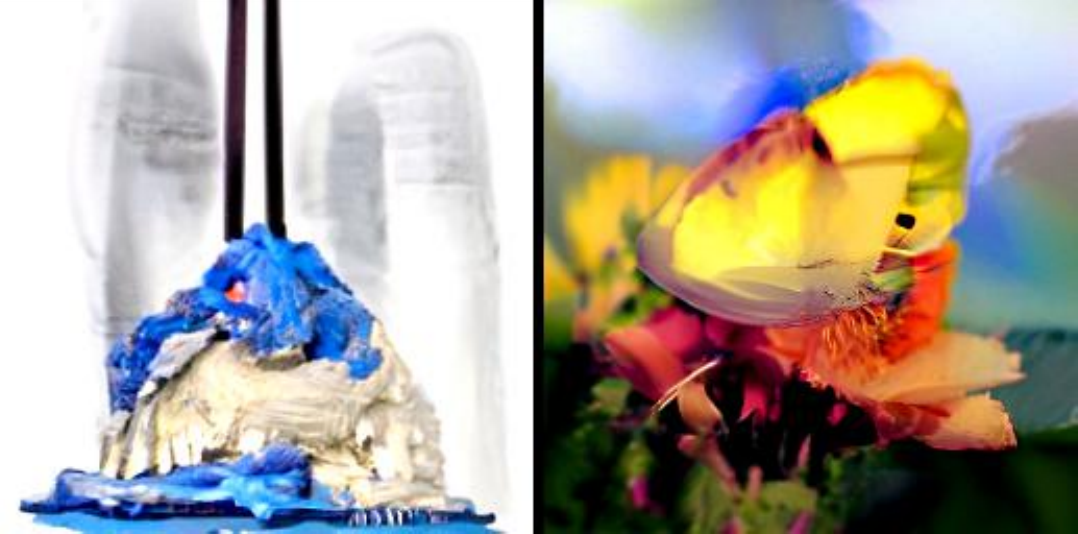} & \includegraphics[width=0.2226\textwidth]{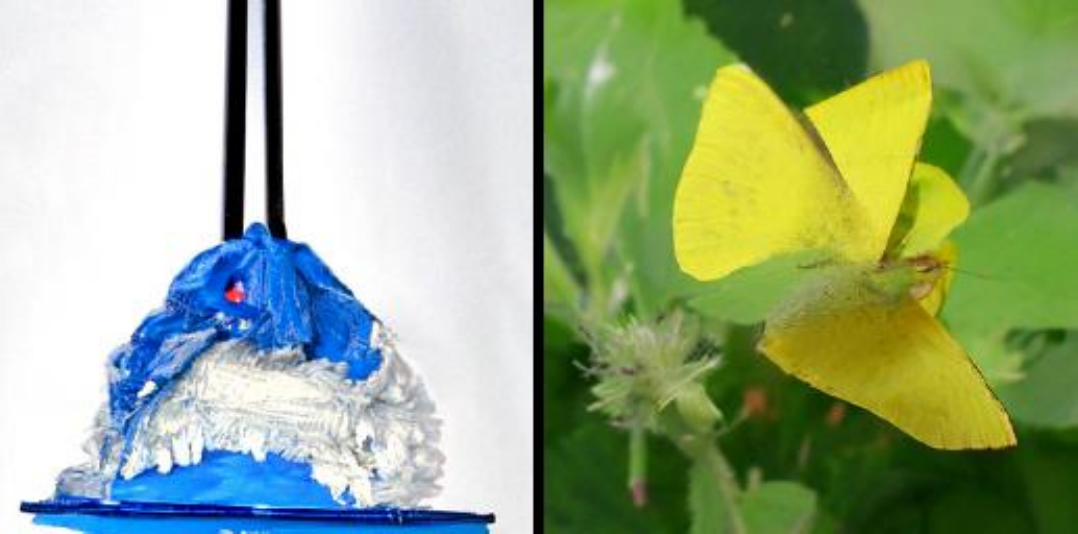} & \includegraphics[width=0.2226\textwidth]{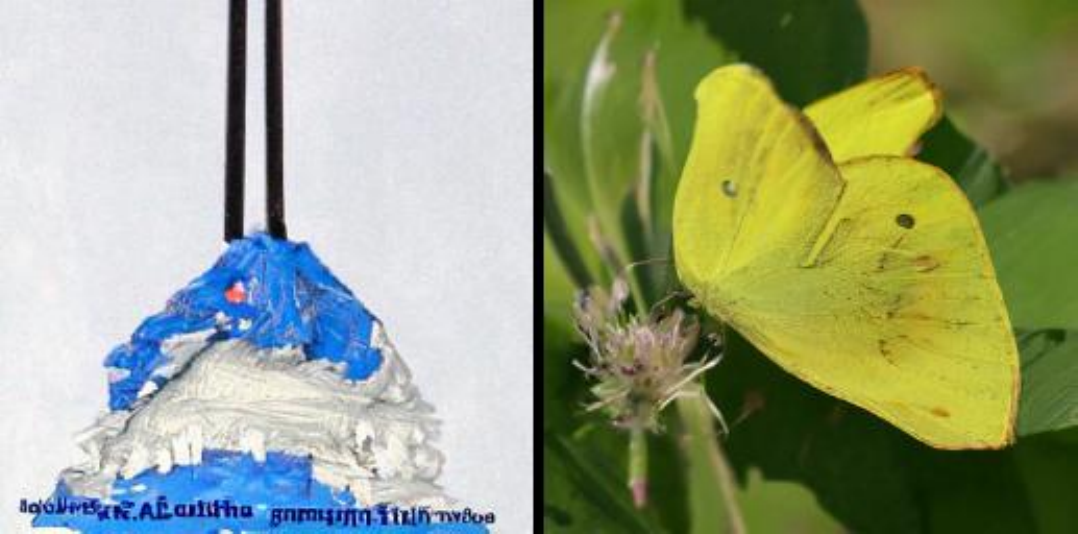} & \includegraphics[width=0.2226\textwidth]{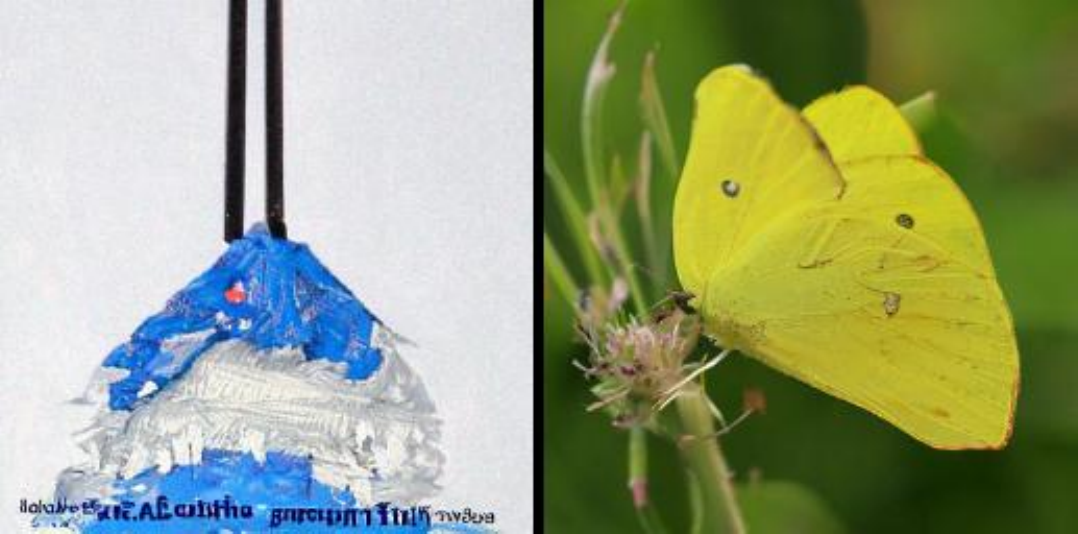}
\\
\multirow{-4.5}{*}{\parbox{0.8cm}{\centering \textbf{SciRE-Solver} (ours)}}
&\includegraphics[width=0.2226\textwidth]{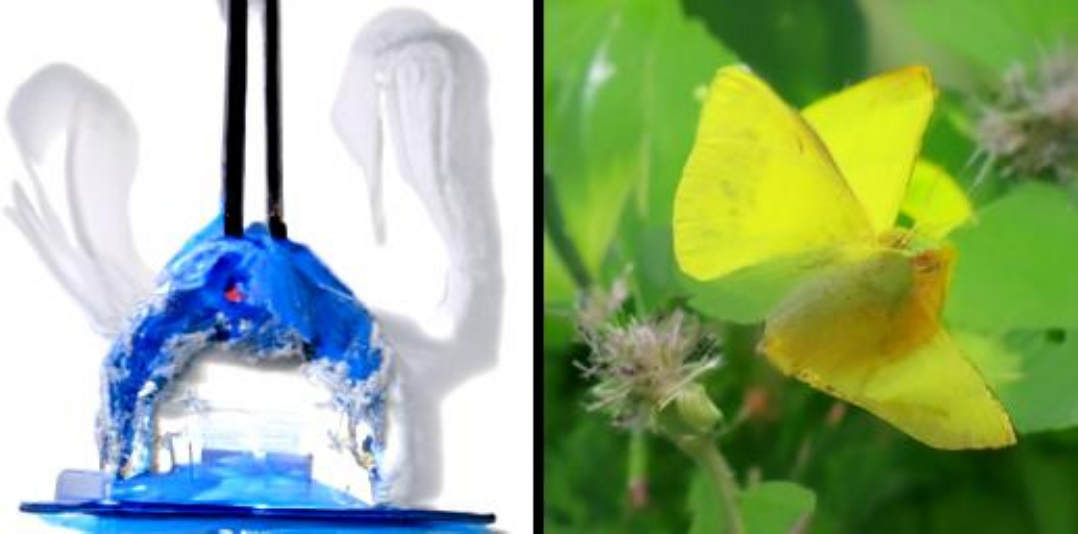} & \includegraphics[width=0.2226\textwidth]{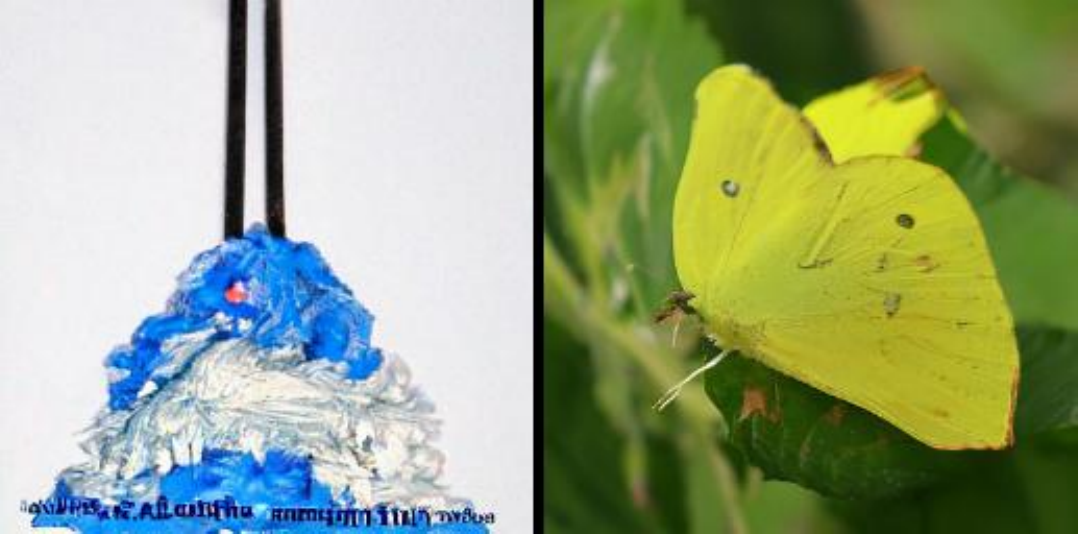} & \includegraphics[width=0.2226\textwidth]{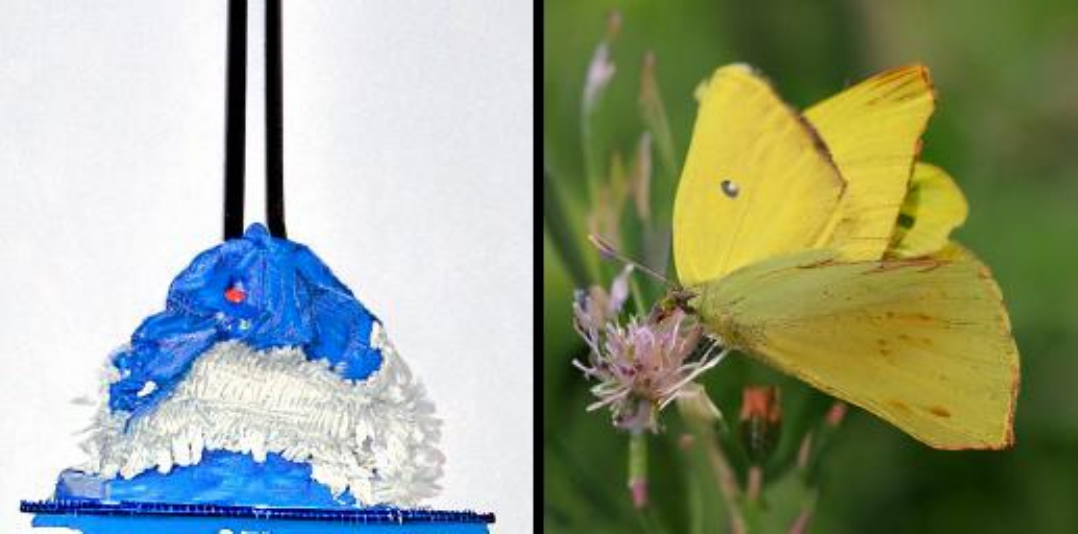} & \includegraphics[width=0.2226\textwidth]{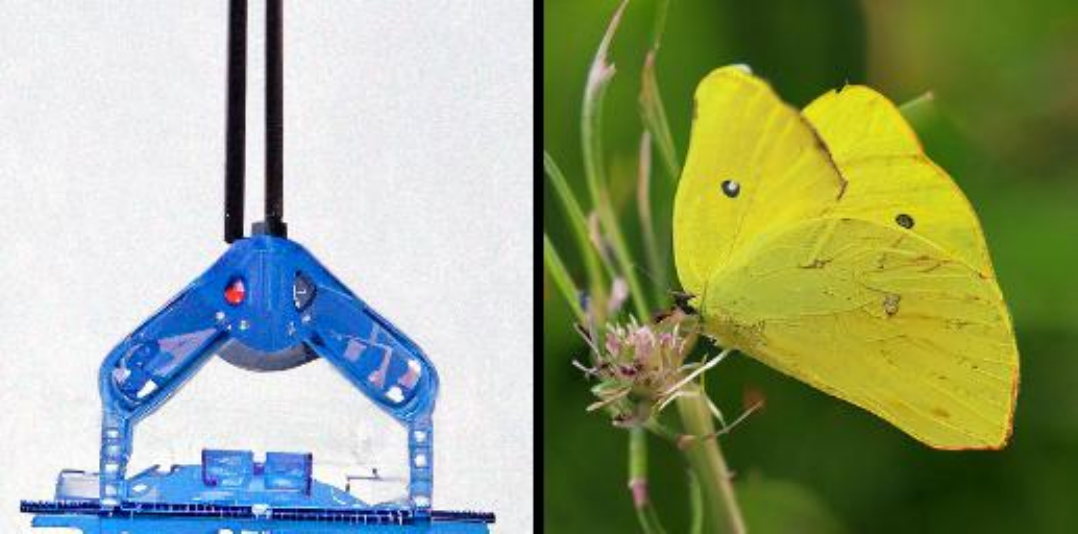} \\
\end{tabular}
\vspace{-0.1cm}
 \caption{
Generated samples of the pre-trained DM on ImageNet 256$\times$256 (classifier scale: 2.5) using 10-50 sampling steps from different sampling methods with the same random seed and codebase.  SciRE-Solver, our algorithm, generates high-quality results in a fewer number of steps.
 }
\label{fig:maintext256}
\vspace{-0.4cm}
\end{figure}
Diffusion models (DMs) \cite{sohl2015deep,ho2020denoising,song2021scorebased}
are a powerful class of generative models and
have recently gained significant attention for their competitive performance on various tasks,
including image generation \cite{dhariwal2021diffusion,meng2022sdedit}, text-to-image generation \cite{ramesh2022hierarchical}, video synthesis \cite{ho2022imagen}, and voice synthesis \cite{chen2021wavegrad,liu2022diffsinger}. DMs are
composed of two diffusion stages, the forward and reverse stages.
The forward stage of DMs is to add randomness with Gaussian noise in order to slowly disrupt the data distribution, without any training. This is distinct from generative adversarial networks  \cite{goodfellow2020generative} and variational auto-encoders \cite{kingma2013auto}, which necessitate training for both forward and reverse processes. The reverse stage of DMs is tasked with recovering the original input data from the diffused (noisy) data by learning to reverse the forward diffusion process, step by step. DMs learn models by emulating the ground-truth inverse process of a fixed forward process.
Consequently, the reverse diffusion plays an indispensable role within DMs,
as it enables the model to perform denoising (generation) on corrupted input data.

One key downside of DMs is their slow iterative sampling process, usually requiring hundreds of iterations to generate high-fidelity samples.
Two distinct categories of methods have arisen
to tackle this challenge: training-based  and training-free methods.
Training-based methods require additional training, such as knowledge distillation \cite{salimans2021progressive,meng2023distillation} and consistency models \cite{song2023consistency},
 noise level learning \cite{nichol2021improved},
or models  combined with other generative models
\cite{xiao2022tackling,vahdat2021score,zhang2021diffusion}.

On the other hand,
training-free methods  strive to accelerate the sampling process
through numerical algorithms without requiring extra training. This offers greater flexibility for
utilization across various models and applications, including both unconditional and conditional sampling. Recent training-free fast sampling methods can be attributed to the design of numerical algorithms for solving diffusion ODEs, benefiting from the fact that the sampling process of DMs can be reformulated as solving the corresponding diffusion ODE, as confirmed by DDIM \cite{song2021denoising} and Score-based models \cite{song2021scorebased}. Leveraging this framework, several high-order numerical algorithms with impressive results on DMs have been suggested, including PNDM \cite{liu2022pseudo}, DPM-Solver \cite{lu2022dpm}, DEIS \cite{zhang2023fast},
UniPC \cite{zhao2023unipc},
and ERA-Solver \cite{li2023era}.  However, these methods lack  structured mathematically guidance
for estimating the derivative of score function, which is a central issue for higher-order algorithms based on Taylor expansion.

In this work, we introduce the recursive difference (RD) method to calculate the derivative of score function required for high-order algorithms, by recursively extracting the hidden lower-order derivative information of the higher-order derivative terms in the Taylor expansion of the score function at required point, as illustrated in Figure \ref{fig:diagram-rdem}. Based on the RD method and the truncated Taylor expansion of score-integrand, we propose \emph{SciRE-Solver} with the convergence order guarantee  for accelerating sampling  of DMs.
In order to further explore the effectiveness of the RD method,
we propose a variant named \emph{SciREI-Solver}, which incorporates the RD method and exponential integrator. The FID-measured numerical experiments demonstrate the effectiveness of our proposed sampling algorithms using the RD method, as shown in Figures \ref{fig:rdefde},  \ref{fig:scireidpm} and \ref{fig:roubust}.
Our proposed sampling algorithms with RD method advance  the sampling efficiency of the training-free sampling method to a new level. Such as, we achieve 3.48 FID with 12 NFE and 2.42 FID with 20 NFE for continuous-time DMs on CIFAR10, respectively. 
Furthermore,
Figure \ref{fig:maintext256} demonstrates that SciRE-Solver also possesses the capability to generate high-quality results in a fewer number of steps on pre-trained models of high-resolution image datasets.
Notably, SciRE-Solver with a small NFEs demonstrates the promising potential to surpass the FIDs achieved in the original papers of some pre-trained models, distinguishing itself from other samplers. For example, we reach SOTA value of $2.40$ FID with $100$ NFE for continuous-time DM  and of $3.15$ FID with $84$ NFE for discrete-time DM on CIFAR-10, as well as of $2.17$ (2.02) FID with $18$ (50) NFE for discrete-time DM on CelebA 64$\times$64.

\section{Background}\label{diffusionmodel}
This section provides a brief outlines of how the sampling process of DMs is equivalent to solving the diffusion ODE, as well as the numerical methods that have been used for diffusion ODEs.

\subsection{Diffusion ODEs}
A Markov sequence $\left\{\mathbf{x}_t\right\}_{t \in[0, T]}$  with $T>0$ starting with $\mathbf{x}_0$, in the forward diffusion process of DMs
for $D$-dimensional data, is defined by the  transition distribution
\begin{equation}\label{transdis}
q\left(\mathbf{x}_t \mid \mathbf{x}_{t-1}\right):=\mathcal{N}\left(\mathbf{x}_t ; \beta_t \mathbf{x}_{t-1},\left(1-\beta_t^2\right) \mathbf{I}\right),
\end{equation}
where $\beta_t \in \mathbb{R}^{+}$ is the variance schedule function \cite{ho2020denoising}, which is differentiable w.r.t $t$ and possesses a bounded derivative.
Given the transition distribution in Eq. (\ref{transdis}), one can formulate the transition kernel of noisy data $\mathbf{x}_{t}$ conditioned on clean data $\mathbf{x}_0$ as
\begin{equation}\label{condix0}
q\left(\mathbf{x}_t \mid \mathbf{x}_0\right)=\mathcal{N}\left(\mathbf{x}_t; \alpha_{t} \mathbf{x}_0, \sigma_{t}^2 \mathbf{I}\right),
\end{equation}
where $\alpha_t=\prod_{i=1}^t \beta_i$, $\sigma_{t}=\sqrt{1-\alpha_t^2}$ for
the variance-preserving setting.
DMs choose \emph{noise schedules} for $\alpha_t$ and $\sigma_t$ to ensure that the \emph{marginal distribution} $q_T\left(\mathbf{x}_T\right)$ approximates $\mathcal{N}\left(\mathbf{x}_T ; \mathbf{0}, \hat{\sigma}^2 \mathbf{I}\right)$ for $\hat{\sigma}>0$.

\cite{kingma2021variational} established the equivalence between the transition kernel of the following stochastic differential equation (SDE) and the one in Eq. (\ref{condix0}) for $\forall t \in[0, T]$:
\begin{equation}\label{sde}
\mathrm{d} \mathbf{x}_t=f(t) \mathbf{x}_t \mathrm{~d} t+g(t) \mathrm{d} \boldsymbol{\omega}_t, \quad \mathbf{x}_0 \sim q_0\left(\mathbf{x}_0\right),
\end{equation}
where $\boldsymbol{\omega}_t \in \mathbb{R}^D$ denotes a standard Wiener process, and
\begin{equation}\label{fg}
f(t)=\frac{\mathrm{d} \log \alpha_t}{\mathrm{~d} t}, \quad g^2(t)=\frac{\mathrm{d} \sigma_t^2}{\mathrm{~d} t}-2 \frac{\mathrm{d} \log \alpha_t}{\mathrm{~d} t} \sigma_t^2 .
\end{equation}
\cite{song2021scorebased} demonstrated with some regularity conditions that
the forward process in Eq. (\ref{condix0}) has the following equivalent reverse process (\emph{reverse SDE}) from time $T$ to $0$:
\begin{equation}\label{rsde}
\mathrm{d} \mathbf{x}_t=\left[f(t) \mathbf{x}_t-g^2(t) \nabla_{\mathbf{x}} \log q_t\left(\mathbf{x}_t\right)\right] \mathrm{d} t+g(t) \mathrm{d} \overline{\boldsymbol{\omega}}_t, \quad \mathbf{x}_T \sim q_T\left(\mathbf{x}_T\right),
\end{equation}
where $\overline{\boldsymbol{\omega}}_t$ represents a standard Wiener process in the reverse time. Based on the reverse SDE in Eq. (\ref{rsde}),
\cite{song2021scorebased} derived the following ODE:
\begin{equation}\label{dsode}
\frac{\mathrm{d} \mathbf{x}_t}{\mathrm{~d} t}=f(t) \mathbf{x}_t-\frac{1}{2} g^2(t) \nabla_{\mathbf{x}} \log q_t\left(\mathbf{x}_t\right), \quad \mathbf{x}_T \sim q_T\left(\mathbf{x}_T\right),
\end{equation}
where $\mathbf{x}_t$ has a marginal distribution $q_t\left(\mathbf{x}_t\right)$, which is equivalent to the marginal distribution of $\mathbf{x}_t$ of the SDE in Eq. (\ref{rsde}). By substituting the trained noise prediction model $\boldsymbol{\epsilon}_\theta\left(\mathbf{x}_t, t\right)$ for the scaled score function: $-\sigma_t \nabla_{\mathbf{x}} \log q_t\left(\mathbf{x}_t\right)$,
\cite{song2021scorebased} defined the \emph{diffusion ODE} for DMs:
\begin{equation}\label{dode}
\frac{\mathrm{d} \mathbf{x}_t}{\mathrm{~d} t}=f(t) \mathbf{x}_t+\frac{g^2(t)}{2 \sigma_t} \boldsymbol{\epsilon}_\theta\left(\mathbf{x}_t, t\right), \quad \mathbf{x}_T \sim \mathcal{N}\left(\mathbf{0}, \hat{\sigma}^2 \boldsymbol{I}\right) .
\end{equation}
\subsection{Numerical Methods of Diffusion ODEs}\label{numericalmdode}
Traditional numerical techniques for solving ODEs find their roots in concepts like Taylor expansions, the trapezoidal rule, and Simpson's rule. These foundational ideas have paved the way for the development of well-known approaches such as \textbf{Euler's method}, \textbf{Runge-Kutta methods}, and \textbf{linear multi-step methods} \cite{suli2010numerical}. In the realm of diffusion ODEs, a similar lineage of inspiration from these classical methods can be observed in the construction of various numerical approaches.

DDIM \cite{song2021denoising} can be accurately interpreted as the forward \textbf{Euler} method from the perspective of the diffusion ODE in Eq. \ref{dode}. Song et al. \cite{song2021scorebased} tested the \textbf{Runge-Kutta} Fehlberg method for diffusion ODEs.
Liu et al. \cite{liu2022pseudo} investigated the \textbf{Runge-Kutta} methods and \textbf{linear multi-step} methods, and based on this, further proposed the PNDM. Lu et al. \cite{lu2022dpm} introduced the exponential integrator with the semi-linear structure from the ODE literature \cite{atkinson2011numerical}, and employed \textbf{Taylor expansion} techniques to handle the remaining integration, resulting in the proposed DPM-Solver.
Zhang et al. \cite{zhang2023fast} proposed DEIS by introducing the exponential integrator and further leveraging the assistance of both \textbf{Runge-Kutta} methods and \textbf{linear multi-step} (Adams-Bashforth) methods. 
Li. \cite{li2023era} explored the use of \textbf{linear multi-step} (implicit Adams) methods with Lagrange interpolation function, and further proposed ERA-Solver.

In this work, our main focus is on algorithms based on Taylor expansions.  We introduce sampling algorithms that are predicated on the recursive difference  method, which stands out as one of the most notable distinctions between our algorithm and the DPM-solver.

\section{Sampling Algorithms based on Recursive Difference for Diffusion Models}
This section introduces
the recursive difference (RD) method, which is employed to compute the derivative of score function  within sampling algorithms for DMs based on Taylor expansion.
Based
on the RD method and the truncated Taylor expansion of the score-integrand,
we propose the \emph{SciRE-Solver} with the
 convergence order guarantee to accelerating sampling of DMs.
\begin{figure}[!ht]
\vspace{-0.1cm}
    \centering
    \includegraphics[width=0.99\linewidth]{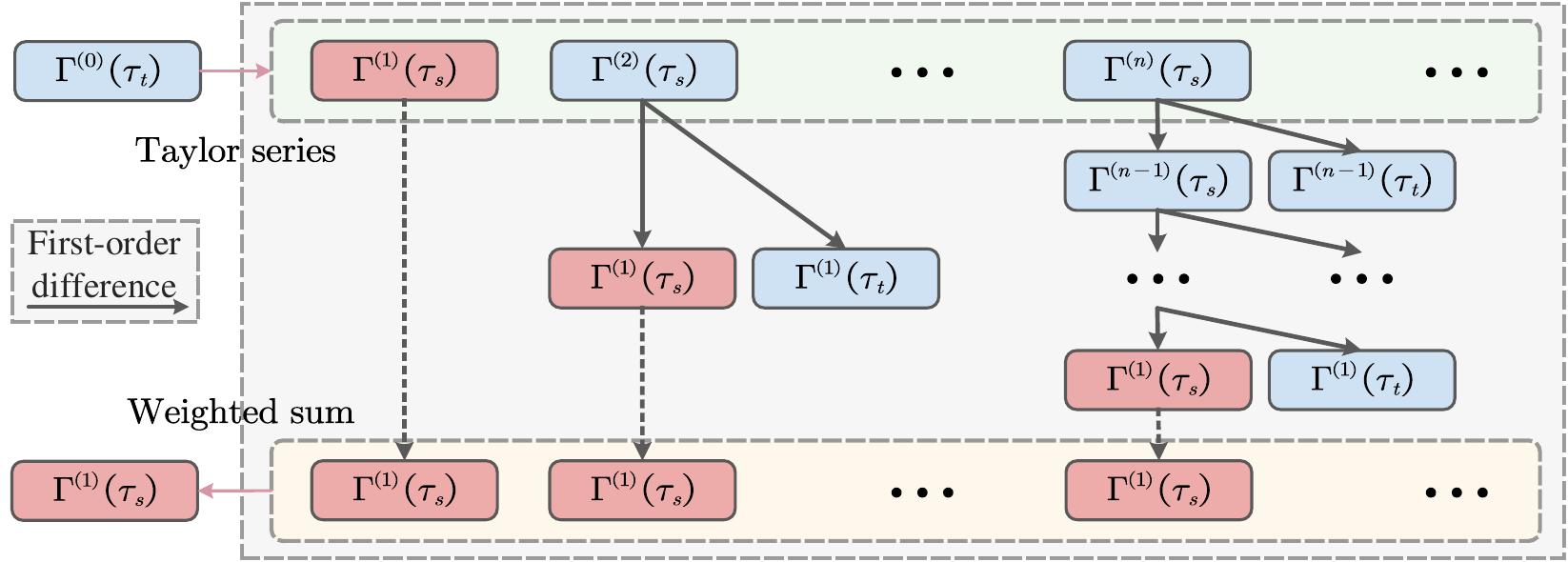}
    \caption{Schematic diagram of the \emph{recursive difference method} tailored for sampling algorithms of diffusion models. Here we denote $\boldsymbol{\epsilon}_\theta^{(k)}\left(\mathbf{x}_{\psi(\tau)}, \psi(\tau)\right)$ as $\Gamma^{(k)} (\tau)$ for simplicity, as explained in Appendix \ref{proof-of-theorem31-corollary1}. The diagram exhibits an example of the derivative process of $\Gamma^{(1)}(\tau_s)$ with $\Gamma^{(0)}(\tau_t)$ given as input. Similarly, we can obtain the $\Gamma^{(k)}(\tau_s), \forall k \in \mathbb{Z}_{+}$ with $\Gamma^{(0)}(\tau_t)$ as input using analogous procedures. }
    \label{fig:diagram-rdem}
     \vspace{-0.2cm}
\end{figure}

\subsection{Recursive Difference Method for Diffusion ODEs}
Since one can generate samples by solving the diffusion ODE in Eq. (\ref{dode}) numerically from $T$ to $0$,  sampling algorithms can be designed from the perspective of numerical solutions to differential equations. By applying the \emph{variation-of-constants} formula \cite{hale2013introduction} to ODE equation (\ref{dode}), we can obtain
\begin{equation}\label{voc}
\mathbf{x}_t=e^{\int_s^t f(\gamma) \mathrm{d} \gamma} \left(\int_s^t h(\gamma)\boldsymbol{\epsilon}_\theta\left(\mathbf{x}_\gamma, \gamma\right)\mathrm{d} \gamma+\mathbf{x}_s\right),
\end{equation}
where $h(\gamma):=e^{-\int_s^\gamma f(z) \mathrm{d} z} \frac{g^2(\gamma)}{2 \sigma_\gamma}$, and $\mathbf{x}_s$ represents the given initial value.
Based on this Eq. (\ref{voc}), we provide the most simplified solution formula for the diffusion ODEs, as follows.
\begin{proposition}\label{score_solution}
Let $\mathbf{x}_s$ be a given initial value at time $s>0$. Then, the diffusion ODE in Eq. (\ref{dode}) has the following solution formula:
\begin{equation}\label{exactsolution}
\frac{\mathbf{x}_t}{\alpha_t}-\frac{\mathbf{x}_s}{\alpha_s}=\int_{{{\rm{NSR}}}(s)}^{{{\rm{NSR}}}(t)}
\boldsymbol{\epsilon}_\theta\left(\mathbf{x}_{{\rm{rNSR}}(\tau)}, {\rm{rNSR}}(\tau)\right)\mathrm{d} \tau,
\end{equation}
where ${\rm{NSR}}(\gamma):= \frac{\sigma_\gamma
 }{\alpha_\gamma}$, we refer to it as the time-dependent
 \textbf{noise-to-signal-ratio (NSR)} function; ${\rm{rNSR}}(\cdot)$ is the inverse function of ${\rm{NSR}}(\cdot)$, satisfying $\gamma={{\rm{rNSR}}}\left({{\rm{NSR}}}(\gamma)\right)$ for any diffusion time   $\gamma$. We provide the detailed derivation in Appendix \ref{proof-of-proposition31} for this solution formula.
\end{proposition}

As the integral term in the r.h.s. of Eq. (\ref{exactsolution}) is solely dependent on the evaluation network $\boldsymbol{\epsilon}_\theta\left(\mathbf{x}_{s}, s\right)$ of scaled score function, we refer to such a simplified solution formula as  ``\emph{score-integrand form}" of diffusion ODEs.
In score-integrand form,
we can solve the diffusion ODE by integrating the $\boldsymbol{\epsilon}_{\theta}\left(\mathbf{x}_{t}, t\right)$.
However, in theory, directly tackling this problem is very challenging because $\boldsymbol{\epsilon}_\theta\left(\mathbf{x}_{t}, t\right)$ is a large-scale, well-trained complex neural network. Nevertheless, we can solve it using numerical methods. For example, we can perform a Taylor expansion on the score-integrand.

Denote $h_{t_i}:={{\rm{NSR}}}(t_{i-1})-{{\rm{NSR}}}(t_i)$, $\tau_{t_{i}}:={{\rm{NSR}}}(t_{i})$, $\psi(\tau):={\rm{rNSR}}(\tau)$, and $\boldsymbol{\epsilon}_\theta^{(k)}\left(\mathbf{x}_{\psi(\tau)}, \psi(\tau)\right):=\frac{\mathrm{d}^k\boldsymbol{\epsilon}_\theta\left(\mathbf{x}_{\psi(\tau)}, \psi(\tau)\right)}{\mathrm{d} ~\tau^k}$ as $k$-th order total derivative of $\boldsymbol{\epsilon}_\theta\left(\mathbf{x}_{\psi(\tau)}, \psi(\tau)\right)$ w.r.t. $\tau$. For $n\geq 1$, the $n$-th order Taylor expansion of $\boldsymbol{\epsilon}_\theta\big(\mathbf{x}_{\psi(\tau_{t_{i-1}})}, \psi(\tau_{t_{i-1}})\big)$ w.r.t. $\tau$ at $\tau_{t_{i}}$ is
\begin{equation}\label{taylor}
\boldsymbol{\epsilon}_\theta\left(\mathbf{x}_{\psi(\tau_{t_{i-1}})}, \psi(\tau_{t_{i-1}})\right)=\sum_{k=0}^{n} \frac{h_{t_i}^k}{k!}
\boldsymbol{\epsilon}_\theta^{(k)}\left(\mathbf{x}_{\psi(\tau_{t_i})}, \psi(\tau_{t_i})\right)
+\mathcal{O}(h_{t_i}^{n+1}).
\end{equation}
By substituting this Taylor expansion into Eq. (\ref{exactsolution}),
we get
\begin{equation}\label{itersolution2}
   \mathbf{x}_{t_{i-1}}=\frac
   {\alpha_{t_{i-1}}
 }{\alpha_{t_i} }\mathbf{x}_{t_i} +\alpha_{t_{i-1}}\sum_{k=0}^{n}
 \frac
 {h_{t_i}^{k+1}}
{(k+1)!}\boldsymbol{\epsilon}_\theta^{(k)}\left(\mathbf{x}_{ \psi(\tau_{t_i})}, \psi(\tau_{t_i})\right)
+\mathcal{O}(h_{t_i}^{n+2}).
\end{equation}
Consequently, Eq. (\ref{itersolution2}) provides an iterative scheme for solving the diffusion ODE.  By following the classical thought path, we can develop an $n$-th order solver for diffusion  ODEs  by omitting the error term $\mathcal{O}(h_{t_i}^{n+1})$ and approximating the first $(n-1)$-order derivatives $\boldsymbol{\epsilon}_\theta^{(k)}\big(\mathbf{x}_{\psi(\tau_{t_i})},  \psi(\tau_{t_i})\big)$ for $k\leq n-1$ in turn \cite{atkinson2011numerical}.
Such as, we can obtain the \emph{first-order iterative algorithm} when $n=1$:
\begin{equation}\label{firstiter}
  \tilde{\mathbf{x}}_{t_{i-1}}=\frac
   {\alpha_{t_{i-1}}
 }{\alpha_{t_i} }\tilde{\mathbf{x}}_{t_i} +\alpha_{t_{i-1}}h_i
\boldsymbol{\epsilon}_\theta\left(\tilde{\mathbf{x}}_{\psi(\tau_{t_i})},  \psi(\tau_{t_i})\right),
\end{equation}
where
$\tilde{\mathbf{x}}$ is an approximation of the true value $\mathbf{x}$,
and $\tilde{\mathbf{x}}_{t_{N}}=\mathbf{x}_{T}$ is the given initial value.

However, beneath all these smooth operations, a key issue we face is how to assess the derivatives in Taylor expansions when dealing with $n\geq2$. There is no any problem with using traditional \emph{finite difference (FD) method} to estimate the derivative of the score function, and it also has the advantage of being easy to obtain the convergence order due to the convenience brought by the Taylor expansion itself.
But, in the realm of diffusion ODE, evaluating the derivative necessitates acquiring additional variables, which can only offer approximate values through lower-order algorithms.
For this reason, evaluating the derivative of the score function directly using conventional FD method may pose a challenge, as errors can propagate easily.
Furthermore, our experiments indicate that using algorithm with FD-based-estimation during the iterative process actually produces suboptimal sampling results, as shown in Figure \ref{fig:rdefde}.  As different methods of derivative estimation can lead to different algorithms and sampling results, estimating the derivative of the score function is a central issue for higher-order algorithms based on Taylor expansion.

In the following, 
we systematically introduce the \emph{recursive difference (RD) method} to calculate the derivative of the score function. This method recursively extracts the hidden lower-order derivative information of the higher-order derivative terms in the Taylor expansion of the score-integrand at the required point, as illustrated in Figure \ref{fig:diagram-rdem}.
Formally, denote ${\rm{NSR}}_{\min}:=\min\limits_{i}\{{\rm{NSR}}(t_i)\}$, ${\rm{NSR}}_{\max}:=\max\limits_{i}\{{\rm{NSR}}(t_i)\}$, we now present the following RD method.
\begin{theorem}\label{recge}
Let $\mathbf{x}_s$ be a given initial value at time $s>0$,
$\mathbf{x}_t$ be the estimated value at time $t$ obtained by the first-order iterative algorithm in Eq. (\ref{firstiter}).
Assume that $\boldsymbol{\epsilon}_\theta\left(\mathbf{x}_{\psi(\tau)}, \psi(\tau)\right)\in \mathbb{C}^{\infty}[{\rm{NSR}}_{\min},{\rm{NSR}}_{\max}]$. Then,  we have
\begin{align}
    \begin{aligned}
\boldsymbol{\epsilon}^{(1)}_\theta\left(\tilde{\mathbf{x}}_{\psi(\tau_s)},\psi(\tau_s)\right) &=\frac{e}{e-1}\frac{ \boldsymbol{\epsilon}_\theta\left(\tilde{\mathbf{x}}_{\psi(\tau_t)}, \psi(\tau_t)\right)-\boldsymbol{\epsilon}_\theta\left(\tilde{\mathbf{x}}_{\psi(\tau_s)}, \psi(\tau_s)\right)}{h_s}\\
&-\frac{\boldsymbol{\epsilon}^{(1)}_\theta\left(\tilde{\mathbf{x}}_{\psi(\tau_t)}, \psi(\tau_t)\right)}{e-1}-\frac{(e-2)h_s}{2(e-1)}\boldsymbol{\epsilon}^{(2)}_\theta\left(\tilde{\mathbf{x}}_{\psi(\tau_t)}, \psi(\tau_t)\right)+\mathcal{O}(h_s^{2}),
    \end{aligned}
\end{align}
where $\mathbb{C}^{\infty}[{\rm{NSR}}_{\min},{\rm{NSR}}_{\max}]$ denotes
$\boldsymbol{\epsilon}_\theta\left(\mathbf{x}_{\psi(\tau)}, \psi(\tau)\right)$ is an infinitely continuously differentiable function w.r.t. $\tau$ over the interval $[{\rm{NSR}}_{\min},{\rm{NSR}}_{\max}]$.
\end{theorem}
We observe that the differentiability constraint imposed by Theorem \ref{recge} appears to be rather restrictive.
To enhance its broad applicability, we further propose the RD method with limited differentiability.
\begin{corollary}\label{reclimited}
Let $\mathbf{x}_s$ be a given initial value at time $s>0$,
$\mathbf{x}_t$ be the estimated value at time $t$ obtained by the first-order iterative algorithm in Eq. (\ref{firstiter}).
Assume that $\boldsymbol{\epsilon}_\theta\left(\mathbf{x}_{\psi(\tau)}, \psi(\tau)\right)\in \mathbb{C}^{m}[{\rm{NSR}}_{\min},{\rm{NSR}}_{\max}]$, i.e., $m$ times continuously differentiable, where $m\geq3$. Then,  we have
\begin{equation}
    \begin{aligned}
\boldsymbol{\epsilon}^{(1)}_\theta\left(\tilde{\mathbf{x}}_{\psi(\tau_s)},\psi(\tau_s)\right)& =\frac{1}{\phi_1(m)}\frac{ \boldsymbol{\epsilon}_\theta\left(\tilde{\mathbf{x}}_{\psi(\tau_t)}, \psi(\tau_t)\right)-\boldsymbol{\epsilon}_\theta\left(\tilde{\mathbf{x}}_{\psi(\tau_s)}, \psi(\tau_s)\right)}{h_s}\\
&-\frac{\phi_2(m)}{\phi_1(m)}\boldsymbol{\epsilon}^{(1)}_\theta\left(\tilde{\mathbf{x}}_{\psi(\tau_t)}, \psi(\tau_t)\right)
-\frac{\phi_3(m)h_s}{\phi_1(m)}\boldsymbol{\epsilon}^{(2)}_\theta\left(\tilde{\mathbf{x}}_{\psi(\tau_t)}, \psi(\tau_t)\right)+\mathcal{O}(h_s^{2}),
    \end{aligned}
\end{equation}
where $\phi_1(m)=\sum\limits_{k=1}^{m}\frac{(-1)^{k-1}}{k!}$, $\phi_2(m)=\sum\limits_{k=2}^{m}\frac{(-1)^{k}}{k!}$, and  $\phi_3(m)=\sum\limits_{k=3}^{m}\frac{(-1)^{k+1}}{k!}$.
\end{corollary}
We provide a detailed derivation of the RD methods as stated in Theorem \ref{recge} and Corollary \ref{reclimited} in Appendix \ref{proof-of-theorem31-corollary1}. Experiments report that replacing the derivative term with the RD method in the algorithm consistently yields better FID results compared to with the FD method, as shown in Figure \ref{fig:rdefde}.

\subsection{Sampling Algorithms based on Recursive Difference Method}\label{sec3.2}
Now, based on the Eq. (\ref{itersolution2}) and the RD methods stated by Corollary \ref{reclimited} and Theorem \ref{recge}, we propose two algorithms named \emph{SciRE-Solver-2} and \emph{SciRE-Solver-3} for $n=2$ and $n=3$, respectively. Under mild assumptions, we provide the convergence order for SciRE-Solver-$k$ ($k=2,3$), as stated in the following theorem. The proof is given in Appendix \ref{proof-of-theorem32}. Due to the typically increased complexity of higher-order algorithms, the treatment of $k\geq4$  will be left for future research.
\begin{algorithm}
	\setstretch{1.0}
	\renewcommand{\algorithmicrequire}{\textbf{Require:}}
	\renewcommand{\algorithmicensure}{\textbf{Return:}}
	\caption{SciRE-Solver-2}
	\label{algorithm:score-based-solver-2}
	\begin{algorithmic}[1]
		\Require initial value $\mathbf{x}_T$, time trajectory $\left\{t_i\right\}_{i=0}^N$, model $\boldsymbol{\epsilon}_\theta, m\geq3$
            \State $\tilde{\mathbf{x}}_{t_N} \leftarrow \mathbf{x}_T,  r_1\leftarrow\frac{1}{2}$
		\For{$i\leftarrow $ $N$ to $0$}
                \State $ h_i~~\leftarrow ~{\rm{NSR}}({t_{i-1}})-{\rm{NSR}}(t_{i})$
                \State $ s_i~~\leftarrow ~{\rm{rNSR}}\left({\rm{NSR}}(t_{i})+r_1 h_i\right)$
                \State $ \tilde{\mathbf{x}}_{s_i} ~~\leftarrow ~\frac{\alpha_{s_i}}{\alpha_{t_{i}}} \tilde{\mathbf{x}}_{t_{i}}+\alpha_{s_i}r_1 h_i\boldsymbol{\epsilon}_\theta\left(\tilde{\mathbf{x}}_{t_{i}}, t_{i}\right)$
                \State $ \tilde{\mathbf{x}}_{t_{i-1}} \leftarrow \frac{\alpha_{t_{i-1}}}{\alpha_{t_{i}}} \tilde{\mathbf{x}}_{t_{i}}+\alpha_{t_{i-1}} h_i\boldsymbol{\epsilon}_\theta\left(\tilde{\mathbf{x}}_{t_{i}}, t_{i}\right)+\alpha_{t_{i-1}}\frac{h_i}{2\phi_1(m)r_1} \left(\boldsymbol{\epsilon}_\theta\left(\tilde{\mathbf{x}}_{s_i}, s_i\right)-\boldsymbol{\epsilon}_\theta\left(\tilde{\mathbf{x}}_{t_{i}}, t_{i}\right)\right)$
            \EndFor
		
		\Ensure $\tilde{\mathbf{x}}_{0}$.
	\end{algorithmic}
\end{algorithm}
\vspace{-14pt}
\begin{algorithm}
	\setstretch{1.0}
	\renewcommand{\algorithmicrequire}{\textbf{Require:}}
	\renewcommand{\algorithmicensure}{\textbf{Return:}}
	\caption{SciRE-Solver-3}
	\label{algorithm:score-based-solver-3}
	\begin{algorithmic}[1]
		\Require initial value $\mathbf{x}_T$, time trajectory $\left\{t_i\right\}_{i=0}^N$, model $\boldsymbol{\epsilon}_\theta, m\geq3$  
            \State $\tilde{\mathbf{x}}_{t_N} \leftarrow \mathbf{x}_T,  r_1\leftarrow\frac{1}{3}, r_2\leftarrow\frac{2}{3}$
		\For{$i\leftarrow $ $N$ to $0$}
                \State $h_i \leftarrow {\rm{NSR}}({t_{i-1}})-{\rm{NSR}}(t_{i})$
                \State $s_{i_1},s_{i_2} \leftarrow {\rm{rNSR}}\left({\rm{NSR}}(t_{i})+r_1 h_i\right), {\rm{rNSR}}\left({\rm{NSR}}(t_{i})+r_2 h_i\right)$
                \State $\mathbf{x}_{s_{i_1}} \leftarrow \frac{\alpha_{s_{i_1}}}{\alpha_{t_{i}}} \mathbf{x}_{t_{i}}+\alpha_{s_{i_1}}r_1 h_i\boldsymbol{\epsilon}_\theta\left(\mathbf{x}_{t_{i}}, t_{i}\right)$
                \State $ \mathbf{x}_{s_{i_2}} \leftarrow \frac{\alpha_{s_{i_2}}}{\alpha_{t_{i}}} \mathbf{x}_{t_{i}}                +\alpha_{s_{i_2}}r_2h_i   \boldsymbol{\epsilon}_\theta\left(\mathbf{x}_{t_{i}}, t_{i}\right)
                + \alpha_{s_{i_2}}\frac{h_i}{\phi_1(m)}\left(\boldsymbol{\epsilon}_\theta\left(\mathbf{x}_{s_{i_1}}, s_{i_1}\right)-\boldsymbol{\epsilon}_\theta\left(\mathbf{x}_{t_{i}}, t_{i}\right)\right)$
                \State $ \mathbf{x}_{t_{i-1}} \leftarrow \frac{\alpha_{t_{i-1}}}{\alpha_{t_{i}}} \mathbf{x}_{t_{i}}+\alpha_{t_{i-1}} h_i\boldsymbol{\epsilon}_\theta\left(\mathbf{x}_{t_{i}}, t_{i}\right)+
                \alpha_{t_{i-1}}\frac{h_i}{2\phi_1(m)r_2} \left(\boldsymbol{\epsilon}_\theta\left(\mathbf{x}_{s_{i_2}}, s_{i_2}\right)-\boldsymbol{\epsilon}_\theta\left(\mathbf{x}_{t_{i}}, t_{i}\right)\right)$
            \EndFor
		
		\Ensure $\mathbf{x}_{0}$.
	\end{algorithmic}
\end{algorithm}
\vspace{-10pt}
\begin{theorem}\label{convergence}
    Assume that $\boldsymbol{\epsilon}_\theta\left(\mathbf{x}_{\psi(\tau)}, \psi(\tau)\right)\in \mathbb{C}^{m}[{\rm{NSR}}_{\min},{\rm{NSR}}_{\max}]$. Then, for $k=2,3$,
the global convergence order of SciRE-Solver-$k$ is no less than $k-1$.
\end{theorem}

\subsection{Parametrizable Time Trajectory}
In SciRE-Solver, it is necessary to specify a time trajectory.
Although SciRE-Solver  can generate high-quality samples in a few steps using existing quadratic and uniform time trajectories,
it has been demonstrated in experiments  in \cite{lu2022dpm} and \cite{zhang2023fast} that the optimal time trajectory can further improve the sampling efficiency.
Here, we present two parametrizable  alternative methods for the $\rm{NSR}$ function to compute the time trajectory, named as \emph{NSR}-type and \emph{Sigmoid}-type time trajectories.
The comparative experiments are provided in Appendix \ref{experiment-details}.
\section{Assessing the Efficacy of the RD Method through Ablation Studies}
This section demonstrates the effectiveness of the RD method from two perspectives: 1. Comparing it with traditional finite difference (FD) method; 2. Introducing the RD method into  the exponential-based calculation formula  and comparing it with its counterpart algorithm, DPM-Solver-2.
\subsection{Comparisons of the RD Method and the FD method}
\begin{figure}[!ht]
\vspace{-0.35cm}
\setlength{\abovecaptionskip}{0.cm}
    \centering
    \subfigure
    [CIFAR-10 (discrete)]
    {\includegraphics[width=0.4\linewidth]{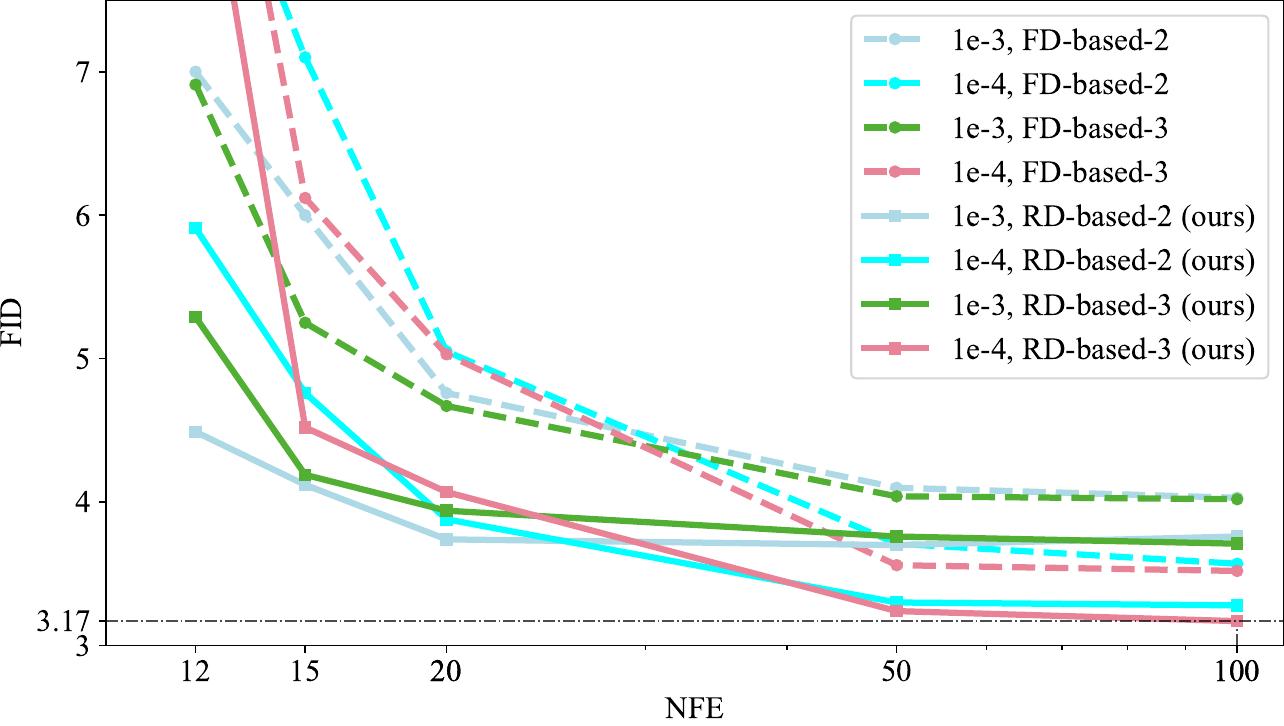}}
    ~~~~~~~~~~
    \subfigure
    [CelebA 64$\times$64 (discrete)]
    {\includegraphics[width=0.4\linewidth]{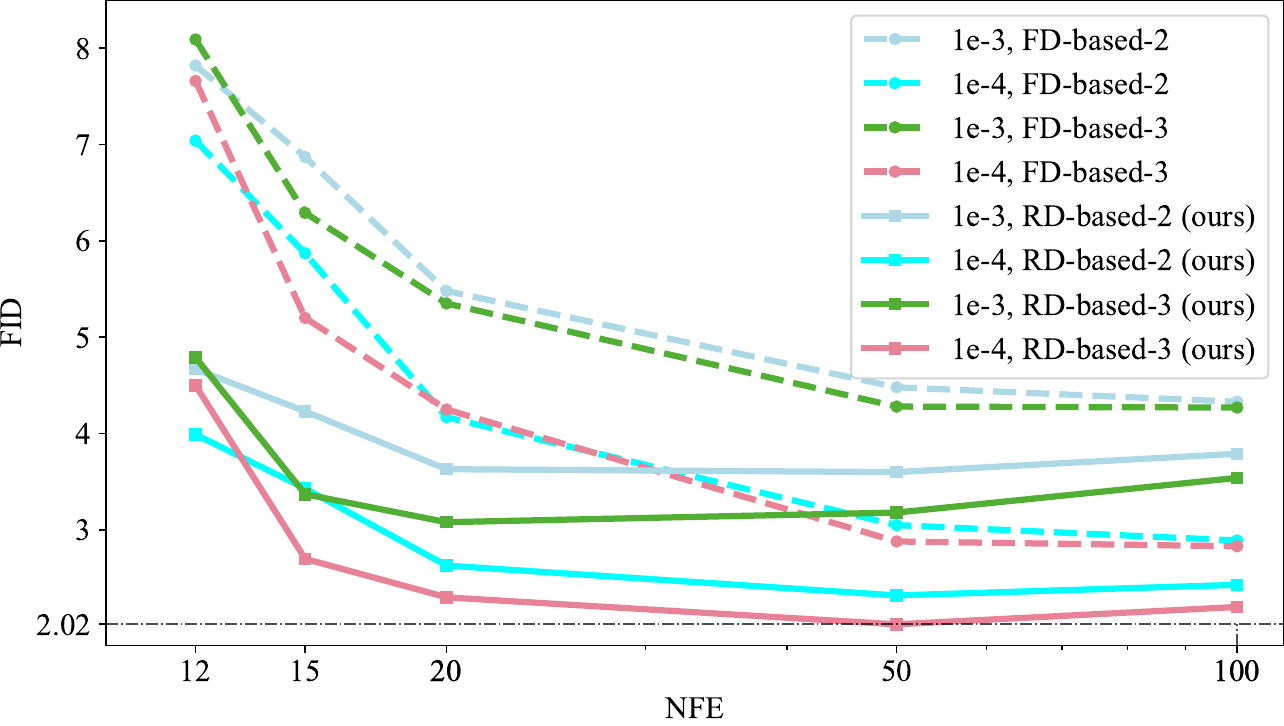}}
    \caption{
  Comparisons of FID $\downarrow$ obtained by employing  RD and FD in SciRE-Solver codebase.
  The RD-based method is consistently superior to the FD-based method across different cases.
    }
    \label{fig:rdefde}
\setlength{\belowcaptionskip}{-0.cm}
\vspace{-0.3cm}
\end{figure}

In Corollary \ref{reclimited}, the RD method degenerates into the FD method, if we set $\phi_1(m)=1$ and drop other terms. Thus,
we set $\phi_1(m)=1$ in our SciRE-Solver codebase to represent the sampling algorithm based on FD method.
Comparative experiments are presented in Figure \ref{fig:rdefde} under identical settings.
\subsection{SciREI-Solver, and compared with DPM-Solver-2}
\begin{figure}[!ht]
\vspace{-0.3cm}
\setlength{\abovecaptionskip}{0.cm}
    \centering
    \subfigure
    [CIFAR-10 ($1e-3$)]
    {\includegraphics[width=0.243\linewidth]{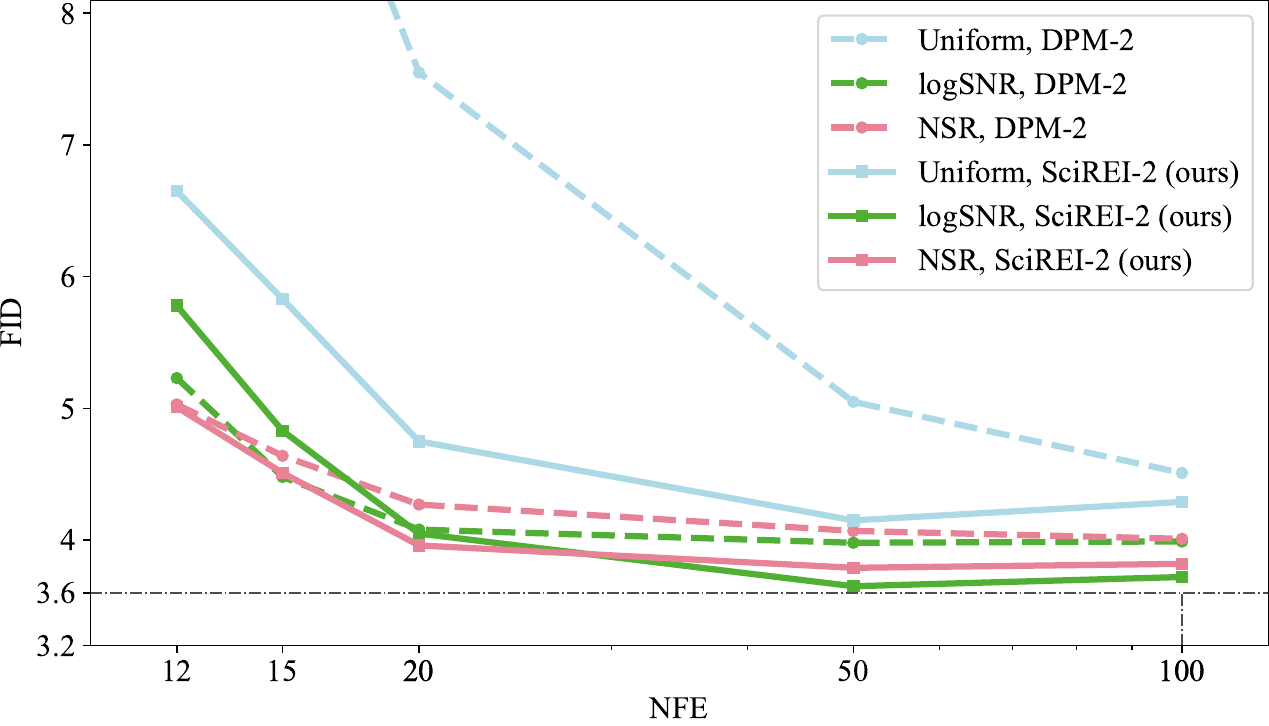}}
    \subfigure
    [CIFAR-10 ($1e-4$)]
    {\includegraphics[width=0.243\linewidth]{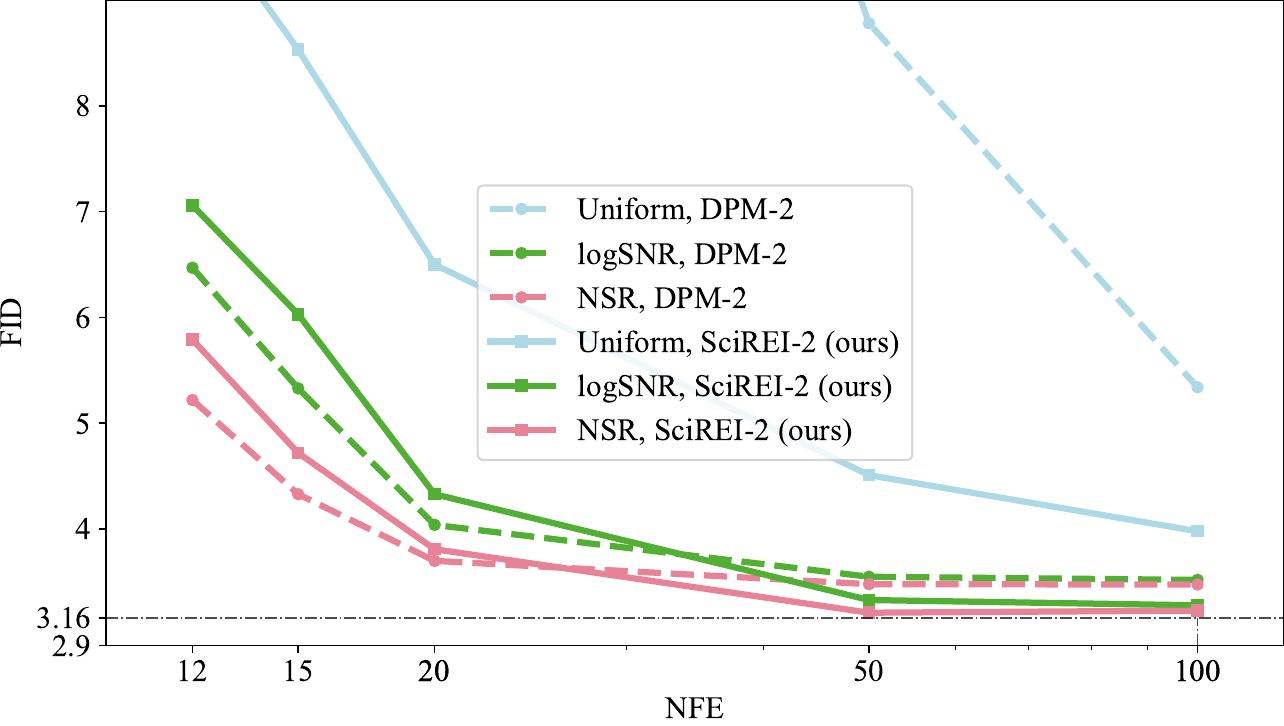}}
    \subfigure
    [CelebA 64 (1e-3)]
    {\includegraphics[width=0.243\linewidth]{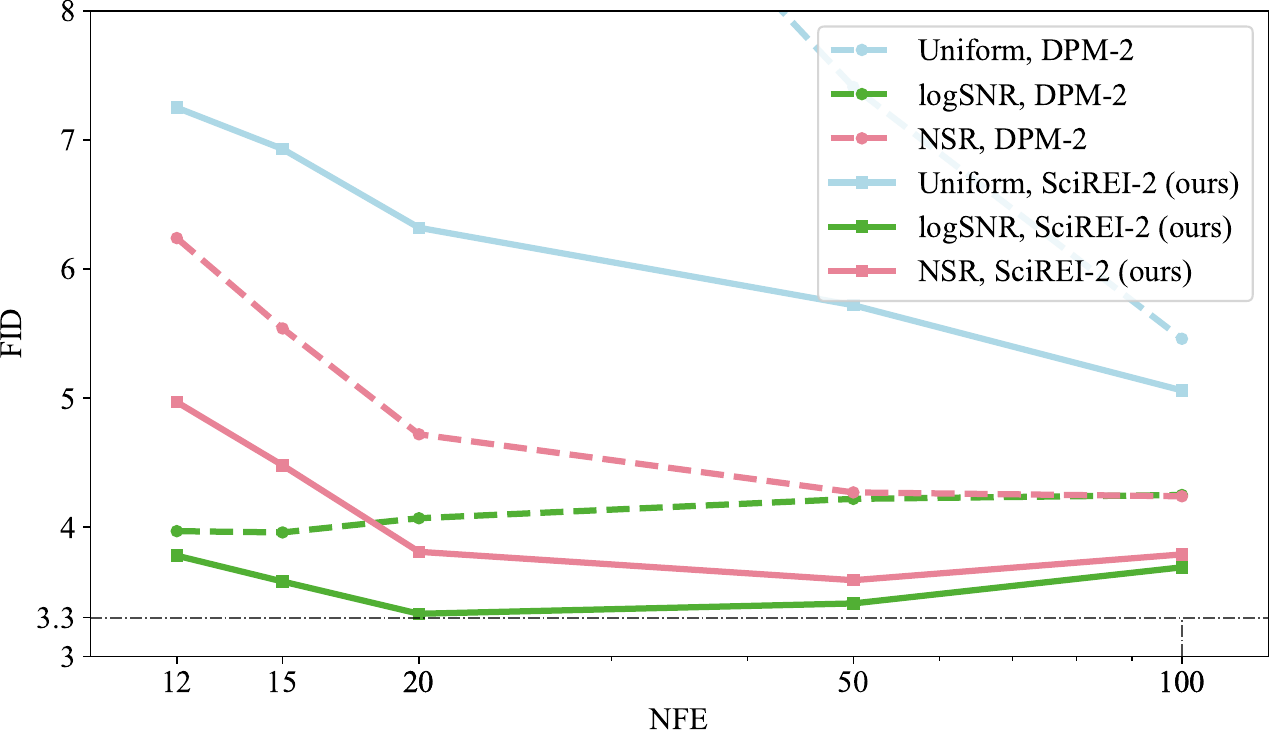}}
    \subfigure
    [CelebA 64 (1e-4)]
    {\includegraphics[width=0.243\linewidth]{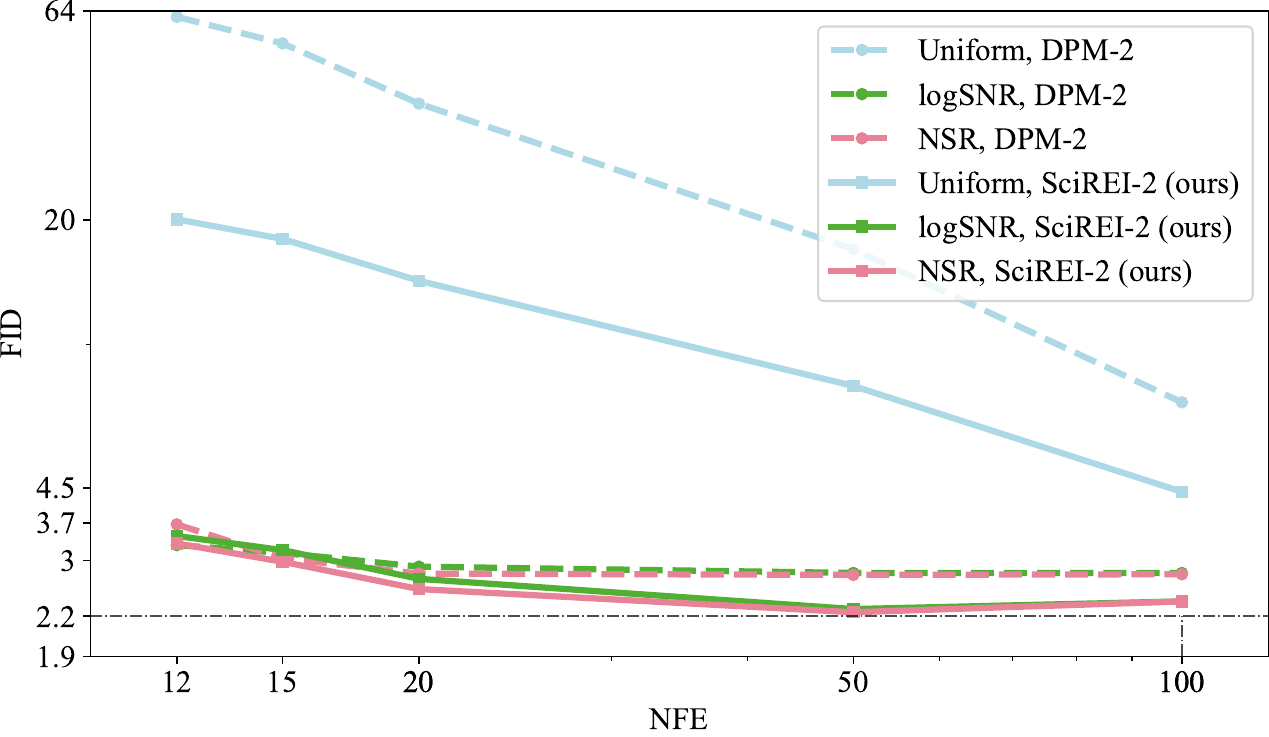}}
    \caption{
  Comparisons of FID $\downarrow$ obtained by SciREI-2 and DPM-2 solvers across different trajectories. SciREI-2 is more robust than DPM-2 across different time trajectories under the same sampling step.
    }
    \label{fig:scireidpm}
\setlength{\belowcaptionskip}{-0.cm}
\vspace{-0.1cm}
\end{figure}

To further investigate the RD method, we introduce \emph{SciREI-Solver} (n=2), a variant combining the RD method and the exponential-based calculation formula from DPM-Solver. Refer to Appendix \ref{gexct} for the details of SciREI-Solver. We compare the generative performance of SciREI-Solver-2 and DPM-Solver-2 with the identical settings on the CIFAR-10 and CelebA 64 datasets using various time trajectories and termination times, the experiment results are presented in Figure \ref{fig:scireidpm}.

More generally, we also provide the sampling comparison between the RD-based sampling algorithms (including SciRE-Solver-2 and SciREI-Solver-2) and the baseline algorithm (DPM-Solver-2) on high-resolution image datasets, as shown in Figure \ref{fig:roubust}. More comparisons are provided in Appendix \ref{gexct}.

\begin{figure}[ht]
\vspace{-0.2cm}
\centering
\begin{tabular}{m{0.7cm}p{2.76cm}p{2.76cm}p{2.76cm}p{2.76cm}}
   ~~&~~~~~~~~~~\,~NFE=$6$& ~~~~\,~~~~~~NFE=$12$  &~~~~\,~~~~~~NFE=$24$ &~~~\,~~~~~~~NFE=$36$ \\
\multirow{-3}{*}{\parbox{0.7cm}{\centering {\small DPM-2} }}
&\includegraphics[width=0.2226\textwidth]{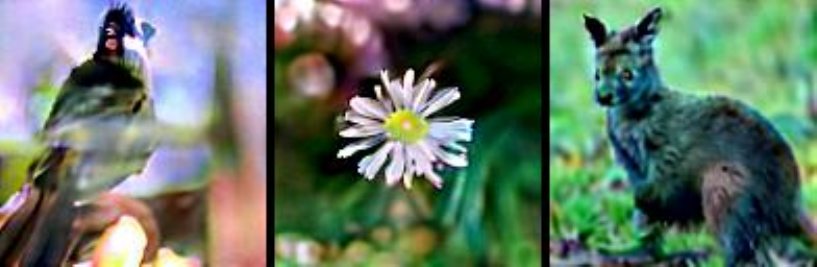} & \includegraphics[width=0.2226\textwidth]{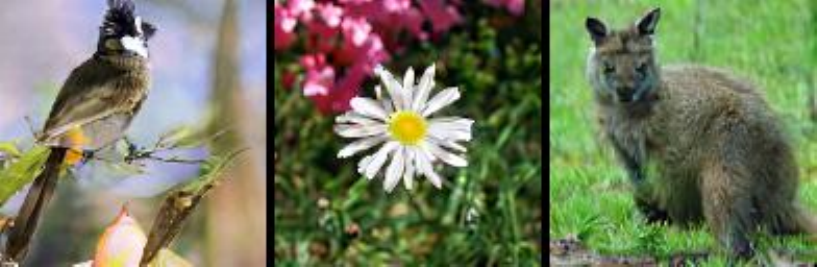} & \includegraphics[width=0.2226\textwidth]{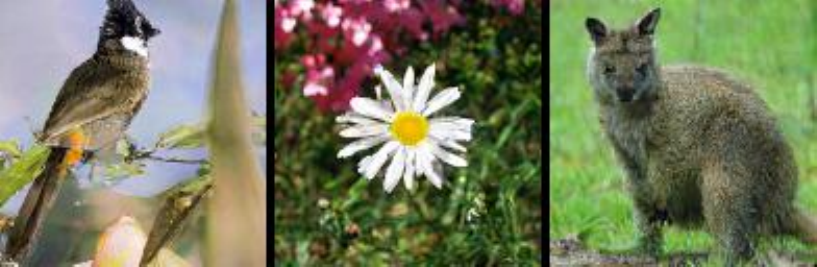} & \includegraphics[width=0.2226\textwidth]{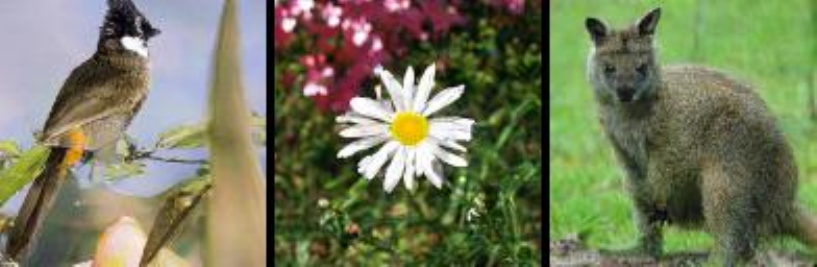}
\\
\multirow{-3}{*}{\parbox{0.6cm}{\centering {\small SciREI-2(\textbf{ours})}}}
& \includegraphics[width=0.2226\textwidth]{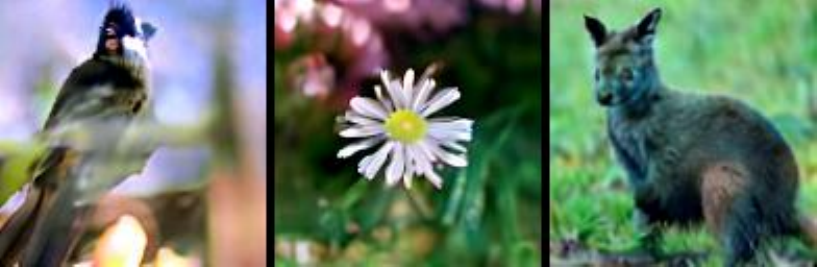} & \includegraphics[width=0.2226\textwidth]{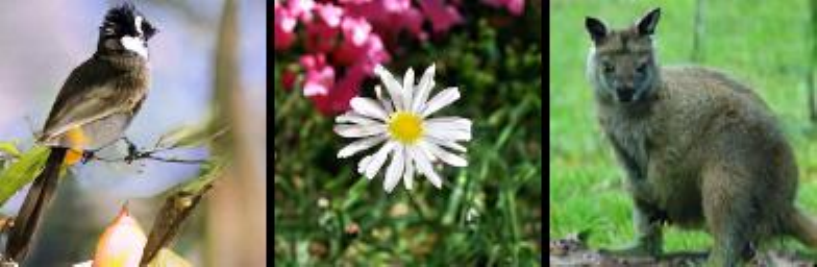} & \includegraphics[width=0.2226\textwidth]{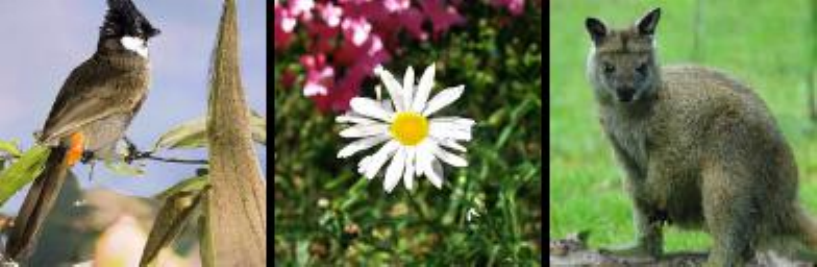} & \includegraphics[width=0.2226\textwidth]{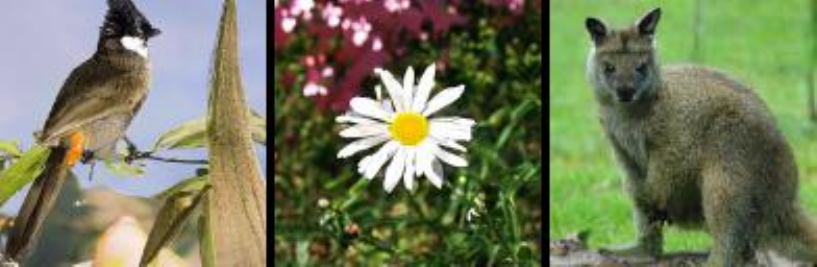}
\\
\multirow{-3}{*}{\parbox{0.7cm}{\centering {\small SciRE-2(\textbf{ours})}}}
& \includegraphics[width=0.2226\textwidth]{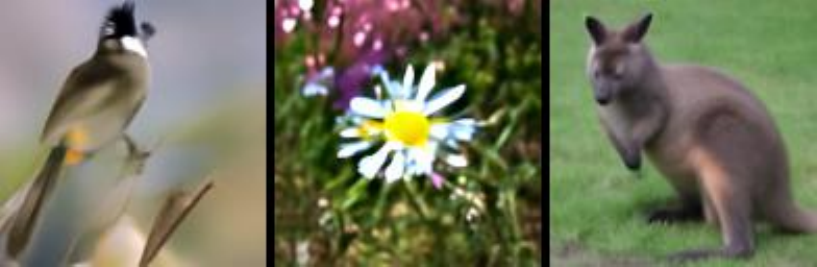} & \includegraphics[width=0.2226\textwidth]{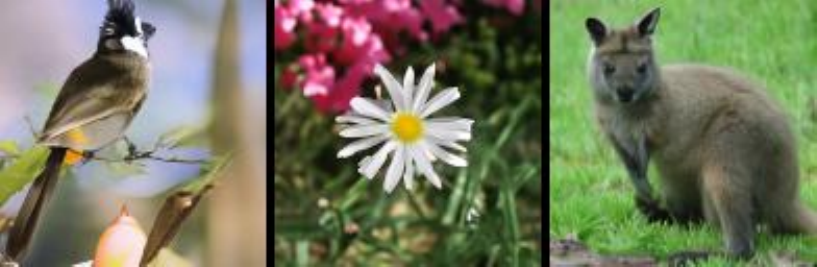} & \includegraphics[width=0.2226\textwidth]{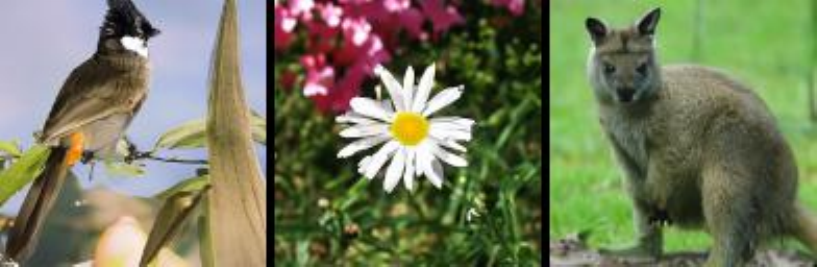} & \includegraphics[width=0.2226\textwidth]{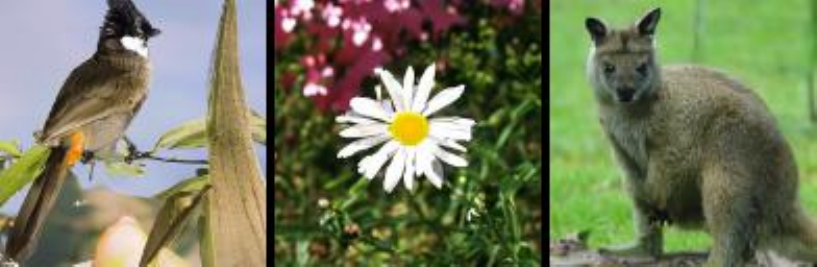}
\\
\multirow{-4.5}{*}{\parbox{0.7cm}{\centering {\small DPM-2} }}
&\includegraphics[width=0.2226\textwidth]{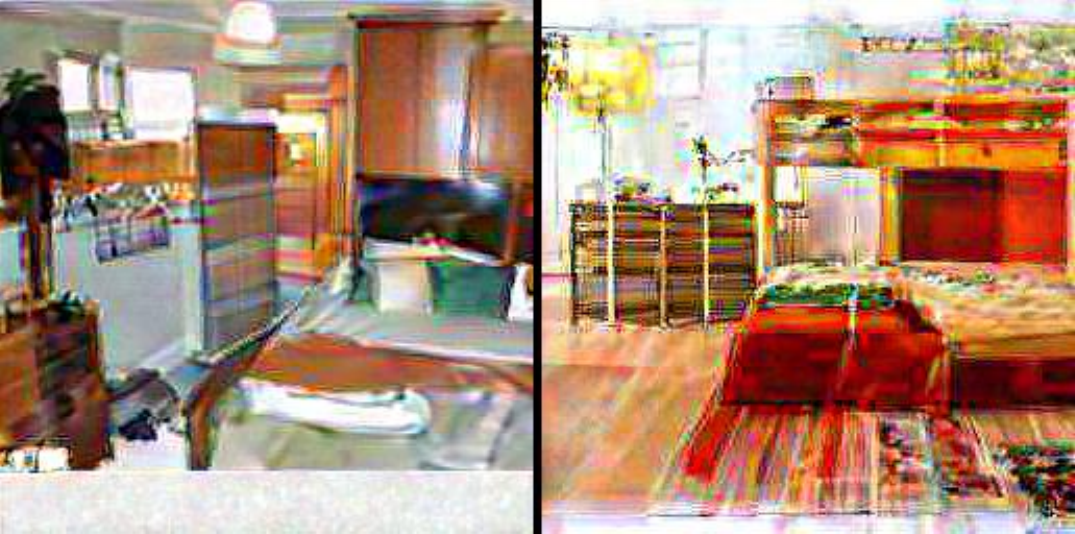} & \includegraphics[width=0.2226\textwidth]{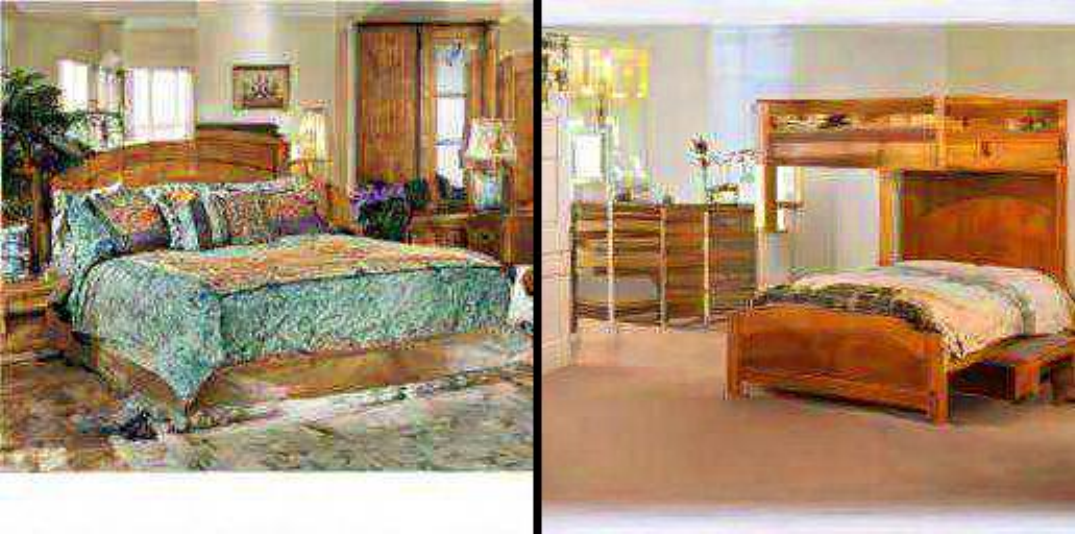} & \includegraphics[width=0.2226\textwidth]{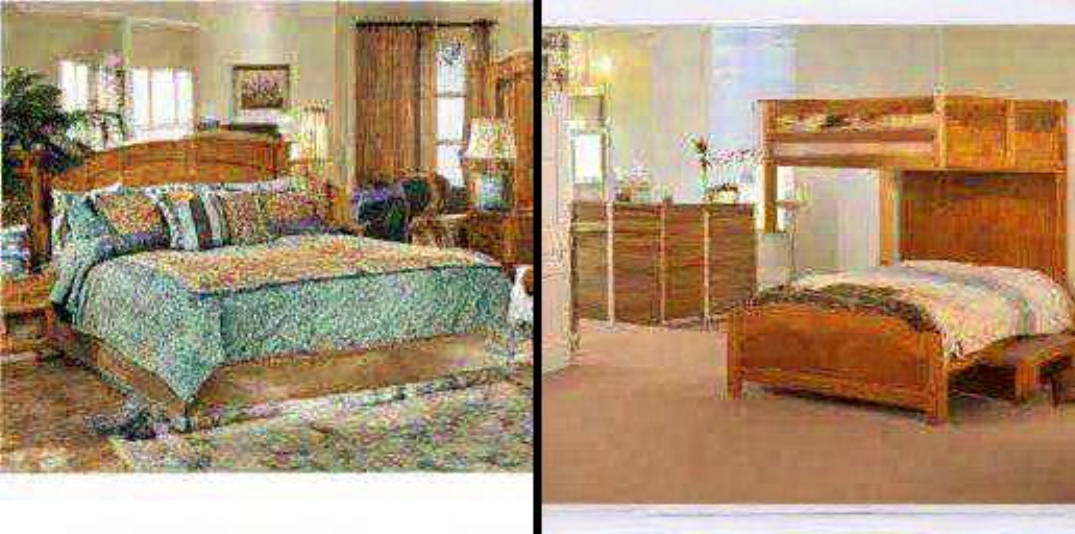} & \includegraphics[width=0.2226\textwidth]{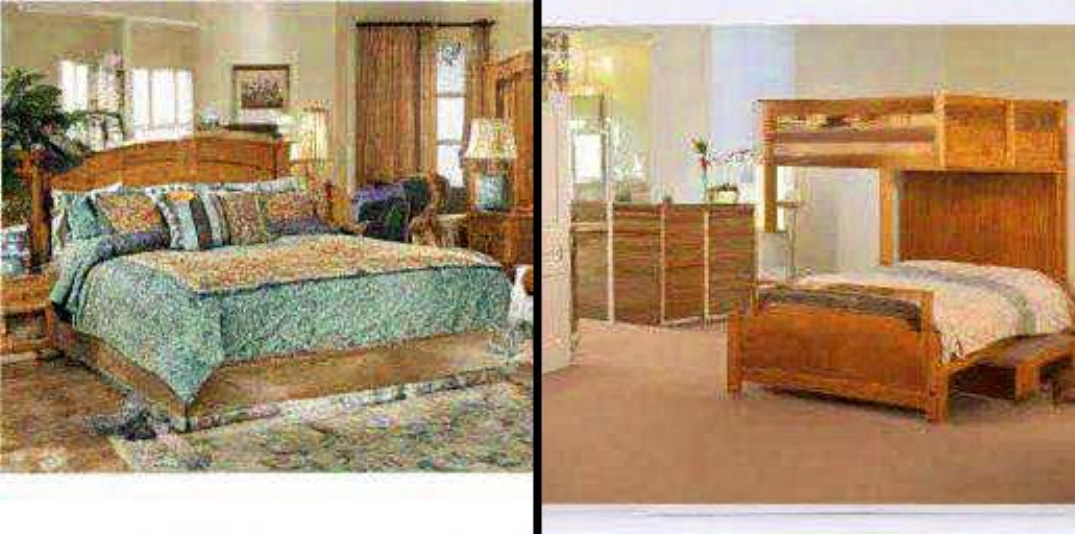}
\\
\multirow{-4.5}{*}{\parbox{0.6cm}{\centering {\small SciREI-2(\textbf{ours})}}}
& \includegraphics[width=0.2226\textwidth]{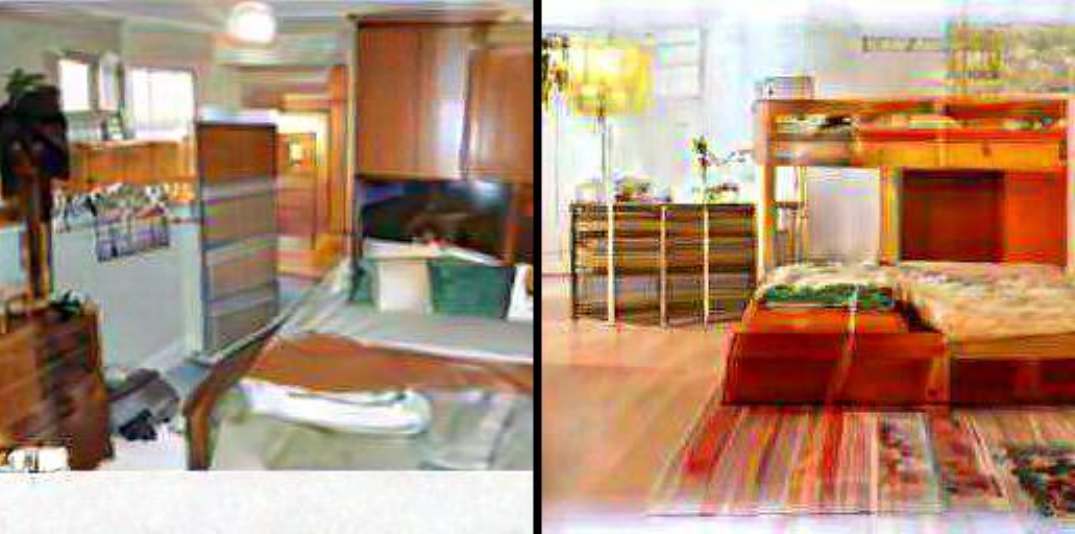} & \includegraphics[width=0.2226\textwidth]{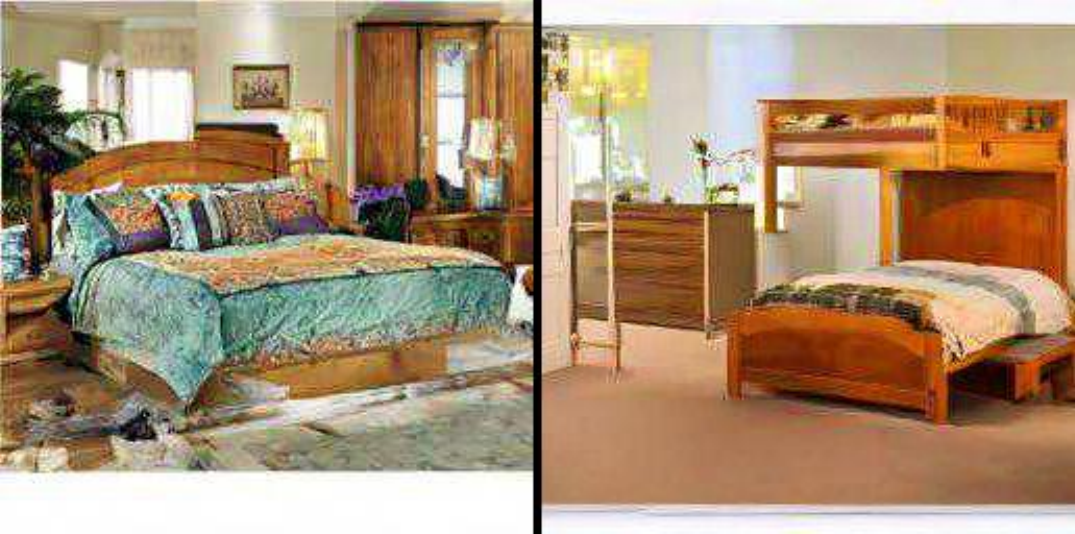} & \includegraphics[width=0.2226\textwidth]{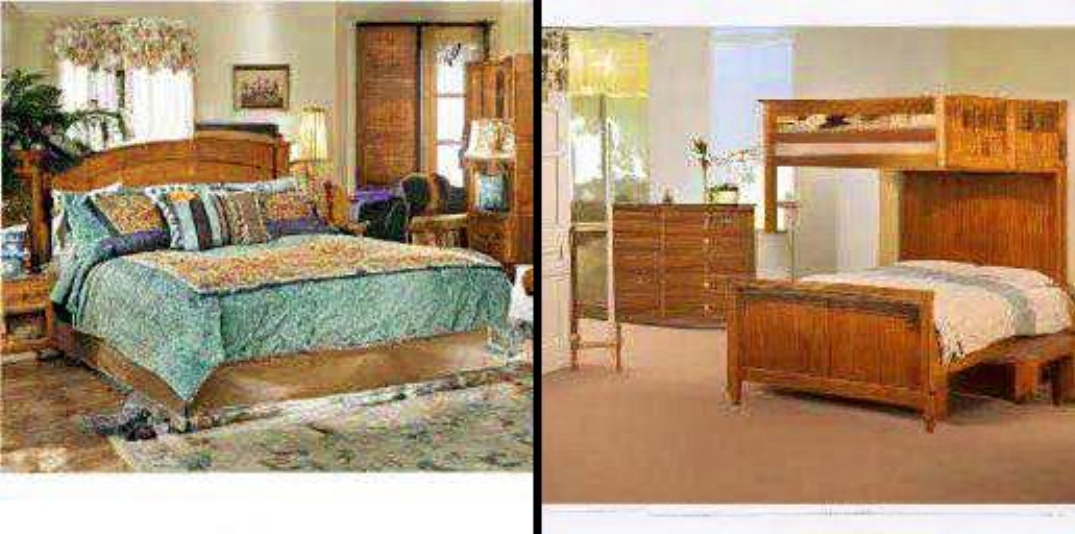} & \includegraphics[width=0.2226\textwidth]{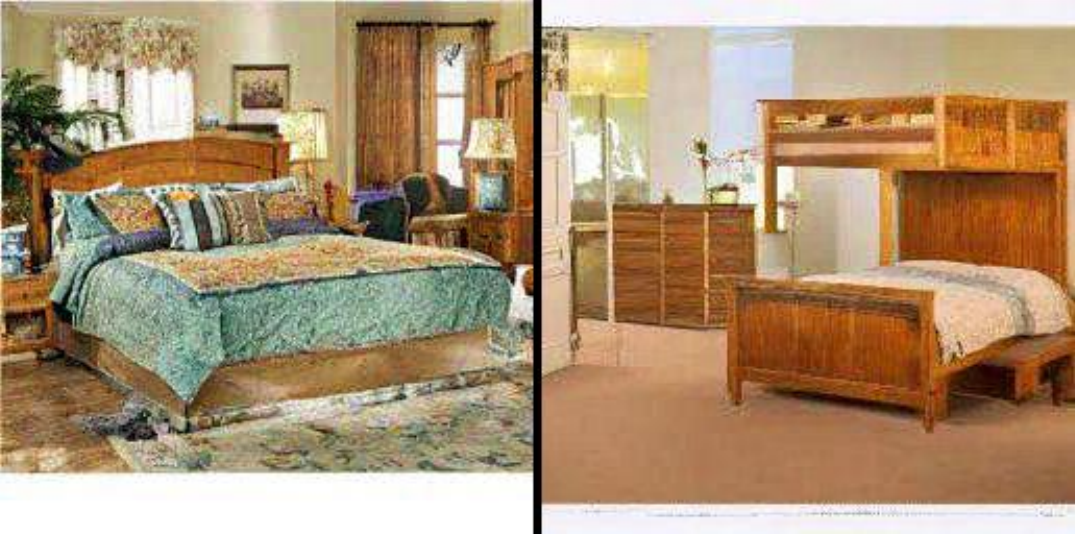}
\\
\multirow{-4.5}{*}{\parbox{0.7cm}{\centering {\small SciRE-2(\textbf{ours})}}}
& \includegraphics[width=0.2226\textwidth]{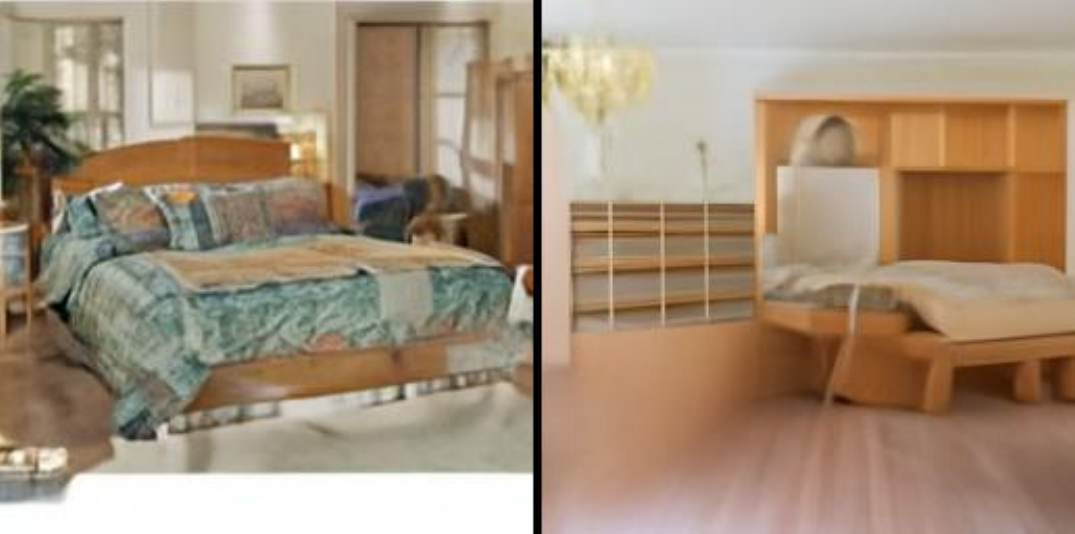} & \includegraphics[width=0.2226\textwidth]{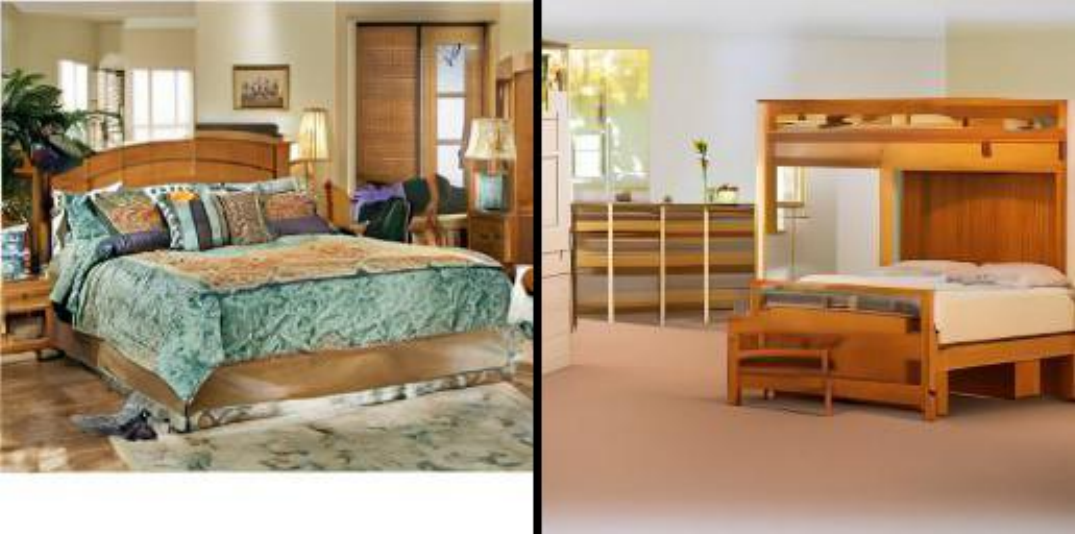} & \includegraphics[width=0.2226\textwidth]{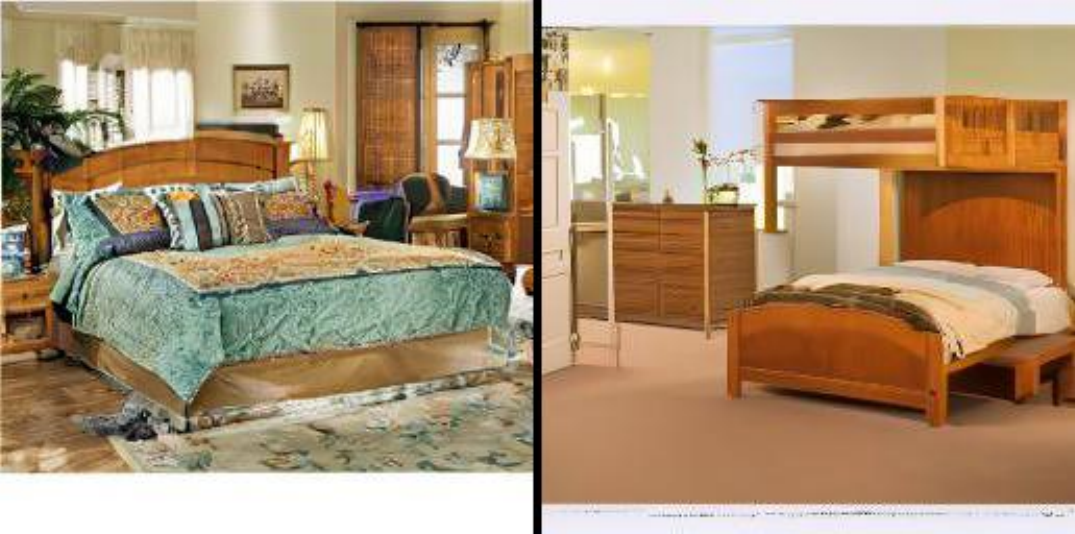} & \includegraphics[width=0.2226\textwidth]{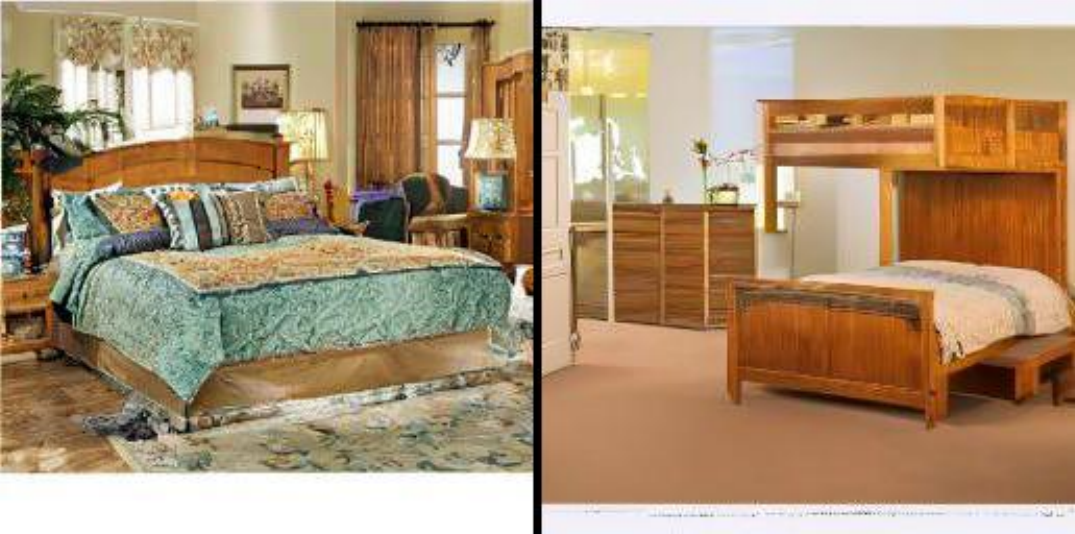}
\\
\end{tabular}
\vspace{-0.15cm}
 \caption{
Compare the generation results of the RD-based methods (Solvers: SciRE-2, SciREI-2) and the baseline method (Solver: DPM-2) using 6-36 sampling steps with the uniform time trajectory and identical  settings, on pre-trained models with ImageNet 128$\times$128 and LSUN bedroom 256$\times$256.
 }
\label{fig:roubust}
\vspace{-0.25cm}
\end{figure}

\section{Experiments}
This section show that SciRE-Solver  can significantly improve the sampling efficiency of pre-trained DPM models, including continuous-time and discrete-time DPMs.  Specifically, we conduct sampling experiments using individual SciRE-Solver-$2$ and SciRE-Solver-$3$ on the pre-trained models of the diffusion models.  For each experiments, we draw 50K samples and assess sample quality using the widely adopted FID score \cite{heusel2017gans}, where lower FID $\downarrow$ generally indicate better sample quality.
In order to facilitate the exploration of more possibilities of the SciRE-Solver our proposed and to fully utilize the given number of score function evaluations (NFE), we defined a simple combinatorial  version based on SciRE-Solver-$k$ and named as SciRE-Solver-agile, as detailed in
Appendix \ref{experiment-details}.
When comparing with other existing fast sampling algorithms, we will compare the best FID values reported by these algorithms in the relevant literature with the FID obtained by our proposed SciRE-Solver under the same NFE, as shown in Table \ref{tab:differdatasets}.
We evaluate the generative performance of DDIM, DPM-Solver, and SciRE-Solver on CIFAR-10 and CelebA 64$\times$64 datasets with the same settings and codebase, the corresponding numerical results are reported in Table \ref{tab:samecodebase}.
\begin{figure}[!ht]
\vspace{-0.3cm}
\setlength{\abovecaptionskip}{0.cm}
    \centering
    \subfigure[CIFAR-10 (discrete)]{\includegraphics[width=0.32\linewidth]{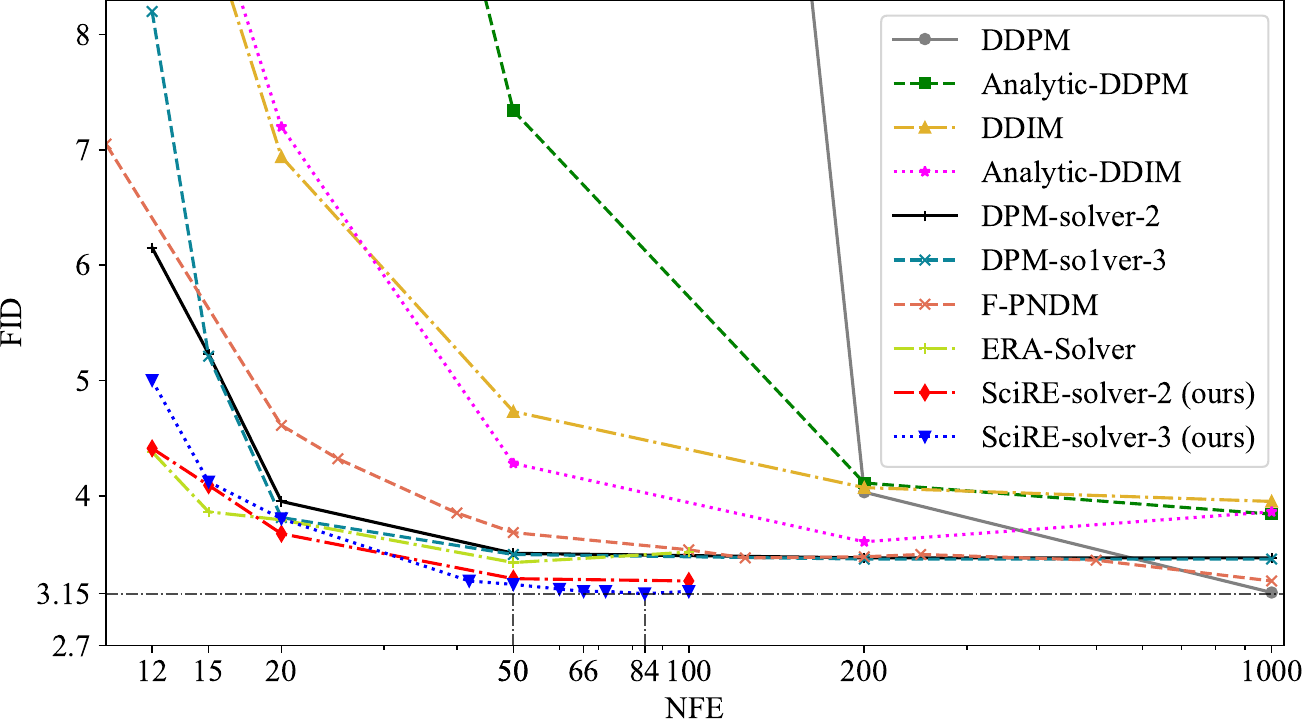}}
    \subfigure[CIFAR-10 (continuous)]{\includegraphics[width=0.32\linewidth]{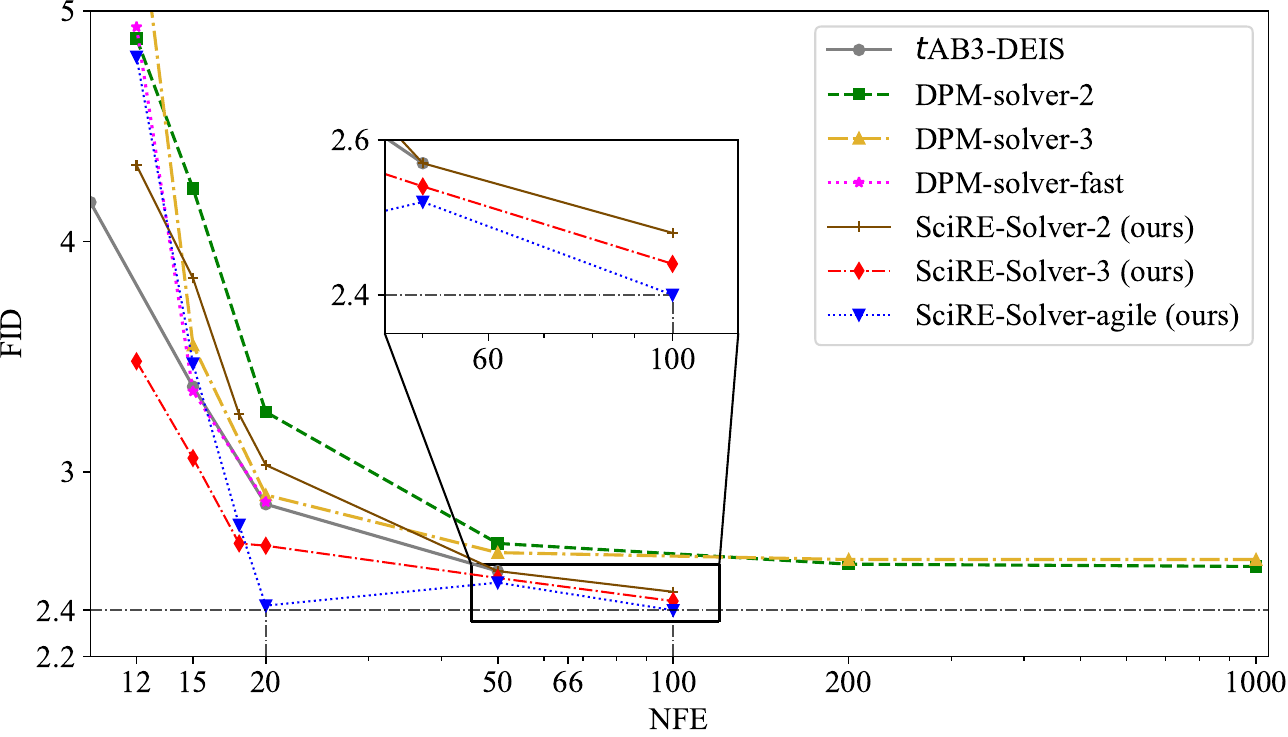}}
    \subfigure[CelebA 64$\times$64 (discrete)]{\includegraphics[width=0.32\linewidth]{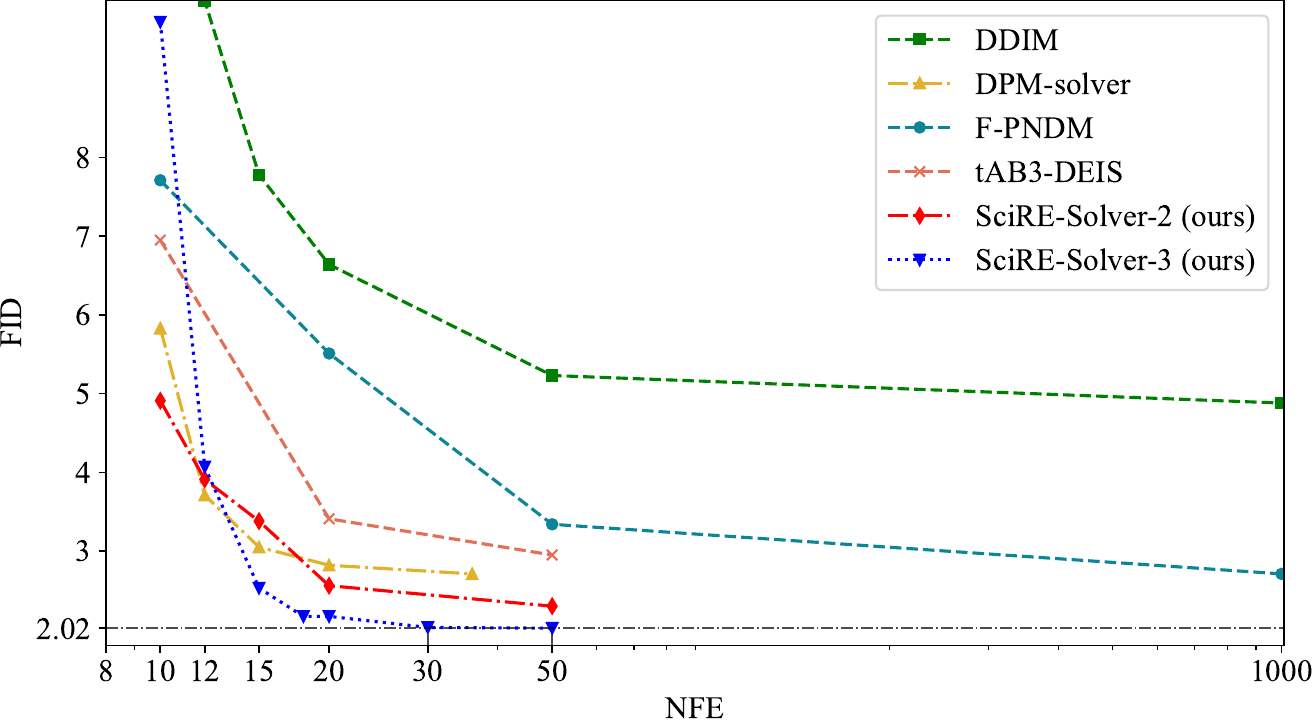}}
    \vspace{-0.05cm}
    \caption{The comparative diagram of FID $\downarrow$ of different training-free sampling methods on the CIFAR-10 and CelebA 64$\times$64 datasets.
    In these three cases, our samplers reach SOTA.}
    \label{fig:dsicrete-continuious-cifar10-results}
\setlength{\belowcaptionskip}{-0.cm}
\end{figure}
\vspace{-0.5cm}
\subsection{Experiment Setting and Ablation Study}
When running our proposed SciRE-Solver-$k$ in Algorithms \ref{algorithm:score-based-solver-2} and \ref{algorithm:score-based-solver-3}, it is necessary to assign a value $m$ to $\phi_1(m)$. As stated in Corollary \ref{reclimited}, when assigning $m$, we need to ensure that $m\geq3$. Considering that the limit of $\phi_1(m)$ is $\frac{e-1}{e}$,
then our experiments
only consider these two extreme cases, i.e., we only choose to allocate $m$ as 3 or directly set $\phi_1(m)=\frac{e-1}{e}$. We provide ablation experiments for these two cases in Appendix \ref{experiment-details}.
Moreover, in Appendix \ref{experiment-details}, we also conducted ablation experiments  on different types of time trajectories (uniform,  quadratic , logSNR, and we proposed NSR and Sigmoid),
different sampling endpoints ($1e-3$ and $1e-4$), and different \emph{NSR}-type time trajectory (setting the parameter $k$ to $3.1$ and $2$ in NSR).
The earlier experiments were all run on TITAN-V GPUs.

\subsection{Comparison with Discrete-Time Sampling Methods}
We compare SciRE-Solver proposed in Section \ref{sec3.2}  with existing discrete-time training-free methods, including DDPM \cite{ho2020denoising}, DDIM \cite{song2021denoising}, Analytic-DDPM and Analytic-DDIM \cite{bao2022analyticdpm}, PNDM \cite{liu2022pseudo}, DPM-Solver \cite{lu2022dpm},  DEIS \cite{zhang2023fast}, and ERA-Solver \cite{li2023era}. Specifically, we use the discrete-time model trained by $L_{\text{simple}}$ in \cite{ho2020denoising} on CIFAR-10 and CelebA 64$\times$64 datasets with linear noise schedule, and assign $m=3$ to $\phi_1(m)$. Under this setting, we use the same NSR-type time trajectory with fixed parameter for both SciRE-Solver-$2$ and SciRE-Solver-$3$,
the details are available in Appendix \ref{experiment-details}.
Experimental results demonstrate that SciRE-solver achieves higher sample quality compared to other samplers in an efficient manner, as shown in (a) and (c) of Fig \ref{fig:dsicrete-continuious-cifar10-results}.
SciRE-solver almost reaches convergence at around $66$ NFE and $18$ NFE, achieving the new SOTA values of $3.15$ FID with $84$ NFE, and of $2.17$ FID with $18$ NFE on CIFAR-10 and CelebA 64$\times$64, respectively.
Table \ref{tab:differdatasets} displays the specific FID scores obtained by different samplers with varying NFEs.

\subsection{Comparison with Continuous-Time Sampling Methods}\label{continu-numer}
We compare SciRE-Solver-$k$ and SciRE-Solver-agile with DPM-Solver-$k$ \cite{lu2022dpm}, DPM-Solver-fast and DEIS \cite{zhang2023fast}, where $k=2,~3$. On CIFAR-10, we use ``VP deep'' model \cite{song2021scorebased} with the linear noise schedule. When $\text{NFE}\geq15$, we employ the identical NSR-type time trajectories with consistent parametric functions for SciRE-Solver-$2$ and SciRE-Solver-$3$, respectively, the details are available in Appendix \ref{experiment-details}. Meanwhile, we consider using the sigmoid-type time trajectory only when NFE is less than $15$.
The superior of SciRE-Solver is particularly evident in its ability to generate high-quality samples with $2.42$ FID in just $20$ NFE, as shown in (b) of Fig \ref{fig:dsicrete-continuious-cifar10-results}. Furthermore, supported by several experimental validations, SciRE-Solver achieves $2.40$ FID in just $100$ NFE, which attains a new SOTA value under the VP-deep model \cite{song2021scorebased} that we used.
Table \ref{tab:differdatasets} displays the specific FID scores obtained by different samplers at varying NFEs.
\begin{SCtable}
\vspace{-0.3cm}
\small
\captionsetup{font=small, skip=1ex}
\centering
\caption{
\fontsize{8.3}{8.1}\selectfont
{
Generation quality measured by FID $\downarrow$ of different sampling methods for DPMs on CIFAR-10
and CelebA 64$\times$64 with \textit{discrete-time} or \textit{continuous-time} pre-training models.
In this Table, we compare the best FID reported in existing literature with the FID achieved by our proposed SciRE-Solver at the same NFE.
The bold black represents the best result obtained under the same NFE (column).
The results with $^{\textbf{\dag}}$ means the actual NFE is smaller than the given NFE because the given NFE cannot be divided by $2$ or $3$.
Some results are missing in their original papers, which are replaced by $``\setminus"$.  Here,
we used the same time trajectory scheme to evaluate the results of SciRE-Solver on CIFAR-10 and  CelebA 64$\times$64 datasets with discrete models. The setting of continuous-time on CIFAR-10 are described in Section \ref{continu-numer}.
More comparisons and additional details are shown in Appendix \ref{experiment-details}.}
}
\begin{tabular}{lrrrrrr}
\toprule
Sampling method \textbackslash NFE   & 12 & 15  & 20 & 50 &200 & 1000 \\
\midrule
\multicolumn{7}{l}{CIFAR-10 (discrete-time model \cite{ho2020denoising}, linear noise schedule)}
 \\
\midrule
DDPM \cite{ho2020denoising}  & $246.3$ &  $197.6$ & $137.3$ & $32.6$ & $4.03$ &  $3.16$\\
Analytic-DDPM \cite{bao2022analyticdpm}  & $27.69$ &  $20.82$& $15.35$ & $7.34$ & $4.11$ &  $3.84$\\
DDIM \cite{song2021denoising} & $11.02$ &  $8.92$& $6.94$ & $4.73$ & $4.07$ & $3.95$\\
Analytic-DDIM \cite{bao2022analyticdpm}  & $11.68$ & $9.16$ & $7.20$ & $4.28$ & $3.60$ & $3.86$\\
$t$AB3-DEIS \cite{zhang2023fast}  &\multicolumn{2}{l}{7.12(10NFE)} &  $4.53$  & $3.78$ & $\setminus$  &$\setminus$ \\
DPM-Solver-$2$ \cite{lu2022dpm} & $6.15$ &
$^{\textbf{\dag}}5.23$
 &  $3.95$  & $3.50$ & $3.46$ & $3.46$\\
DPM-Solver-$3$ \cite{lu2022dpm}  & $8.20$ & $5.21$ &  $^{\textbf{\dag}}3.81$  & $^{\textbf{\dag}}3.49$ & $^{\textbf{\dag}}3.45$ & $^{\textbf{\dag}}3.45$\\
F-PNDM \cite{liu2022pseudo} & \multicolumn{2}{l}{$7.03$(10NFE)}  & $4.61$ & $3.68$ & $3.47$ & $3.26$\\
ERA-Solver \cite{li2023era} & $\textbf{4.38}$  & $\textbf{3.86}$ & $3.79$ & $3.42$ & \multicolumn{2}{c}{$3.51$(100NFE)}\\
\midrule
SciRE-Solver-$2$ (ours) & $4.41$ & $^{\textbf{\dag}}4.09$  & $\textbf{3.67}$  & $3.28$ & \multicolumn{2}{c}{$3.26$(100NFE)}   \\
SciRE-Solver-$3$ (ours) & $5.00$ & $4.12$ & $^{\textbf{\dag}}3.80$ & $^{\textbf{\dag}}\textbf{3.23}$ &  \multicolumn{2}{c}{$\textbf{3.15}$ (84NFE)}  \\
\midrule
\multicolumn{7}{l}{CIFAR-10 (VP deep continuous-time model \cite{song2021scorebased})}
 \\
\midrule
DPM-Solver-$2$ \cite{lu2022dpm} & $4.88$ & $^{\textbf{\dag}}4.23$ &  $3.26$  & $2.69$ & $2.60$& $2.59$\\
DPM-Solver-$3$ \cite{lu2022dpm} & $5.53$ &$3.55$  &  $^{\textbf{\dag}}2.90$  & $^{\textbf{\dag}}2.65$ &$^{\textbf{\dag}}2.62$& $^{\textbf{\dag}}2.62$ \\
DPM-Solver-fast \cite{lu2022dpm} & $4.93$ &  $3.35$& $2.87$ & $\setminus$ & $\setminus$ & $\setminus$ \\
$t$AB3-DEIS \cite{zhang2023fast} & $\setminus$ &  $3.37$& $2.86$ & $2.57$ & $\setminus$ &  $\setminus$\\
\midrule
SciRE-Solver-$2$ (ours) & $4.33$ &  $^{\textbf{\dag}}3.84$ &$3.03$  & $2.57$ &   \multicolumn{2}{c}{$2.48$ (100NFE)}  \\
SciRE-Solver-$3$ (ours) & $\textbf{3.48}$&  $\textbf{3.06}$  & $^{\textbf{\dag}}2.68$ &  $^{\textbf{\dag}}2.54$ &  \multicolumn{2}{c}{$^{\textbf{\dag}}2.44$~~(100NFE)}  \\
SciRE-Solver-agile (ours) &$4.80$ &  $3.47$ & $\textbf{2.42}$ &  $\textbf{2.52}$ &\multicolumn{2}{c}{$\textbf{2.40}$ (100NFE)} \\
\toprule
\multicolumn{7}{l}{CelebA 64$\times$64 (discrete-time model \cite{song2021denoising}, linear noise schedule)}
 \\
\midrule
Sampling method  \textbackslash  NFE   & 10 & 12  & 15 & 20 &50 & 1000 \\
\midrule
DDIM \cite{song2021denoising} & $10.85$ & $9.99$ &  $7.78$  & $6.64$ & $5.23$& $4.88$\\
DPM-Solver \cite{lu2022dpm} & $5.83$ & $\textbf{3.71}$ &  $3.05$  & $2.82$ &
\multicolumn{2}{c}{$2.71$~~(36NFE)}
\\
F-PNDM \cite{liu2022pseudo} & $7.71$  & $\setminus$ & $\setminus$ & $5.51$& $3.34$ & $2.71$\\
$t$AB3-DEIS \cite{zhang2023fast} & $6.95$ & $\setminus$& $\setminus$ & $3.41$& $2.95$ &  $\setminus$\\
\midrule
SciRE-Solver-$2$ (ours) & $\textbf{4.91}$ &  $3.91$ &$^{\textbf{\dag}}3.38$  & $2.56$ &  $2.30$ & $-$\\
SciRE-Solver-$3$ (ours) & $^{\textbf{\dag}}9.72$&  $4.07$  & $\textbf{2.53}$ &  $^{\textbf{\dag}}\textbf{2.17}$ & $^{\textbf{\dag}}\textbf{2.02}$ & $-$ \\
\bottomrule
\end{tabular}
\label{tab:differdatasets}
\vspace{-0.3cm}
\end{SCtable}
\section{Conclusions}
In this work, we introduce the recursive difference (RD) method to calculate the derivative of the score function evaluations in the realm of diffusion models. By applying the RD method to the truncated Taylor expansion of the score-integrand, we propose the \emph{SciRE-Solver} with the convergence order guarantee  to accelerate the sampling process of DMs.
The effectiveness of the RD method in evaluating the derivative of score function is confirmed through comparative experiments
with both the finite-based difference algorithm and the popular DPM-Solver-2 algorithm.
Numerical experiments indicate that SciRE-Solver not only can generates high-quality samples across various datasets using fewer-steps but also,  using a small NFEs demonstrates
promising potential to surpass the FID achieved by some pre-trained models in
their original papers using no fewer than $1000$ NFEs.

\paragraph{Limitations and broader impact}
SciRE-Solver has demonstrated the ability to surpass the FID scores shown in the original papers of some pre-trained models on CIFAR-10 and CelebA 64$\times$64 datasets with a small NFEs. This indicates that developing efficient sampling algorithms can further enhance the generative capacity of existing pre-trained models.
However, fast sampling
algorithms may overly rely on local information during the acceleration process, resulting in excessive noise or bias.
In high-resolution image datasets, it is a open question whether there exists an optimal time trajectory for SciRE-Solver and other fast samplers.

\begin{ack}
The first author is very grateful for the valuable feedback received from the authors of the DPM-Solver paper during the early stage of this work regarding program-related issues.
This work was supported in
part by grants from National Science Foundation of
China (61571005), the
fundamental research program of Guangdong, China (2020B1515310023,2023A1515011281).

\end{ack}
\bibliographystyle{splncs}
\bibliography{SciRE}






\clearpage
\begin{appendices}
\section{Diffusion SDEs and ODEs}
\label{appendix:}
\subsection{Diffusion SDEs}
In the forward diffusion process of DPMs
for $D$-dimensional data, a Markov sequence $\left\{\mathbf{x}_t\right\}_{t \in[0, T]}$  with $T>0$ starting with $\mathbf{x}_0$ is defined by the transition distribution
\begin{equation}\label{Atransdis}
q\left(\mathbf{x}_t \mid \mathbf{x}_{t-1}\right):=\mathcal{N}\left(\mathbf{x}_t ; \beta_t \mathbf{x}_{t-1},\left(1-\beta_t^2\right) \mathbf{I}\right),
\end{equation}
where $\beta_t \in \mathbb{R}^{+}$ is the variance schedule function.
With the transition distribution in Eq. (\ref{transdis}), one can formulate the transition kernel of noisy data $\mathbf{x}_{t}$ conditioned on clean data $\mathbf{x}_0$ as
\begin{equation}\label{Acondix0}
q\left(\mathbf{x}_t \mid \mathbf{x}_0\right)=\mathcal{N}\left(\mathbf{x}_t; \alpha_{t} \mathbf{x}_0, \sigma_{t}^2 \mathbf{I}\right),
\end{equation}
where $\alpha_t=\prod_{i=1}^t \beta_i$, $\sigma_{t}=\sqrt{1-\alpha_t^2}$ for
the variance-preserving setting.
Clearly, $\alpha_t, \sigma_t\in \mathbb{R}^{+}$ are also differentiable function of time $t$ with  bounded derivatives like $\beta_t$.
$\alpha_t^2 / \sigma_t^2$ is strictly decreasing w.r.t. $t$, and called the signal-to-noise-ratio (SNR) in \cite{kingma2021variational}.
In \cite{kingma2021variational},
Kingma et al. established the equivalence between the transition kernel of the following
SDE and the one in Eq. (\ref{Acondix0}) for  $\forall t \in[0, T]$ :

\begin{equation}\label{Asde}
\mathrm{d} \mathbf{x}_t=f(t) \mathbf{x}_t \mathrm{~d} t+g(t) \mathrm{d} \boldsymbol{\omega}_t, \quad \mathbf{x}_0 \sim q_0\left(\mathbf{x}_0\right),
\end{equation}
where $\boldsymbol{\omega}_t \in \mathbb{R}^D$ denotes a standard Wiener process, and
\begin{equation}\label{Afg}
f(t)=\frac{\mathrm{d} \log \alpha_t}{\mathrm{~d} t}, \quad g^2(t)=\frac{\mathrm{d} \sigma_t^2}{\mathrm{~d} t}-2 \frac{\mathrm{d} \log \alpha_t}{\mathrm{~d} t} \sigma_t^2 .
\end{equation}
Further, Song et al. \cite{song2021scorebased} demonstrated with some regularity conditions that
the forward process in Eq. (\ref{Acondix0}) has the following equivalent reverse process (\emph{reverse SDE}) from time $T$ to $0$:
\begin{equation}\label{Arsde}
\mathrm{d} \mathbf{x}_t=\left[f(t) \mathbf{x}_t-g^2(t) \nabla_{\mathbf{x}} \log q_t\left(\mathbf{x}_t\right)\right] \mathrm{d} t+g(t) \mathrm{d} \overline{\boldsymbol{\omega}}_t, \quad \mathbf{x}_T \sim q_T\left(\mathbf{x}_T\right),
\end{equation}
where $\overline{\boldsymbol{\omega}}_t$ represents a standard Wiener process in the reverse time. Since $f(t)$ and $g(t)$ are determined by the noise schedule ($\alpha_t$, $\sigma_t$) in the reverse SDE (\ref{Arsde}), the sole term that remains unknown
is the score function $\nabla_{\mathbf{x}} \log q_t\left(\mathbf{x}_t\right)$ at each time $t$.
Therefore, DPMs train the model by using a neural network $\boldsymbol{\epsilon}_\theta\left(\mathbf{x}_t, t\right)$ parameterized by $\theta$ to approximate  the scaled score function: $-\sigma_t \nabla_{\mathbf{x}} \log q_t\left(\mathbf{x}_t\right)$, where the parameter $\theta$ is trained by a re-weighted variant of the evidence lower bound (ELBO)
\cite{ho2020denoising,song2021scorebased}:
\begin{align}\label{Aoptimized}
\begin{aligned}
\mathcal{L}(\theta ; \lambda(t)) &=\frac{1}{2} \int_0^T \lambda(t) \mathbb{E}_{q_t\left(\mathbf{x}_t\right)}\left[\left\|\boldsymbol{\epsilon}_\theta\left(\mathbf{x}_t, t\right)+\sigma_t \nabla_{\mathbf{x}} \log q_t\left(\mathbf{x}_t\right)\right\|_2^2\right] \mathrm{d} t \\
&=\frac{1}{2} \int_0^T \lambda(t) \mathbb{E}_{q_0\left(\mathbf{x}_0\right)}\mathbb{E}_{q(\boldsymbol{\epsilon})}\left[\left\|\boldsymbol{\epsilon}_\theta\left(\mathbf{x}_t, t\right)-\boldsymbol{\epsilon}\right\|_2^2\right] \mathrm{d} t+C,
\end{aligned}
\end{align}
where $\boldsymbol{\epsilon} \sim q(\boldsymbol{\epsilon})=\mathcal{N}(\boldsymbol{\epsilon}; \mathbf{0}, \boldsymbol{I}), \mathbf{x}_t=\alpha_t \mathbf{x}_0+\sigma_t \boldsymbol{\epsilon}$, $\lambda(t)$ represents a weighting function, and $C$ is a $\theta$-independent constant.
After the model is trained, DPMs replace the score function in Eq. (\ref{Asde}) with $-\boldsymbol{\epsilon}_\theta\left(\mathbf{x}_t, t\right) / \sigma_t$ and define the following \emph{diffusion SDE}:
\begin{equation}\label{Adsde}
\mathrm{d} \mathbf{x}_t=\left[f(t) \mathbf{x}_t+\frac{g^2(t)}{\sigma_t} \boldsymbol{\epsilon}_\theta\left(\mathbf{x}_t, t\right)\right] \mathrm{d} t+g(t) \mathrm{d} \overline{\boldsymbol{\omega}}_t, \quad \mathbf{x}_T \sim \mathcal{N}\left(\mathbf{0}, \hat{\sigma}^2 \boldsymbol{I}\right) .
\end{equation}
DPMs can generate samples by numerically solving the diffusion SDE stated in Eq. (\ref{Adsde}) using discretization methods that span from $T$ to $0$.

\subsection{ Diffusion ODEs}
Based on the reverse SDE in Eq. (\ref{Arsde}),
Song et al. \cite{song2021scorebased} derived a Liouville equation by investigating the evolution equation (Fokker-Planck Equation) of the probability density function of the variable $\mathbf{x}_t$. 
This Liouville equation has the same probability density function w.r.t.  the variable $\mathbf{x}_t$ as that of the reverse SDE. 
As a result, the reverse SDE can be transformed into the following ODE:
\begin{equation}\label{Adsode}
\frac{\mathrm{d} \mathbf{x}_t}{\mathrm{~d} t}=f(t) \mathbf{x}_t-\frac{1}{2} g^2(t) \nabla_{\mathbf{x}} \log q_t\left(\mathbf{x}_t\right), \quad \mathbf{x}_T \sim q_T\left(\mathbf{x}_T\right),
\end{equation}
where $\mathbf{x}_t$ has a marginal distribution $q_t\left(\mathbf{x}_t\right)$, which is equivalent to the marginal distribution of $\mathbf{x}_t$ of the reverse SDE in Eq.  (\ref{Arsde}).
Since the $\boldsymbol{\epsilon}_\theta\left(\mathbf{x}_t, t\right)$ trained in Eq. (\ref{Aoptimized}) can also be thought of as predicting the Gaussian noise added to $\mathbf{x}_t$,
it is commonly referred to as the  \emph{noise prediction model}. By substituting the trained noise prediction model for the score function in Eq. (\ref{Adsode}), Song et al. \cite{song2021scorebased} defined the following \emph{diffusion ODE} for DPMs:
\begin{equation}\label{Adode}
\frac{\mathrm{d} \mathbf{x}_t}{\mathrm{~d} t}=f(t) \mathbf{x}_t+\frac{g^2(t)}{2 \sigma_t} \boldsymbol{\epsilon}_\theta\left(\mathbf{x}_t, t\right), \quad \mathbf{x}_T \sim \mathcal{N}\left(\mathbf{0}, \hat{\sigma}^2 \boldsymbol{I}\right) .
\end{equation}
Therefore, one can also generate samples by solving the diffusion ODE from $T$ to $0$.

\section{Proof of Proposition 3.1}
\label{proof-of-proposition31}
Consider the following stochastic differential equation (SDE):
\begin{equation}\label{sde-appendix}
\mathrm{d} \mathbf{x}_t=f(t) \mathbf{x}_t \mathrm{~d} t+g(t) \mathrm{d} \boldsymbol{\omega}_t, \quad \mathbf{x}_0 \sim q_0\left(\mathbf{x}_0\right),
\end{equation}
where $\boldsymbol{\omega}_t \in \mathbb{R}^D$ is the standard Wiener process, and
\begin{equation}\label{fg-appendix}
f(t)=\frac{\mathrm{d} \log \alpha_t}{\mathrm{~d} t}, \quad g^2(t)=\frac{\mathrm{d} \sigma_t^2}{\mathrm{~d} t}-2 \frac{\mathrm{d} \log \alpha_t}{\mathrm{~d} t} \sigma_t^2 .
\end{equation}
Here $\alpha_t, \sigma_t\in \mathbb{R}^{+}$ are monotonic and differentiable functions of time $t$ with  bounded derivatives. 
Song et al. \cite{song2021scorebased} defined the following \emph{diffusion ODE} for Eq. (\ref{sde-appendix}):
\begin{equation}\label{dode-appendix}
\frac{\mathrm{d} \mathbf{x}_t}{\mathrm{~d} t}=f(t) \mathbf{x}_t+\frac{g^2(t)}{2 \sigma_t} \boldsymbol{\epsilon}_\theta\left(\mathbf{x}_t, t\right), \quad \mathbf{x}_T \sim \mathcal{N}\left(\mathbf{0}, \hat{\sigma}^2 \boldsymbol{I}\right) ,
\end{equation}
where $\boldsymbol{\epsilon}_\theta\left(\mathbf{x}_t, t\right)$ is the trained scale score function. The exact solution of the above semi-linear ODE can be formulated by the \emph{variation-of-constants} formula \cite{hale2013introduction}:
\begin{equation}\label{voc-appendix}
\mathbf{x}_t=e^{\int_s^t f(\gamma) \mathrm{d} \gamma} \left(\int_s^t h(\gamma)\boldsymbol{\epsilon}_\theta\left(\mathbf{x}_\gamma, \gamma\right)\mathrm{d} \gamma+\mathbf{x}_s\right),
\end{equation}
where $h(\gamma):=e^{-\int_s^\gamma f(z) \mathrm{d} z} \frac{g^2(\gamma)}{2 \sigma_\gamma}$, and $\mathbf{x}_s$ represents the given initial value. Since $f(\gamma)=\frac{\mathrm{d} \log\left(\alpha_\gamma\right)}{\mathrm{~d} \gamma}$, thus
$h(\gamma)=\frac{\alpha_s}{\alpha_\gamma}\frac{g^2(\gamma)}{2 \sigma_\gamma}$.
We observe that $h(\gamma)$ can be rewritten as
\begin{equation}\label{nsr-appendix}
  h(\gamma)=\frac{\alpha_s}{2\alpha_\gamma\sigma_\gamma}\left(\frac{\mathrm{d} \sigma_\gamma^2}{\mathrm{~d} \gamma}-2 \frac{\mathrm{d} \log \alpha_\gamma}{\mathrm{~d} \gamma}\sigma_\gamma^2\right)=\frac{\alpha_s}{\alpha_\gamma}\left(
 \frac{\mathrm{d} \sigma_\gamma}{\mathrm{~d} \gamma}
 - \frac{\sigma_\gamma
 }{\alpha_\gamma}
\frac{\mathrm{d} \alpha_\gamma}{\mathrm{~d} \gamma}
 \right)
 =\alpha_s\frac{\mathrm{d}{\rm{NSR}}(\gamma)}{\mathrm{~d} \gamma},
\end{equation}
where ${\rm{NSR}}(\gamma):= \frac{\sigma_\gamma
 }{\alpha_\gamma}$, and we refer to it as the time-dependent
 \emph{noise-to-signal-ratio (NSR)} function. Note that the NSR function defined above differs from the signal-to-noise-ratio (SNR) function defined in \cite{kingma2021variational}, but there is a relationship between them: $\rm{SNR}=\frac{1}{\rm{NSR}^2}$.
Then, based on Eq. (\ref{nsr-appendix}), we can rewrite Eq. (\ref{voc-appendix}) as
\begin{equation}\label{rewritevoc-appendix}
\mathbf{x}_t=\frac{\alpha_t
 }{\alpha_s}\mathbf{x}_s +\alpha_t\int_s^t
\frac{\mathrm{d} {\rm{NSR}}(\gamma)}{\mathrm{~d} \gamma}
\boldsymbol{\epsilon}_\theta\left(\mathbf{x}_\gamma, \gamma\right)\mathrm{d} \gamma.
\end{equation}

Since $\rm{NSR}(\cdot)$ is a monotonically function w.r.t. time, we can define its reverse function as ${\rm{rNSR}}(\cdot)$, such that  $\gamma={{\rm{rNSR}}}\left({{\rm{NSR}}}(\gamma)\right)$ for any diffusion time $\gamma$. Thus, using the \emph{change-of-variable}  for  ${\rm{NSR}}(\gamma)$ to Eq. (\ref{rewritevoc-appendix}), we can obtain
\begin{equation} \label{exactsolution-appendix}
    \begin{aligned}
        \mathbf{x}_t&=\frac{\alpha_t}{\alpha_s}\mathbf{x}_s +\alpha_t\int_s^t \frac{\mathrm{d} {\rm{NSR}}(\gamma)}{\mathrm{~d} \gamma}\boldsymbol{\epsilon}_\theta\left(\mathbf{x}_\gamma, \gamma\right)\mathrm{d} \gamma \\
        &=\frac{\alpha_t}{\alpha_s}\mathbf{x}_s+\alpha_t\int_s^t \boldsymbol{\epsilon}_\theta\left(\mathbf{x}_\gamma, \gamma\right) \mathrm{d} {\rm{NSR}}(\gamma) \\
        &=\frac{\alpha_t}{\alpha_s}\mathbf{x}_s+\alpha_t\int_{{{\rm{NSR}}}(s)}^{{{\rm{NSR}}}(t)}
\boldsymbol{\epsilon}_\theta\left(\mathbf{x}_{{\rm{rNSR}}(\tau)}, {\rm{rNSR}}(\tau)\right)\mathrm{d} \tau.
    \end{aligned}
\end{equation}
Now we complete the proof the Proposition 3.1.

\paragraph{Hints} Eq. (\ref{exactsolution-appendix})  implies that the solution of the diffusion ODE can be decomposed into a linear part and a nonlinear part, and this structure arises from the use of the \emph{variation-of-constants} formula \cite{hale2013introduction}. The linear part can be computed analytically, while the remaining nonlinear part is an integral involving the neural network of score function evaluations. Compared with directly numerically solving the diffusion ODE, such decomposition method can reduce numerical errors and improve calculation accuracy because the linear part can be analytically computed, as demonstrated by DPM-Solver \cite{lu2022dpm}. Now, we observe that the integral term on the r.h.s. of Eq. (\ref{exactsolution-appendix}) appears to be a traditional integration problem, involving the score function as the integrand solely. Thus, we could use conventional numerical methods for solving integrals to evaluate it.
However, caution must be exercised when employing these methods, as the integrand is merely an approximation of the scaled score function, and its explicit expression remains unknown, while the integrand involves some large-scale neural networks.
Therefore, using traditional techniques to accelerate the sampling process of diffusion models may amplify  the numerical error in such  scenarios. 
Nonetheless, in the realm of diffusion models, we can draw inspiration from traditional numerical techniques to develop fast sampling algorithms suitable for diffusion models.

\begin{figure}[ht]
\vspace{-0.4cm}
\centering
\begin{tabular}{m{0.8cm}p{2.7cm}p{2.7cm}p{2.7cm}p{2.7cm}}
   ~~&~~~~~~~~~~\,~NFE=$6$& ~~~~\,~~~~~~NFE=$12$  &~~~~\,~~~~~~NFE=$24$ &~~~\,~~~~~~~NFE=$36$ \\
\multirow{-4.6}{*}{\parbox{0.7cm}{\centering {\small DPM-2} }}
&\includegraphics[width=0.216\textwidth]{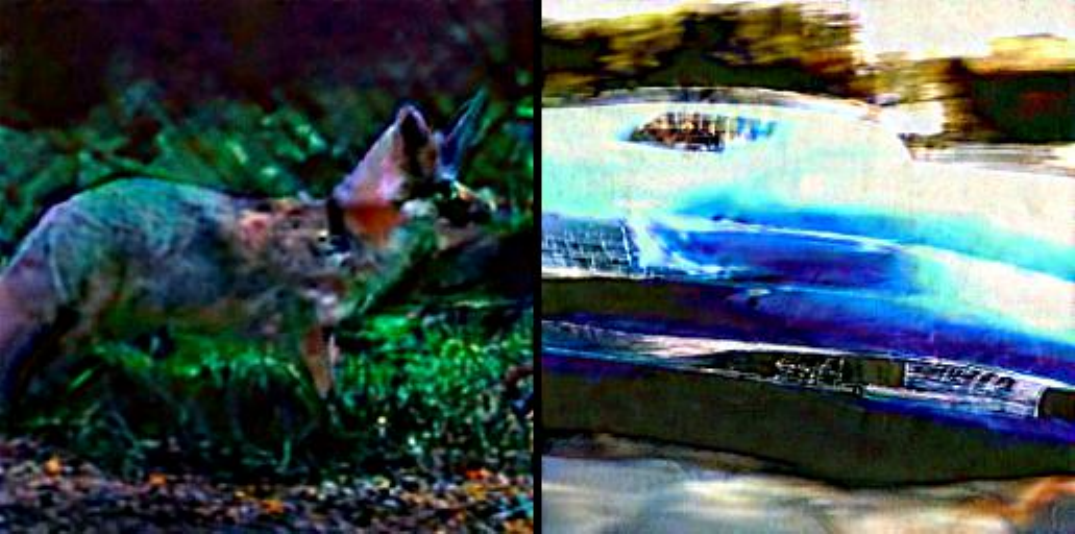} & \includegraphics[width=0.216\textwidth]{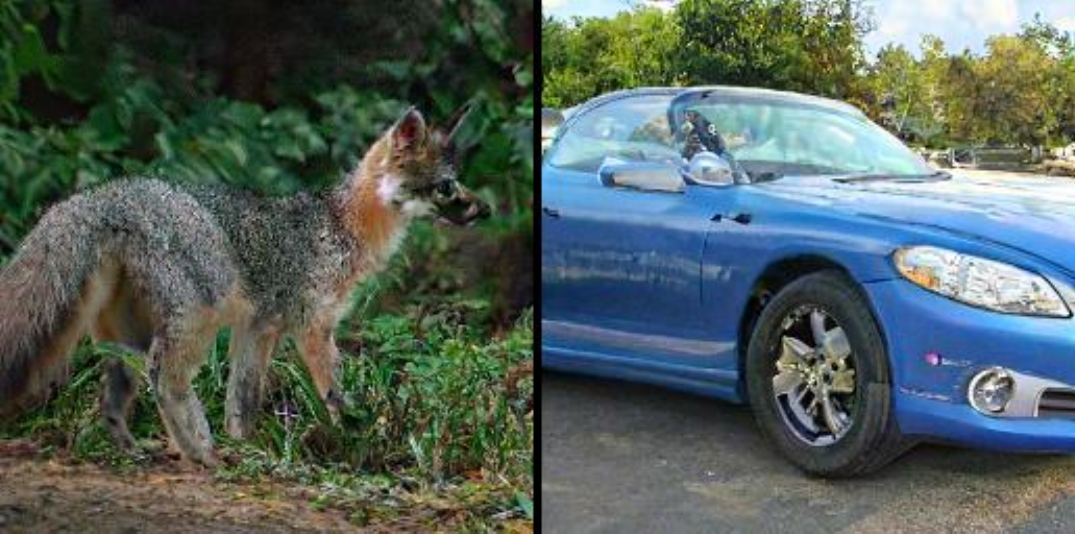} & \includegraphics[width=0.216\textwidth]{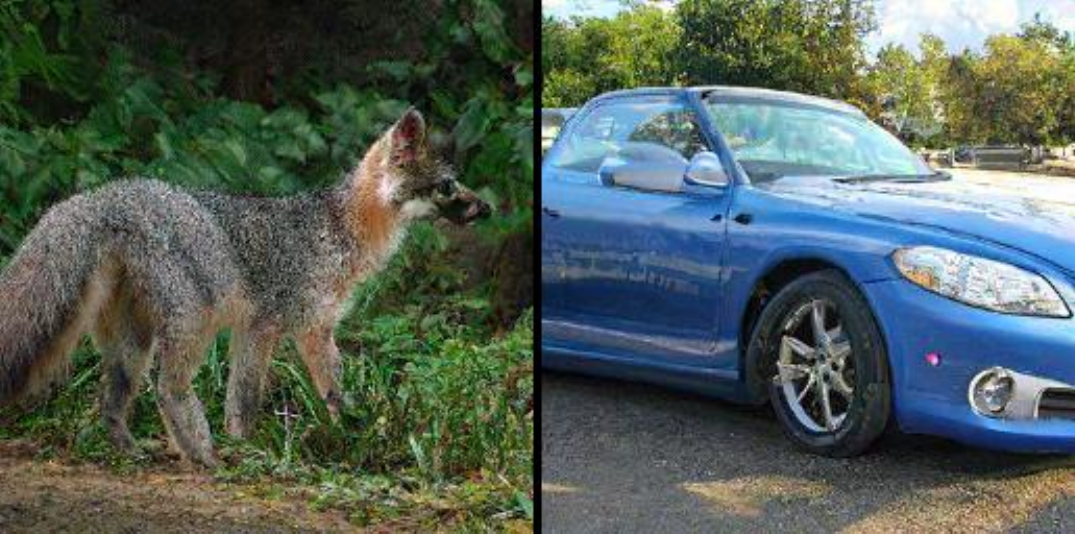} & \includegraphics[width=0.216\textwidth]{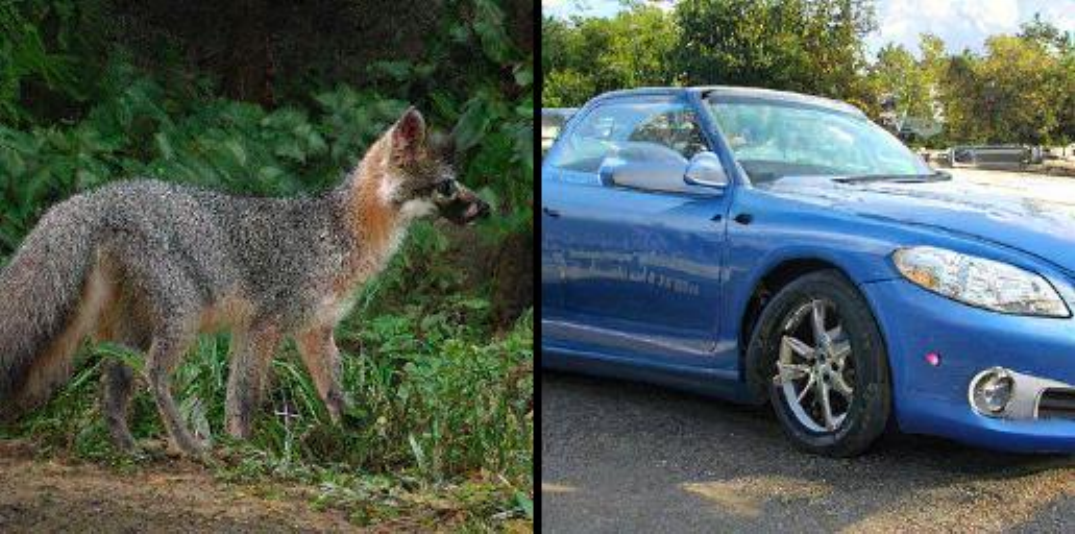}
\\
\multirow{-4.6}{*}{\parbox{0.7cm}{\centering {\small SciREI-2(\textbf{ours})}}}
&\includegraphics[width=0.216\textwidth]{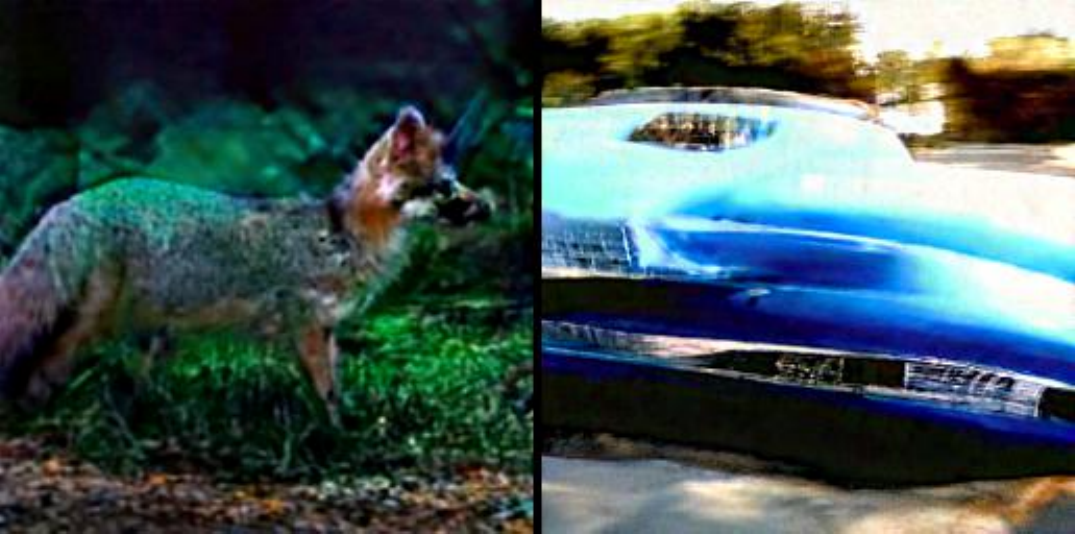} & \includegraphics[width=0.216\textwidth]{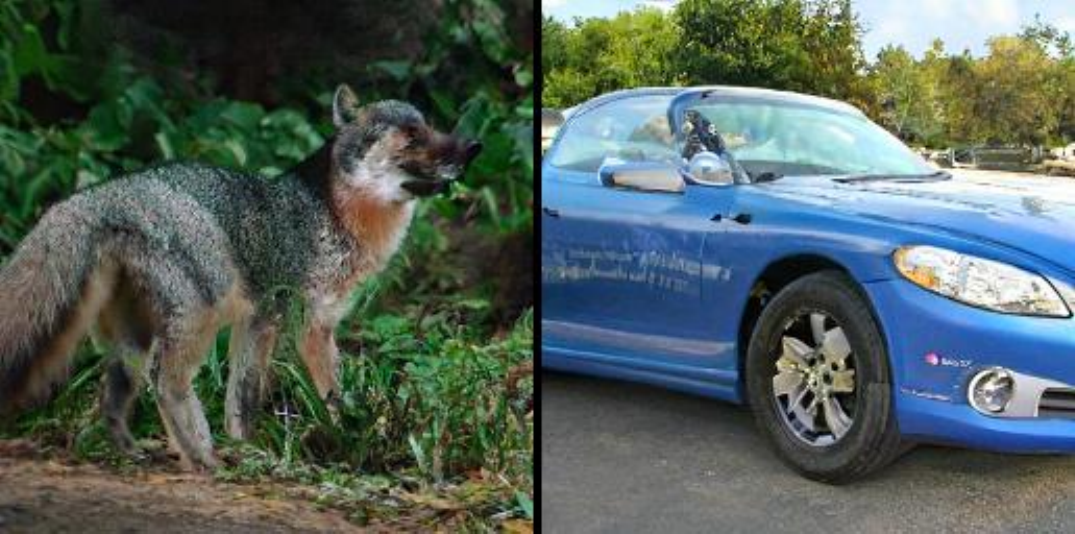} & \includegraphics[width=0.216\textwidth]{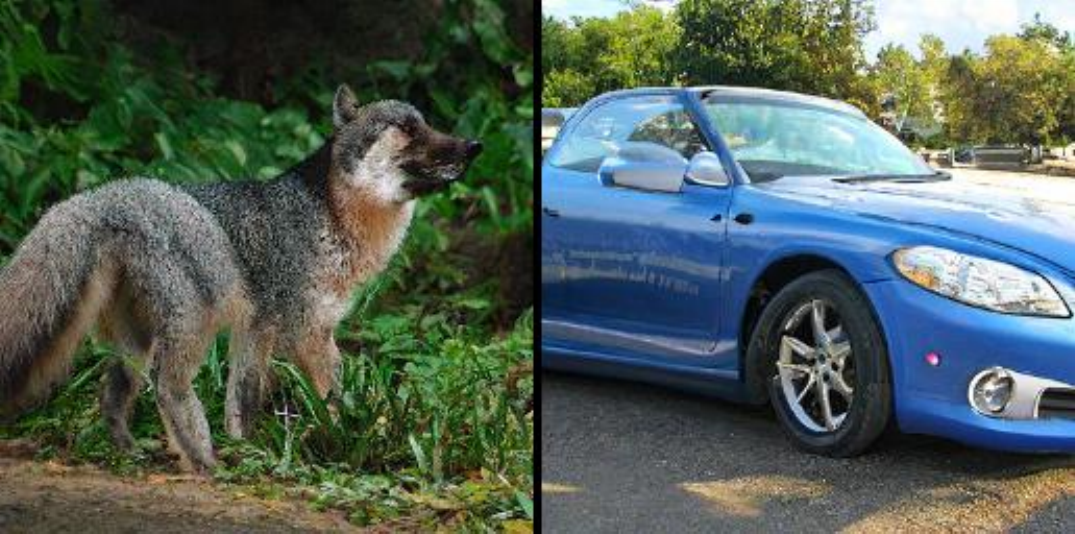} & \includegraphics[width=0.216
\textwidth]{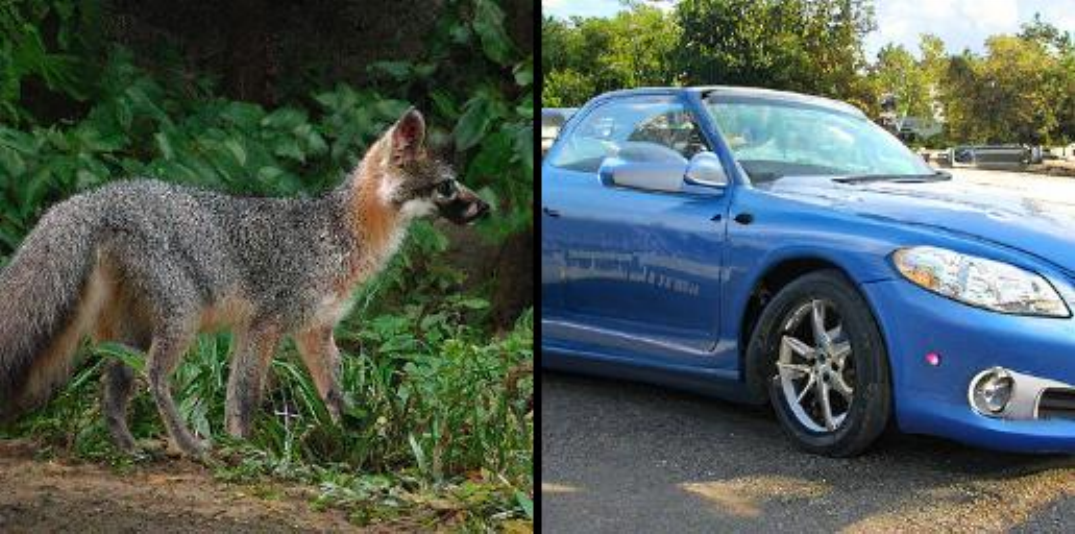}
\\
\multirow{-4.6}{*}{\parbox{0.7cm}{\centering {\small SciRE-2(\textbf{ours})}}}
& \includegraphics[width=0.216\textwidth]{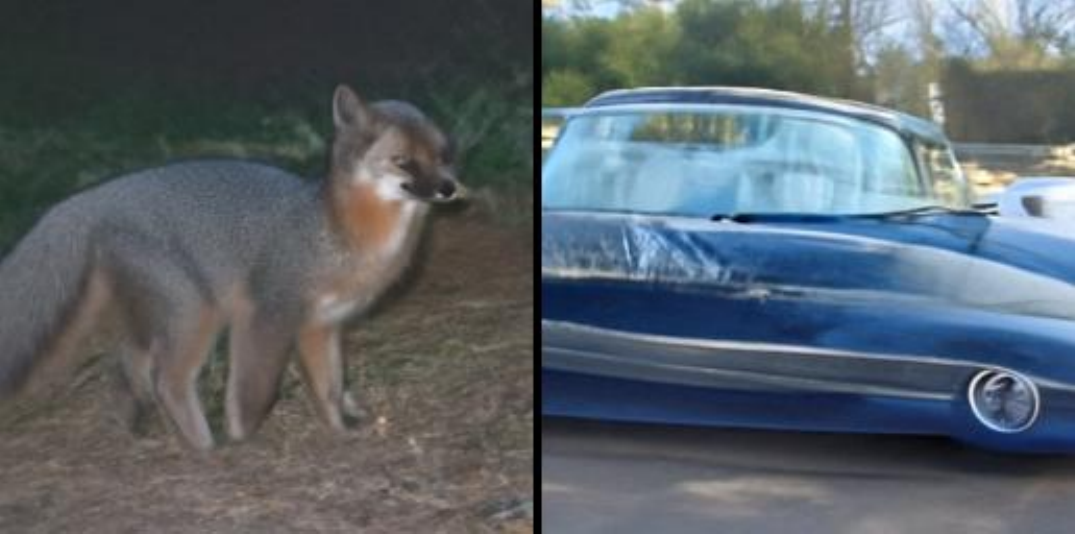} & \includegraphics[width=0.216\textwidth]{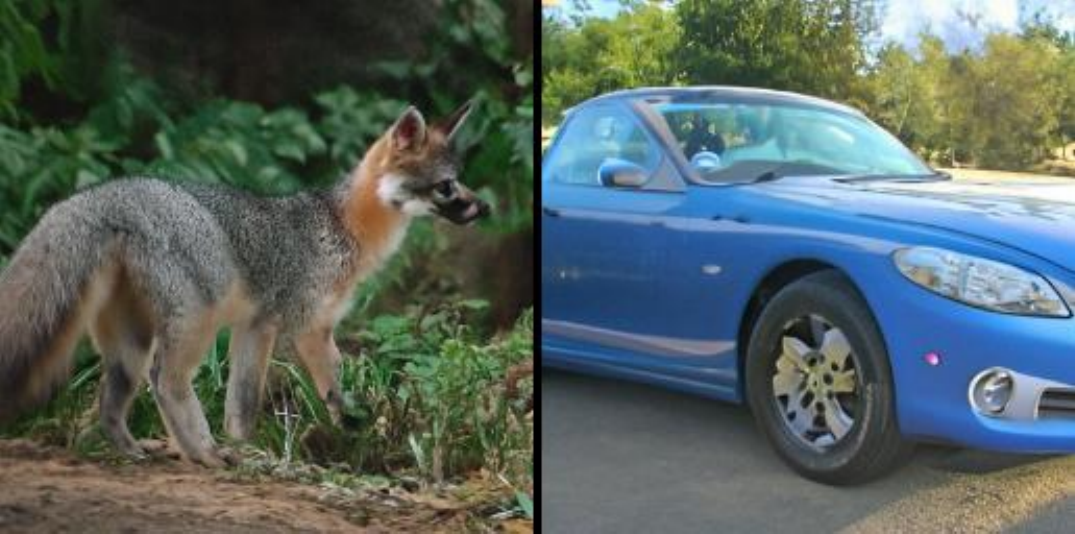} & \includegraphics[width=0.216\textwidth]{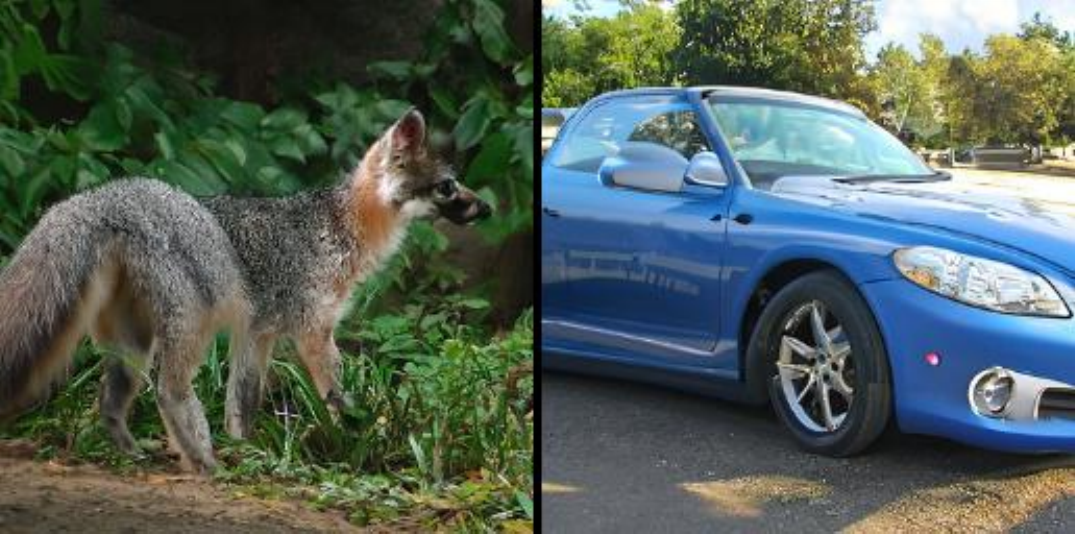} & \includegraphics[width=0.216\textwidth]{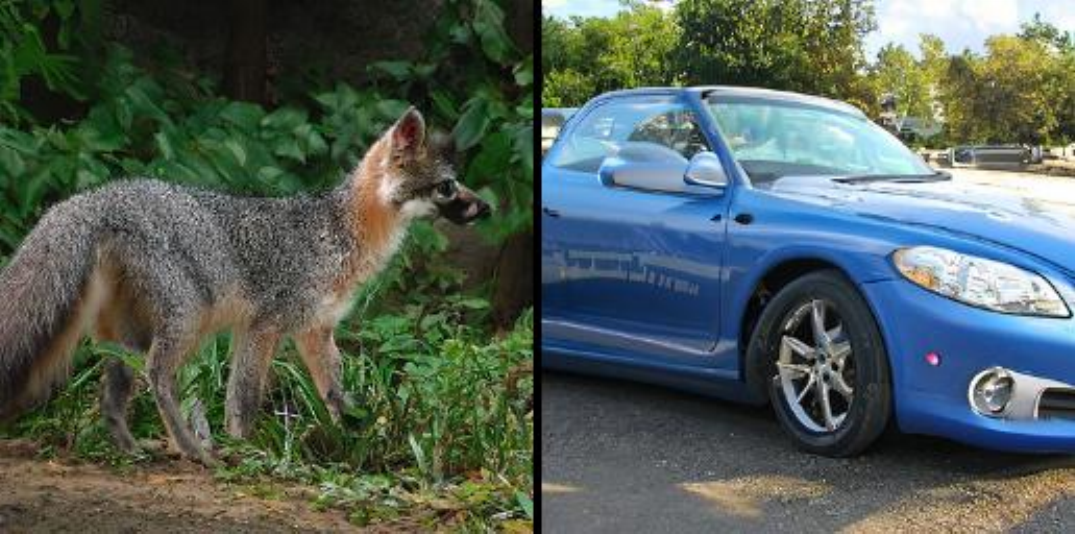}
\\
\multirow{-9}{*}{\parbox{0.7cm}{\centering {\small DPM-2} }}
&\includegraphics[width=0.216\textwidth]{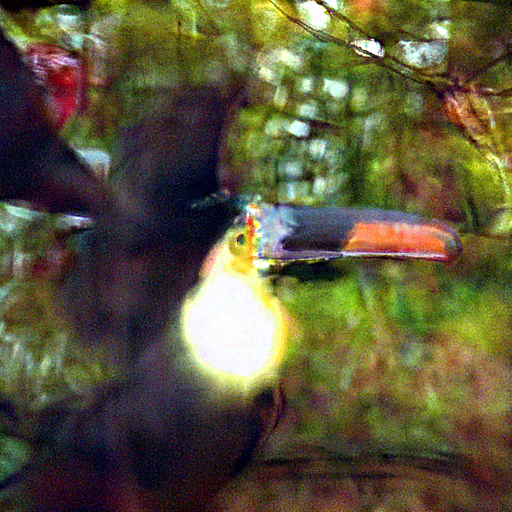} & \includegraphics[width=0.216\textwidth]{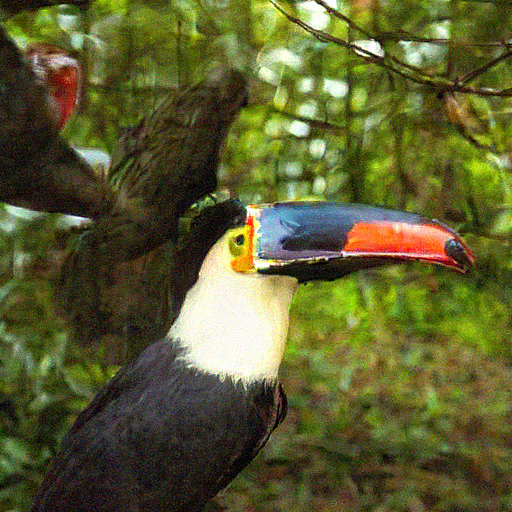} & \includegraphics[width=0.216\textwidth]{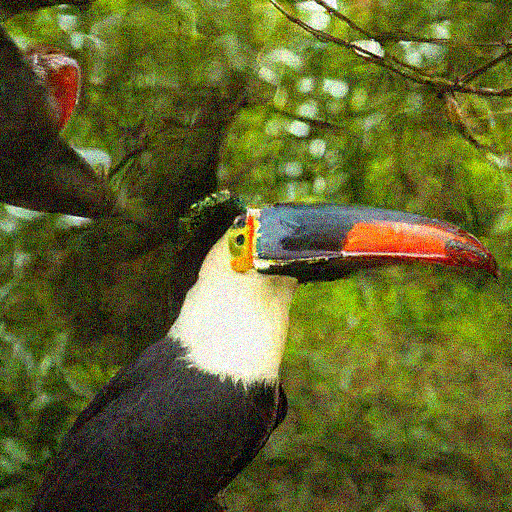} & \includegraphics[width=0.216\textwidth]{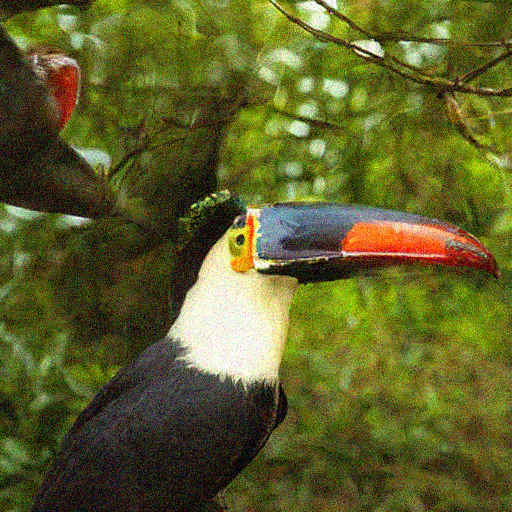}
\\
\multirow{-9}{*}{\parbox{0.6cm}{\centering {\small SciREI-2(\textbf{ours})}}}
& \includegraphics[width=0.216\textwidth]{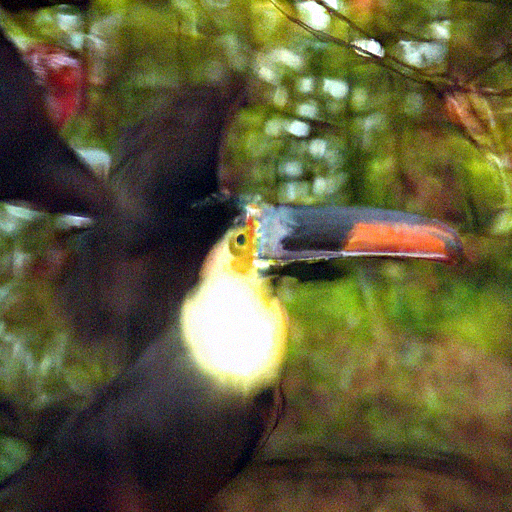}
& \includegraphics[width=0.216\textwidth]{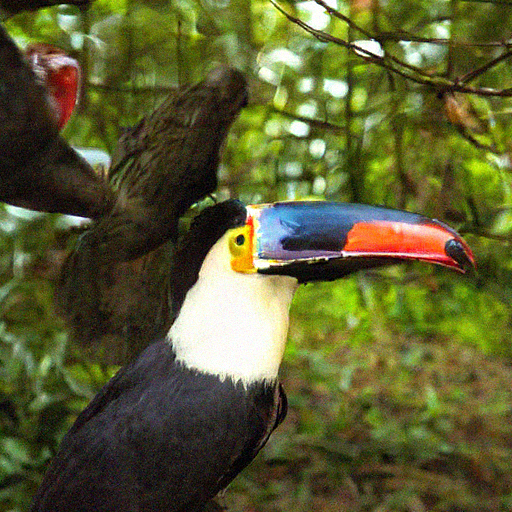}
& \includegraphics[width=0.216\textwidth]{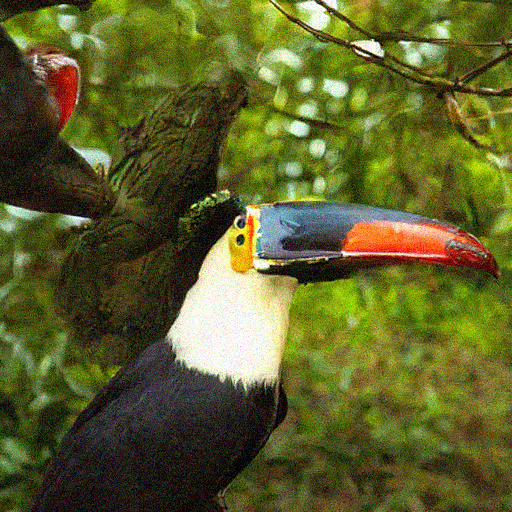} & \includegraphics[width=0.216\textwidth]{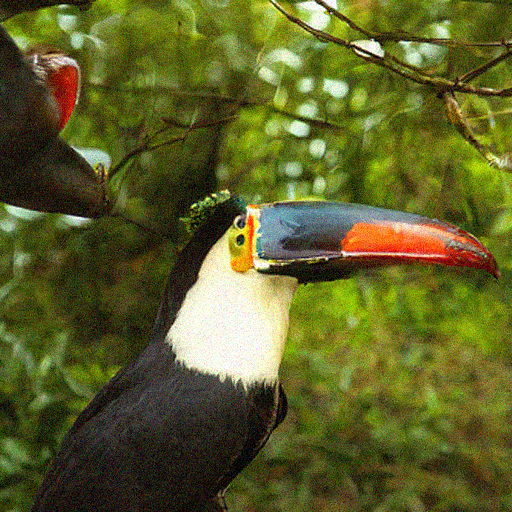}
\\
\multirow{-9}{*}{\parbox{0.7cm}{\centering {\small SciRE-2(\textbf{ours})}}}
& \includegraphics[width=0.216\textwidth]{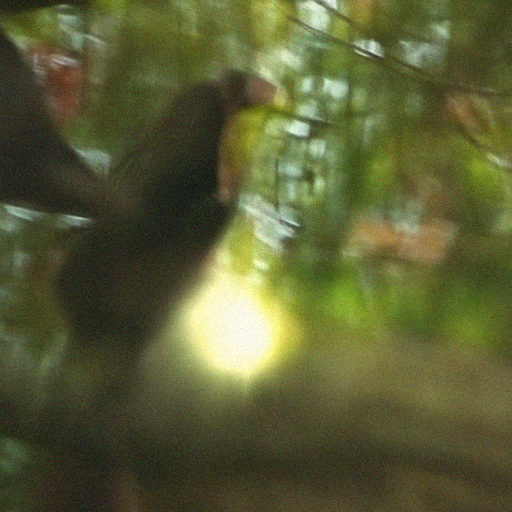}
& \includegraphics[width=0.216\textwidth]{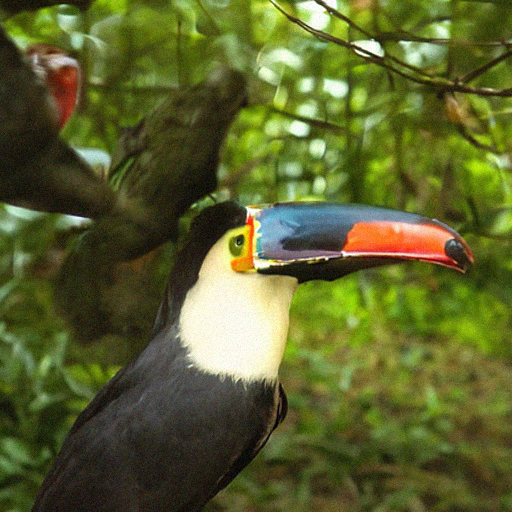}
& \includegraphics[width=0.216\textwidth]{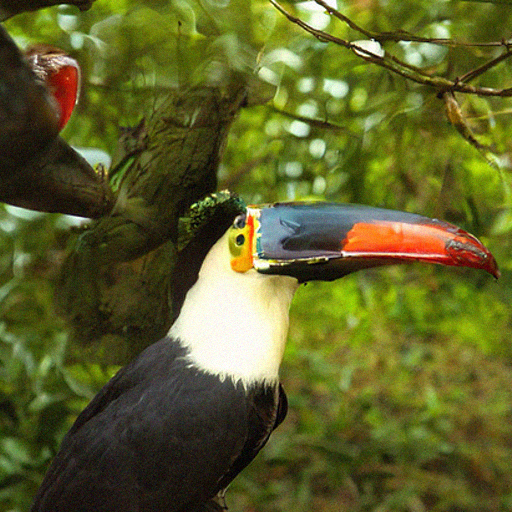}
& \includegraphics[width=0.216\textwidth]{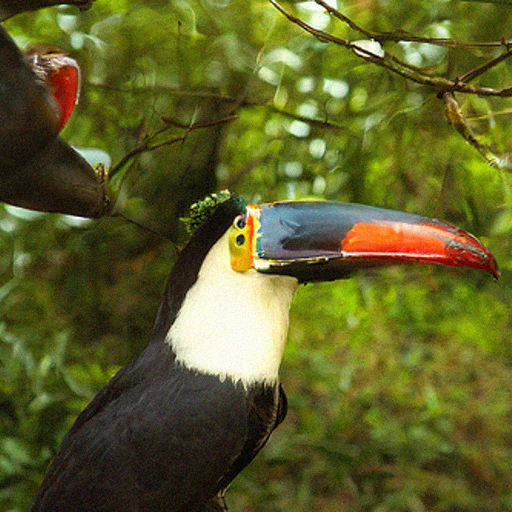}
\\
\end{tabular}
 \caption{
Compare the generation results of the RD-based methods (Solvers: SciRE-2, SciREI-2) and the baseline method (Solver: DPM-2) using 6-36 sampling steps with the uniform time trajectory and identical  settings, on pre-trained models with ImageNet 256$\times$256 and 512$\times$512 datasets.
 }
\label{fig:fulu}
\end{figure}
\section{
SciREI-Solver, and Compared to DPM-Solver-2 for Assessing the Benefits with RD
}\label{gexct}
In order to  further explore the effectiveness of the RD method, we  propose a variant named \emph{SciREI-Solver}, which incorporates the RD method and the exponential-based calculation formula provided by DPM-Solver. We provide numerical experiments to demonstrate the benefits of the RD method.
\subsection{SciREI-Solver}\label{ascireisolver}
\begin{table}
\centering
\caption{
Comparison of Quality Generation between FD-based and RD-based Algorithms.
We use consistent \emph{NSR} trajectory (\textbf{k=2}) and the same codebase.}
\label{tab:rdenonrde}
\begin{tabular}{clrrrrrrr}
\toprule
 FD or RD & Initial time$\backslash$ NFE & 12 & 15 & 20 & 50 & 100 \\
\midrule \multicolumn{6}{l}{The discrete-time model of CIFAR-10 dataset \cite{ho2020denoising} }   \\
\midrule
\multirow{2}{*}{FD  (Solver-2)}&$1e-3$ & $7.00$ &  $^{\textbf{\dag}}6.00$ & $4.76$ & $4.10$  & $4.03$\\
&$1e-4$ & $9.03$ & $^{\textbf{\dag}}7.10$ &  $5.05$&  $3.71$ & $3.57$ \\
\hline
\multirow{2}{*}{RD  (SciRE-Solver-2)}&$1e-3$ & $\textbf{4.49}$& $^{\textbf{\dag}}\textbf{4.12}$ & $\textbf{3.74}$  & $3.70$ &$3.76$ \\
&$1e-4$ & $5.91$ &  $^{\textbf{\dag}}4.76$ & $3.88$ & $\textbf{3.30}$ & $\textbf{3.28}$\\
\midrule
\multirow{2}{*}{FD  (Solver-3)}&$1e-3$ & $6.91$ &  $5.25$& $^{\textbf{\dag}}4.67$&  $^{\textbf{\dag}}4.04$ & $^{\textbf{\dag}}4.02$\\
&$1e-4$ & $10.19$ & $6.12$& $^{\textbf{\dag}}5.03$ &  $^{\textbf{\dag}}3.56$ &$^{\textbf{\dag}}3.52$  \\
\hline
\multirow{2}{*}{RD  (SciRE-Solver-3)}&$1e-3$ &$\textbf{5.29}$ & $\textbf{4.19}$ &  $^{\textbf{\dag}}\textbf{3.94}$& $^{\textbf{\dag}}3.76$& $^{\textbf{\dag}}3.71$ \\
&$1e-4$ & $9.10$ & $4.52$  & $^{\textbf{\dag}}4.07$  & $^{\textbf{\dag}}\textbf{3.24}$  & $^{\textbf{\dag}}\textbf{3.17}$ \\
\midrule
\multicolumn{6}{l}{The discrete-time model of CelebA 64$\times$64 dataset \cite{ho2020denoising} }   \\
\midrule
\multirow{2}{*}{FD  (Solver-2)}&$1e-3$ & $7.82$ & $^{\textbf{\dag}}6.87$ & $5.48$& $4.48$  & $4.33$\\
&$1e-4$ & $7.04$  &$^{\textbf{\dag}}5.87$  & $4.17$ & $3.05$  & $2.89$ \\
\hline
\multirow{2}{*}{RD  (SciRE-Solver-2)}&$1e-3$ & $4.67$ & $^{\textbf{\dag}}4.23$ & $3.63$  &$3.60$  & $3.79$\\
&$1e-4$ &$\textbf{3.99}$ & $^{\textbf{\dag}}\textbf{3.43}$ & $\textbf{2.63}$& $\textbf{2.32}$ & $\textbf{2.43}$\\
\midrule
\multirow{2}{*}{FD  (Solver-3)}&$1e-3$ & $8.09$ & $6.29$ &$^{\textbf{\dag}}5.35$ &   $^{\textbf{\dag}}4.28$& $^{\textbf{\dag}}4.27$\\
&$1e-4$ & $7.66$ & $5.20$ & $^{\textbf{\dag}}4.25$ & $^{\textbf{\dag}}2.88$  & $^{\textbf{\dag}}2.83$ \\
\hline
\multirow{2}{*}{RD  (SciRE-Solver-3)}&$1e-3$ & $4.79$& $3.37$ & $^{\textbf{\dag}}3.08$ & $^{\textbf{\dag}}3.18$& $^{\textbf{\dag}}3.54$ \\
&$1e-4$ & $\textbf{4.50}$ & $\textbf{2.70}$  & $^{\textbf{\dag}}\textbf{2.30}$  &  $^{\textbf{\dag}}\textbf{2.02}$ &$\textbf{2.20}$  \\
\bottomrule
\end{tabular}
\vspace{-0.2cm}
\end{table}
In this work, we introduced the RD method to evaluating the derivative of the scaled score function, in light of these results, we proposed the SciRE-Solver with the truncated Taylor expansion of the score-integrand. To further investigate  the effectiveness the RD method in the realm of sampling for diffusion models, we apply the RD method to the exponential-based contextualisation provided by the DPM-Solver \cite{lu2022dpm}, and proposed the \emph{SciREI-Solver}.
We specifically investigate the RD method in the context of ``the generalized version of DPM-Solver-2, i.e., the Algorithm 4 in the Appendix of the DPM-Solver paper" (referred to as DPM-Solver-2 throughout this paper for simplicity).
Since both SciREI-Solver and DPM-Solver-2 are derived when $n=2$, we occasionally refer to \emph{SciREI-Solver} as \emph{SciREI-Solver-2} for to enhance clarity in comparisons.
\begin{algorithm}
	\setstretch{1.0}
	\renewcommand{\algorithmicrequire}{\textbf{Require:}}
	\renewcommand{\algorithmicensure}{\textbf{Return:}}
	\caption{SciREI-Solver ~(or SciREI-Solver-2)}
	\label{algorithm:scirei-solver}
	\begin{algorithmic}[1]
		\Require initial value $\mathbf{x}_T$, time trajectory $\left\{t_i\right\}_{i=0}^N$, model $\boldsymbol{\epsilon}_\theta, m\geq3$
            \State $\tilde{\mathbf{x}}_{t_N} \leftarrow \mathbf{x}_T,  r_1\leftarrow\frac{1}{2}$
		\For{$i\leftarrow $ $N$ to $0$}
                \State $ h_i~~\leftarrow ~\lambda_{t_{i-1}} - \lambda_{t_i}$
                \State $ s_i~~\leftarrow ~t_\lambda\left(\lambda_{t_{i}}+r_1 h_i\right)$
                \State $ \tilde{\mathbf{x}}_{s_i} ~\leftarrow ~\frac{\alpha_{s_i}}{\alpha_{t_{i}}} \tilde{\mathbf{x}}_{t_{i}}-\sigma_{s_i}\left(e^{r_1 h_i}-1\right) \boldsymbol{\epsilon}_\theta\left(\tilde{ \mathbf{x}}_{t_{i}}, t_{i}\right)$
                \State $\tilde{\mathbf{x}}_{t_{i-1}} \leftarrow
                \frac{\alpha_{t_{i-1}}}{\alpha_{t_i}} \tilde{\mathbf{x}}_{t_i}-\sigma_{t_{i-1}}(e^{h_i}-1) \boldsymbol{\epsilon}_\theta(\tilde{\mathbf{x}}_{t_i}, t_i)
                - \frac{\sigma _{t_{i-1}}}{\phi_1(m) r_1 h_i }(e^{h_i}-h_i-1)\left(\boldsymbol{\epsilon}_\theta(\tilde{\mathbf{x}}_{s_i}, s_i)-\boldsymbol{\epsilon}_\theta(\tilde{\mathbf{x}}_{t_i}, t_i)\right)$

            \EndFor
		
		\Ensure $\tilde{\mathbf{x}}_{0}$.
	\end{algorithmic}
\end{algorithm}

Formally, \cite{lu2022dpm} provides an exponential contextualized solution formula for the diffusion ODE: 
\begin{equation}\label{dpmsolver}
    \mathbf{x}_t=\frac{\alpha_t}{\alpha_s} \mathbf{x}_s-\alpha_t \int_{\lambda_s}^{\lambda_t} e^{-\lambda} \hat{\boldsymbol{\epsilon}}_\theta\left(\hat{\mathbf{x}}_\lambda, \lambda\right) \mathrm{d} \lambda
\end{equation}
where $\lambda_t=\log\frac{\alpha_t}{\sigma_t}$, $\hat{\boldsymbol{\epsilon}}_\theta\left(\hat{\mathbf{x}}_\lambda, \lambda\right) = \boldsymbol{\epsilon}_\theta\left(\mathbf{x}_{t(\lambda)}, t(\lambda)\right)$, and $t(\lambda)$ is the inverse function of $\lambda_t$ w.r.t. time $t$.
Under this solution formula of exponential-based contextualisation,
the formula below is obtained by Taylor expansion around $\lambda$:
\begin{equation}\label{dpmsolver1}
\mathbf{x}_t=\frac{\alpha_t}{\alpha _s} \mathbf{x}_s-\sigma _t \sum _{k=0}^n h^{k+1} \varphi _{k+1}(h) \hat{\epsilon} _\theta^{(k)}\left(\hat{\mathbf{x}}_{\lambda _s}, \lambda _s\right)+\mathcal{O}(h^{n+2}),
\end{equation}
where
$\varphi_k(h)=\int_0^1 e^{(1-\delta)h} \frac{\delta^{k-1}}{(k-1) !} \mathrm{d} \delta$, $\varphi_0(h)=e^h$.

When $n=1$ in Eq. (\ref{dpmsolver1}), we have then
\begin{equation}\label{dpmsolver2}
\mathbf{x}_t=\frac{\alpha _t}{\alpha _s} \mathbf{x}_s-\sigma _t h \varphi _1(h) \hat{\boldsymbol{\epsilon}}_\theta(\hat{\mathbf{x}}_{\lambda_s}, {\lambda_s})-\sigma _t h^2 \varphi _2(h) \hat{\boldsymbol{\epsilon}}_\theta^{(1)}(\hat{\mathbf{x}} _{\lambda_s}, \lambda_s) + \mathcal{O}(h^3)
\end{equation}
where
\begin{equation}
\varphi _1(h)=\frac{e^h-1}{h}, ~~\varphi _2(h)=\frac{e^h-h-1}{h^2}.
\end{equation}
The following iteration is obtained by DPM-Solver-2:
\begin{equation}\label{dpmsolver3}
\mathbf{x}_t=\frac{\alpha _t}{\alpha_s} \mathbf{x}_s-\sigma _t(e^h-1) \hat{\boldsymbol{\epsilon}}_\theta(\hat{\mathbf{x}}_{\lambda_s}, {\lambda_s})-\textcolor{blue}{\frac{\sigma _t}{2r_1}(e^h-1)}\left(\hat{\boldsymbol{\epsilon}}_\theta(\hat{\mathbf{x}}_{\lambda_{s_1}}, \lambda_{s_1})-\hat{\boldsymbol{\epsilon}}_\theta(\hat{\mathbf{x}}_{\lambda_s}, {\lambda_s})\right).
\end{equation}
With our proposed the recursive difference (RDE) method to evaluate $\hat{\boldsymbol{\epsilon}}_\theta^{(1)}(\hat{\mathbf{x}} _{\lambda_s}, \lambda_s)$, we get the following new iteration:
\begin{equation}\label{scirei-solver}
\mathbf{x}_t=\frac{\alpha _t}{\alpha_s} \mathbf{x}_s-\sigma_t(e^h-1) \hat{\boldsymbol{\epsilon}}_\theta(\hat{\mathbf{x}}_{\lambda_s}, {\lambda_s})- \textcolor{orange}{\frac{\sigma _t}{\phi_1(m) r _1 h }(e^h-h-1)}\left(\hat{\boldsymbol{\epsilon}}_\theta(\hat{\mathbf{x}}_{\lambda_{s_1}}, \lambda_{s_1})-\hat{\boldsymbol{\epsilon}}_\theta(\hat{\mathbf{x}}_{\lambda_s}, {\lambda_s})\right),
\end{equation}
where the definition of $\phi_1(m) $ is referred to in Corollary \ref{reclimited}. Thus,
we will refer this new iteration algorithm to as \emph{SciREI-Solver} shown in Algorithm \ref{algorithm:scirei-solver}, which incorporates the RD method and the exponential-based calculation formula recommended by DPM-Solver.

\subsection{Differences with DPM-Solver}
Clearly, there are differences between SciREI-Solver and DPM-Solver-2, as indicated by the \textbf{blue} and \textbf{orange}  labels in Eq. (\ref{dpmsolver3}) and Eq. (\ref{scirei-solver}).

In the following, we present a straightforward comparison between SciRE-Solver, proposed by us in the main content of this paper, and DPM-Solver. Firstly, the score-integrand form in Eq. (\ref{exactsolution}) is different the solution formula of exponential-based contextualisation in Eq. (\ref{dpmsolver}). Secondly, different solution forms of diffusion ODE result in different integrands and distinct Taylor series expansions around different function spaces.
Specifically, we expand $\boldsymbol{\epsilon}_\theta\big(\mathbf{x}_{\psi(\tau)}, \psi(\tau)\big)$ in a Taylor series around $\tau$, which is distinct from DPM-Solver where $\boldsymbol{\epsilon}_\theta\big(\mathbf{x}_{t(\lambda)}, t(\lambda)\big)$ is expanded w.r.t. $\lambda$, as $\tau\neq\lambda$. Such differences lead to different results,  as expressed in  Eq. \ref{itersolution2} and Eq. \ref{dpmsolver1}. The differences of Eq. \ref{itersolution2} and Eq. \ref{dpmsolver1} illustrate that, despite reparameterizing the diffusion ODE in both cases, different changes-of-variable have resulted in distinct algorithmic sources based on Taylor expansion concept. Finally, and most importantly, SciRE-Solver is a numerical algorithm based on the RD method we introduced, which fundamentally distinguishes it from the DPM-Solver, much like the difference (the blue
and orange labels) between SciREI-Solver and DPM-Solver-2.

\subsection{The benefits with RD: Effectiveness
 and Robustness}
 
In order to validate the benefits of the RD method, we compare the FID scores obtained for generated
samples from the RD-based methods and other methods with the same settings and codebase  on the
CIFAR-10 and CelebA 64$\times$64 datasets. Specifically, we compare the RD method with the traditional finite difference (FD) method, and we also compare the RD-based SciREI-Solver-2 with its counterpart algorithm, DPM-Solver-2. 
For high-resolution image datasets, we conduct sampling comparisons under the same settings and codebase for ImageNet at resolutions of 128$\times$128, 256$\times$256, and 512$\times$512, as well as for the LSUN bedroom dataset at a resolution of 256$\times$256, due to server limitations. To ensure fairness in our experiments, we maintain the same settings and codebase for each sampling algorithm to evaluate the various methods of RD-based and RD-none.

Firstly, we use FID to measure the sampling performance of the sampling algorithms when estimating derivatives using finite difference (\emph{FD}) method and recursive derivative (\emph{RD}) method, respectively.
Here, we set $\phi_1(m)=1$ in SciRE-Solver to represent the sampling algorithm based on FD method.
Based on the FID metric of generated samples, with the same codebase, we assess the performance of these two derivative estimation methods using discrete diffusion models trained on the CIFAR-10 and CelebA 64$\times$64 datasets.
Without loss of generality, we use a consistent NSR trajectory with $k=2$, because SciRE-Solver can achieve the better quality of generated samples for $k\in[2,7]$, as mentioned in Section \ref{timetra}.
Our numerical experiments demonstrate that, across different initial times,
the quality of generated samples achieved by the SciRE-Solver using RD method consistently outperforms that of the solver using FD method, as shown in Table \ref{tab:rdenonrde}.  These numerical experiments measured by FID demonstrate that in the domain of diffusion ODEs, the use of the RD method for estimating derivatives of score function evaluation networks consistently outperforms the traditional FD method.

 \begin{table}
\vspace{-0.8cm}
\centering
\caption{
Comparison of Quality Generation between DPM-Solver-2 and SciREI-Solver-2 (our) Algorithms.
We use consistent \emph{NSR} trajectory (\textbf{k=3.1}) and the same codebase.}
\label{tab:rdedpm-2}
\fontsize{9.7}{9}\selectfont
\begin{tabular}{lllrrrrr}
\toprule
Trajectory & Initial Time & Sampling method \textbackslash NFE   & 12 & 15  & 20 & 50 &100\\
\midrule
\multicolumn{8}{l}{CIFAR-10 (discrete-time model \cite{ho2020denoising}, linear noise schedule)}
 \\
\midrule
\multirow{2}{*}{Uniform time}&\multirow{2}{*}{$\epsilon=10^{-3}$}&
DPM-Solver-$2$   & $11.81$ & $^{\textbf{\dag}}10.16$ & $7.55$ & $5.05$ & $4.51$\\
& &SciREI-Solver-$2$ (ours) & $\textbf{6.65}$ & $^{\textbf{\dag}}\textbf{5.83}$ & $\textbf{4.75}$  &$\textbf{4.15}$  & $4.29$\\
\hline
\multirow{2}{*}{Uniform time}&\multirow{2}{*}{$\epsilon=10^{-4}$}&
DPM-Solver-$2$  & $33.67$ & $^{\textbf{\dag}}29.16$ & $20.50$ & $8.78$ & $5.34$\\
& &SciREI-Solver-$2$ (ours) & $9.93$ & $^{\textbf{\dag}}8.53$ & $6.50$& $4.51$ & $\textbf{3.98}$\\
\midrule
\multirow{2}{*}{logSNR}&\multirow{2}{*}{$\epsilon=10^{-3}$}&
DPM-Solver-$2$   & $\textbf{5.23}$ & $^{\textbf{\dag}}\textbf{4.48}$ & $4.08$ & $3.98$ & $3.99$\\
& &SciREI-Solver-$2$ (ours) & $5.78$ & $^{\textbf{\dag}}4.83$ & $4.05$  &$3.65$  & $3.72$\\
\hline
\multirow{2}{*}{logSNR }&\multirow{2}{*}{$\epsilon=10^{-4}$}&
DPM-Solver-$2$ & $6.47$ & $^{\textbf{\dag}}5.33$ & $\textbf{4.04}$ & $3.55$ & $3.52$\\
& &SciREI-Solver-$2$ (ours) & $7.06$ & $^{\textbf{\dag}}6.03$ & $4.33$& $\textbf{3.33}$ & $\textbf{3.28}$\\
\midrule
\multirow{2}{*}{NSR ($k=3.1$)}&\multirow{2}{*}{$\epsilon=10^{-3}$}&
DPM-Solver-$2$   & $5.03$ & $^{\textbf{\dag}}4.64$ & $4.27$ & $4.07$ & $4.01$\\
& &SciREI-Solver-$2$ (ours) & $\textbf{5.01}$ & $^{\textbf{\dag}}4.51$ & $3.96$  &$3.79$  & $3.82$\\
\hline
\multirow{2}{*}{NSR ($k=3.1$)}&\multirow{2}{*}{$\epsilon=10^{-4}$}&
DPM-Solver-$2$  & $5.22$ & $^{\textbf{\dag}}\textbf{4.33}$ & $\textbf{3.70}$ & $3.48$ & $3.476$\\
& &SciREI-Solver-$2$ (ours) & $5.79$ & $^{\textbf{\dag}}4.72$ & $3.81$& $\textbf{3.21}$ & $\textbf{3.23}$\\
\toprule
\multicolumn{8}{l}{CelebA 64$\times$64 (discrete-time model \cite{song2021denoising}, linear noise schedule)}
 \\
\midrule
\multirow{2}{*}{Uniform time}&\multirow{2}{*}{$\epsilon=10^{-3}$}&
DPM-Solver-$2$   & $15.23$ & $^{\textbf{\dag}}13.63$ & $10.99$ & $7.41$ & $5.46$\\
& &SciREI-Solver-$2$ (ours) & $\textbf{7.25}$ & $^{\textbf{\dag}}\textbf{6.93}$ & $\textbf{6.32}$  &$\textbf{5.72}$  & $5.06$\\
\hline
\multirow{2}{*}{Uniform time}&\multirow{2}{*}{$\epsilon=10^{-4}$}&
DPM-Solver-$2$  & $61.90$ & $^{\textbf{\dag}}53.40$ & $38.19$ & $16.98$ & $7.24$\\
& &SciREI-Solver-$2$ (ours) & $20.05$ & $^{\textbf{\dag}}17.98$ & $14.24$& $7.93$ & $\textbf{4.40}$\\
\midrule
\multirow{2}{*}{logSNR}&\multirow{2}{*}{$\epsilon=10^{-3}$}&
DPM-Solver-$2$   & $3.97$ & $^{\textbf{\dag}}3.96$ & $4.07$ & $4.22$ & $4.25$\\
& &SciREI-Solver-$2$ (ours) & $3.78$ & $^{\textbf{\dag}}3.58$ & $3.33$  &$3.41$  & $3.69$\\
\hline
\multirow{2}{*}{logSNR }&\multirow{2}{*}{$\epsilon=10^{-4}$}&
DPM-Solver-$2$ & $\textbf{3.27}$ & $^{\textbf{\dag}}\textbf{3.13}$ & $2.90$ & $2.80$ & $2.799$\\
& &SciREI-Solver-$2$ (ours) & $3.44$ & $^{\textbf{\dag}}3.18$ & $\textbf{2.71}$& $\textbf{2.29}$ & $\textbf{2.39}$\\
\midrule
\multirow{2}{*}{NSR ($k=3.1$)}&\multirow{2}{*}{$\epsilon=10^{-3}$}&
DPM-Solver-$2$   & $6.24$ & $^{\textbf{\dag}}5.54$ & $4.72$ & $4.27$ & $4.24$\\
& &SciREI-Solver-$2$ (ours) & $4.97$ & $^{\textbf{\dag}}4.48$ & $3.81$  &$3.59$  & $3.79$\\
\hline
\multirow{2}{*}{NSR ($k=3.1$)}&\multirow{2}{*}{$\epsilon=10^{-4}$}&
DPM-Solver-$2$  & $3.67$ & $^{\textbf{\dag}}3.04$ & $2.79$ & $2.77$ & $2.78$\\
& &SciREI-Solver-$2$ (ours) & $\textbf{3.30}$ & $^{\textbf{\dag}}\textbf{2.98}$ & $\textbf{2.56}$& $\textbf{2.25}$ & $\textbf{2.39}$\\
\bottomrule
\end{tabular}
\vspace{-0.4cm}
\end{table}
Secondly, for further investigation of the RD method, we introduce SciREI-Solver ($n=2$) in section \ref{ascireisolver},
referring to it as SciREI-Solver-2 to align it in form with its counterpart, DPM-Solver-2.
We compare the generative performance of SciREI-Solver-2 and DPM-Solver-2 with the identical settings on the CIFAR-10 and CelebA 64$\times$64 datasets using various time trajectories and termination times, as illustrated in  Table \ref{tab:rdedpm-2}. Table \ref{tab:rdedpm-2} demonstrates that  SciREI-Solver-2 based on the RD method exhibits greater robustness than the DPM-Solver-2 across different time trajectories, especially on the CelebA 64$\times$64 dataset. These experiments measured by FID also simultaneously demonstrate that as NFE increases, SciREI-Solver-2 based on the RD method consistently outperforms its counterpart DPM-Solver-2. Next, we will conduct some sampling comparison experiments for both of SciREI-Solver-2 and the DPM-Solver-2.

Thirdly, we provide the sampling comparisons between the RD-based sampling algorithms
(including SciRE-Solver-2 and SciREI-Solver-2) and the baseline algorithm (DPM-Solver-2) on
high-resolution image datasets. In Figure \ref{fig:roubust}, we compare the generation results of the RD-based methods (Solvers: SciRE-2, SciREI-2)
and the baseline method (Solver: DPM-2) using 6-36 sampling steps with the uniform time trajectory
and identical settings, on pre-trained models with ImageNet 128$\times$128 and LSUN bedroom 256$\times$256.
Here, we further compare the generation results of the RD-based methods
and the baseline method on pre-trained models with ImageNet 256$\times$256 and 512$\times$512 datasets, using 6-36 sampling steps with the uniform time trajectory
and identical settings. In these experiments, we can observe that when using 36 NFEs, samples generated by the popular DPM-Solver-2 still exhibit more noise compared to our proposed SciREI-Solver-2 and SciRE-Solver-2 based on the RD method. Therefore, all these sampling experiments on
high-resolution image datasets also demonstrate the effectiveness of the RD method.

In summary, all experiments-above under the same settings and codebase indicate that the RD method brings benefits to Taylor-based numerical algorithms in the realm of diffusion ODEs. Therefore, we strongly recommend using the RDE method, if the sampling algorithms require evaluating the derivative of the score function evaluation networks.

\section{Proof of Theorem 3.1 and Corollary 1}
\label{proof-of-theorem31-corollary1}
\subsection{Preliminaries}
Throughout this section, we denote ${\rm{NSR}}_{\min}:=\min\limits_{i}\{{\rm{NSR}}(t_i)\}$, ${\rm{NSR}}_{\max}:=\max\limits_{i}\{{\rm{NSR}}(t_i)\}$, and assume that $\boldsymbol{\epsilon}_\theta\left(\mathbf{x}_{\psi(\tau)}, \psi(\tau)\right)\in \mathbb{C}^{\infty}[{\rm{NSR}}_{\min},{\rm{NSR}}_{\max}]$, which means that the total derivatives $\frac{\mathrm{d}^{k} \boldsymbol{\epsilon }_{\theta }(\mathbf{x}_{\psi(\tau)}, \psi(\tau) )}{\mathrm{d} \tau^{k}} $ exist and are continuous for $k \in \mathbb{Z}_{+}$. Notice that $\tau:={\rm{NSR}}(t)$, $\psi(\tau):={\rm{rNSR}}(\tau)$ and the reverse function of ${\rm{NSR}}$, i.e. ${\rm{rNSR}}$, satisfying $t={\rm{rNSR}}({\rm{NSR}}(t))=\psi(\tau)$. Denote $h_{s}:={\rm{NSR}}(t)-{\rm{NSR}}(s)=\tau_{t}-\tau_{s}$, and $\boldsymbol{\epsilon}_\theta^{(k)}\left(\mathbf{x}_{\psi(\tau)}, \psi(\tau)\right):=\frac{\mathrm{d}^k\boldsymbol{\epsilon}_\theta\left(\mathbf{x}_{\psi(\tau)}, \psi(\tau)\right)}{\mathrm{d} \tau^k}$ as $k$-th order total derivative of $\boldsymbol{\epsilon}_\theta\left(\mathbf{x}_{\psi(\tau)}, \psi(\tau)\right)$ w.r.t.  $\tau$. For $n\geq 1$, the $n$-th order Taylor expansion of $\boldsymbol{\epsilon}_\theta\left(\mathbf{x}_{\psi(\tau_{t})}, \psi(\tau_{t})\right)$ w.r.t. $\tau$ at $\tau_{s}$ is
\begin{equation}\label{taylor-appendix}
\boldsymbol{\epsilon}_\theta\left(\mathbf{x}_{\psi(\tau_{t})}, \psi(\tau_{t})\right)=\sum_{k=0}^{n} \frac{h_{s}^k}{k!}
\boldsymbol{\epsilon}_\theta^{(k)}\left(\mathbf{x}_{\psi(\tau_{s})}, \psi(\tau_{s})\right)
+\mathcal{O}(h_{s}^{n+1}).
\end{equation}

For any $k\geq 0$, we can approximate the $k$-th order total derivative term $\boldsymbol{\epsilon}_\theta^{(k)}\left(\mathbf{x}_{\psi(\tau_{s})}, \psi(\tau_{s})\right)$ in Eq. (\ref{taylor-appendix}) by using the first-order difference formula:
\begin{equation} \label{difference-appendix}
\boldsymbol{\epsilon}_\theta^{(k)}\left(\mathbf{x}_{\psi(\tau_{s})}, \psi(\tau_{s})\right) = \frac{\boldsymbol{\epsilon}_\theta^{(k-1)}\left(\mathbf{x}_{\psi(\tau_{t})}, \psi(\tau_{t})\right)-\boldsymbol{\epsilon}_\theta^{(k-1)}\left(\mathbf{x}_{\psi(\tau_{s})}, \psi(\tau_{s})\right)}{h_{s}} -\mathcal{O}(h_{s}).
\end{equation}
For ease of notation, we denote $\boldsymbol{\epsilon}_\theta^{(k)}\left(\mathbf{x}_{\psi(\tau)}, \psi(\tau)\right)$ as $\Gamma^{(k)} (\tau)$. Notice that $\boldsymbol{\epsilon}_\theta\left(\mathbf{x}_{\psi(\tau)}, \psi(\tau)\right)=\Gamma^{(0)} (\tau)$. Then Eq. (\ref{difference-appendix}) can be represented as:
\begin{equation} \label{difference-simple-appendix}
    \Gamma^{(k)} (\tau_{s})=\frac{\Gamma^{(k-1)} (\tau_{t})-\Gamma^{(k-1)} (\tau_{s})}{h_{s}} - \mathcal{O}(h_{s}).
\end{equation}

\subsection{Proof of Theorem 3.1}
\begin{proof}
While $n\to \infty$, Eq. (\ref{taylor-appendix}) becomes:
\begin{equation}
\begin{aligned}
\Gamma^{(0)} (\tau_{t})&=\sum_{k=0}^{\infty} \frac{h_{s}^k}{k!}
\Gamma^{(k)} (\tau_{s}) \\
&=\Gamma^{(0)} (\tau_{s})+ \sum_{k=1}^{\infty} \frac{h_{s}^k}{k!}\Gamma^{(k)} (\tau_{s}).
\end{aligned}
\end{equation}
Moving $\Gamma^{(0)} (\tau_{s})$ from the right-hand side of the above equation to the left-hand side and then dividing both sides of the equation by $h_{s}$, we can obtain:
\begin{equation} \label{first-order-difference-appendix}
\begin{aligned}
    & \quad \frac{\Gamma^{(0)} (\tau_{t})-\Gamma^{(0)} (\tau_{s})}{h_{s}} \\
    &=\sum_{k=1}^{\infty} \frac{h_{s}^{k-1}}{k!}\Gamma^{(k)} (\tau_{s}) \\
    &=\Gamma^{(1)} (\tau_{s}) + \sum_{k=2}^{\infty} \frac{h_{s}^{k-1}}{k!}\Gamma^{(k)} (\tau_{s}) \\
    &=\Gamma^{(1)} (\tau_{s}) + \sum_{k=2}^{\infty} \frac{h_{s}^{k-1}}{k!}\left ( \frac{\Gamma^{(k-1)} (\tau_{t})-\Gamma^{(k-1)} (\tau_{s})}{h_{s}} - \mathcal{O}(h_{s}) \right ) \\
    &=\Gamma^{(1)} (\tau_{s}) + \sum_{k=2}^{\infty} \frac{h_{s}^{k-2}}{k!} \left( \Gamma^{(k-1)} (\tau_{t})-\Gamma^{(k-1)} (\tau_{s}) \right ) - \underbrace{\sum_{k=2}^{\infty}\frac{h_{s}^{k-1}}{k!}\mathcal{O}(h_{s})}_{{\rm{remainder}} \ Q} \\
    &=\Gamma^{(1)} (\tau_{s}) - \sum_{k=2}^{\infty} \frac{h_{s}^{k-2}}{k!} \Gamma^{(k-1)} (\tau_{s}) + \underbrace{\sum_{k=2}^{\infty} \frac{h_{s}^{k-2}}{k!} \Gamma^{(k-1)} (\tau_{t})}_{R_1} - Q \\
    &=\left ( 1-\frac{1}{2} \right ) \Gamma^{(1)} (\tau_{s}) - \sum_{k=3}^{\infty}\frac{h_{s}^{k-2}}{k!} \Gamma^{(k-1)} (\tau_{s}) + R_1 - Q \\
    &=\left ( 1-\frac{1}{2} \right ) \Gamma^{(1)} (\tau_{s}) - \sum_{k=3}^{\infty}\frac{h_{s}^{k-3}}{k!}\left( \Gamma^{(k-2)} (\tau_{t})-\Gamma^{(k-2)} (\tau_{s}) - \mathcal{O}(h_{s}^2) \right )+ R_1 - Q \\
    &=\left ( 1-\frac{1}{2} \right ) \Gamma^{(1)} (\tau_{s}) + \sum_{k=3}^{\infty}\frac{h_{s}^{k-3}}{k!}\Gamma^{(k-2)} (\tau_{s})\underbrace{-\sum_{k=3}^{\infty}\frac{h_{s}^{k-3}}{k!}\Gamma^{(k-2)} (\tau_{t})}_{R_2} -\mathcal{O}(h_{s}^2)+R_1 \\
    &=\left ( 1-\frac{1}{2}+\frac{1}{6} \right ) \Gamma^{(1)} (\tau_{s}) + \sum_{k=4}^{\infty}\frac{h_{s}^{k-3}}{k!}\Gamma^{(k-2)} (\tau_{s})+R_2 + R_1 - \mathcal{O}(h_{s}^2) \\
    &=\left ( 1-\frac{1}{2!}+\frac{1}{3!}-\frac{1}{4!} \right ) \Gamma^{(1)} (\tau_{s}) - \sum_{k=5}^{\infty}\frac{h_{s}^{k-4}}{k!}\Gamma^{(k-3)} (\tau_{s}) + R_1 + R_2 + R_3 - \mathcal{O}(h_{s}^2) \\
    & \qquad \qquad \qquad \qquad \qquad \qquad \quad \quad \quad \quad \quad \quad \cdots \\
    &=\sum_{k=1}^{\infty}\frac{(-1)^{k-1}}{k!}\Gamma^{(1)} (\tau_{s})+\sum_{i=1}^{\infty}R_i - \mathcal{O}(h_{s}^2),
\end{aligned}
\end{equation}
where
\begin{equation}
\begin{aligned}
    R_1&=\sum_{k=2}^{\infty} \frac{h_{s}^{k-2}}{k!} \Gamma^{(k-1)} (\tau_{t}) \\
    R_2&=-\sum_{k=3}^{\infty}\frac{h_{s}^{k-3}}{k!}\Gamma^{(k-2)} (\tau_{t}) \\
    R_3&=\sum_{k=4}^{\infty}\frac{h_{s}^{k-4}}{k!}\Gamma^{(k-3)} (\tau_{t})\\
    & \qquad \qquad \cdots\\
    R_i&=(-1)^{i+1}\sum_{k=i+1}^{\infty}\frac{h_{s}^{k-i-1}}{k!}\Gamma^{(k-i)} (\tau_{t}), \quad i\in \mathbb{Z}_{+}.
\end{aligned}
\end{equation}
Adding the first term and second term of $R_{i}, i\in\mathbb{Z}_{+}$  separately, we can derive
\begin{equation}
    \sum_{i=1}^{\infty}R_i=\sum_{k=2}^{\infty}\frac{(-1)^{k}}{k!}\Gamma^{(1)}(\tau_{t}) + \sum_{k=3}^{\infty}\frac{(-1)^{k+1}h_{s}}{k!}\Gamma^{(2)}(\tau_{t}) + \underbrace{\sum_{i=3}^{\infty}\sum_{k=i+1}^{\infty}\frac{(-1)^{k+i-1}h_{s}^{i-1}}{k!}\Gamma^{(i)}(\tau_{t})}_{\mathcal{O}(h_{s}^{2})}.
\end{equation}
Notice that
\begin{equation} \label{Leibniz}
    \sum_{i=3}^{\infty}\sum_{k=i+1}^{\infty}\frac{(-1)^{k+i-1}h_{s}^{i-1}}{k!}\Gamma^{(i)}(\tau_{t})=  \sum_{i=3}^{\infty}(-1)^{i-1}h_{s}^{i-1}\Gamma^{(i)}(\tau_{t})\sum_{k=i+1}^{\infty} \frac{(-1)^{k}}{k!} =\mathcal{O}(h_{s}^{2}),
\end{equation}
because $\sum\limits_{k=i+1}^{\infty} \frac{(-1)^{k}}{k!}, \forall i \in \mathbb{Z}_{+}$ are all convergent alternating series which can be easily proved with Leibniz's test. Then Eq. (\ref{first-order-difference-appendix}) can be shown as:
\begin{equation}
    \frac{\Gamma^{(0)}(\tau_{t})-\Gamma^{(0)}(\tau_{s})}{h_{s}}=\sum_{k=1}^{\infty}\frac{(-1)^{k-1}}{k!}\Gamma^{(1)}(\tau_{s})+\sum_{i=1}^{2}\sum_{k=i+1}^{\infty}\frac{(-1)^{k+i-1}h_{s}^{i-1}}{k!}\Gamma^{(i)}(\tau_{t}) -\mathcal{O}(h_{s}^{2}).
\end{equation}
Consequently, we can get $\Gamma^{(1)}(\tau_{s})$ by simple manipulation of rearranging and affine transformation applied to above equation:
\begin{equation} \label{results-before-replace-appendix}
\begin{aligned}
    \Gamma^{(1)}(\tau_{s})&=\frac{e}{e-1}\frac{\Gamma^{(0)}(\tau_{t})-\Gamma^{(0)}(\tau_{s})}{h_{s}}-\frac{e}{e-1}\sum_{i=1}^{2}\sum_{k=i+1}^{\infty}\frac{(-1)^{k+i-1}h_{s}^{i-1}}{k!}\Gamma^{(i)}(\tau_{t})+\mathcal{O}(h_{s}^{2}) \\
    &=\frac{e}{e-1}\frac{\Gamma^{(0)}(\tau_{t})-\Gamma^{(0)}(\tau_{s})}{h_{s}}-\frac{e}{e-1}\left ( \frac{1}{e} \Gamma^{(1)}(\tau_{t})+\frac{e-2}{2e}h_{s}\Gamma^{(2)}(\tau_{t}) \right )+\mathcal{O}(h_{s}^{2}) \\
    &=\frac{e}{e-1}\frac{\Gamma^{(0)}(\tau_{t})-\Gamma^{(0)}(\tau_{s})}{h_{s}}-\frac{\Gamma^{(1)}(\tau_{t})}{e-1}-\frac{(e-2)h_s}{2(e-1)}\Gamma^{(2)}(\tau_{t}) + \mathcal{O}(h_{s}^{2}),
\end{aligned}
\end{equation}
where
\begin{equation}
\begin{aligned}
    \sum_{k=1}^{\infty}\frac{(-1)^{k-1}}{k!}&=1-e^{-1} \\
    \sum_{k=2}^{\infty}\frac{(-1)^{k}}{k!}&=e^{-1} \\
    \sum_{k=3}^{\infty}\frac{(-1)^{k+1}}{k!}&=-e^{-1}+\frac{1}{2}=\frac{e-2}{2e},
\end{aligned}
\end{equation}
for $e^{x}=\sum\limits_{k=0}^{\infty}\frac{x^{k}}{k!}$ with $x=-1$.

Notice that we denote $\boldsymbol{\epsilon}_\theta^{(k)}\left(\mathbf{x}_{\psi(\tau)}, \psi(\tau)\right)$ as $\Gamma^{(k)}(\tau)$. While using $\tilde{\mathbf{x}}$ to approximate $\mathbf{x}$ and replacing the terms like $\Gamma^{(k)}(\tau)$ in Eq. (\ref{results-before-replace-appendix}) with terms like $\boldsymbol{\epsilon}_\theta^{(k)}\left(\mathbf{x}_{\psi(\tau)}, \psi(\tau)\right)$, we can get the results shown in Theorem \ref{recge}.
\end{proof}

\subsection{Proof of Corollary 1}
We observe that the differentiability constraint imposed by Theorem \ref{recge} appears to be rather restrictive.
In order to enhance its broad applicability, we further propose a recursive derivative estimation method under the assumption of limited differentiability. The corresponding proof process is as follows:

\begin{proof}
Assume that $\boldsymbol{\epsilon}_\theta\left(\mathbf{x}_{\psi(\tau)}, \psi(\tau)\right)\in \mathbb{C}^{n}[{\rm{NSR}}_{\min},{\rm{NSR}}_{\max}]$. While $n$ is a finite positive integer, Eq. (\ref{taylor-appendix}) becomes:
\begin{equation}
\begin{aligned}
\Gamma^{(0)}(\tau_{t})&=\sum_{k=0}^{n} \frac{h_{s}^k}{k!}
\Gamma^{(k)}(\tau_{s}) +\mathcal{O}(h_{s}^{n+1})\\
&=\Gamma^{(0)}(\tau_{s})+ \sum_{k=1}^{n} \frac{h_{s}^k}{k!}\Gamma^{(k)}(\tau_{s})+\mathcal{O}(h_{s}^{n+1}).
\end{aligned}
\end{equation}
Same as the derivation process in (\ref{first-order-difference-appendix}), we can obtain
\begin{equation}  \label{first-order-difference-discrete-appendix}
    \begin{aligned}
        & \quad \frac{\Gamma^{(0)}(\tau_{t})-\Gamma^{(0)}(\tau_{s})}{h_{s}} \\
        &=\sum_{k=1}^{n} \frac{h_{s}^{k-1}}{k!}\Gamma^{(k)}(\tau_{s}) +\mathcal{O}(h_{s}^{n})\\
        &=\Gamma^{(1)}(\tau_{s}) + \sum_{k=2}^{n} \frac{h_{s}^{k-1}}{k!}\Gamma^{(k)}(\tau_{s}) +\mathcal{O}(h_{s}^{n})\\
        &=\Gamma^{(1)}(\tau_{s}) - \sum_{k=2}^{n} \frac{h_{s}^{k-2}}{k!} \Gamma^{(k-1)}(\tau_{s})+ \underbrace{\sum_{k=2}^{n} \frac{h_{s}^{k-2}}{k!} \Gamma^{(k-1)}(\tau_{t})}_{R_1} - \underbrace{\mathcal{O}(h_{s}^{2})+\mathcal{O}(h_{s}^{n})}_{\mathcal{O}(h_{s}^{2})-\mathcal{O}(h_{s}^{n})=\mathcal{O}(h_{s}^{2})} \\
        &=\left ( 1-\frac{1}{2} \right ) \Gamma^{(1)}(\tau_{s}) + \sum_{k=3}^{n}\frac{h_{s}^{k-3}}{k!}\Gamma^{(k-2)}(\tau_{s})\underbrace{-\sum_{k=3}^{n}\frac{h_{s}^{k-3}}{k!}\Gamma^{(k-2)}(\tau_{t})}_{R_2} -\mathcal{O}(h_{s}^2)+R_1 \\
        &=\left ( 1-\frac{1}{2}+\frac{1}{6} \right ) \Gamma^{(1)}(\tau_{s}) - \sum_{k=4}^{n}\frac{h_{s}^{k-4}}{k!}\Gamma^{(k-3)}(\tau_{s})\underbrace{+\sum_{k=4}^{n}\frac{h_{s}^{k-4}}{k!}\Gamma^{(k-3)}(\tau_{t})}_{R_3} -\mathcal{O}(h_{s}^2)+R_1+R_2 \\
        & \qquad \qquad \qquad \qquad \qquad \qquad \quad \quad \quad \quad \quad \quad \cdots \\
        &=\sum_{k=1}^{n}\frac{(-1)^{k-1}}{k!}\Gamma^{(1)}(\tau_{s})+\sum_{i=1}^{n-1}R_i - \mathcal{O}(h_{s}^2),
    \end{aligned}
\end{equation}
where
\begin{equation}
\begin{aligned}
    R_1&=\sum_{k=2}^{n} \frac{h_{s}^{k-2}}{k!} \Gamma^{(k-1)}(\tau_{t}) \\
    R_2&=-\sum_{k=3}^{n}\frac{h_{s}^{k-3}}{k!}\Gamma^{(k-2)}(\tau_{t}) \\
    R_3&=\sum_{k=4}^{n}\frac{h_{s}^{k-4}}{k!}\Gamma^{(k-3)}(\tau_{t})\\
    & \qquad \qquad \cdots\\
    R_{n-1}&=(-1)^{n}\sum_{k=n}^{n}\frac{h_{s}^{k-n}}{k!}\Gamma^{(k-n+1)} (\tau_{t}).
\end{aligned}
\end{equation}
Hence adding the first term and second term of $R_{i}, i=1, 2, \ldots, n-1$ separately, we also have
\begin{equation} \label{R-terms-discrete-appendix}
\begin{aligned}
    \sum_{i=1}^{n-1}R_i&=\sum_{k=2}^{n}\frac{(-1)^{k}}{k!}\Gamma^{(1)}(\tau_{t}) + \sum_{k=3}^{n}\frac{(-1)^{k+1}h_{s}}{k!}\Gamma^{(2)}(\tau_{t}) + \sum_{i=3}^{n-1}\sum_{k=i+1}^{n}\frac{(-1)^{k+i-1}h_{s}^{i-1}}{k!}\Gamma^{(i)}(\tau_{t}) \\
    &=\sum_{k=2}^{n}\frac{(-1)^{k}}{k!}\Gamma^{(1)}(\tau_{t}) + \sum_{k=3}^{n}\frac{(-1)^{k+1}h_{s}}{k!}\Gamma^{(2)}(\tau_{t}) + \mathcal{O}(h_{s}^{2}).
\end{aligned}
\end{equation}
Denote $\phi_1(n)=\sum\limits_{k=1}^{n}\frac{(-1)^{k-1}}{k!}$, $\phi_2(n)=\sum\limits_{k=2}^{n}\frac{(-1)^{k}}{k!}$, and  $\phi_3(n)=\sum\limits_{k=3}^{n}\frac{(-1)^{k+1}}{k!}$. Combining (\ref{first-order-difference-discrete-appendix}) and (\ref{R-terms-discrete-appendix}), we can easily derive $\Gamma(\tau_{s},1)$ by
\begin{equation}\label{results-before-replace-discrete-appendix}
    \Gamma^{(1)}(\tau_{s})=\frac{1}{\phi_1(n)}\frac{\Gamma^{(0)}(\tau_{t})-\Gamma^{(0)}(\tau_{s})}{h_{s}}-\frac{\phi_2(n)}{\phi_1(n)}\Gamma^{(1)}(\tau_{t}) - \frac{\phi_3(n)h_s}{\phi_1(n)}\Gamma^{(2)}(\tau_{t}) + \mathcal{O}(h_{s}^{2}).
\end{equation}
Similarly, while using $\tilde{\mathbf{x}}$ to approximate $\mathbf{x}$ and replacing the terms like $\Gamma^{(k)}(\tau)$ in Eq. (\ref{results-before-replace-discrete-appendix}) with terms like $\boldsymbol{\epsilon}_\theta^{(k)}\left(\mathbf{x}_{\psi(\tau)}, \psi(\tau)\right)$, we can get the results shown in Corollary \ref{reclimited}.
\end{proof}

\section{Proof of Theorem 3.2}
\label{proof-of-theorem32}
Moreover, under reasonable assumptions, SciRE-Solver-$k$ is a $k$-th order solver.
\subsection{Preliminaries}
\begin{assumption} \label{assumption-continuous-differential}
    The function $\boldsymbol{\epsilon}_\theta\left(\mathbf{x}_{t}, t\right)$ is set to be in $  \mathbb{C}^{m}$, meaning that  $\boldsymbol{\epsilon}_\theta\left(\mathbf{x}_{t}, t\right)$ is $m$ times continuously differentiable. Specifically, $m\geq 3$ in this paper.
\end{assumption}
\begin{assumption}\label{assumption-lipschitz}
    $\boldsymbol{\epsilon}_\theta\left(\mathbf{x}_{t}, t\right)$ is a Lipschitz continuous function w.r.t. $\mathbf{x}_{t}$.
\end{assumption}
\begin{assumption} \label{assumption-neighbourhood}
    For $\forall t$, there exists $\delta > 0$ such that $\exists \rho\in \mathbb{U}(t, \delta), (l+\frac{1}{2^n})\boldsymbol{\epsilon}_\theta^{(1)}\left(\mathbf{x}_{t}, t\right)=l(1+\frac{1}{2^n l})\boldsymbol{\epsilon}_\theta^{(1)}\left(\mathbf{x}_{t}, t\right)=l\boldsymbol{\epsilon}_\theta^{(1)}\left(\mathbf{x}_{\rho}, \rho\right)$ if $n\in \mathbb{N}$ is large enough. Here $\mathbb{U}$ denotes the neighbourhood of $t$.
\end{assumption}
As in Appendix \ref{proof-of-theorem31-corollary1}, we denote $\tau:={\rm{NSR}}(t)$, $\psi(\tau):={\rm{rNSR}}(\tau)$ and $h_{s}:=\tau_{t}-\tau_{s}$. In Appendix \ref{proof-of-proposition31}, we transform the form  of the solution to diffusion ODEs as in Eq. (\ref{rewritevoc-appendix}) to the form as in (\ref{exactsolution-appendix}) by using the change of variable formula. The resulting solution can be formed as:
\begin{equation} \label{exactsolution-appendixD}
\mathbf{x}_t=\frac{\alpha_t}{\alpha_s}\mathbf{x}_s+\alpha_t\int_{\tau_{s}}^{\tau_{t}}
\boldsymbol{\epsilon}_\theta\left(\mathbf{x}_{\psi(\tau)}, \psi(\tau)\right)\mathrm{d} \tau.
\end{equation}

Then by substituting the $n$-th order Taylor expansion of $\boldsymbol{\epsilon}_\theta\left(\mathbf{x}_{\psi(\tau_{t})}, \psi(\tau_{t})\right)$ w.r.t. $\tau$ at $\tau_{s}$
\begin{equation}\label{taylor-appendix-D}
\boldsymbol{\epsilon}_\theta\left(\mathbf{x}_{\psi(\tau_{t})}, \psi(\tau_{t})\right)=\sum_{k=0}^{n} \frac{h_{s}^k}{k!}
\boldsymbol{\epsilon}_\theta^{(k)}\left(\mathbf{x}_{\psi(\tau_{s})}, \psi(\tau_{s})\right)
+\mathcal{O}(h_{s}^{n+1}),
\end{equation}
for the $\boldsymbol{\epsilon}_\theta\left(\mathbf{x}_{\psi(\tau_{t})}, \psi(\tau_{t})\right)$ in Eq. (\ref{exactsolution-appendixD}), we can derive the exact solution of $\mathbf{x}_{t}$ in Eq. (\ref{exactsolution-appendixD}) as follows:
\begin{equation}\label{itersolution2-appendix}
\begin{aligned}
\mathbf{x}_{t}&=\frac
   {\alpha_{t}
 }{\alpha_{s} }\mathbf{x}_{s} +\alpha_{t}\sum_{k=0}^{n}
 \frac
 {h_{s}^{k+1}}
{(k+1)!}\boldsymbol{\epsilon}_\theta^{(k)}\left(\mathbf{x}_{ \psi(\tau_{s})}, \psi(\tau_{s})\right)
+\mathcal{O}(h_{s}^{n+2}) \\
&=\frac
   {\alpha_{t}
 }{\alpha_{s} }\mathbf{x}_{s} +\alpha_{t}\sum_{k=0}^{n}
 \frac
 {h_{s}^{k+1}}
{(k+1)!}\boldsymbol{\epsilon}_\theta^{(k)}\left(\mathbf{x}_{s}, s\right)
+\mathcal{O}(h_{s}^{n+2}).
\end{aligned}
\end{equation}

\subsection{Proof of Theorem 3.2 when \texorpdfstring{$k=2$}{}}
\label{proof-theorem-32-2}
In this subsection, we prove the global convergence order of SciRE-Solver-2 is no less than 1 and is 2 under reasonable assumptions.

\begin{proof}
For each iteration step, we first update according to:
\begin{align}
    h_s&=\tau_{t}-\tau_{s}, \\
    s_1&=\psi(\tau_{s}+r_1 h_s), \\
    \mathbf{u}_1 &=\frac{\alpha_{s_1}}{\alpha_{s}} \mathbf{x}_{s}+\alpha_{s_1}r_1 h_s\boldsymbol{\epsilon}_\theta\left(\mathbf{x}_{s}, s\right), \\
    \tilde{\mathbf{x}}_{t} &= \frac{\alpha_{t}}{\alpha_{s}} \mathbf{x}_{s}+\alpha_{t} h_s\boldsymbol{\epsilon}_\theta\left(\mathbf{x}_{s}, s\right)+\alpha_{t}\frac{h_s^2}{2\phi_1(m)r_1 h_s} \left(\boldsymbol{\epsilon}_\theta\left(\mathbf{u}_1, s_1\right)-\boldsymbol{\epsilon}_\theta\left(\mathbf{x}_{s}, s\right)\right),\label{approximation-appendix}
\end{align}
where $\tilde{\mathbf{x}}_{t}$ denotes the approximate solution of $\mathbf{x}_{t}$ computed by SciRE-Solver-2, $r_1\in(0, 1)$ is a hyperparameter so that $s_1 \in (t, s)$.

Next, taking $n=1$ in Eq. (\ref{itersolution2-appendix}), we can get the exact solution of $\mathbf{x}_{t}$ as follows:
\begin{equation}
    \mathbf{x}_t=\frac{\alpha_t}{\alpha_s}\mathbf{x}_s
    + \alpha_t h_s \boldsymbol{\epsilon}_\theta\left(\mathbf{x}_{s}, s\right)
    + \alpha_{t}\frac{h_{s}^{2}}{2}\boldsymbol{\epsilon}_\theta^{(1)}\left(\mathbf{x}_{s}, s\right)
    + \mathcal{O}(h_{s}^{3})
\end{equation}

Then by subtracting the last equation from Eq. (\ref{approximation-appendix}) and using $\mathbf{u}_1 - \mathbf{x}_{s_1}=\mathcal{O}(h_s^2)$, we have
\begin{equation}\label{truncation-error-appendxi}
\begin{aligned}
\frac{\mathbf{x}_t-\tilde{\mathbf{x}}_{t}}{\alpha_{t}}&=\frac{h_{s}^{2}}{2}\boldsymbol{\epsilon}_\theta^{(1)}\left(\mathbf{x}_{s}, s\right)-
\frac{h_s^2}{2\phi_1(m)r_1 h_s} \left(\boldsymbol{\epsilon}_\theta\left(\mathbf{u}_1, s_1\right)-\boldsymbol{\epsilon}_\theta\left(\mathbf{x}_{s}, s\right)\right)+ \mathcal{O}(h_{s}^{4})  \\
&=\frac{h_s^2}{2}\left ( \boldsymbol{\epsilon}_{\theta}^{(1)}\left(\mathbf{x}_{s}, s\right) - \frac{\boldsymbol{\epsilon}_\theta\left(\mathbf{x}_{s_1}, s_1\right)-\boldsymbol{\epsilon}_\theta\left(\mathbf{x}_{s}, s\right)}{\phi_1(m)r_1 h_s} \right )  + \mathcal{O}(h_s^3),
\end{aligned}
\end{equation}
where $\| \boldsymbol{\epsilon}_\theta\left(\mathbf{u}_1, s_1\right) - \boldsymbol{\epsilon}_\theta\left(\mathbf{x}_{s_1}, s_1\right) \|=\mathcal{O}(\| \mathbf{u}_1 - \mathbf{x}_{s_1} \|)=\mathcal{O}(h_s^2)$ under Assumption \ref{assumption-lipschitz}.

By Assumption \ref{assumption-lipschitz} and Lagrange's mean value theorem, we find that
\begin{equation} \label{bounded-epsilon-appendix}
    \| \boldsymbol{\epsilon}_\theta\left(\mathbf{x}_{s_1}, s_1\right)-\boldsymbol{\epsilon}_\theta\left(\mathbf{x}_{s}, s\right) \| \leq L\| \mathbf{x}_{s_1} - \mathbf{x}_{s} \|=L \| \mathbf{x}_{\eta}' (s_1 - s) \|=L\|\mathbf{x}_{\eta}'\psi'(\tau_{\xi}) (\tau_{s_1} - \tau_s)  \|,
\end{equation}
where $\eta \in (\psi(\tau_{s_1}), \psi(\tau_{s}))$, $\tau_{\xi} \in (\tau_{s_1}, \tau_{s})$ and $L$ is the Lipschitz constant. Since $\| \tau_{s_1} - \tau_s \| \leq \| \tau_{t} - \tau_s \|=\mathcal{O}(h_s)$, the r.h.s. of the above inequation is $\mathcal{O}(h_s)$.

Besides, by Assumption \ref{assumption-continuous-differential}, we also have
\begin{equation}
    \boldsymbol{\epsilon}_{\theta}^{(1)}\left(\mathbf{x}_{s}, s\right) - \frac{\boldsymbol{\epsilon}_\theta\left(\mathbf{x}_{s_1}, s_1\right)-\boldsymbol{\epsilon}_\theta\left(\mathbf{x}_{s}, s\right)}{\phi_1(m)r_1 h_s}= \mathcal{O}(1)- \frac{\mathcal{O}(h_s)}{h_s}=\mathcal{O}(1).
\end{equation}
Hence $\mathbf{x}_t-\tilde{\mathbf{x}}_{t}=\mathcal{O}(h_s^2)$. We prove that SciRE-Solver-2 is \textbf{at least} a first order solver.

Furthermore, we will prove that \textbf{SciRE-Solver-2 is a second order solver under mild condition.}

Specifically, while $\phi_1 (m)=\phi_1 (3)=2/3$, by the Lagrange's mean value theorem, there exists $\xi \in (s_1, s)$ such that
\begin{equation}
    \frac{\boldsymbol{\epsilon}_\theta\left(\mathbf{x}_{s_1}, s_1\right)-\boldsymbol{\epsilon}_\theta\left(\mathbf{x}_{\xi}, \xi\right)}{\phi_1(m)r_1 h_s}=\boldsymbol{\epsilon}_\theta^{(1)}\left(\mathbf{x}_{\xi}, \xi\right) + \mathcal{O}(h_s).
\end{equation}
Note that $\phi_1 (m)r_1 h_s=\tau_{s1}-\tau_{\xi}$ and $(1-\phi_1 (m))r_1 h_s=\tau_{\xi}-\tau_{s}$, where $\phi_1 (m)=1/3$,  hence
\begin{equation}
\begin{aligned}
    \frac{\boldsymbol{\epsilon}_\theta\left(\mathbf{x}_{s_1}, s_1\right)-\boldsymbol{\epsilon}_\theta\left(\mathbf{x}_{s}, s\right)}{\phi_1(m)r_1 h_s} &= \frac{\boldsymbol{\epsilon}_\theta\left(\mathbf{x}_{s_1}, s_1\right)-\boldsymbol{\epsilon}_\theta\left(\mathbf{x}_{\xi}, \xi\right)}{\phi_1(m)r_1 h_s}+\frac{\boldsymbol{\epsilon}_\theta\left(\mathbf{x}_{\xi}, \xi\right)- \boldsymbol{\epsilon}_\theta\left(\mathbf{x}_{s}, s\right)}{\phi_1(m)r_1 h_s} \\
    &=\boldsymbol{\epsilon}_\theta^{(1)}\left(\mathbf{x}_{\xi}, \xi\right) + \mathcal{O}(h_s) + \frac{1}{2} \frac{\boldsymbol{\epsilon}_\theta\left(\mathbf{x}_{\xi}, \xi\right)- \boldsymbol{\epsilon}_\theta\left(\mathbf{x}_{s}, s\right)}{(1-\phi_1(m))r_1 h_s} \\
    &=\boldsymbol{\epsilon}_\theta^{(1)}\left(\mathbf{x}_{\xi}, \xi\right) + \frac{1}{2}\boldsymbol{\epsilon}_\theta^{(1)}\left(\mathbf{x}_{s}, s\right)+ \mathcal{O}(h_s).
\end{aligned}
\end{equation}

Combining the above equation with Eq. (\ref{truncation-error-appendxi}), we have
\begin{equation}
\begin{aligned}
    \frac{\mathbf{x}_t-\tilde{\mathbf{x}}_{t}}{\alpha_{t}}&=\frac{h_{s}^{2}}{2}\boldsymbol{\epsilon}_\theta^{(1)}\left(\mathbf{x}_{s}, s\right)-
    \frac{h_s^2}{2} \frac{\boldsymbol{\epsilon}_\theta\left(\mathbf{x}_{s_1}, s_1\right)-\boldsymbol{\epsilon}_\theta\left(\mathbf{x}_{s}, s\right)}{\phi_1(m)r_1 h_s}+ \mathcal{O}(h_{s}^{3})  \\
    &=\frac{h_s^2}{2}\boldsymbol{\epsilon}_\theta^{(1)}\left(\mathbf{x}_{s}, s\right)-\frac{h_s^2}{2}\left ( \boldsymbol{\epsilon}_\theta^{(1)}\left(\mathbf{x}_{\xi}, \xi\right) + \frac{1}{2}\boldsymbol{\epsilon}_\theta^{(1)}\left(\mathbf{x}_{s}, s\right) \right )+ \mathcal{O}(h_{s}^{3})\\
    &=\frac{h_s^2}{4}\left(\boldsymbol{\epsilon}_\theta^{(1)}\left(\mathbf{x}_{s}, s\right) - 2\boldsymbol{\epsilon}_\theta^{(1)}\left(\mathbf{x}_{\xi}, \xi\right) \right)+ \mathcal{O}(h_{s}^{3}),
\end{aligned}
\end{equation}
where $\xi \in (s_1, s)$. By Assumption \ref{assumption-continuous-differential},  $\boldsymbol{\epsilon}_\theta^{(2)}\left(\mathbf{x}_{s}, s\right)$ is bounded hence the first term in the r.h.s. of above equation is $\mathcal{O}(h_s^3)$. While for the second term, we find that
\begin{equation}
\begin{aligned}
    & \quad \boldsymbol{\epsilon}_\theta^{(1)}\left(\mathbf{x}_{s}, s\right) - 2\boldsymbol{\epsilon}_\theta^{(1)}\left(\mathbf{x}_{\xi}, \xi\right) \\
    &=\left [\boldsymbol{\epsilon}_\theta^{(1)}\left(\mathbf{x}_{s}, s\right) - (1+\frac{1}{2})\boldsymbol{\epsilon}_\theta^{(1)}\left(\mathbf{x}_{s_1}, s_1\right) \right ] + \left [(1+\frac{1}{2})\boldsymbol{\epsilon}_\theta^{(1)}\left(\mathbf{x}_{s_1}, s_1\right) - 2\boldsymbol{\epsilon}_\theta^{(1)}\left(\mathbf{x}_{\xi}, \xi\right) \right ] \\
    &=\left [\boldsymbol{\epsilon}_\theta^{(1)}\left(\mathbf{x}_{s}, s\right)-(1+\frac{1}{4})\boldsymbol{\epsilon}_\theta^{(1)}\left(\mathbf{x}_{\xi}, \xi\right) \right ] + \left [(1+\frac{1}{4})\boldsymbol{\epsilon}_\theta^{(1)}\left(\mathbf{x}_{\xi}, \xi\right) - (1+\frac{1}{2})\boldsymbol{\epsilon}_\theta^{(1)}\left(\mathbf{x}_{s_1}, s_1\right) \right ] \\
    &+\left [(1+\frac{1}{2})\boldsymbol{\epsilon}_\theta^{(1)}\left(\mathbf{x}_{s_1}, s_1\right)-(1+\frac{1}{2}+\frac{1}{4})\boldsymbol{\epsilon}_\theta^{(1)}\left(\mathbf{x}_{s}, s\right) \right ] + \left [ (1+\frac{1}{2}+\frac{1}{4})\boldsymbol{\epsilon}_\theta^{(1)}\left(\mathbf{x}_{s}, s\right) - 2\boldsymbol{\epsilon}_\theta^{(1)}\left(\mathbf{x}_{\xi}, \xi\right) \right ] \\
    &\qquad \qquad \cdots
\end{aligned}
\end{equation}
We now define \textit{group}, for example, $\boldsymbol{\epsilon}_\theta^{(1)}\left(\mathbf{x}_{s}, s\right) - (1+\frac{1}{2})\boldsymbol{\epsilon}_\theta^{(1)}\left(\mathbf{x}_{s_1}, s_1\right)$, which is grouped by ``$[]$''. We also find that for each group, if the coefficient of the first term is $l$, then the coefficient of the second term is $l+\frac{1}{2^n}$ after using dichotomy for $n$ times. Note that $l \in [1, 2)$ such that $l+\frac{1}{2^n} \in (1,2], \forall n \in \mathbb{N}$. Besides, $t$ takes two different values in $\{s, s_1, \xi\}$. By Eq. (\ref{bounded-epsilon-appendix}) and Assumption \ref{assumption-neighbourhood}, if $n$ is large enough, each group is $\mathcal{O}(h_s)$, for example,
\begin{equation}
    l\boldsymbol{\epsilon}_\theta^{(1)}\left(\mathbf{x}_{s_1}, s_1\right)-\left ( l+\frac{1}{2^n} \right ) \boldsymbol{\epsilon}_\theta^{(1)}\left(\mathbf{x}_{s}, s\right)=l\boldsymbol{\epsilon}_\theta^{(1)}\left(\mathbf{x}_{s_1}, s_1\right)-l\boldsymbol{\epsilon}_\theta^{(1)}\left(\mathbf{x}_{\rho}, \rho\right)=\mathcal{O}(h_s).
\end{equation}
Hence $\boldsymbol{\epsilon}_\theta^{(1)}\left(\mathbf{x}_{s}, s\right) - 2\boldsymbol{\epsilon}_\theta^{(1)}\left(\mathbf{x}_{\xi}, \xi\right)$ is $\mathcal{O}(2^nh_s)$. In practice, $n$ is finite, meaning that $2^n$ is bounded and $\mathcal{O}(2^nh_s)=\mathcal{O}(h_s)$. Subquently, the proof is completed by
\begin{equation}
    \frac{\mathbf{x}_t-\tilde{\mathbf{x}}_{t}}{\alpha_{t}}=\mathcal{O}(h_s^3)+ \frac{h_s^2}{4}\mathcal{O}(h_s)+\mathcal{O}(h_s^3)=\mathcal{O}(h_s^3).
\end{equation}

\end{proof}

\subsection{Proof of Theorem 3.2 when \texorpdfstring{$k=3$}{}}
In this subsection, we prove the global convergence order of SciRE-Solver-3 is no less than 2.

\begin{proof}
For each iteration step, we first update according to:
\begin{align}
    h_s&=\tau_{t}-\tau_{s}, \\
    s_1&=\psi(\tau_{s}+r_1 h_s), \\
    s_2&=\psi(\tau_{s}+r_2 h_s), \\
    \mathbf{u}_1 &=\frac{\alpha_{s_1}}{\alpha_{s}} \mathbf{x}_{s}+\alpha_{s_1}r_1 h_s\boldsymbol{\epsilon}_\theta\left(\mathbf{x}_{s}, s\right), \\
    \mathbf{u}_2 &=\frac{\alpha_{s_2}}{\alpha_{s}} \mathbf{x}_{s}+\alpha_{s_2}r_2 h_s\boldsymbol{\epsilon}_\theta\left(\mathbf{x}_{s}, s\right)+\alpha_{s_2}\frac{h_s}{\phi_1 (m)}\left(\boldsymbol{\epsilon}_\theta\left(\mathbf{u}_1, s_1\right)-\boldsymbol{\epsilon}_\theta\left(\mathbf{x}_{s}, s\right)\right), \\
    \tilde{\mathbf{x}}_{t} &= \frac{\alpha_{t}}{\alpha_{s}} \mathbf{x}_{s}+\alpha_{t} h_s\boldsymbol{\epsilon}_\theta\left(\mathbf{x}_{s}, s\right)+\alpha_{t}\frac{h_s^2}{2\phi_1(m)r_2 h_s} \left(\boldsymbol{\epsilon}_\theta\left(\mathbf{u}_2, s_2\right)-\boldsymbol{\epsilon}_\theta\left(\mathbf{x}_{s}, s\right)\right),\label{approximation-3-appendix}
\end{align}
where $\tilde{\mathbf{x}}_{t}$ denotes the approximate solution of $\mathbf{x}_{t}$ computed by SciRE-Solver-3, $r_1\in(0, 1)$ and $r_2=1-r_1$ are hyperparameters so that $s_1, s_2 \in (t, s)$.

We firstly prove that $\mathbf{u}_2 - \mathbf{x}_{s_2}=\mathcal{O}(h_s^3)$:
\begin{equation}
\begin{aligned}
    \mathbf{u}_2 &=\frac{\alpha_{s_2}}{\alpha_{s}} \mathbf{x}_{s}+\alpha_{s_2}r_2 h_s\boldsymbol{\epsilon}_\theta\left(\mathbf{x}_{s}, s\right)+\alpha_{s_2}\frac{h_s}{\phi_1 (m)}\left(\boldsymbol{\epsilon}_\theta\left(\mathbf{u}_1, s_1\right)-\boldsymbol{\epsilon}_\theta\left(\mathbf{x}_{s}, s\right)\right) \\
    &=\underbrace{\frac{\alpha_{s_2}}{\alpha_{s}} \mathbf{x}_{s}+\alpha_{s_2}r_2 h_s\boldsymbol{\epsilon}_\theta\left(\mathbf{x}_{s}, s\right)+\alpha_{s_2}\frac{h_s}{\phi_1 (m)}\left(\boldsymbol{\epsilon}_\theta\left(\mathbf{x}_{s_1}, s_1\right)-\boldsymbol{\epsilon}_\theta\left(\mathbf{x}_{s}, s\right)\right)}_{\mathbf{x}_{s_2}} + \mathcal{O}(h_s^3),
\end{aligned}
\end{equation}
by $\mathbf{u}_1 - \mathbf{x}_{s_1}=\mathcal{O}(h_s^2)$ and Assumption \ref{assumption-lipschitz}.

Next, taking $n=2$ in Eq. (\ref{itersolution2-appendix}), we can get the exact solution of $\mathbf{x}_{t}$ as follows:
\begin{equation}
    \mathbf{x}_t=\frac{\alpha_t}{\alpha_s}\mathbf{x}_s
    + \alpha_t h_s \boldsymbol{\epsilon}_\theta\left(\mathbf{x}_{s}, s\right)
    + \alpha_{t}\sum_{k=1}^{2}\frac{h_{s}^{k+1}}{(k+1)!}\boldsymbol{\epsilon}_\theta^{(k)}\left(\mathbf{x}_{s}, s\right)
    + \mathcal{O}(h_{s}^{4})
\end{equation}

Then by subtracting the last equation from Eq. (\ref{approximation-3-appendix}) and using $\mathbf{u}_1 - \mathbf{x}_{s_1}=\mathcal{O}(h_s^2)$, we have
\begin{equation}\label{truncation-error-3-appendxi}
\begin{aligned}
\frac{\mathbf{x}_t-\tilde{\mathbf{x}}_{t}}{\alpha_{t}}&=\sum_{k=1}^{2}\frac{h_{s}^{k+1}}{(k+1)!}\boldsymbol{\epsilon}_\theta^{(k)}\left(\mathbf{x}_{s}, s\right)-
\frac{h_s^2}{2\phi_1(m)r_2 h_s} \left(\boldsymbol{\epsilon}_\theta\left(\mathbf{u}_2, s_2\right)-\boldsymbol{\epsilon}_\theta\left(\mathbf{x}_{s}, s\right)\right)+ \mathcal{O}(h_{s}^{4})  \\
&=\frac{h_s^2}{2}\boldsymbol{\epsilon}_{\theta}^{(1)}\left(\mathbf{x}_{s}, s\right) + \frac{h_s^3}{3!}\boldsymbol{\epsilon}_{\theta}^{(2)}\left(\mathbf{x}_{s}, s\right) - \frac{h_s^2}{2}\frac{\boldsymbol{\epsilon}_\theta\left(\mathbf{x}_{s_2}, s_2\right)-\boldsymbol{\epsilon}_\theta\left(\mathbf{x}_{s}, s\right)}{\phi_1(m)r_2 h_s} + \mathcal{O}(h_s^4),
\end{aligned}
\end{equation}
where $\| \boldsymbol{\epsilon}_\theta\left(\mathbf{u}_2, s_2\right) - \boldsymbol{\epsilon}_\theta\left(\mathbf{x}_{s_2}, s_2\right) \|=\mathcal{O}(\| \mathbf{u}_2 - \mathbf{x}_{s_2} \|)=\mathcal{O}(h_s^3)$.
Similar to the proof in \ref{proof-theorem-32-2}, we have
\begin{equation}
\begin{aligned}
    \frac{\mathbf{x}_t-\tilde{\mathbf{x}}_{t}}{\alpha_{t}}&=\sum_{k=1}^{2}\frac{h_{s}^{k+1}}{(k+1)!}\boldsymbol{\epsilon}_\theta^{(k)}\left(\mathbf{x}_{s}, s\right)-
    \frac{h_s^2}{2} \frac{\boldsymbol{\epsilon}_\theta\left(\mathbf{x}_{s_2}, s_2\right)-\boldsymbol{\epsilon}_\theta\left(\mathbf{x}_{s}, s\right)}{\phi_1(m)r_2 h_s}+ \mathcal{O}(h_{s}^{4})  \\
    &= \frac{h_s^3}{3!}\boldsymbol{\epsilon}_\theta^{(2)}\left(\mathbf{x}_{s}, s\right) + \frac{h_s^2}{2}\boldsymbol{\epsilon}_\theta^{(1)}\left(\mathbf{x}_{s}, s\right)-\frac{h_s^2}{2}\left ( \boldsymbol{\epsilon}_\theta^{(1)}\left(\mathbf{x}_{\xi}, \xi\right) + \frac{1}{2}\boldsymbol{\epsilon}_\theta^{(1)}\left(\mathbf{x}_{s}, s\right) \right )+ \mathcal{O}(h_{s}^{3})\\
    &=\frac{h_s^3}{3!}\boldsymbol{\epsilon}_\theta^{(2)}\left(\mathbf{x}_{s}, s\right)+\frac{h_s^2}{4}\left(\boldsymbol{\epsilon}_\theta^{(1)}\left(\mathbf{x}_{s}, s\right) - 2\boldsymbol{\epsilon}_\theta^{(1)}\left(\mathbf{x}_{\xi}, \xi\right) \right)+ \mathcal{O}(h_{s}^{3}) \\
    &=\mathcal{O}(h_{s}^{3})+2^n\mathcal{O}( h_{s}^{3})+\mathcal{O}(h_{s}^{3})=\mathcal{O}(h_{s}^{3}),
\end{aligned}
\end{equation}
where $\xi \in (s_2, s)$ and $n\in \mathbb{N}$ is finite. Hence we prove that the global convergence order of SciRE-Solver-3 is no less than 2.
\end{proof}

\section{Algorithms of SciRE-Solver}
In Section \ref{sec3.2}, we propose  the \emph{recursive derivative estimation} (RDE) method for approximating derivatives, the conclusion as shown in Theorem \ref{recge} and Corollary \ref{reclimited}. In this section, we discuss in detail the SciRE-Solver based on RDE.

We review that the $n$-th order Taylor expansion of $\boldsymbol{\epsilon}_\theta\big(\mathbf{x}_{\psi(\tau_s)}, \psi(\tau_s)\big)$ w.r.t. $\tau$ at $\tau_t$ is
\begin{equation}\label{taylor-a}
\boldsymbol{\epsilon}_\theta\left(\mathbf{x}_{\psi(\tau_s)}, \psi(\tau_s)\right) =
\sum_{k=0}^{n} \frac{h_{t}^k}{k!}
\boldsymbol{\epsilon}_\theta^{(k)}\left(\mathbf{x}_{\psi(\tau_t)}, \psi(\tau_t)\right)
+\mathcal{O}(h_t^{n+1}).
\end{equation}
After substituting this Taylor expansion into the score-integrand form presented in Eq. (\ref{score_solution}) for the diffusion ODEs, we derive
\begin{equation}\label{itersolution2-a}
   \mathbf{x}_{s}=\frac
   {\alpha_{s}
 }{\alpha_{t} }\mathbf{x}_{t} +\alpha_{s}\sum_{k=0}^{n}
 \frac
 {h_{t}^{k+1}}
{(k+1)!}\boldsymbol{\epsilon}_\theta^{(k)}\left(\mathbf{x}_{ \psi(\tau_{t})}, \psi(\tau_{t})\right)
+\mathcal{O}(h_{t}^{n+2}).
\end{equation}
When $n=2$, we have then the following truncation formula:
\begin{equation}\label{trunc-a}
\tilde{\mathbf{x}}_{s}=\frac
   {\alpha_{s}
 }{\alpha_{t} }\mathbf{x}_{t} +\alpha_{s}
 \left(h_{t} \boldsymbol{\epsilon}_\theta(\mathbf{x}_{ \psi(\tau_{t})}, \psi(\tau_{t})) + \frac{h_{t}^{2}}{2}\boldsymbol{\epsilon}_\theta^{(1)}(\mathbf{x}_{ \psi(\tau_{t})}, \psi(\tau_{t}))
 + \frac{h_{t}^{3}}{6}\boldsymbol{\epsilon}_\theta^{(2)}(\mathbf{x}_{ \psi(\tau_{t})}, \psi(\tau_{t}))
 + \mathcal{O}(h_{t}^{4})\right),
\end{equation}
where $\tilde{\mathbf{x}}_{s}$ represents the approximate value of $\mathbf{x}_{s}$.

By the RDE method in Corollary \ref{reclimited}, we have
\begin{equation}\label{rde-a}
    \begin{aligned}
\boldsymbol{\epsilon}^{(1)}_\theta\left(\mathbf{x}_{\psi(\tau_t)},\psi(\tau_t)\right)& =\frac{1}{\phi_1(m)}\frac{ \boldsymbol{\epsilon}_\theta\left(\tilde{\mathbf{x}}_{\psi(\tau_{s1})}, \psi(\tau_{s1})\right)-\boldsymbol{\epsilon}_\theta\left(\mathbf{x}_{\psi(\tau_t)}, \psi(\tau_t)\right)}{r_1h_t}\\
&-\frac{\phi_2(m)}{\phi_1(m)}\boldsymbol{\epsilon}^{(1)}_\theta\left(\mathbf{x}_{\psi(\tau_{s1})}, \psi(\tau_{s1})\right)
-\frac{\phi_3(m)r_1h_t}{\phi_1(m)}\boldsymbol{\epsilon}^{(2)}_\theta\left(\mathbf{x}_{\psi(\tau_{s1})}, \psi(\tau_{s1})\right)+\mathcal{O}(h_t^{2}).
    \end{aligned}
\end{equation}
where $\tau_{s1}-\tau_t=r_1h_t$.
Combining (\ref{trunc-a}) with Eq. (\ref{rde-a}), we have
\begin{equation}\label{trunca-1-order}
    \begin{aligned}
\tilde{\mathbf{x}}_{s}&={\frac
   {\alpha_{s}
 }{\alpha_{t} }\mathbf{x}_{t} +\alpha_{s}h_{t} \boldsymbol{\epsilon}_\theta\left(\mathbf{x}_{ \psi(\tau_{t})}, \psi(\tau_{t})\right)+\alpha_{s}\frac{h_{t}^{2}}{2}\frac{ \boldsymbol{\epsilon}_\theta\left(\tilde{\mathbf{x}}_{\psi(\tau_{s1})}, \psi(\tau_{s1})\right)-\boldsymbol{\epsilon}_\theta\left(\mathbf{x}_{\psi(\tau_t)}, \psi(\tau_t)\right)}{\phi_1(m)r_1h_t} }\\
& - {\alpha_{s}\frac{h_{t}^{2}}{2}\frac{\phi_2(m)}{\phi_1(m)}\boldsymbol{\epsilon}^{(1)}_\theta\left(\mathbf{x}_{\psi(\tau_{s1})}, \psi(\tau_{s1})\right)
+\alpha_{s}h_t^3\left(\frac{1}{6}-\frac{\phi_3(m)r_1}{2\phi_1(m)}\right) \boldsymbol{\epsilon}^{(2)}_\theta\left(\mathbf{x}_{\psi(\tau_{s1})}, \psi(\tau_{s1})\right)}\\
&+{\mathcal{O}(h_{t}^{4}).}
     \end{aligned}
\end{equation}

By truncating the term containing $h_t^2$ in Eq.  (\ref{trunca-1-order}), we obtain the algorithm shown in Algorithm \ref{algorithm:score-based-solver-2}:
\begin{equation}\label{algoritem-1}
    \begin{aligned}
\tilde{\mathbf{x}}_{s}\leftarrow\frac
   {\alpha_{s}
 }{\alpha_{t} }\mathbf{x}_{t} +\alpha_{s}h_{t} \boldsymbol{\epsilon}_\theta\left(\mathbf{x}_{ \psi(\tau_{t})}, \psi(\tau_{t})\right)+\alpha_{s}\frac{h_t}{2}\frac{ \boldsymbol{\epsilon}_\theta\left(\tilde{\mathbf{x}}_{\psi(\tau_{s1})}, \psi(\tau_{s1})\right)-\boldsymbol{\epsilon}_\theta\left(\mathbf{x}_{\psi(\tau_t)}, \psi(\tau_t)\right)}{\phi_1(m)r_1}.
     \end{aligned}
\end{equation}
According to Eq. (\ref{trunca-1-order}), it appears that we can easily conclude from Eq. (\ref{algoritem-1}) that the preliminary result has a local truncation error of $\mathcal O(h^2_t)$. Nevertheless, our numerical experiments conducted on different datasets demonstrate that the algorithm \ref{algorithm:score-based-solver-2} derived through this truncation method can generate high-quality samples with a restricted number of score function evaluations (NFE). Further details can be found in Appendix \ref{experiment-details}. After careful observation, in fact, this technique partially eliminates the dependence on derivatives while still containing derivative-related information (such as the difference between two score function evaluations), thereby mitigating the error propagation caused by derivative estimation to a certain extent. By repeatedly using this technique, we can derive our SciRE-Solver-$3$ in Algorithm \ref{algorithm:score-based-solver-3}.

Note that if we have an even better estimate for $\boldsymbol{\epsilon}^{(1)}_\theta\left(\mathbf{x}_{\psi(\tau_{s1})}, \psi(\tau_{s1})\right)$, we can further truncate the term containing $h_t^3$, leading to a generalized SciRE-Solver.
We leave it for future study.

\subsection{Analytical Formulation of the function $\rm{rNSR}(\cdot)$}
The computational costs associated with computing ${\rm{rNSR}}(\cdot)$ are negligible. This is due to the fact that for the noise schedules of $\alpha_t$ and $\sigma_t$ employed in previous DPMs (referred to as ``linear'' and ``cosine'' in \cite{ho2020denoising,nichol2021improved}), both ${\rm{rNSR}}(\cdot)$ and its inverse function ${\rm{NSR}}(t)$ have analytic formulations. We mainly consider the variance preserving type here, since it is the most widely-used type. The functions for other types (variance exploding and sub-variance preserving type) can be similarly derived.

\paragraph{Linear Noise Schedule \cite{ho2020denoising}.} In fact,
$$
\log \alpha_t=-\frac{\left(\beta_1-\beta_0\right)}{4} t^2-\frac{\beta_0}{2} t,
$$
where $\beta_0=0.1$ and $\beta_1=20$, following \cite{song2021scorebased,lu2022dpm}. As $\sigma_t=\sqrt{1-\alpha_t^2}$, we can compute $\rm{NSR}(t)$ analytically. Moreover, the inverse function is
$$
t={\rm{rNSR}}(\tau)
=\frac{1}{\beta_1-\beta_0}\left(\sqrt{\beta_0^2+2\left(\beta_1-\beta_0\right) \log \left(1+\tau^2\right)}-\beta_0\right),
$$
where $\tau={\rm{NSR}}(t)$.
To reduce the influence of numerical issues, we can compute $t$ by the following equivalent formulation:
$$
t={\rm{rNSR}}(\tau)
=\frac{2 \log \left(1+\tau^2\right)}{\sqrt{\beta_0^2+2\left(\beta_1-\beta_0\right) \log \left(1+\tau^2\right)}+\beta_0} .
$$
And we solve diffusion ODEs between $[\epsilon, T]$, where $T=1$.

\paragraph{Cosine Noise Schedule \cite{nichol2021improved}.}
 Denote
$$
\log \alpha_t=\log \left(\cos \left(\frac{\pi}{2} \cdot \frac{t+s}{1+s}\right)\right)-\log \left(\cos \left(\frac{\pi}{2} \cdot \frac{s}{1+s}\right)\right),
$$
where $s=0.008$, following \cite{nichol2021improved}. As $\sigma_t=\sqrt{1-\alpha_t^2}$, we can compute $\rm{NSR}(t)$ analytically. Denote $\tau={\rm{NSR}}(t)$, let
$$
\varphi(\tau)=-\frac{1}{2} \log \left(1+\tau^2\right)
$$
which computes the corresponding $\log \alpha$ for $\tau$. Then the inverse function is
$$
t={\rm{rNSR}}(\tau)=\frac{2(1+s)}{\pi} \arccos \left(e^{\varphi(\tau)+\log \cos \left(\frac{\pi s}{2(1+s)}\right)}\right)-s .
$$
And we solve diffusion ODEs between $[\epsilon, T]$, where $T=0.9946$, following \cite{lu2022dpm}.

\subsection{SciRE-Solver-agile}
In order to facilitate the exploration of more possibilities of the SciRE-Solver our proposed and to fully utilize the given number of score function evaluations (NFE), we defined a simple combinatorial  version based on SciRE-Solver-$k$ and named as SciRE-Solver-agile.
This version is based on whether the given NFE is divisible by $k$. If it is not divisible, solver-$k$ is used as much as possible first, and then smaller order SciRE-Solver or DDIM are used to supplement.

To achieve this, when given a fixed budget $N$ for the number of score function evaluations, we evenly divide the given interval into $M=(\lfloor N / 3\rfloor+1)$ segments. Subsequently, we carry out $M$ sampling steps, adjusting based on the remainder $R$ when dividing $N$ by 3 to ensure a precise total of $N$ evaluations.

When $R=0$, we initiate $M-2$ SciRE-Solver-3 steps, succeeded by 1 SciRE-Solver-2 step and 1 DDIM step. This results in a total of $3\cdot \left(\frac{N}{3}-1\right)+2+1=N$ evaluations.

In the case of $R=1$, we begin with $M-1$ SciRE-Solver-3 steps, followed by 1 DDIM step. This yields a total of $3\cdot \left(\frac{N-1}{3}\right)+1=N$ evaluations.

Lastly, when $R=2$, we conduct $M-1$ SciRE-Solver-3 steps, succeeded by 1 SciRE-Solver-2 step. This leads to a cumulative count of $3\cdot \left(\frac{N-2}{3}\right)+2=N$ score function evaluations.

Our empirical observations show that using this time step design can enhance the quality of image generation. With the implementation of the SciRE-Solver algorithm, high-quality samples can be generated in just $20$ steps, such as achieving a $2.42$ FID result on CIFAR-10 with just $20$ NFE.

\subsection{Sampling from Discrete-Time DPMs}
SciRE-Solver aims to solve continuous-time diffusion ODEs. For DPMs trained on discrete-time labels, we need to firstly wrap the model function to a noise prediction model that accepts the continuous time as the input.
In the subsequent discussion, we examine the broader scenario of discrete-time DPMs, specifically focusing on two variants: the $1000$-step DPMs \cite{ho2020denoising} and the $4000$-step DPMs \cite{nichol2021improved}.
Discrete-time DPMs \cite{ho2020denoising} train the noise prediction model at $N$ fixed time steps $\left\{t_n\right\}_{n=1}^N$, and the value of $N$ is typically set to either $1000$ or $4000$ in practice.
The implementation of the $4000$-step DPMs \cite{nichol2021improved} entails mapping the time steps of the $4000$-step DPMs to the range of the $1000$-step DPMs. Specifically, the noise prediction model is parameterized as $\tilde{\boldsymbol{\epsilon}}_\theta\left(\mathbf{x}_n, \frac{1000 n}{N}\right)$, where $\mathbf{x}_n$ is corresponding to the value at time $t_{n+1}$, and $n$ ranges from 0 to $N-1$. In practice, these discrete-time DPMs commonly employ  uniform time steps between $[0, T]$, then $t_n=\frac{n T}{N}$, for $n=1, \ldots, N$.

As sated by Lu et al. \cite{lu2022dpm}, the discrete-time noise prediction model is limited in predicting noise levels for times less than the smallest time $t_1$. Given that $t_1=\frac{T}{N}$ and the corresponding discrete-time noise prediction model at time $t_1$ is $\tilde{\boldsymbol{\epsilon}}_\theta\left(\mathbf{x}_0, 0\right)$, it is necessary to "scale" the discrete time steps from $\left[t_1, t_N\right]=\left[\frac{T}{N}, T\right]$ to the continuous time range $[\epsilon, T]$. However, the question of which scaling approach would be beneficial to the corresponding sampling algorithm remains an open problem.

In our codebase, we employ two types of scaling recommended by Lu et al. \cite{lu2022dpm} as follows.

\textbf{Discrete-1.} Let $\boldsymbol{\epsilon}_\theta(\cdot, t)=\boldsymbol{\epsilon}_\theta\left(\cdot, \frac{T}{N}\right)$ for $t \in\left[\epsilon, \frac{T}{N}\right]$, and scale the discrete time steps $\left[t_1, t_N\right]=\left[\frac{T}{N}, T\right]$ to the continuous time range $\left[\frac{T}{N}, T\right]$. Then, the continuous-time noise prediction model is defined by
$$
\boldsymbol{\epsilon}_\theta(\mathbf{x}, t)=\tilde{\boldsymbol{\epsilon}}_\theta\left(\mathbf{x}, 1000 \cdot \max \left(t-\frac{T}{N}, 0\right)\right),
$$
where the continuous time $t \in\left[\epsilon, \frac{T}{N}\right]$ maps to the discrete input 0 , and the continuous time $T$ maps to the discrete input $\frac{1000(N-1)}{N}$.

\textbf{Discrete-2.} Scale the discrete time steps $\left[t_1, t_N\right]=\left[\frac{T}{N}, T\right]$ to the continuous time range $[0, T]$. In this case, the continuous-time noise prediction model is defined by
$$
\boldsymbol{\epsilon}_\theta(\mathbf{x}, t)=\tilde{\boldsymbol{\epsilon}}_\theta\left(\mathbf{x}, 1000 \cdot \frac{(N-1) t}{N T}\right),
$$
where the continuous time 0 maps to the discrete input 0 , and the continuous time $T$ maps to the discrete input $\frac{1000(N-1)}{N}$.

By such reparameterization, the noise prediction model can adopt the continuous-time steps as input, which enables SciRE-Solver to perform  sampling not only for continuous-time DPMs but also for discrete-time DPMs.

\subsection{Conditional Sampling by SciRE-Solver}
With a simple modification, following the settings provided by Lu et al. \cite{lu2022dpm}, SciRE-Solver can be used for conditional sampling. The conditional generation requires sampling from a conditional diffusion ODE, as stated in  \cite{song2021scorebased,dhariwal2021diffusion}.
Specifically, by following the classifier guidance method \cite{dhariwal2021diffusion}, the conditional noise prediction model can defied as $\boldsymbol{\epsilon}_\theta\left(\mathbf{x}_t, t, y\right):=\boldsymbol{\epsilon}_\theta\left(\mathbf{x}_t, t\right)-s \cdot \sigma_t \nabla_{\mathbf{x}} \log p_t\left(y \mid \mathbf{x}_t ; \theta\right)$. Here, $p_t\left(y \mid \mathbf{x}_t ; \theta\right)$ represents a pre-trained classifier, and $s$ denotes the classifier guidance scale.
Thus, one can utilize SciRE-Solver to solve this diffusion ODE for fast conditional sampling.

\subsection{Supported Models}
SciRE-solver support four types of diffusion probabilistic models, including the noise prediction model $\boldsymbol{\epsilon}_\theta$ \cite{ho2020denoising,rombach2021highresolution},  the data prediction model $\mathbf{x}_\theta$  \cite{ramesh2022hierarchical}, the velocity prediction model $\mathbf{v}_\theta$ \cite{ho2022imagen} and marginal score function $\mathbf{s}_\theta$ \cite{song2021scorebased}.
Here, we follow the configurations provided by Lu et al. in \cite{lu2022dpm,lu2022dpm++}.
\section{Experiment Details}
\label{experiment-details}
In this section, we provide more details on SciRE-Solver and further demonstrate the performance of SciRE-Solver on both discrete-time DPMs and continuous-time DPMs. Specifically,  we consider the
$1000$-step DPMs \cite{ho2020denoising} and the $4000$-step DPMs \cite{nichol2021improved},  and consider the end time $\epsilon$ and time trajectory for sampling.  We test our method for sampling the most widely-used variance-preserving (VP) type DPMs \cite{sohl2015deep,song2021scorebased}. In this case, we have $\alpha_t^2+\sigma_t^2=1$ for all $t \in[0, T]$. In spite of this, our method and theoretical results are general and independent of the choice of the noise schedule $\alpha_t$ and $\sigma_t$. In all experiments, the number of NFE represents the sampling steps. For early experiments, we evaluate SciRE-Solver on NVIDIA TITAN X GPUs.

\subsection{End Time of Sampling}
Theoretically, we need to solve diffusion ODEs from time $T$ to time $0$ to generate samples. Practically,
the training and evaluation for the noise prediction model $\boldsymbol{\epsilon}_\theta\left(\mathbf{x}_{t}, t\right)$ usually start from time $T$ to time
$\epsilon$ to avoid numerical issues for t near to $0$, where $\epsilon\geq0$ is a hyperparameter \cite{song2021scorebased}.
In contrast to the sampling methods based on diffusion SDEs \cite{ho2020denoising,song2021scorebased}, we, like DPM-Solver \cite{lu2022dpm}, do not incorporate the ``denoising''
trick (i.e., setting the noise variance to zero) in the final step at time $\epsilon$. Instead, we solely solve diffusion
ODEs from T to $\epsilon$ using the SciRE-Solver.

\subsection{Time trajectories}\label{timetra}
Let $\left\{t_i\right\}_{i=0}^N$ be the time trajectory of diffusion probabilistic models, where $t_{N}=T$ and $t_0=\epsilon\geq0$.
In the context of fast sampling, it is always desirable for the number $N$ of time points in the time trajectory to be as small as possible.
However, the selection of the optimal time trajectory remains an open problem for the few-step sampling regime of diffusion probabilistic models. In this work, we hypothesize  that selecting a time trajectory with sparser time points in the middle and relatively denser time points at the two ends would be beneficial for improving the quality of sample generation. To validate this hypothesis, inspired by the logarithmic and sigmoid functions, we propose two parametrizable alternative methods for the $\rm{NSR}$ function to compute the time trajectory,  named as \emph{NSR}-type and \emph{Sigmoid}-type time trajectories, respectively.

\paragraph{\emph{NSR}-type:}
For a given starting time $t_T$ and ending time $t_0$ of the sampling, the time values at the intermediate endpoints $t_i$ of \emph{NSR}-type time trajectory are obtained as follows:
\begin{enumerate}
  \item ~$\text{trans}_T=-\log({\rm{NSR}}(t_T)+k\cdot{\rm{NSR}}(t_0))$,
  \item ~$\text{trans}_0=-\log({\rm{NSR}}(t_0)+k\cdot{\rm{NSR}}(t_0))$,
  \item ~$\text{trans}_i = \text{trans}_T + i\cdot\frac{\text{trans}_0-\text{trans}_T}{N}$,
  \item ~$t_i={\rm{rNSR}}(e^{-\text{trans}_i}-k\cdot{\rm{NSR}}(t_0))$,
\end{enumerate}
where $k$ is a hyperparameter that controls the flexibility of \emph{NSR}-type time trajectory.

In our experiments, we found that relatively good results can be obtained when $k\in[2,7]$. This means that when using this kind of time trajectory, one can consider setting the value of $k$ within this range.

\paragraph{\emph{Sigmoid}-type:}
For a given starting time $t_T$ and ending time $t_0$ of the sampling, the time values at the intermediate endpoints $t_i$ of \emph{Sigmoid}-type time trajectory are obtained as follows:
\begin{enumerate}
  \item ~$\text{trans}_T=-\log({\rm{NSR}}(t_T)), ~\text{trans}_0=-\log({\rm{NSR}}(t_0))$,
  \item ~$\text{central} = k\cdot\text{trans}_T+(1-k)\cdot\text{trans}_0$,
  \item ~$\text{shift}_T = \text{trans}_T -\text{central},~\text{shift}_0 = \text{trans}_0 -\text{central}$,
  \item ~$\text{scale} = \text{shift}_T + \text{shift}_0$,
  \item ~$\text{sigm}_T = sigmoid\left(\frac{\text{shift}_T}{\text{scale}}\right),~ \text{sigm}_0 = sigmoid\left(\frac{\text{shift}_0}{\text{scale}}\right)$,
  \item ~$\text{sigm}_i = \text{sigm}_T + i\cdot\frac{\text{sigm}_0-\text{sigm}_T}{N}$,
  \item ~$\text{trans}_i = \text{scale}\cdot logistic(\text{sigm}_i)+\text{central}$,
  \item ~$t_i={\rm{rNSR}}(e^{-\text{trans}_i})$,
\end{enumerate}
where $k$ is a hyperparameter that controls the flexibility of \emph{Sigmoid}-type time trajectory.

Empirically, we suggest using the \emph{NSR}-type time trajectory. However, when NFE is less than or equal to $15$, it is recommended to try using the \emph{Sigmoid}-type time trajectory. The generation quality measured by FID of \emph{NSR}-type time trajectory and \emph{Sigmoid}-type time trajectory are shown in Table (\ref{tab:snrtrajectory}) and Table (\ref{tab:snrtrajectorycontious}), respectively. Besides, we also demonstrate the efficiency of our proposed algorithms by using conventional time-quadratic trajectory in Table (\ref{tab:timequadratictra}). In these experimental results, \emph{NSR}-type time trajectory is better than time-quadratic trajectory.

\begin{table}[!ht]
\centering
\caption{
Generation quality measured by FID $\downarrow$ of different sampling methods for DPMs on the pre-trained discrete-time models \cite{ho2020denoising,song2021denoising}  of CIFAR-10 and CelebA 64$\times$64.
}
\label{tab:samecodebase}
\fontsize{9.7}{9}\selectfont
\begin{tabular}{lllrrrrr}
\toprule
Trajectory & Initial Time & Sampling method \textbackslash NFE   & 12 & 15  & 20 & 50 &100\\
\midrule
\multicolumn{8}{l}{CIFAR-10 (discrete-time model \cite{ho2020denoising}, linear noise schedule)}
 \\
 \midrule
\multirow{5}{*}{logSNR}&\multirow{5}{*}{$\epsilon=10^{-3}$}&
DDIM      & $16.08$ & $12.43$ & $9.28$ & $5.36$ &  $4.55$\\
& &DPM-Solver-$2$  & $\textbf{5.18}$ & $^{\textbf{\dag}}\textbf{4.42}$ & $4.05$ & $3.97$ & $3.97$\\
& &DPM-Solver-$3$  & $7.39$ & $4.60$ & $^{\textbf{\dag}}4.33$ & $^{\textbf{\dag}}3.98$ &$^{\textbf{\dag}}3.97$ \\
& &SciRE-Solver-$2$ (ours) & $5.48$ & $^{\textbf{\dag}}4.55$ & $3.96$ & $3.66$ & $3.71$ \\
& &SciRE-Solver-$3$ (ours) & $8.53$ & $5.00$ & $^{\textbf{\dag}}4.34$ & $^{\textbf{\dag}}3.66$ & $^{\textbf{\dag}}3.62$ \\
\midrule
\multirow{5}{*}{logSNR}& \multirow{5}{*}{$\epsilon=10^{-4}$}&
DDIM     &$17.40$  & $13.12$ & $9.54$ & $5.03$ & $4.13$\\
& &DPM-Solver-$2$ & $6.40$ &  $^{\textbf{\dag}}5.26$ & $4.02$ & $3.56$ & $3.51$\\
& &DPM-Solver-$3$  & $9.52$ & $5.17$ & $^{\textbf{\dag}}\textbf{3.80}$ & $^{\textbf{\dag}}3.53$ &  $^{\textbf{\dag}}3.50$\\
& &SciRE-Solver-$2$ (ours) & $6.48$ & $^{\textbf{\dag}}5.35$  &  $4.01$ & $3.34$ & $3.27$ \\
& &SciRE-Solver-$3$ (ours) & $11.71$ & $5.99$ & $^{\textbf{\dag}}4.15$ & $^{\textbf{\dag}}\textbf{3.30}$ & $^{\textbf{\dag}}\textbf{3.163}$\\
\midrule
\multirow{5}{*}{NSR ($k=2$)}&\multirow{5}{*}{$\epsilon=10^{-3}$}&
DDIM     & $13.58$ & $10.63$ & $8.12$ &  $5.03$&  $4.40$\\
& &DPM-Solver-$2$   & $4.91$ & $^{\textbf{\dag}}4.51$ & $4.19$ & $4.00$ & $3.96$\\
& &DPM-Solver-$3$   & $7.33$ & $4.97$ & $^{\textbf{\dag}}4.56$ & $^{\textbf{\dag}}4.00$ & $^{\textbf{\dag}}3.96$\\
& &SciRE-Solver-$2$ (ours) & $\textbf{4.49}$ & $^{\textbf{\dag}}4.12$ & $3.74$ & $3.70$ & $3.76$\\
& &SciRE-Solver-$3$ (ours) & $5.29$ & $4.19$ & $^{\textbf{\dag}}3.94$ & $^{\textbf{\dag}}3.76$ & $^{\textbf{\dag}}3.71$ \\
\midrule
\multirow{5}{*}{NSR ($k=2$)}&\multirow{5}{*}{$\epsilon=10^{-4}$}&
DDIM     & $15.51$ & $11.86$ & $8.77$ & $4.86$ & $4.07$\\
& &DPM-Solver-$2$  & $5.38$ & $^{\textbf{\dag}}4.46$ & $3.78$ & $3.53$ & $3.51$\\
& &DPM-Solver-$3$  & $7.29$ & $\textbf{4.03}$ & $^{\textbf{\dag}}\textbf{3.66}$ & $^{\textbf{\dag}}3.52$ &  $^{\textbf{\dag}}3.50$\\
& &SciRE-Solver-$2$ (ours) & $5.91$ & $^{\textbf{\dag}}4.76$  &  $3.88$ & $3.30$ & $3.28$ \\
& &SciRE-Solver-$3$ (ours) & $9.10$ & $4.52$ & $^{\textbf{\dag}}4.07$ & $^{\textbf{\dag}}\textbf{3.24}$ & $^{\textbf{\dag}}\textbf{3.167}$\\
\toprule
\multicolumn{8}{l}{CelebA 64$\times$64 (discrete-time model \cite{song2021denoising}, linear noise schedule)}
 \\
 \midrule
\multirow{5}{*}{logSNR}&\multirow{5}{*}{$\epsilon=10^{-3}$}&
DDIM     & $14.37$ & $11.91$ & $9.66$ & $6.13$ &  $5.15$\\
& &DPM-Solver-$2$   & $3.952$ & $^{\textbf{\dag}}3.953$ & $4.05$ & $4.21$ & $4.24$\\
& &DPM-Solver-$3$ & $3.79$ & $3.91$ & $^{\textbf{\dag}}4.05$ & $^{\textbf{\dag}}4.26$ &$^{\textbf{\dag}}4.25$ \\
& &SciRE-Solver-$2$ (ours) & $5.39$ & $^{\textbf{\dag}}4.51$ & $3.76$ & $3.49$ & $3.71$\\
& &SciRE-Solver-$3$ (ours) & $4.91$ & $3.65$ & $^{\textbf{\dag}}3.29$ & $^{\textbf{\dag}}3.09$ & $^{\textbf{\dag}}3.41$ \\
\midrule
\multirow{5}{*}{logSNR}& \multirow{5}{*}{$\epsilon=10^{-4}$}&
DDIM     &$12.81$  & $10.28$ & $7.98$ & $4.52$ & $3.59$\\
& &DPM-Solver-$2$  & $\textbf{3.26}$ &  $^{\textbf{\dag}}3.14$ & $2.92$ & $2.82$ & $2.82$\\
& &DPM-Solver-$3$  & $3.93$ & $\textbf{2.91}$ & $^{\textbf{\dag}}2.85$ & $^{\textbf{\dag}}2.82$ &  $^{\textbf{\dag}}2.81$\\
& &SciRE-Solver-$2$ (ours) & $4.29$ & $^{\textbf{\dag}}3.70$  &  $2.87$ & $2.37$ & $2.43$ \\
& &SciRE-Solver-$3$ (ours) & $5.04$ & $3.43$ & $^{\textbf{\dag}}\textbf{2.58}$ & $^{\textbf{\dag}}\textbf{2.06}$ & $^{\textbf{\dag}}\textbf{2.20}$\\
\midrule
\multirow{5}{*}{NSR ($k=2$)}&\multirow{5}{*}{$\epsilon=10^{-3}$}&
DDIM      & $13.08$ & $10.99$ & $8.96$ &  $5.88$&  $4.40$\\
& &DPM-Solver-$2$   & $5.39$ & $^{\textbf{\dag}}4.93$ & $4.37$ & $4.24$ & $4.236$\\
& &DPM-Solver-$3$   & $6.14$ & $4.77$ & $^{\textbf{\dag}}4.41$ & $^{\textbf{\dag}}4.24$ & $^{\textbf{\dag}}4.24$\\
& &SciRE-Solver-$2$ (ours) & $4.61$ & $^{\textbf{\dag}}4.20$ & $3.58$  &$3.56$  & $3.76$\\
& &SciRE-Solver-$3$ (ours) & $4.75$& $3.33$ & $^{\textbf{\dag}}3.04$ & $^{\textbf{\dag}}3.15$& $^{\textbf{\dag}}3.51$ \\
\midrule
\multirow{5}{*}{NSR ($k=2$)}&\multirow{5}{*}{$\epsilon=10^{-4}$}&
DDIM     & $11.88$ & $9.59$ & $7.53$ & $4.38$ & $3.54$\\
& &DPM-Solver-$2$  & $3.11$ & $^{\textbf{\dag}}2.91$ & $2.88$ & $2.79$ & $2.81$\\
& &DPM-Solver-$3$  & $\textbf{2.94}$ & $2.88$ & $^{\textbf{\dag}}2.87$ & $^{\textbf{\dag}}2.80$ &  $^{\textbf{\dag}}2.81$\\
& &SciRE-Solver-$2$ (ours) & $3.95$ & $^{\textbf{\dag}}3.39$ & $2.61$& $2.31$ & $2.43$\\
& &SciRE-Solver-$3$ (ours) & $4.47$ & $\textbf{2.68}$  & $^{\textbf{\dag}}\textbf{2.29}$  &  $^{\textbf{\dag}}\textbf{2.03}$ &$^{\textbf{\dag}}\textbf{2.20}$\\
\bottomrule
\end{tabular}
\end{table}

\subsection{Comparing sample quality with different samplers}
We show the detailed FID results of different sampling methods for DPMs on CIFAR-10 and CelebA 64$\times$64 with discrete-time or continuous-time pre-trained models in Table \ref{tab:differdatasets}. We utilize the code and checkpoint provided in \cite{ho2020denoising,song2021scorebased,nichol2021improved}. Specifically, we employ their \emph{checkpoint\_8} of the ``VP deep” type. In this table, we compare the FID achieved by our proposed SciRE-Solver with the best FID reported in existing literature at the same NFE.
We consistently use the NSR-type time trajectory with parameter $k=3.1$ for SciRE-Solver on the discrete models of CIFAR-10 and CelebA 64$\times$64 datasets.
For continuous models on the CIFAR-10 dataset, we use a Sigmoid-type time trajectory with parameter $k=0.65$ for the SciRE-Solver when the NFE is less than $15$. When NFE is greater than or equal to $15$, we consistently use an NSR-type time trajectory with $k=3.1$.
In order to objectively compare the quality of generated samples for the CelebA 64$\times$64 dataset, given the presence of different FID statistical data, we utilized the FID stats employed by Liu et al. \cite{liu2022pseudo} in Tables \ref{tab:differdatasets}, \ref{tab:timequadratictra} and \ref{tab:snrtrajectory}, 
and utilized the FID stats employed by Lu et al. \cite{lu2022dpm} in Tables  \ref{tab:samecodebase}, \ref{tab:rdedpm-2} and \ref{tab:rdenonrde}. Figure \ref{fig:dsicrete-continuious-cifar10-results}
 illustrates the FIDs achieved by different samplers at various NFE levels. Moreover, in Table \ref{tab:imagenet128u}, we also evaluate SciRE-Solver, DPM-Solver and DDIM with the same settings on the pre-trained model of high-resolution ImageNet 128$\times$128 dataset \cite{dhariwal2021diffusion}, refer to Figures \ref{fig:imagenet2128samecomparison} and \ref{fig:imagenet128samecomparison} for the  comparisons of generated samples. In all tables, the results $^{\textbf{\dag}}$ means the actual NFE is smaller than the given NFE.

In Table \ref{tab:samecodebase}, in order to ensure fairness, we compare the generation performance of SciRE-Solver with DPM-Solver and DDIM on discrete models \cite{ho2020denoising,song2021denoising} of CIFAR-10 and CelebA 64$\times$64 datasets using the same trajectories, settings and codebase.
In this experiment, we employ different time trajectories to evaluate the sampling performance of each sampling algorithm, such as the NSR trajectory and the logNSR trajectory \cite{lu2022dpm}.
Unlike in Table \ref{tab:differdatasets} with parameter $k=3.1$,
we consistently use parameter $k=2$ for the NSR time trajectory in Table \ref{tab:samecodebase},
in order to showcase the impact of different $k$ values on the samplers. Meanwhile, we also compare the performance of generative samples  for these three samplers at different sampling endpoints, such as $1e-3$ and $1e-4$.

Tables \ref{tab:differdatasets} and \ref{tab:samecodebase} demonstrate that the SciRE-Solver  attains SOTA sampling performance  with limited NFE on both discrete-time and continuous-time DPMs in comparison to existing training-free sampling algorithms. Such as, in Table \ref{tab:differdatasets}, we achieve $3.48$ FID with $12$ NFE and $2.42$ FID with $20$ NFE for continuous-time DPMs on CIFAR10, respectively.
Furthermore, with fewer NFE, SciRE-Solver surpass the benchmark values
demonstrated in the original paper of the proposed pre-trained model.
For example, we  reach SOTA value of $2.40$ FID with no more than $100$ NFE for continuous-time DPMs and of $3.15$ FID with $84$ NFE for discrete-time DPMs on CIFAR-10, as well as of $2.17$ FID with $18$ NFE for discrete-time DPMs on CelebA 64$\times$64. Moreover, SciRE-Solver can also achieve SOTA sampling performance within $100$ NFE for both the NSR time trajectory with different parameter values $k$ and the logSNR time trajectory, as shown in Tables  \ref{tab:differdatasets} and \ref{tab:samecodebase}. Especially, in Table \ref{tab:samecodebase}, DPM-Solver is more likely to achieve better sampling performance within $15$ NFE for the logSNR time trajectory and the NSR time trajectory with $k=2$.
However, when NFE exceeds $15$, ScrRE-Solver becomes more advantageous. Moreover, when the endpoint of the sampling is set at $1e-4$, both with the logSNR time trajectory and NSR time trajectory ($k=2$), SciRE-Solver can achieve SOTA sampling performance between $50$ NFE and $100$ NFE.

In Table \ref{tab:imagenet128u}, we also evaluate SciRE-Solver, DPM-Solver and DDIM on the high-resolution ImageNet 128$\times$128 dataset \cite{dhariwal2021diffusion}.
For the sake of fairness, we use the same uniform time trajectory, the same codebase, and the same settings to evaluate SciRE-Solver-2, DPM-Solver-2, and DDIM for 10, 12, 15, 20, and 50 NFEs. The numerical experiment results report that SciRE-Solver-2 achieved 5.58 FID with 10 NFE and 3.67 FID with 20 NFE, respectively, while DMP-Solver-2 only achieved 4.17 FID with 50 NFE. In all these different NFEs, SciRE-Solver-2 outperforms DPM-Solver-2.

In summary, within $20$ NFE, SciRE-Solver with NSR trajectory ($k=3.1$) achieves better FID than existing training-free solvers \cite{bao2022analyticdpm,song2021denoising,zhang2023fast,lu2022dpm,liu2022pseudo,li2023era} for  CIFAR-10 and CelebA 64$\times$64 datasets, as shown Table \ref{tab:differdatasets}.
Meanwhile, within 100 NFE (or even 1000 NFE), existing solvers in the context of discrete models on CIFAR-10 dataset are hardly able to achieve an FID below $3.45$, as shown in Table \ref{tab:differdatasets}. On the other hand, SciRE-Solver, with different time trajectories such as logNSR trajectory and NSR trajectory, can achieve an FID below $3.17$, and even surpass the $3.16$ FID obtained by DDPM at $1000$ NFE, as shown in Tables  \ref{tab:differdatasets} and \ref{tab:samecodebase}. For the continuous VP-type model on CIFAR-10, SciRE-Solver also surpasses the $2.41$ FID obtained by Song et al. \cite{song2021scorebased} using SDE solver with $1000$ NFE.
In Table \ref{tab:samecodebase}, under the time trajectories, settings and the same codebase, SciRE-Solver outperforms the DPM-Solver \cite{lu2022dpm} widely used in stable diffusion \cite{rombach2021highresolution}. Specifically, SciRE-Solver achieves an FID of $3.16$ and $2.03$ within $100$ NFE on CIFAR-10 and CelebA 64$\times$64 datasets respectively, whereas DPM-Solver  struggles to achieve FID values lower than $3.50$ and $2.79$ respectively on the same datasets.
Furthermore, the FID comparison on the high-resolution 128$\times$128 dataset presented in Table \ref{tab:imagenet128u} suggests that SciRE-Solver also possesses advantages in sample generation tasks involving high-resolution image datasets. For more random sampling sample comparisons on different high-resolution ($\geq$128$\times$128) image datasets, please refer to Figures \ref{fig:imagenet2128samecomparison}, \ref{fig:imagenet128samecomparison},
\ref{fig:imagenet512samecomparison},
\ref{fig:imagenet128comparison}, \ref{fig:lsunbedroom256comparison}, \ref{fig:imagenet256comparison}, and \ref{fig:lsunbedroomsamecomparison}.

\begin{table}[!ht]
\centering
\caption{
Generation quality measured by FID $\downarrow$ of different sampling methods with the same codebase
for DPMs on the pre-trained discrete-time model of Imagenet 128$\times$128 \cite{dhariwal2021diffusion}.
}
\label{tab:imagenet128u}
\fontsize{9.8}{9}\selectfont
\begin{tabular}{lllrrrrr}
\toprule
Trajectory & Initial Time & Sampling method \textbackslash NFE   & 10 & 12  & 15 & 20 &50\\
\midrule
\multicolumn{8}{l}{Imagenet 128$\times$128 (with classifier guidance: scale=1.25, under the same codebase)}
 \\
 \midrule
\multirow{3}{*}{Uniform time}&\multirow{3}{*}{$\epsilon=10^{-3}$}&
DDIM & $9.33$& $7.55$&  $6.07$  & $4.91$ &  $3.51$\\
& & DPM-Solver-2  & $10.17$& $7.78$	&  $^{\textbf{\dag}}6.68$  & $5.46$ &  $4.17$\\
& & SciRE-Solver-2 (our)& $5.58$ & $4.73$& $^{\textbf{\dag}}4.27$  & $3.67$ & $3.36$ \\
\bottomrule
\end{tabular}
\vspace{-0.1cm}
\end{table}

\subsection{Ablations study}
\subsubsection{Different orders and Starting times}
\paragraph{Order} We compare the sample quality with different orders of SciRE-Solver-2,3. However, in practice, the actual NFE may be smaller than the given NFE, for example, given the NFE=15, the actucal NFE of SciRE-Solver-2 is 14. To mitigate this problem, we propose the SciRE-Solver-agile method for continuous models. We compare the results of models with different orders on CIFAR-10 and CelebA 64$\times$64 datasets.
Our results indicate that if NFE is less than 20, SciRE-Solver-2 outperforms SciRE-Solver-3, or the latter variant is superior – depending on the specific use case.
\begin{table}
\vspace{-0.5cm}
\centering
\caption{SciRE-Solver with time-quadratic trajectory}
\label{tab:timequadratictra}
\begin{tabular}{clrrrrrrr}
\toprule
Initial time& Sampling method $\backslash$ NFE & 12 & 15 & 20 & 50 & 100 \\
\midrule \multicolumn{4}{l}{ CIFAR-10 (discrete-time model \cite{ho2020denoising})}  & \\
\midrule
\multirow{2}{*}{$\epsilon=10^{-3}$}&SciRE-Solver-2&   $\textbf{4.86}$ & $^{\textbf{\dag}}\textbf{4.10}$ & $\textbf{3.56}$ & $3.74$ & $3.83$  \\
& SciRE-Solver-3  & $19.37$ & $11.18$ & $^{\textbf{\dag}}7.48$ & $^{\textbf{\dag}}3.94$ & $^{\textbf{\dag}}3.85$ \\
\midrule
\multirow{2}{*}{$\epsilon=10^{-4}$}&SciRE-Solver-2&  $6.13$ &  $^{\textbf{\dag}}5.12$&$3.83$  & \textbf{3.31} & 3.27  \\
& SciRE-Solver-3  & $22.39$ &$13.09$ & $^{\textbf{\dag}}8.54$ &$^{\textbf{\dag}}3.45$  &$^{\textbf{\dag}}\textbf{3.22}$  \\
\midrule \multicolumn{4}{l}{ CIFAR-10 (VP deep continuous-time model \cite{song2021scorebased}) }  & \\
\midrule
& SciRE-Solver-2 & $\textbf{5.00}$ &  ${ }^{\dagger}$4.24 & 3.23 & 2.59 & 2.53 \\
$\epsilon=10^{-4}$ & SciRE-Solver-3&  12.53 & 7.33 & ${ }^{\dagger}$5.43 & ${ }^{\dagger}$2.64 & ${ }^{\dagger}$\textbf{2.50}\\
& SciRE-Solver-agile &  $5.03$ & $\textbf{4.24}$ & $\textbf{3.21}$ & $\textbf{2.59}$ & $2.51$ \\
\toprule
Initial time& Sampling method $\backslash$ NFE & 12 & 15 & 20 & 30& 50 \\
\midrule  \multicolumn{4}{l}{ CelebA 64$\times$64 (discrete-time model\cite{song2021denoising}) }  & \\
\midrule
\multirow{2}{*}{$\epsilon=10^{-3}$}&SciRE-Solver-2&   $5.83$& $^{\textbf{\dag}}4.67$ & $3.92$ & $3.77$ & $3.86$  \\
& SciRE-Solver-3  &$8.72$  &$5.06$ &$^{\textbf{\dag}}3.81$  & $^{\textbf{\dag}}3.31$ & $^{\textbf{\dag}}3.56$ \\
\midrule
\multirow{2}{*}{$\epsilon=10^{-4}$}&SciRE-Solver-2&  $\textbf{4.24}$  &  $^{\textbf{\dag}}\textbf{3.27}$& $\textbf{2.46}$  & $2.23$ & $2.20$  \\
& SciRE-Solver-3  & $11.08$ &$5.62$ &  $^{\textbf{\dag}}3.53$& $^{\textbf{\dag}}\textbf{2.13}$ & $^{\textbf{\dag}}\textbf{2.03}$ \\
\bottomrule
\end{tabular}
\vspace{-0.1cm}
\end{table}
\paragraph{Starting time} We also compare SciRE-Solver-2,3 with different starting times $\epsilon=10^{-3}$ and $\epsilon=10^{-4}$. Corresponding results are placed in Tables {\ref{tab:timequadratictra}} and \ref{tab:snrtrajectory}. We use time-quadratic trajectory and \emph{NSR}-type time trajectory for both SciRE-Solver-2 and SciRE-Solver-3 on CIFAR-10 and CelebA 64 $\times$64 datasets.
In our study on the CIFAR-10 dataset, we have observed that employing a sampling method with $\epsilon=10^{-3}$ results in superior sample quality for both continuous and discrete models when NFE is restricted to either 12 or 15. However, for NFE values greater than 15, we recommend opting for $\epsilon=10^{-4}$ to ensure the generation of high-quality samples.
Moreover, in our analysis of the CelebA 64 $\times$64 dataset, we have found that $\epsilon=10^{-4}$ consistently yields better results than $\epsilon=10^{-3}$ across different orders and NFEs. It is noteworthy that for NFE=20, SciRE-Solvers-2,3 show promising results that are on par with the former.

\begin{table}
\centering
\caption{SciRE-Solver-agile with NSR trajectory and starting time $1e-4$.}
\label{tab:snrtrajectoryee}
\begin{tabular}{p{2cm}p{2.5cm}rrrrrrr}
\toprule
$k$& $\phi_1(m)$$\backslash$ NFE & 12 & 15 & 20 & 50 & 100 \\
\midrule \multicolumn{6}{l}{ CIFAR-10 (VP deep continuous-time model \cite{song2021scorebased})}   \\
\midrule
\multirow{2}{*}{$k=3.1$}&$\phi_1(m)=\phi_1(3)$ & $6.93$& $3.73$ & $\textbf{2.42}$ & $2.52$ & $2.48$\\
&$\phi_1(m)=\frac{e-1}{e}$ & $6.79$ & $\textbf{2.57}$  & $2.48$ & $2.61$  & $2.41$\\
\midrule
\multirow{2}{*}{$k=2.2$}&$\phi_1(m)=\phi_1(3)$ &$4.06$   & $3.34$ & $2.54$ & $\textbf{2.51}$  & $2.42$\\
&$\phi_1(m)=\frac{e-1}{e}$ & $6.15$ & $3.39$ & $2.57$ &  $2.61$ & $\textbf{2.40}$ \\
\bottomrule
\end{tabular}
\end{table}

\begin{table}
\vspace{-0.25cm}
\centering
\caption{
Comparison between different time trajectories, starting time is  $1e-3$.
}
\label{tab:snrtrajectorycontious}
\begin{tabular}{clrrrrrrr}
\toprule
 \multicolumn{6}{l}{ CIFAR-10 (VP deep continuous-time model \cite{song2021scorebased})}  \\
 \midrule
Sampling method & Sampling method$\backslash$ NFE & 12& 15 \\
\midrule
\multirow{2}{*}{SciRE-Solver-$3$}& \emph{NSR}-type($k=3.2-0.005\cdot$NFE) & $4.41$& $\textbf{3.06}$& & \\
& \emph{Sigmoid}-type ($k=0.65$) & $\textbf{3.48}$ & $3.47$ & &\\
\bottomrule
\end{tabular}
\vspace{-0.3cm}
\end{table}

\begin{table}
\vspace{-0.4cm}
\centering
\caption{SciRE-Solver  with NSR trajectory $(k=3.1)$.}
\label{tab:snrtrajectory}
\begin{tabular}{clrrrrrrr}
\toprule
Initial time& Sampling method $\backslash$ NFE & 12 & 15 & 20 & 50 & 100 \\
\midrule \multicolumn{4}{l}{ CIFAR-10 (discrete-time model \cite{ho2020denoising}) }  & \\
\midrule
\multirow{2}{*}{$\epsilon=10^{-3}$}&SciRE-Solver-2 & $\textbf{4.41}$
 &$^{\textbf{\dag}}4.09$  & $\textbf{3.67}$ & $3.70$ & $3.80$  \\
& SciRE-Solver-3  & $4.68$ & $\textbf{4.00}$ & $^{\textbf{\dag}}3.72$& $^{\textbf{\dag}}3.84$ & $^{\textbf{\dag}}3.77$ \\
\midrule
\multirow{2}{*}{$\epsilon=10^{-4}$}&SciRE-Solver-2& $5.86$  &$^{\textbf{\dag}}4.77$  & $3.87$ &  $3.28$ & $3.27$  \\
& SciRE-Solver-3  & $8.28$ &$4.51$ &  $^{\textbf{\dag}}3.96$& $^{\textbf{\dag}}\textbf{3.23}$ & $^{\textbf{\dag}}\textbf{3.17}$ \\
\toprule
Initial time& Sampling method $\backslash$ NFE & 12 & 15 & 20 & 30& 50 \\
\midrule
\multicolumn{4}{l}{ CIFAR-10 (VP deep continuous-time model \cite{song2021scorebased}) }  \\
\midrule
& SciRE-Solver-2 & $\textbf{5.49}$ &$^{\textbf{\dag}}4.19$ &  $3.02$& $2.55$ &$2.47$  \\
$\epsilon=10^{-4}$ & SciRE-Solver-3& $ 6.29$ &  $\textbf{3.39}$&  $^{\textbf{\dag}}2.68$& $^{\textbf{\dag}}2.56$ & $^{\textbf{\dag}}\textbf{2.44}$\\
& SciRE-Solver-agile &  $6.93$& $3.73$ & $\textbf{2.42}$ & $\textbf{2.52}$ & $2.48$ \\
\midrule  \multicolumn{4}{l}{ CelebA 64$\times$64 (discrete-time model\cite{song2021denoising}) }  & \\
\midrule
\multirow{2}{*}{$\epsilon=10^{-3}$}&SciRE-Solver-2&  $4.79$ & $^{\textbf{\dag}}4.28$ & $3.86$ & $3.69$&   $3.82$  \\
& SciRE-Solver-3  & $5.01$ & $3.32$&  $^{\textbf{\dag}}3.12$&  $^{\textbf{\dag}}3.09$ & $^{\textbf{\dag}}3.40$ \\
\midrule
\multirow{2}{*}{$\epsilon=10^{-4}$}&SciRE-Solver-2&  $\textbf{3.91}$ & $^{\textbf{\dag}}3.38$ & $2.56$ & $2.41$ & $2.30$  \\
& SciRE-Solver-3  & $4.07$ & $\textbf{2.53}$&  $^{\textbf{\dag}}\textbf{2.17}$&  $^{\textbf{\dag}}\textbf{2.03}$& $^{\textbf{\dag}}\textbf{2.02}$ \\
\bottomrule
\end{tabular}
\end{table}

\subsubsection{\texorpdfstring{$\phi_1(m)=\phi_1(3)$}{} or \texorpdfstring{$\phi_1(m)=\frac{e-1}{e}$}{} }
When running our proposed SciRE-Solver-$k$ in Algorithm \ref{algorithm:score-based-solver-2} and Algorithm \ref{algorithm:score-based-solver-3}, it is necessary to assign a value $m$ to $\phi_1(m)$. As stated in Corollary \ref{reclimited}, when assigning $m$, we need to ensure that $m\geq3$. Considering that the limit of $\phi_1(m)$ is $\frac{e-1}{e}$,
i.e., $\lim\limits_{m\rightarrow\infty}\phi_1(m)=
\lim\limits_{m\rightarrow\infty}\sum\limits_{k=1}^{m}\frac{(-1)^{k-1}}{k!}=\frac{e-1}{e}$,
then our experiments
only consider these two extreme cases, i.e., we only choose to allocate $m$ as 3 or directly set $\phi_1(m)=\frac{e-1}{e}$. We provide ablation experiments for these two cases in Table \ref{tab:snrtrajectoryee}. In case of $\phi_1(m)=\frac{e-1}{e}$, we reach $2.40$ FID SOTA value with $100$ NFE on CIFAR-10 dataset.

\section{
Samples generated on different datasets and some comparisons}
\begin{figure}[ht]
\vspace{-0.45cm}
\centering
\begin{tabular}{m{0.65cm}p{2.7cm}p{2.7cm}p{2.7cm}p{2.7cm}}
   ~~ &~~~~~~~~~~NFE=$6$& ~~~~~~~~~~NFE=$8$  &~~~~~~~~~~NFE=$10$ &~~~~~~~~~~NFE=$12$ \\
\multirow{-18.7}{*}{\parbox{0.8cm}{\centering DDIM \cite{song2021denoising}}}
& \includegraphics[width=0.221\textwidth]{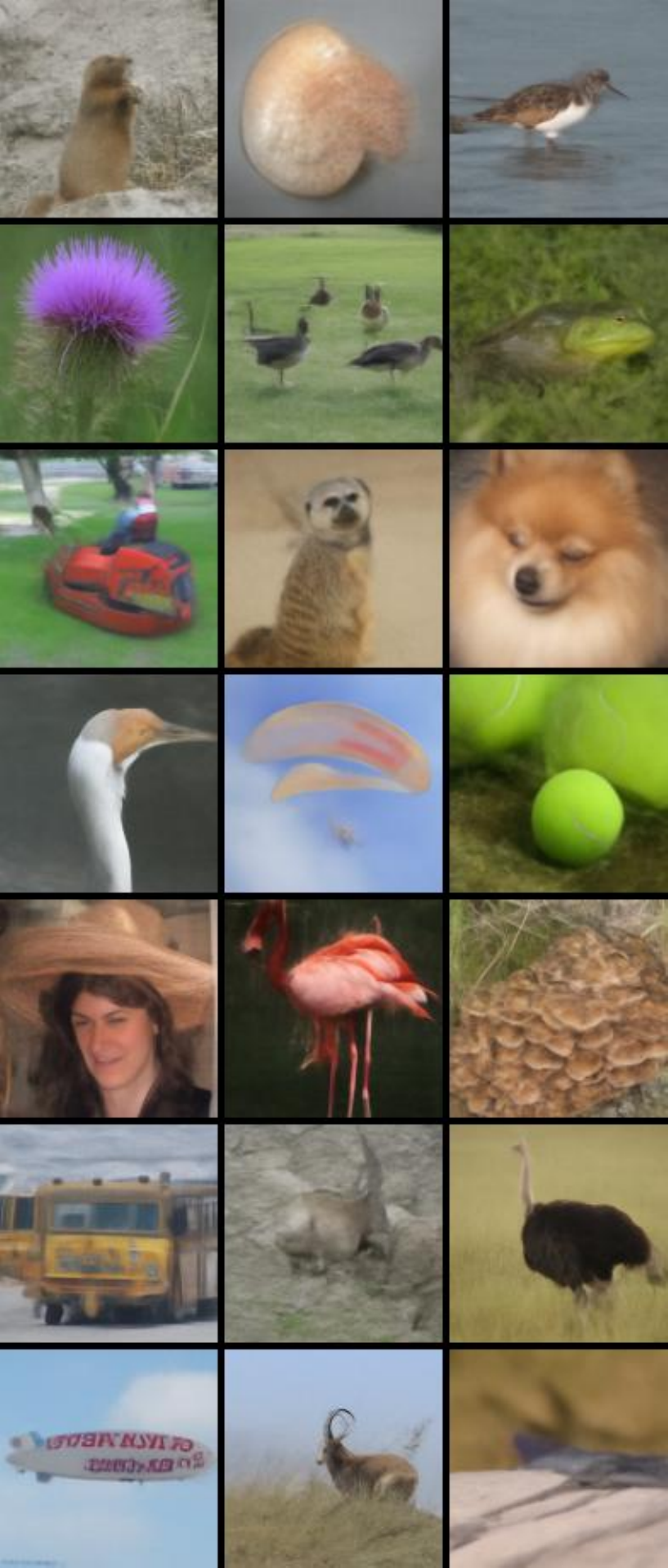} & \includegraphics[width=0.221\textwidth]{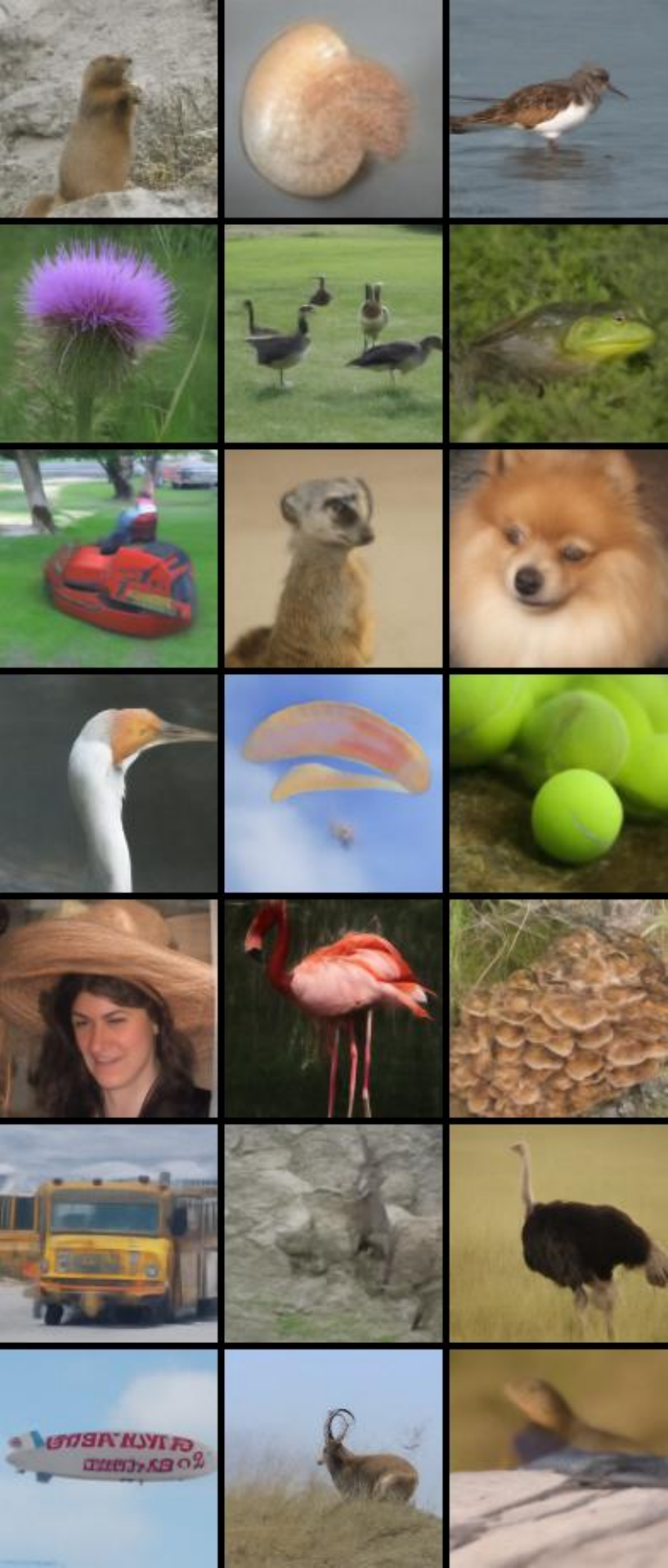} & \includegraphics[width=0.221\textwidth]{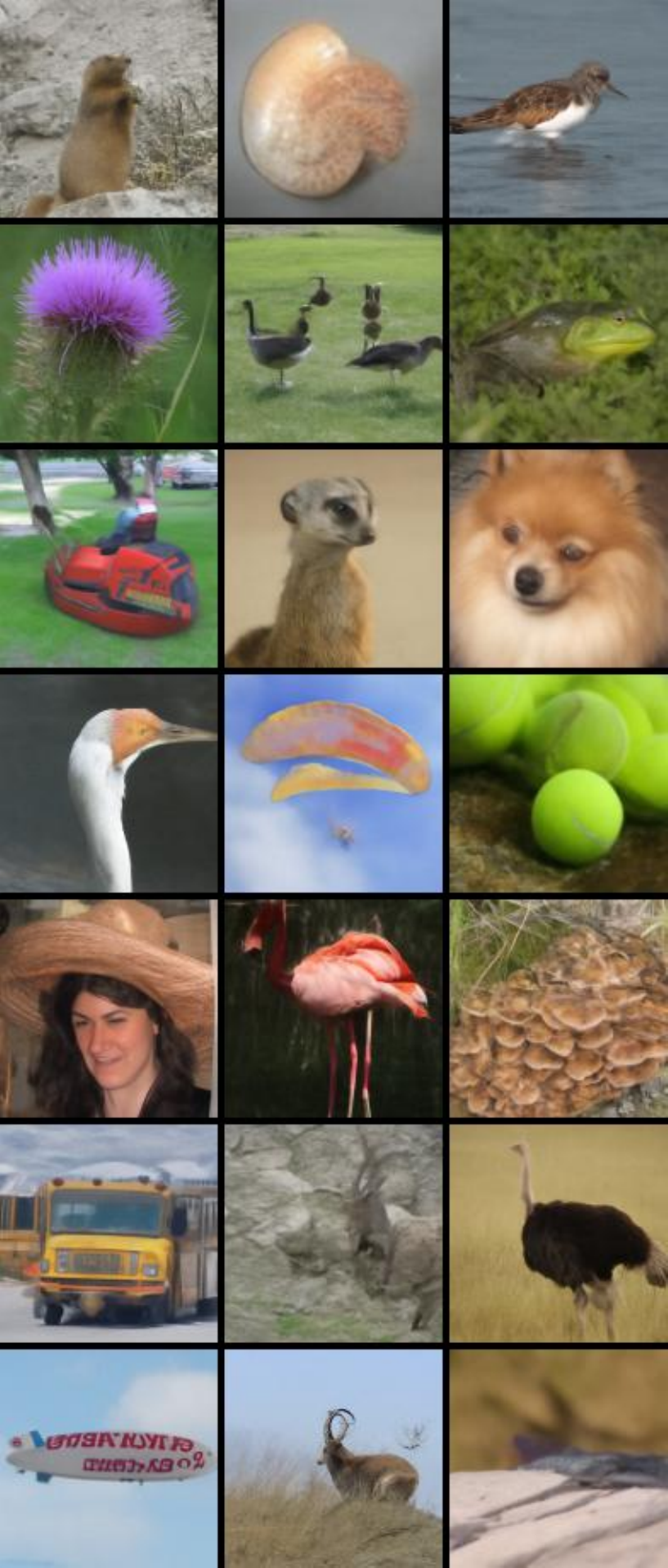} & \includegraphics[width=0.221\textwidth]{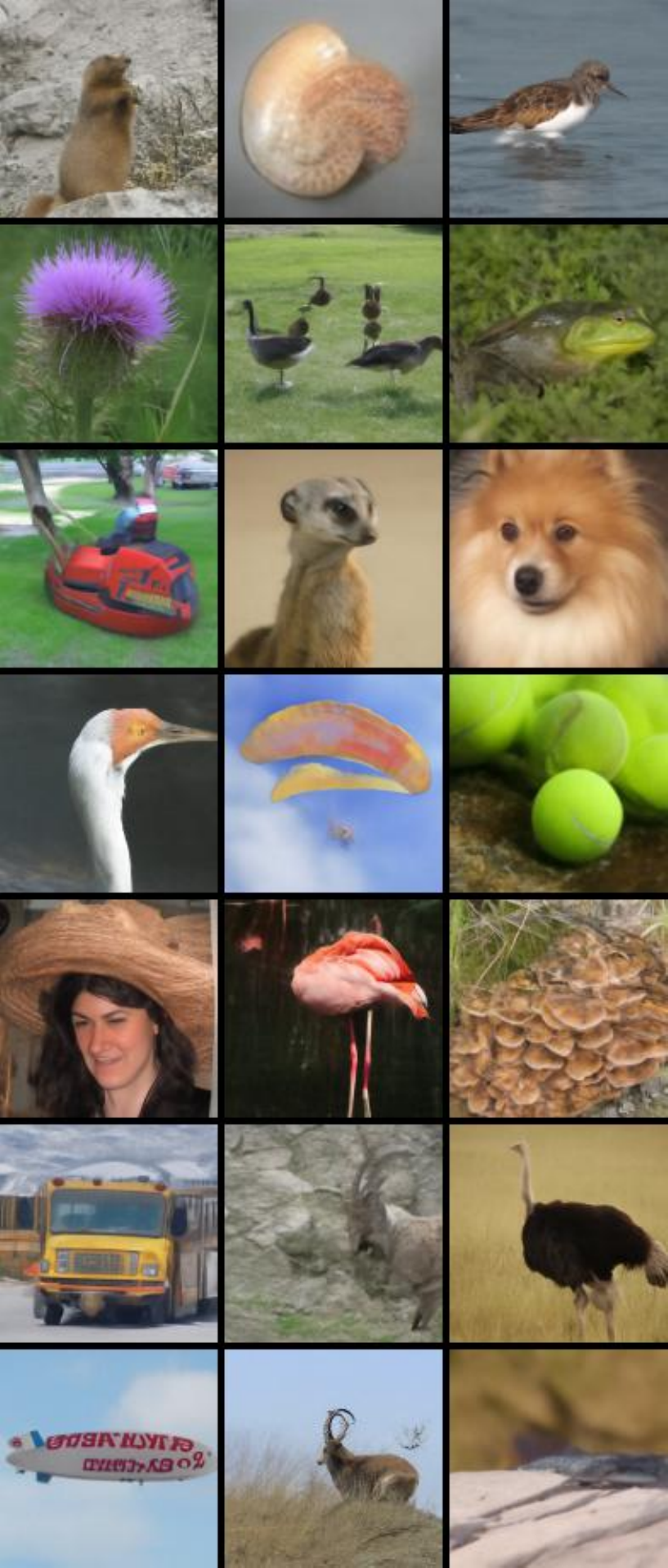}
\\
\multirow{-18.7}{*}{\parbox{0.8cm}{\centering DPM-Solver \cite{lu2022dpm}}}
& \includegraphics[width=0.221\textwidth]
{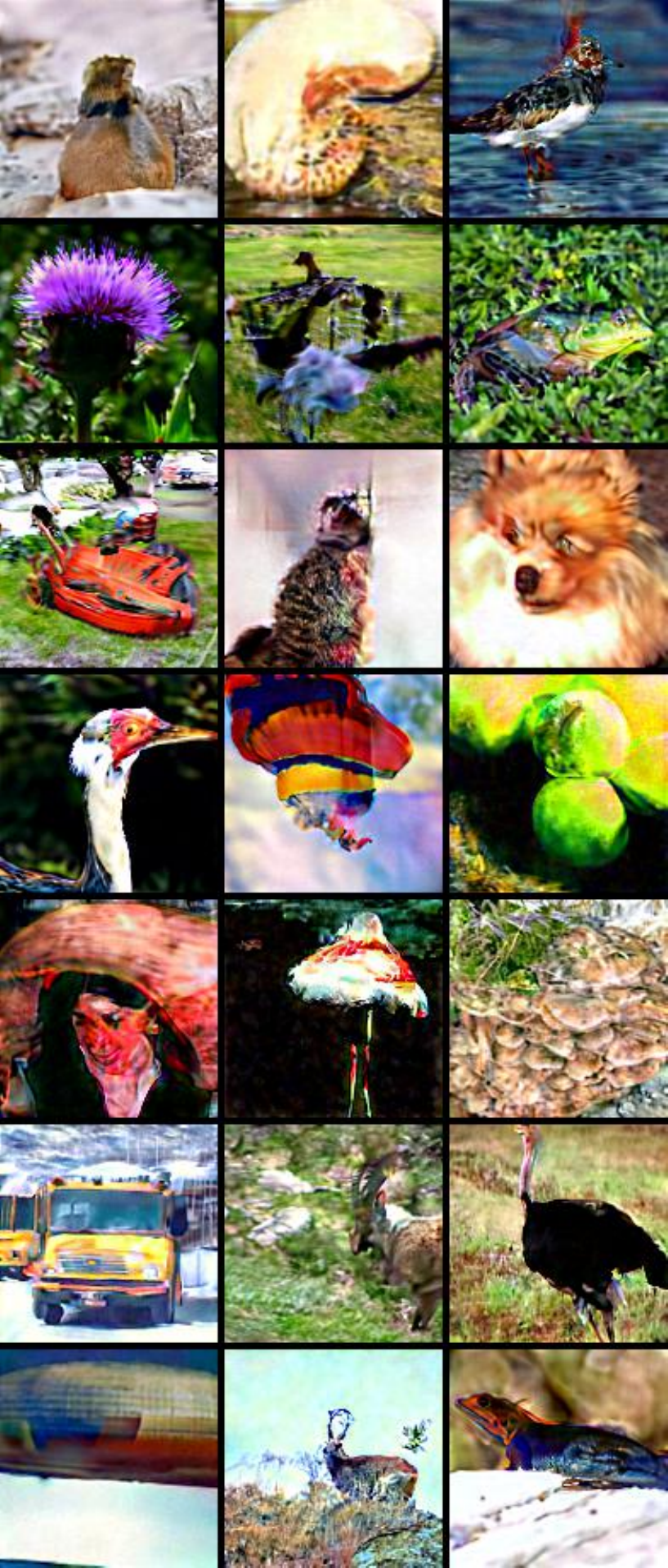} & \includegraphics[width=0.221\textwidth]{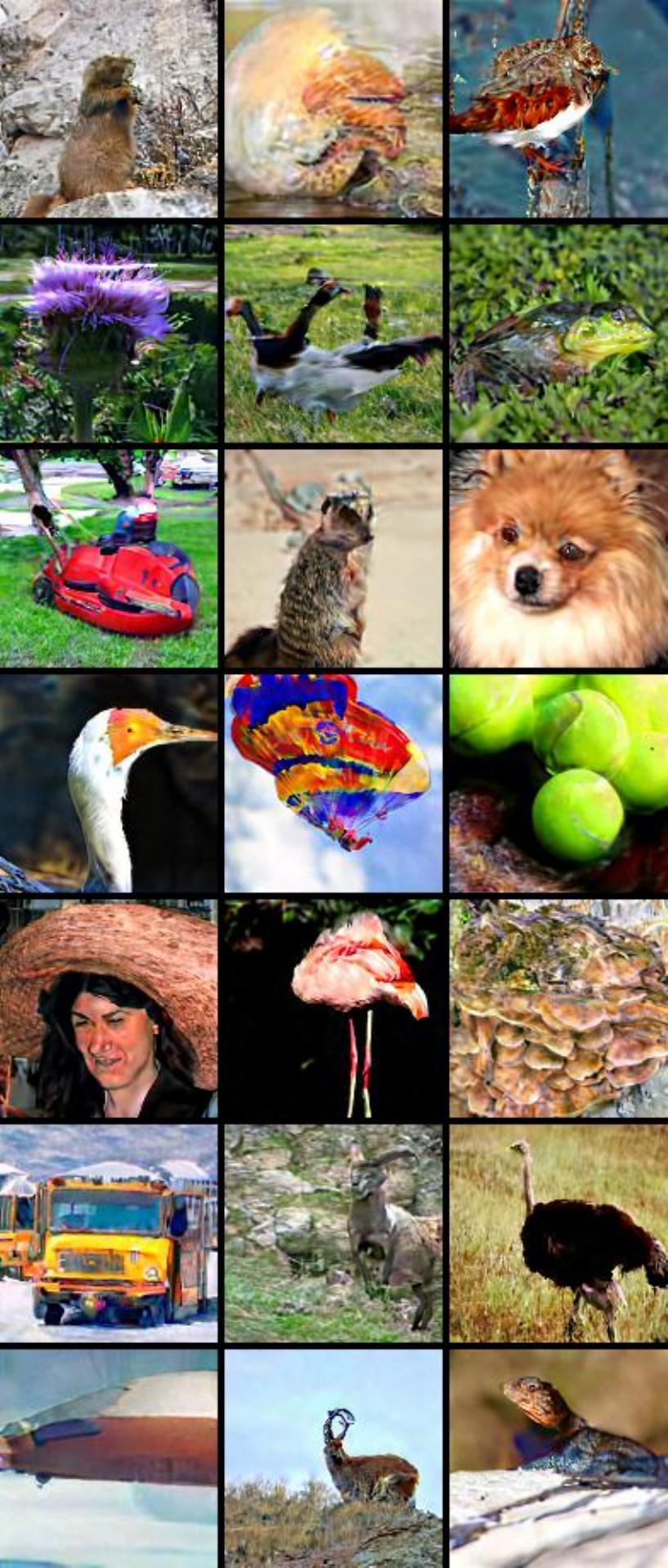} & \includegraphics[width=0.221\textwidth]{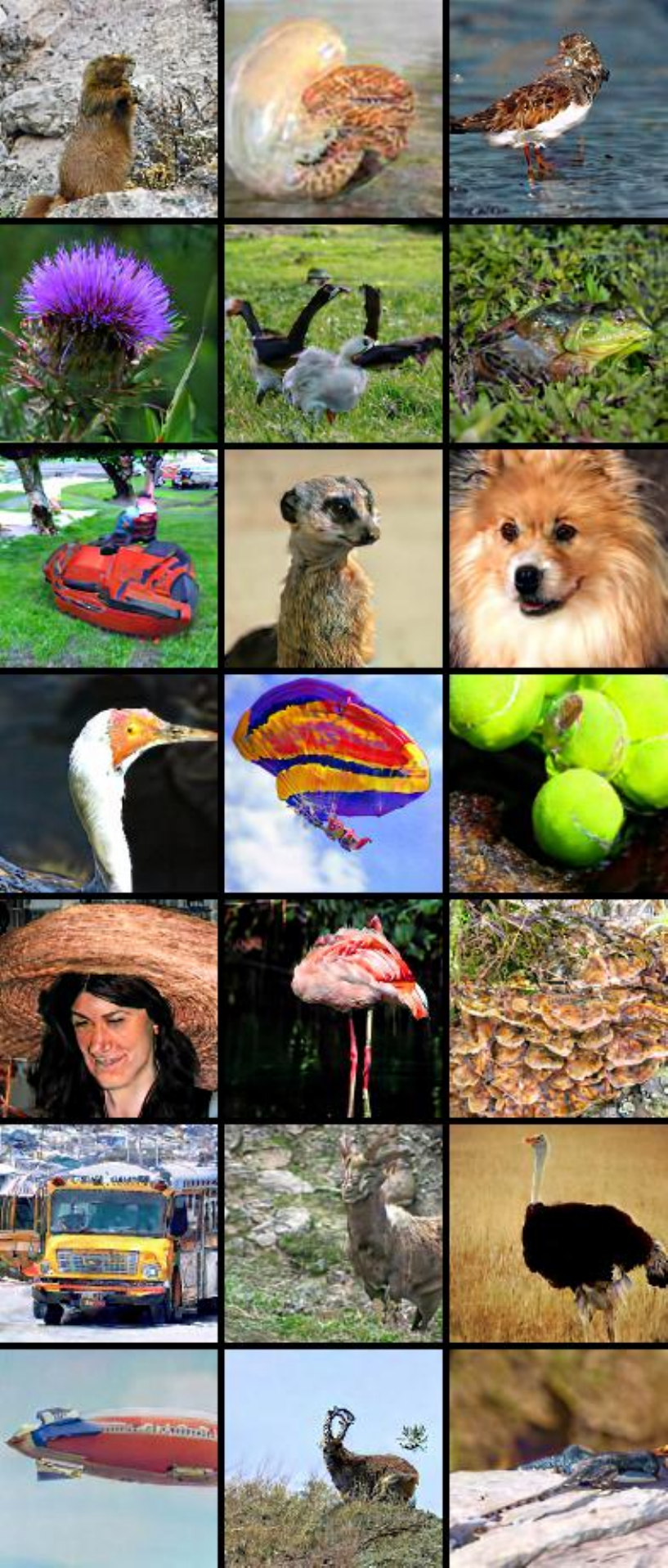} & \includegraphics[width=0.221\textwidth]{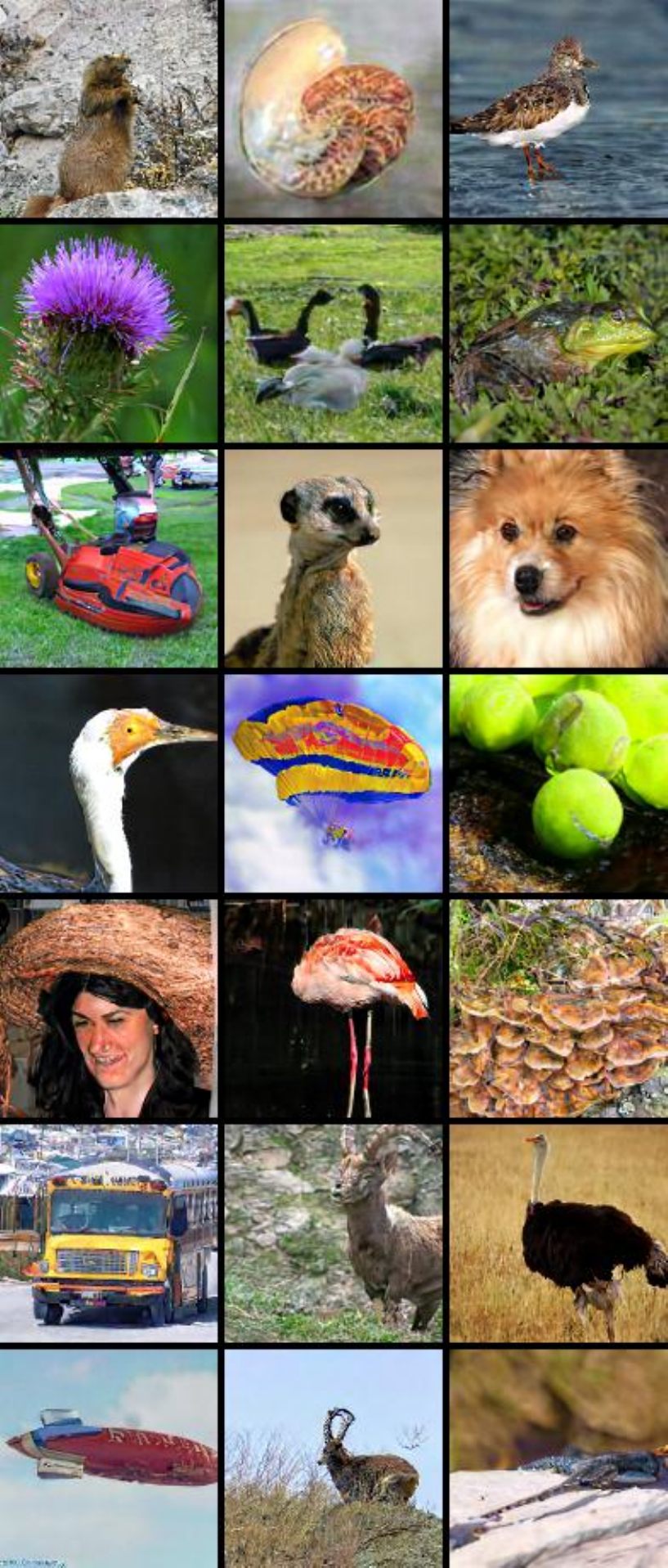}
\\
\multirow{-18.7}{*}{\parbox{0.8cm}{\centering SciRE-Solver (ours)}}
&\includegraphics[width=0.221\textwidth]{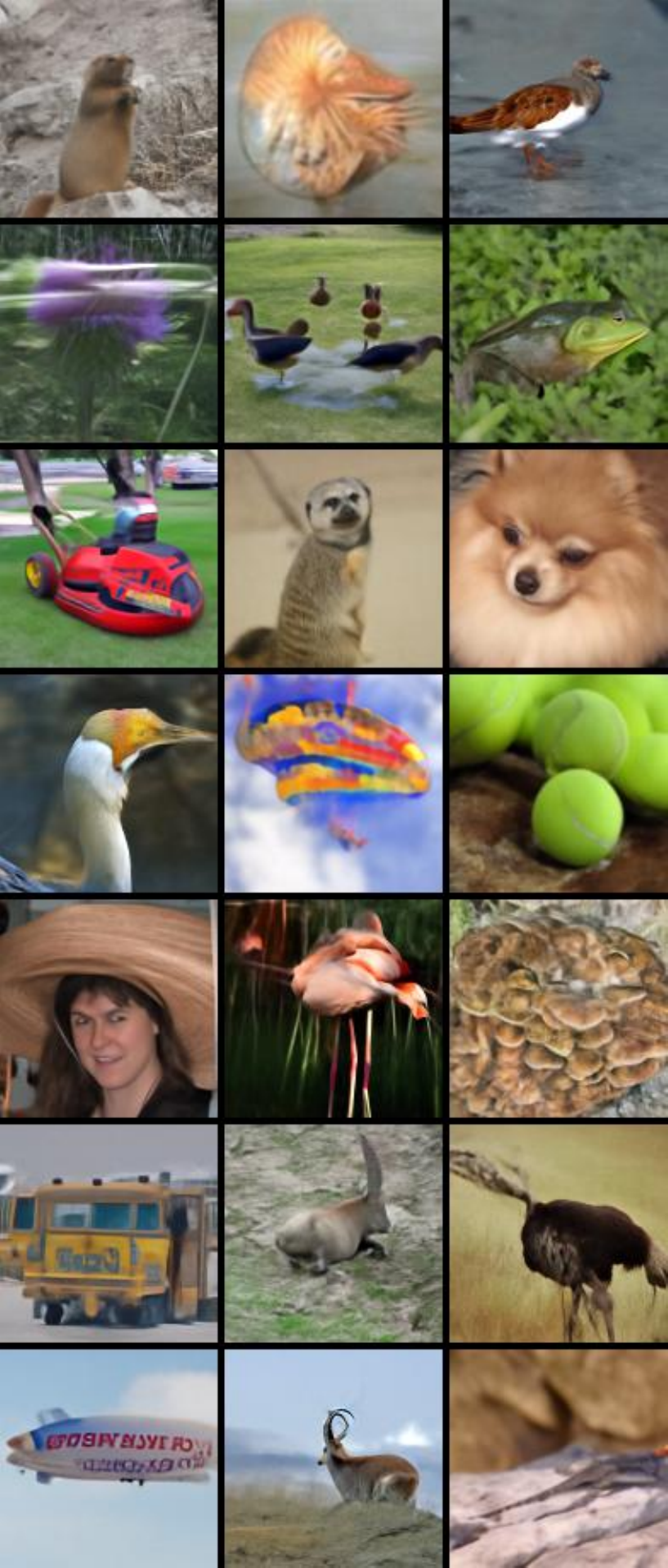} & \includegraphics[width=0.221\textwidth]{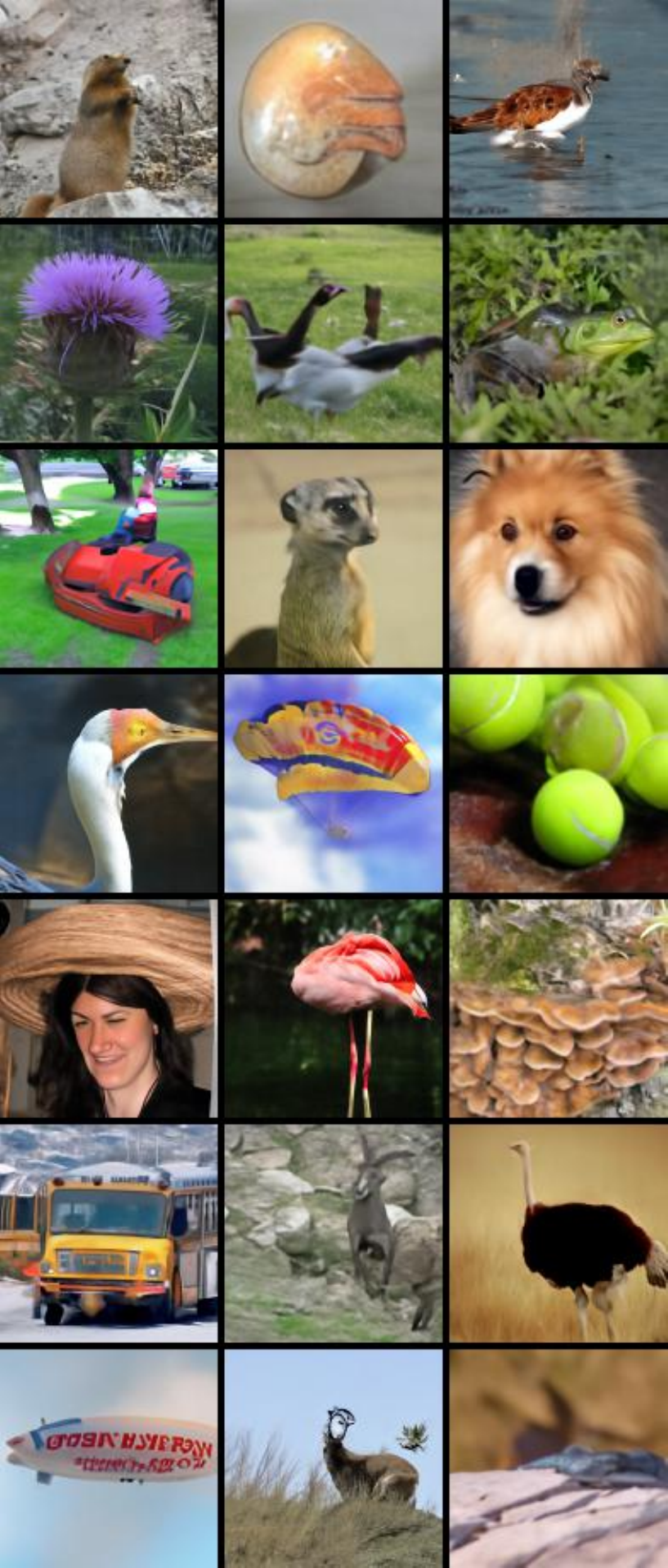} & \includegraphics[width=0.221\textwidth]{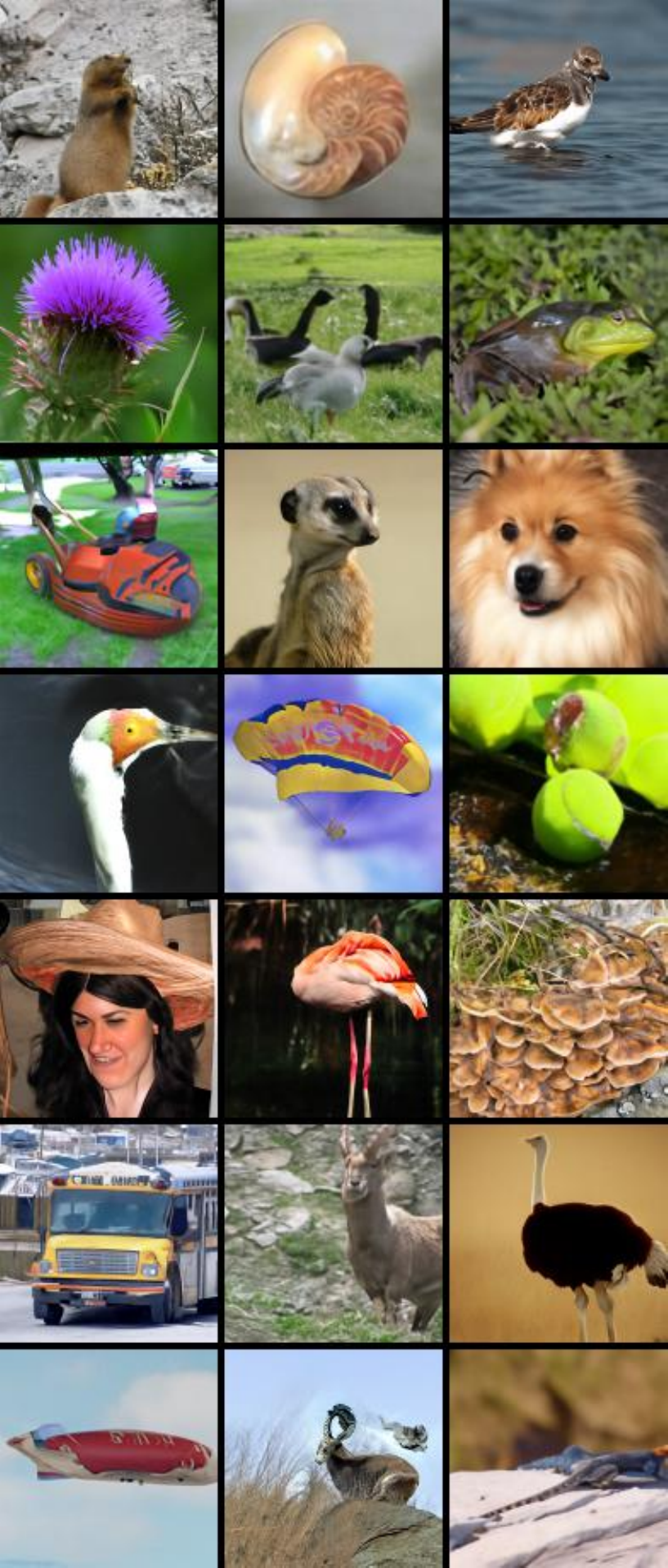} & \includegraphics[width=0.221\textwidth]{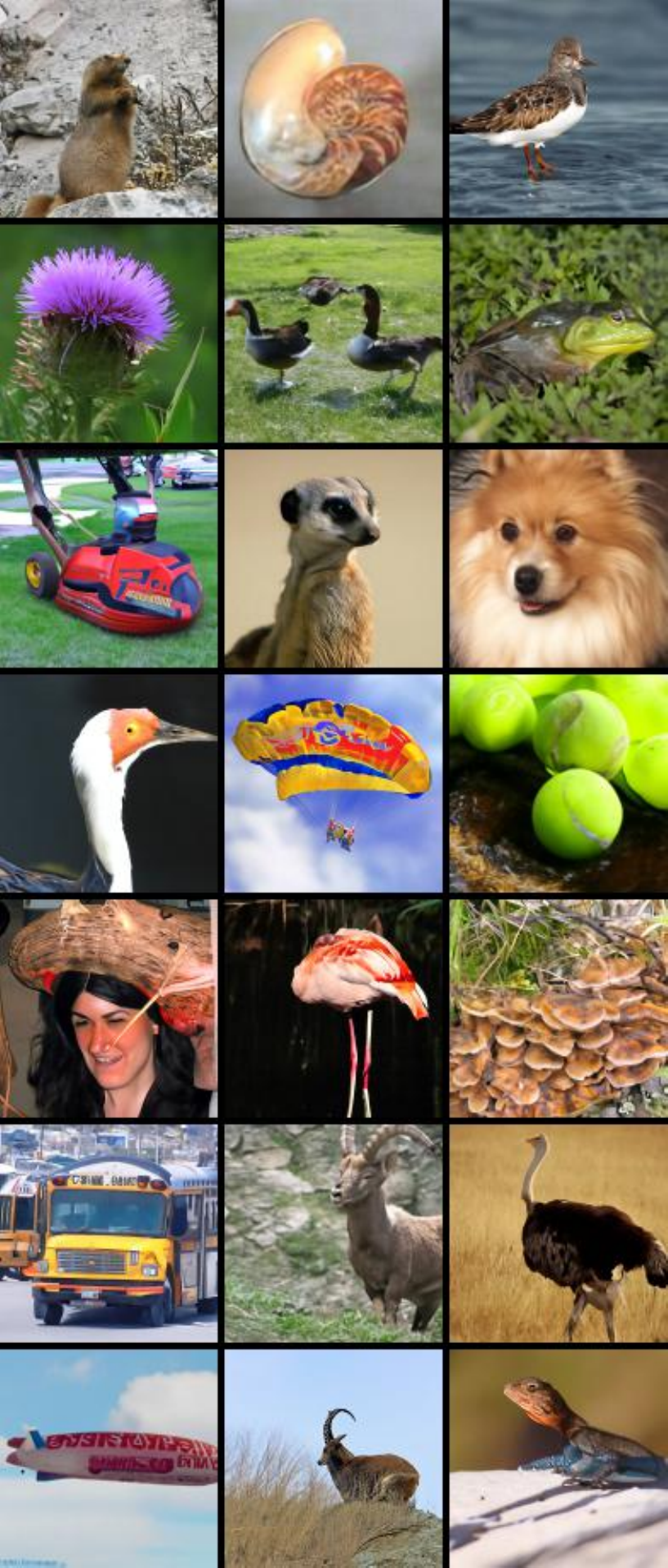} \\
\end{tabular}
 \caption{
Generated samples of the pre-trained DM \cite{dhariwal2021diffusion} on ImageNet 128$\times$128 (classifier scale: 1.25) using
6-12 sampling steps from different sampling methods with the same settings and codebase.
 }
\label{fig:imagenet2128samecomparison}
\end{figure}

\begin{figure}[ht]
\vspace{-0.45cm}
\centering
\begin{tabular}{m{0.65cm}p{2.7cm}p{2.7cm}p{2.7cm}p{2.7cm}}
   ~~ &~~~~~~~~~~NFE=$10$& ~~~~~~~~~~NFE=$15$  &~~~~~~~~~~NFE=$20$ &~~~~~~~~~~NFE=$50$ \\
\multirow{-18.7}{*}{\parbox{0.8cm}{\centering DDIM \cite{song2021denoising}}}
& \includegraphics[width=0.221\textwidth]{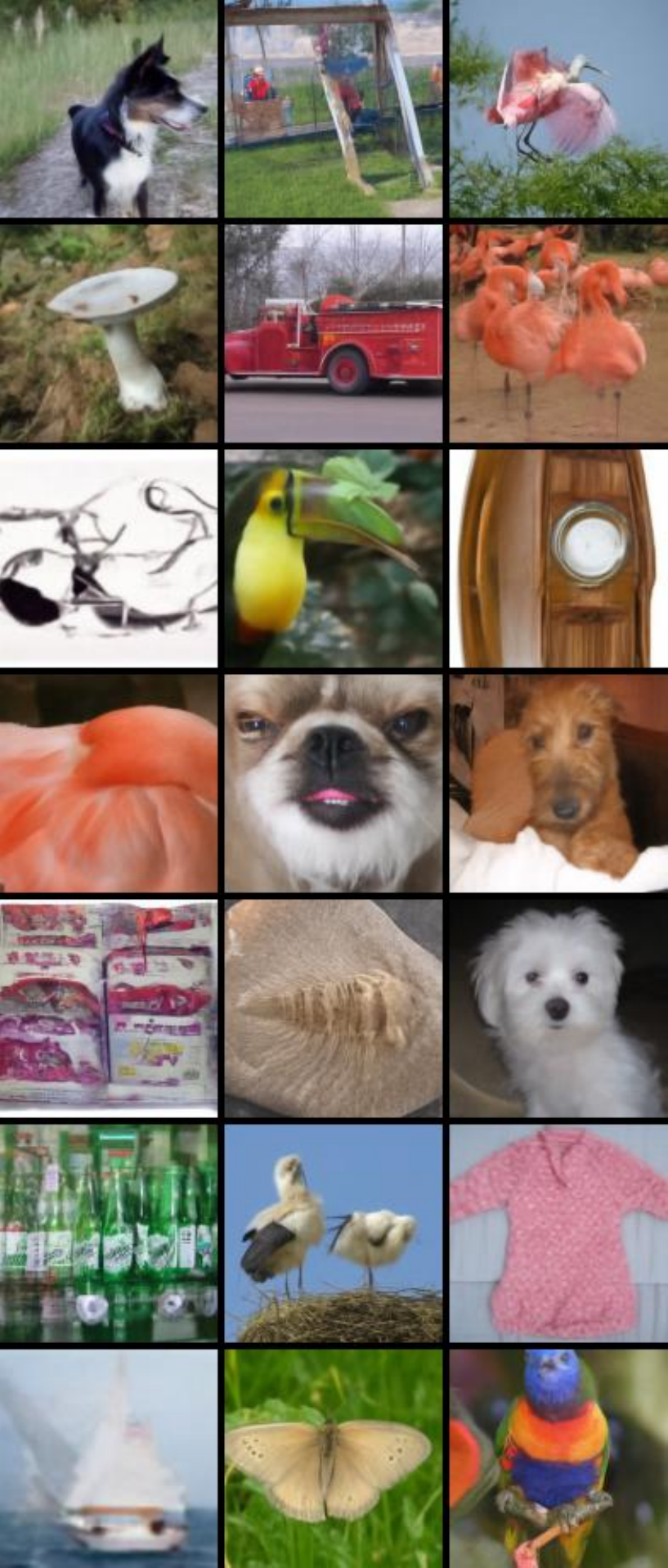} & \includegraphics[width=0.221\textwidth]{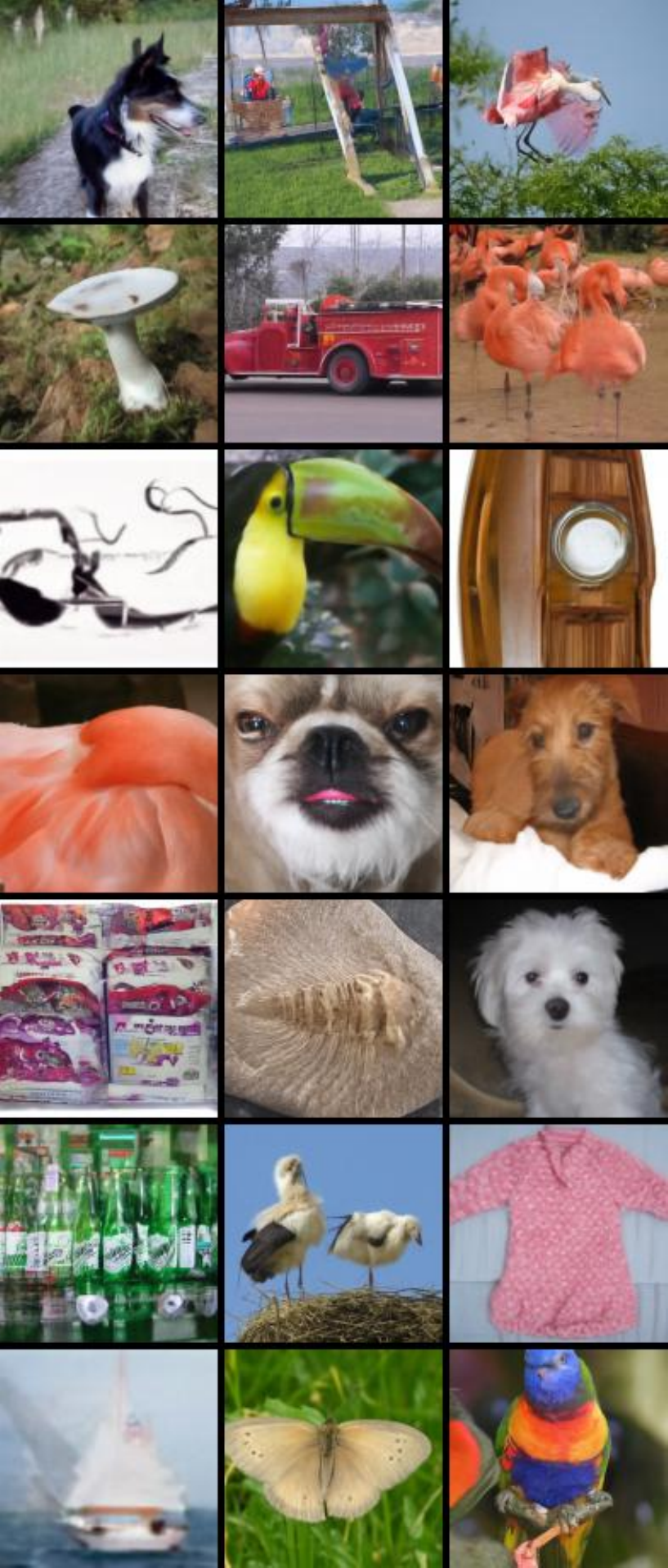} & \includegraphics[width=0.221\textwidth]{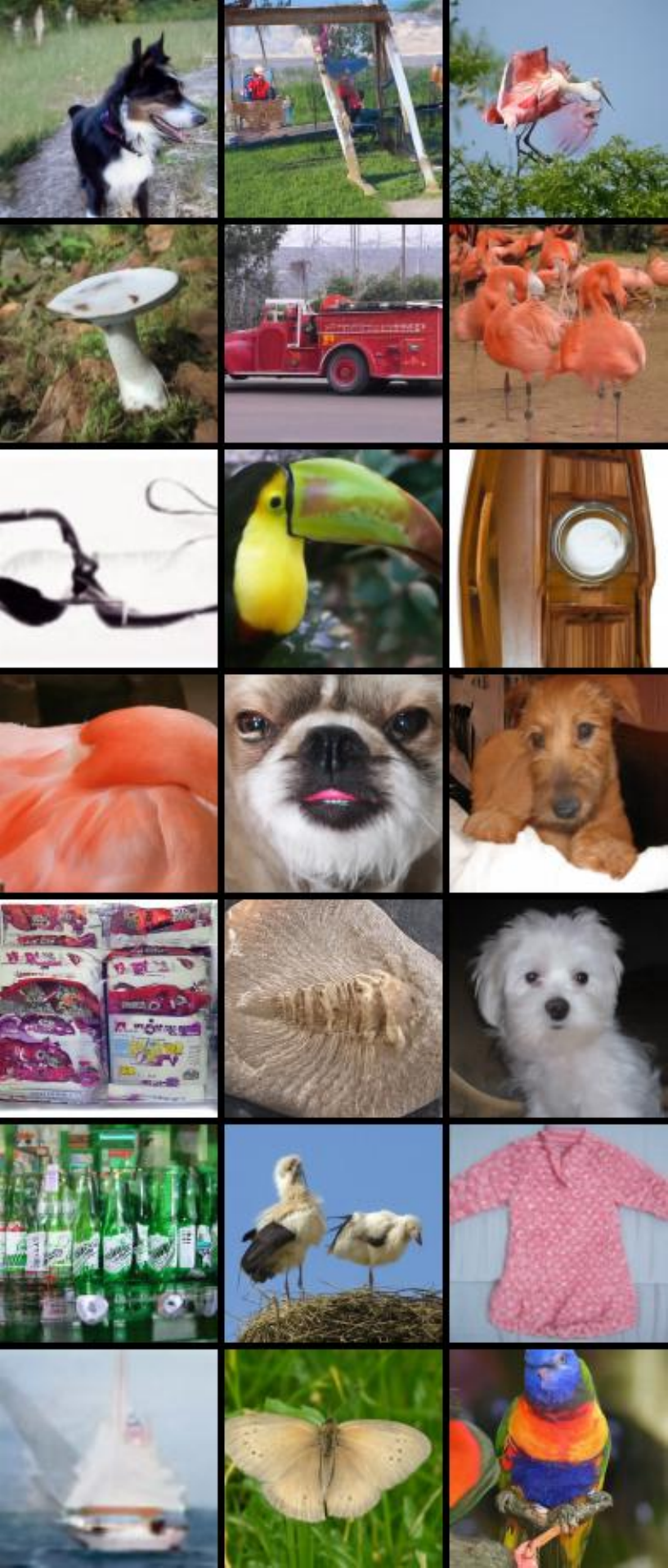} & \includegraphics[width=0.221\textwidth]{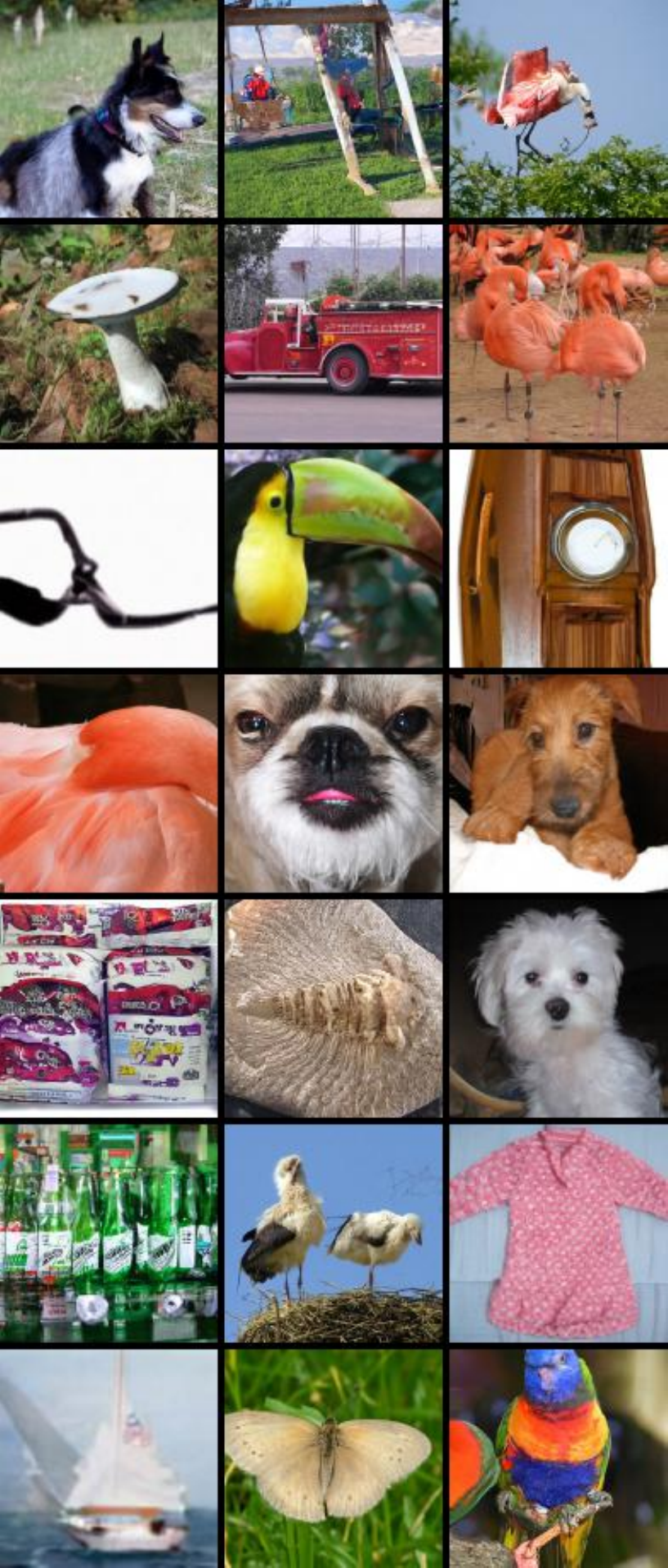}
\\
\multirow{-18.7}{*}{\parbox{0.8cm}{\centering DPM-Solver \cite{lu2022dpm}}}
& \includegraphics[width=0.221\textwidth]{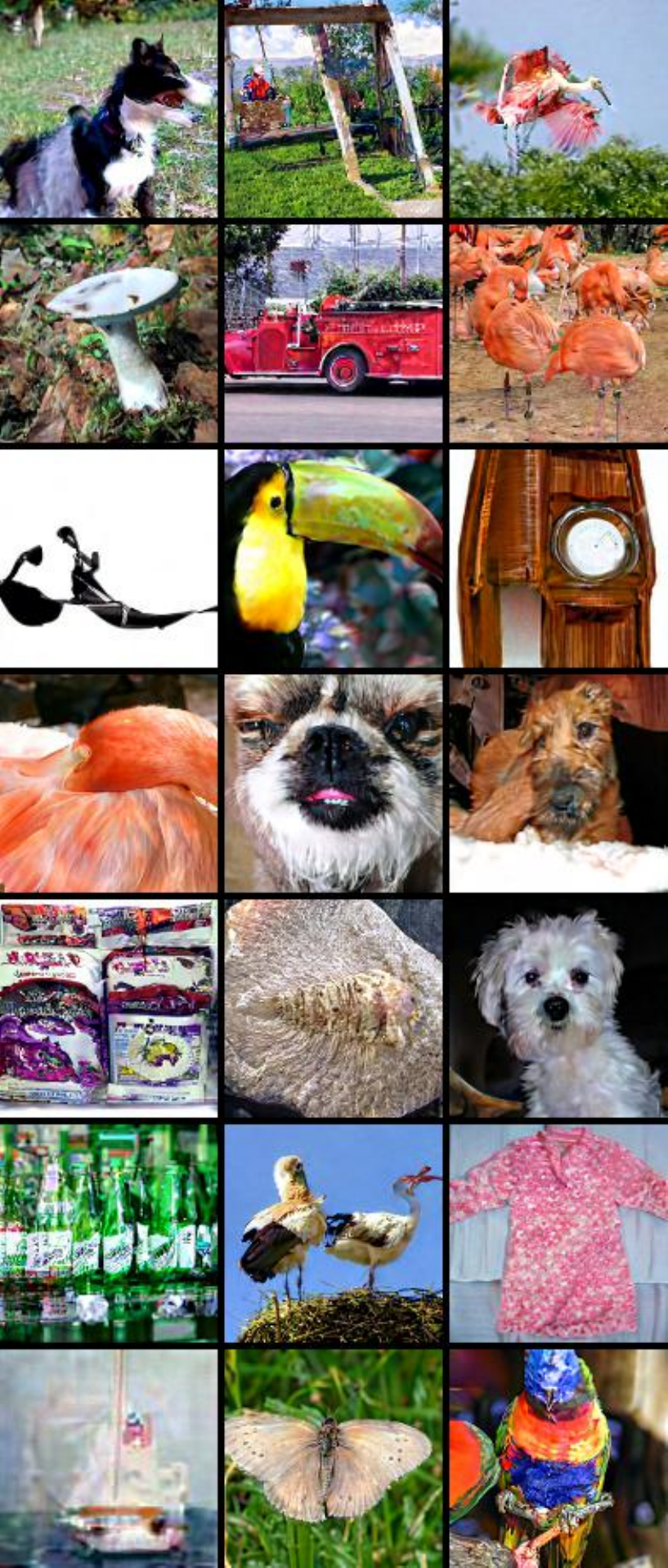} & \includegraphics[width=0.221\textwidth]{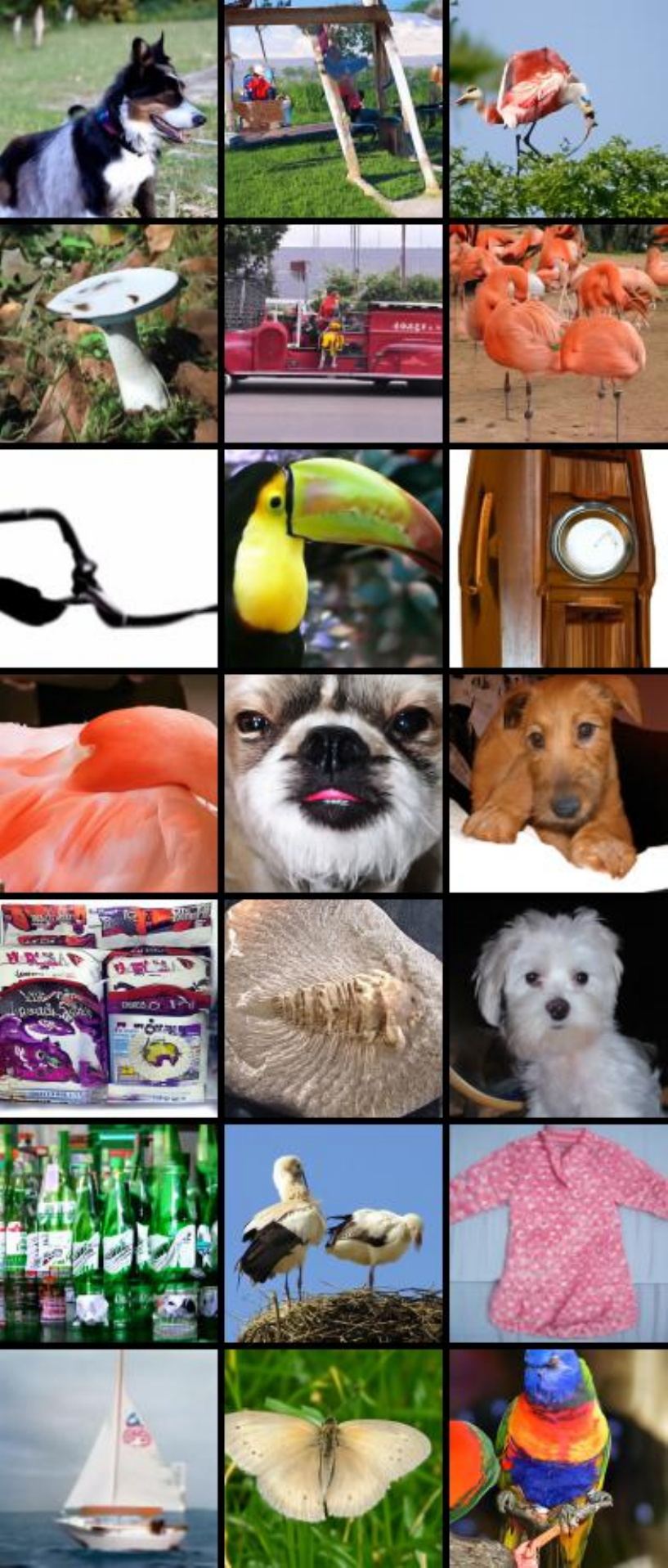} & \includegraphics[width=0.221\textwidth]{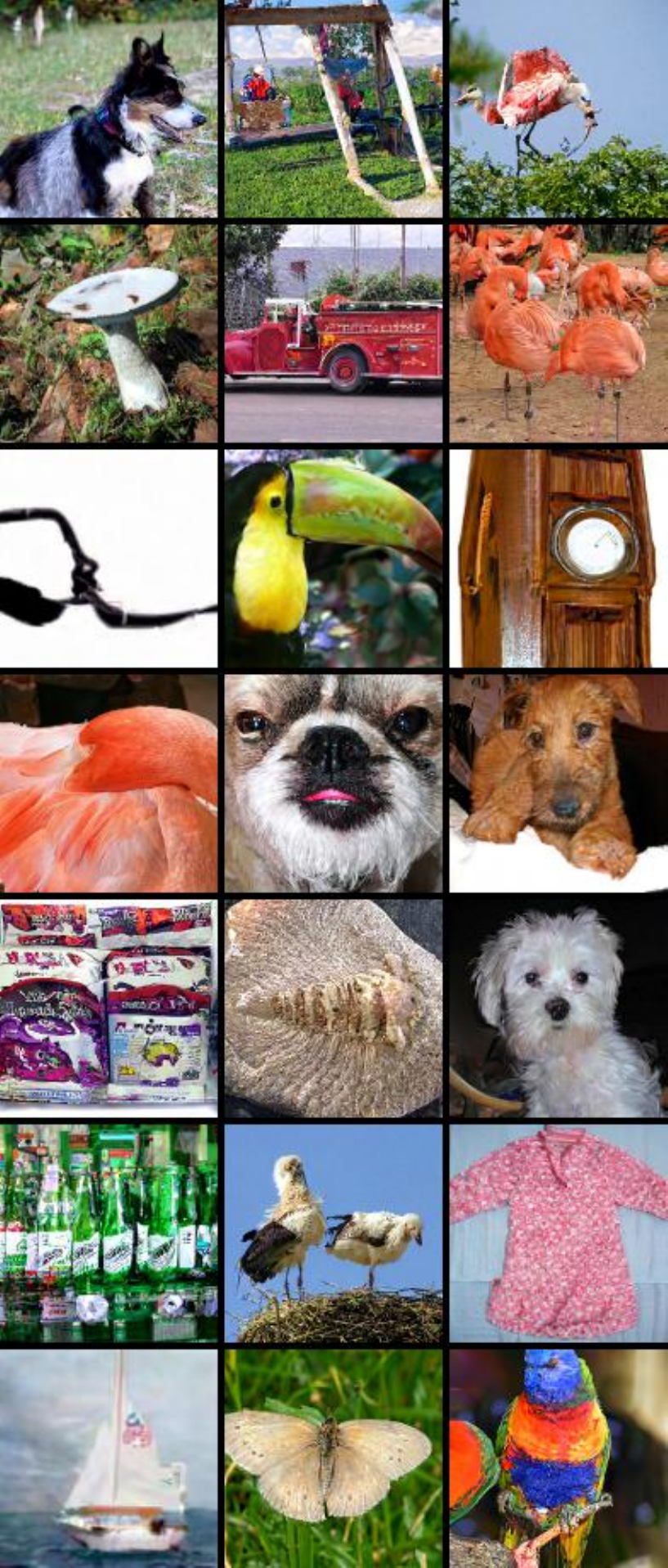} & \includegraphics[width=0.221\textwidth]{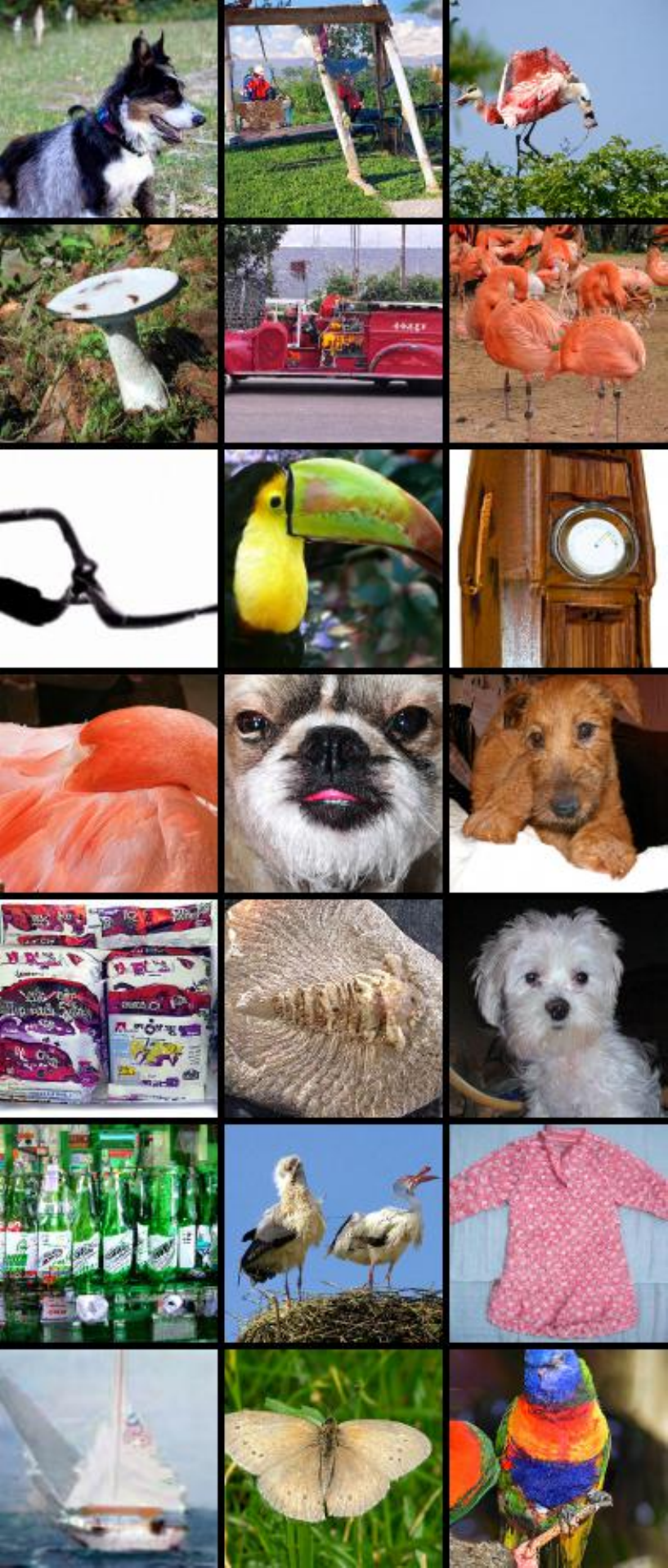}
\\
\multirow{-18.7}{*}{\parbox{0.8cm}{\centering SciRE-Solver (ours)}}
&\includegraphics[width=0.221\textwidth]{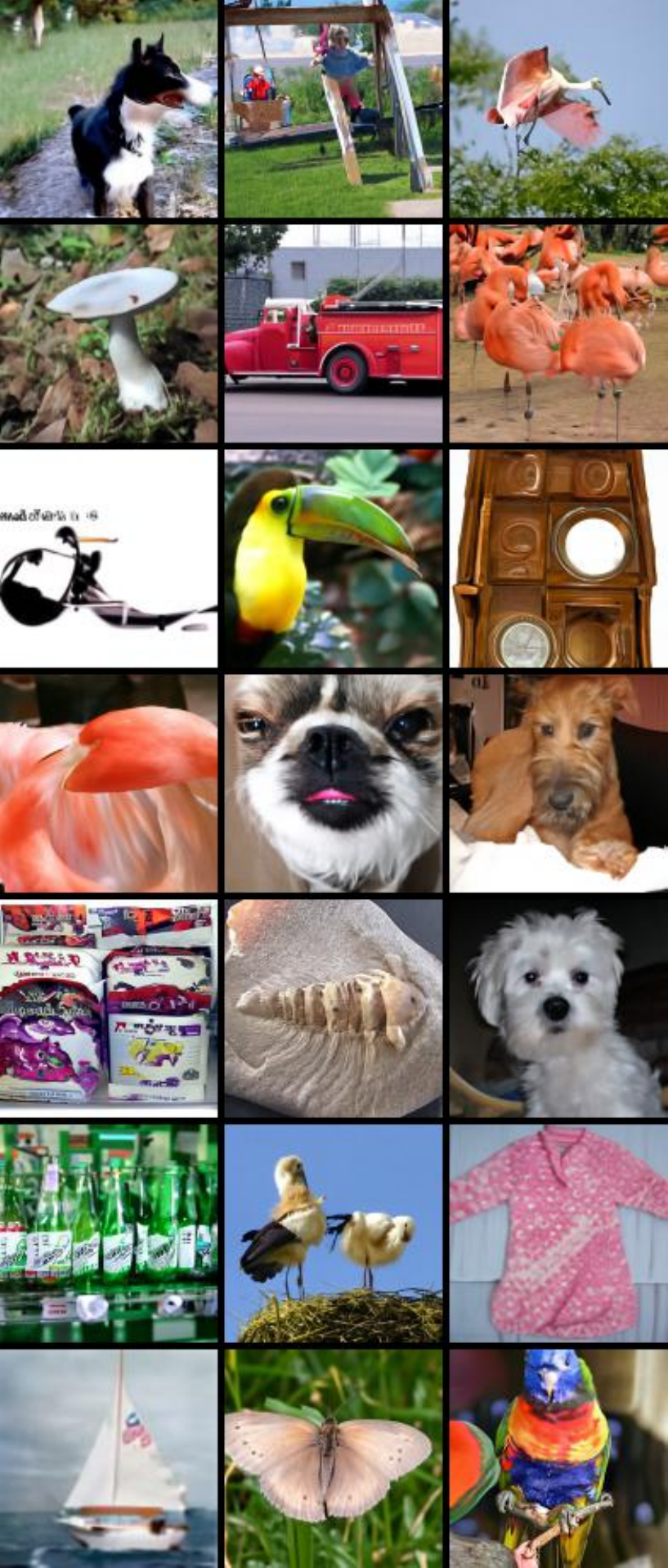} & \includegraphics[width=0.221\textwidth]{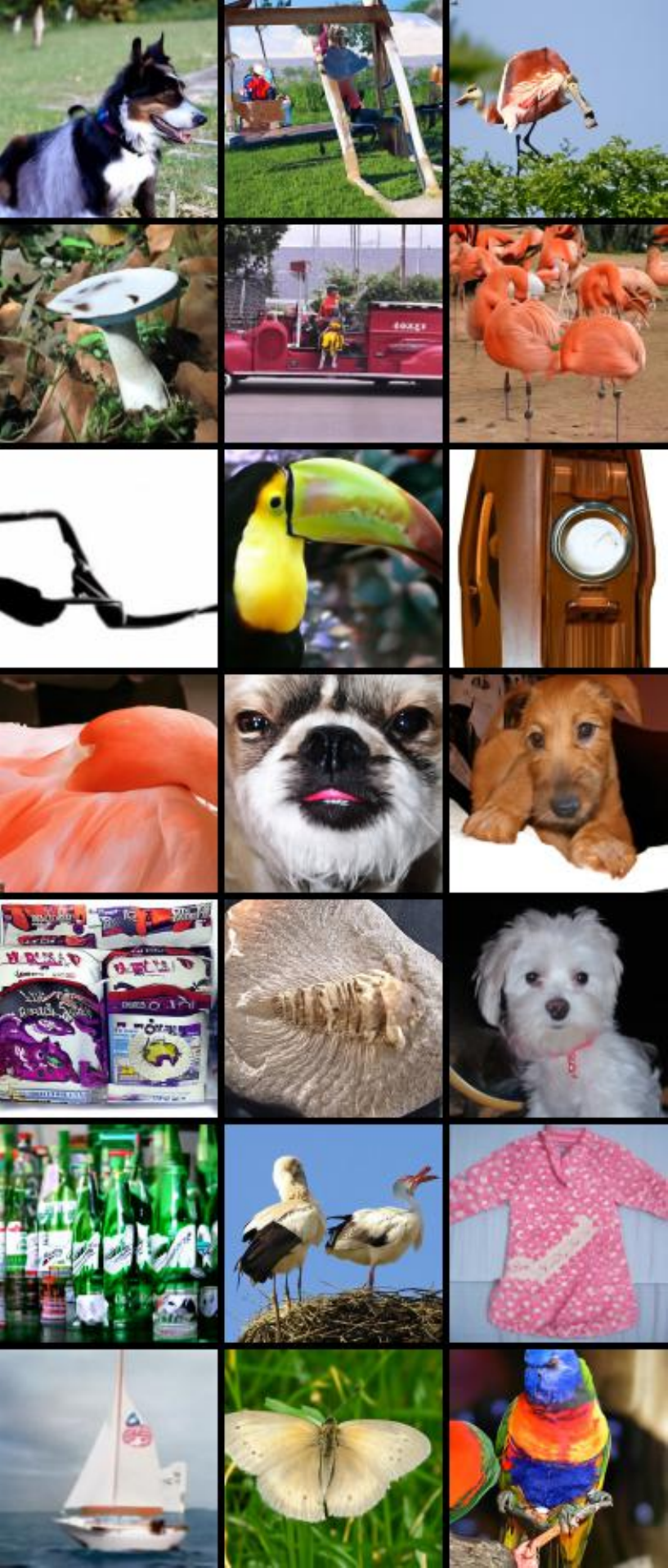} & \includegraphics[width=0.221\textwidth]{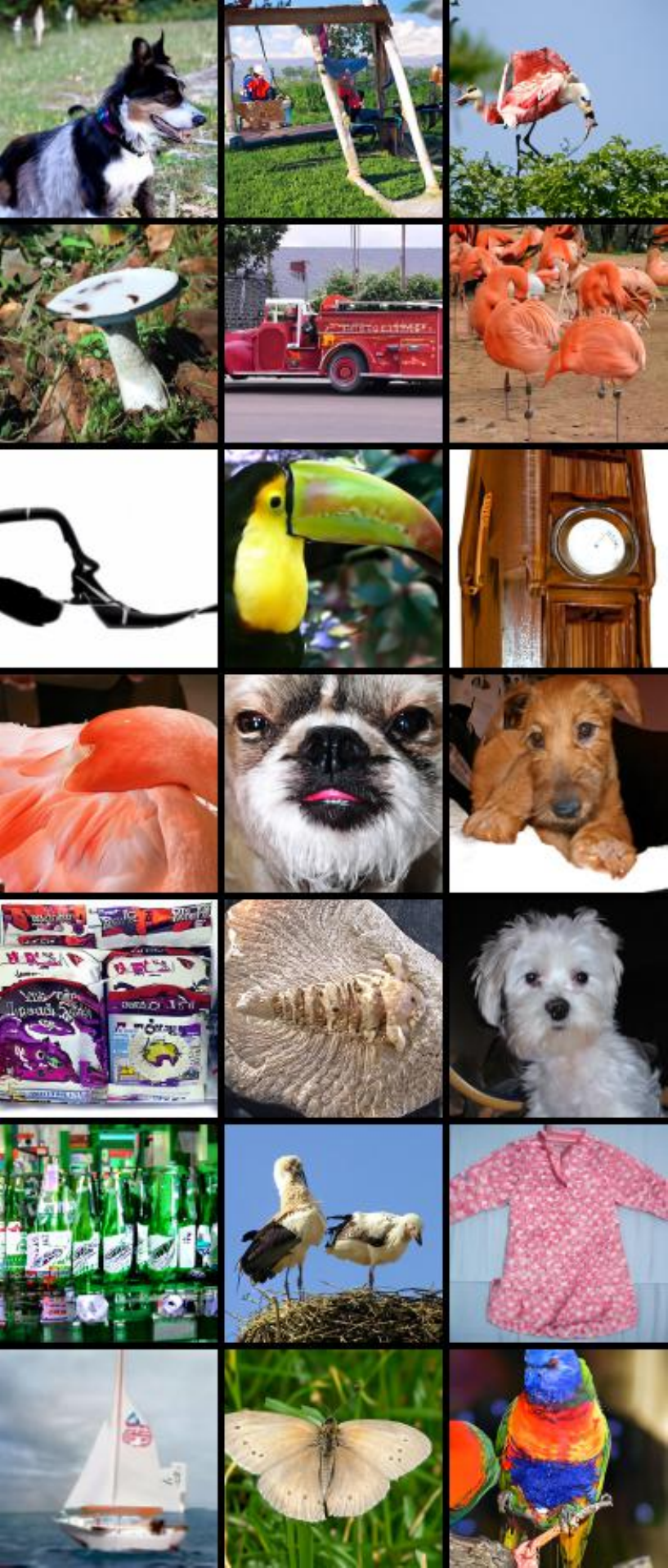} & \includegraphics[width=0.221\textwidth]{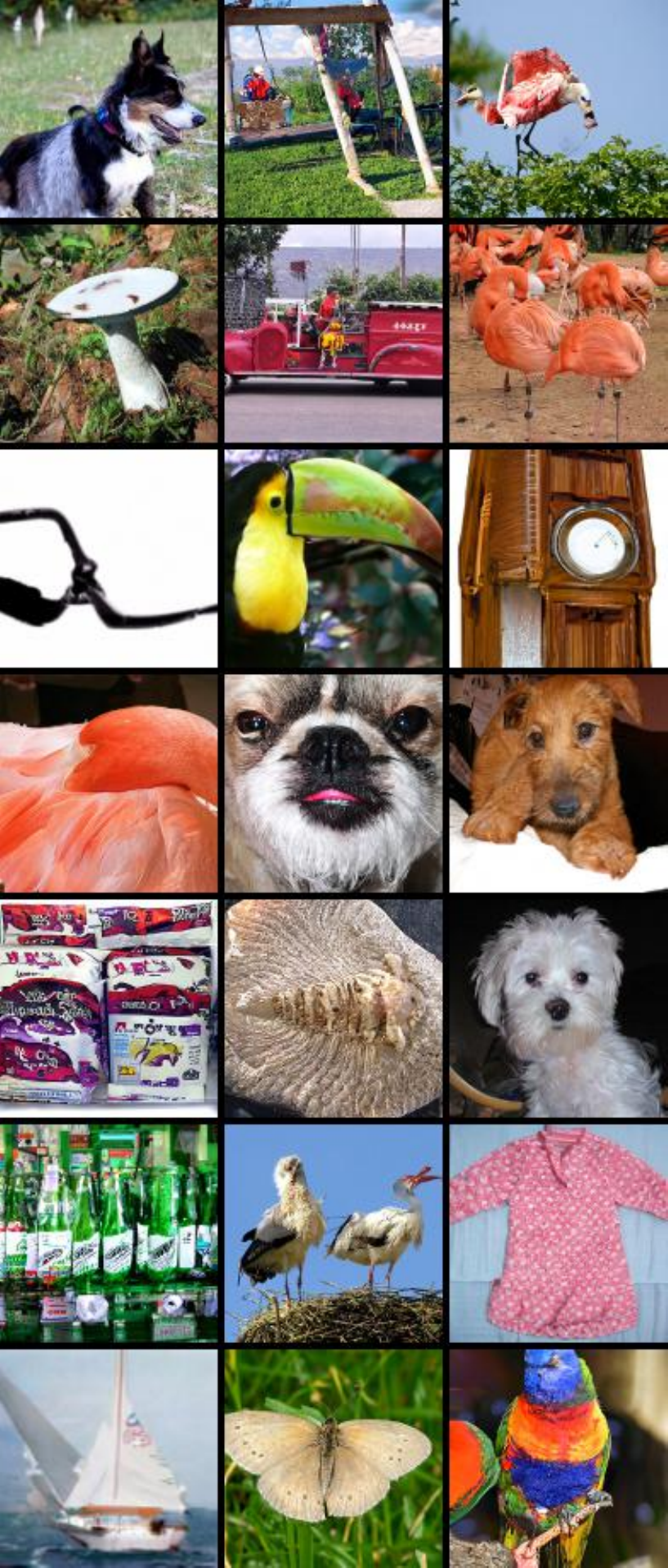} \\
\end{tabular}
 \caption{
Generated samples of the pre-trained DM \cite{dhariwal2021diffusion} on ImageNet 128$\times$128 (classifier scale: 1.25) using
10-50 sampling steps from different sampling methods with the same settings and codebase.
 }
\label{fig:imagenet128samecomparison}
\end{figure}

\begin{figure}[ht]
\vspace{-0.45cm}
\centering
\begin{tabular}{m{0.65cm}p{2.7cm}p{2.7cm}p{2.7cm}p{2.7cm}}
   ~~ &~~~~~~~~~~NFE=$6$& ~~~~~~~~~~NFE=$12$  &~~~~~~~~~~NFE=$24$ &~~~~~~~~~~NFE=$30$ \\
\multirow{-18.7}{*}{\parbox{0.8cm}{\centering DDIM \cite{song2021denoising}}}
& \includegraphics[width=0.221\textwidth]{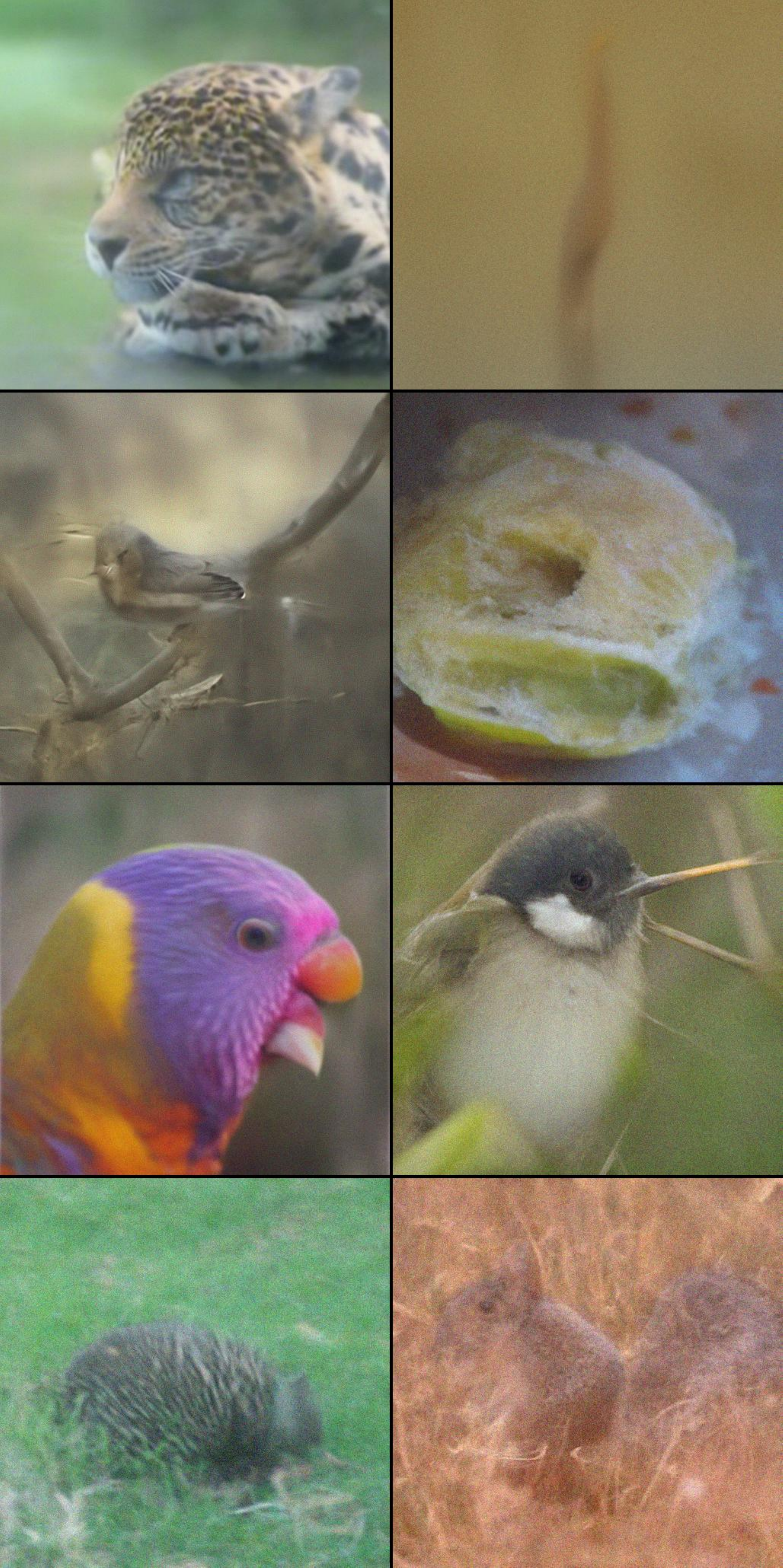} & \includegraphics[width=0.221\textwidth]{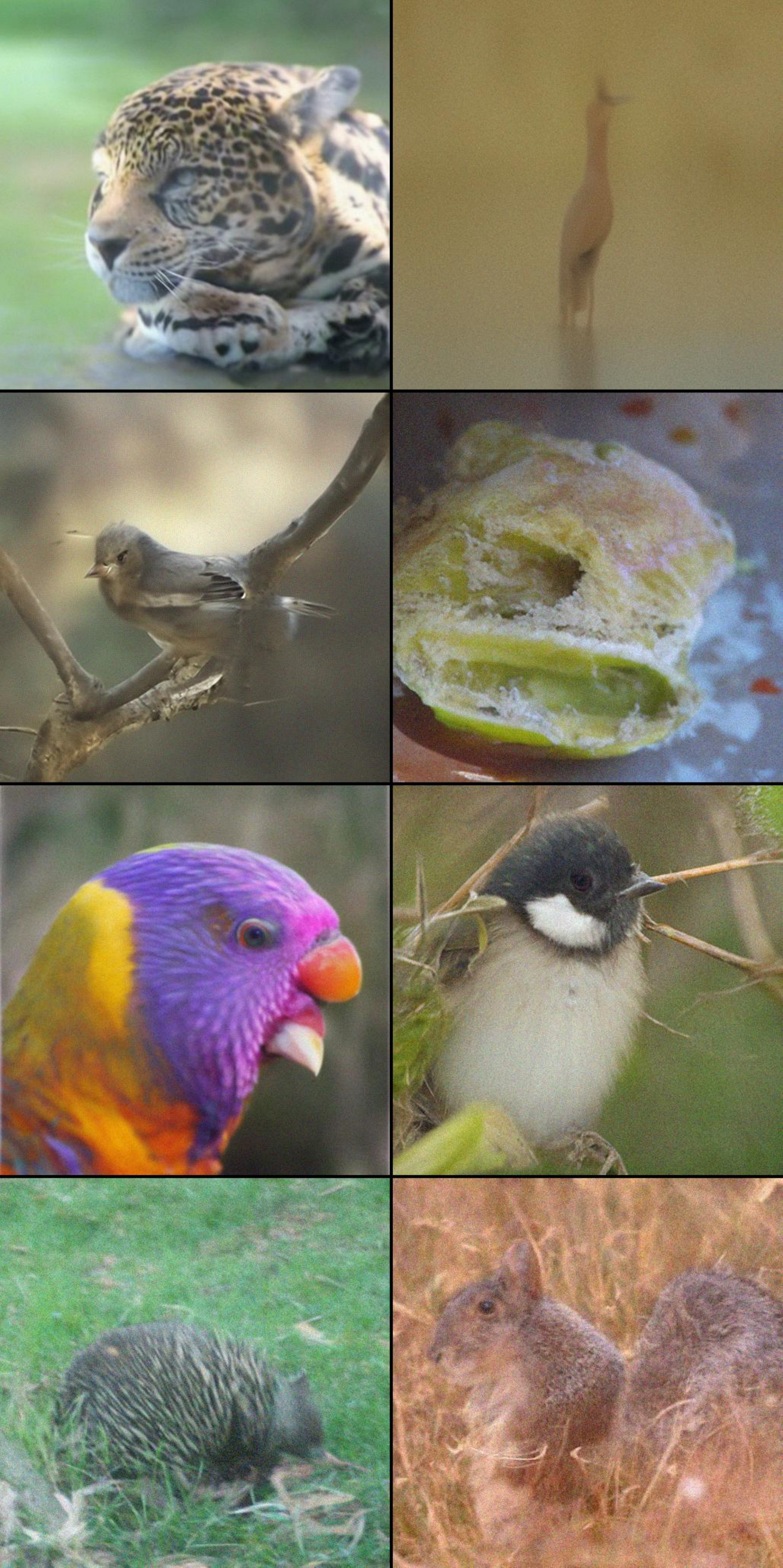} & \includegraphics[width=0.221\textwidth]{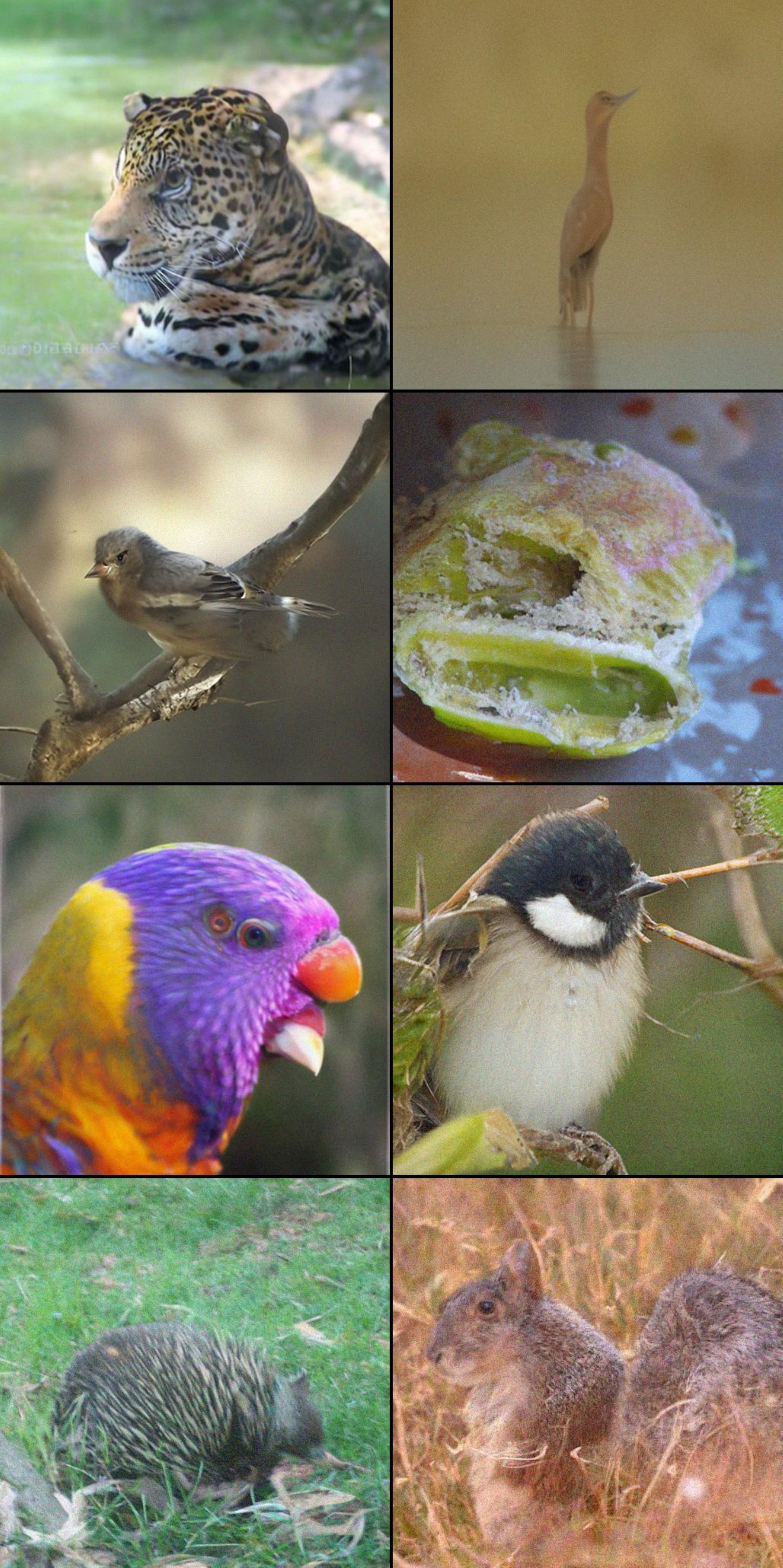} & \includegraphics[width=0.221\textwidth]{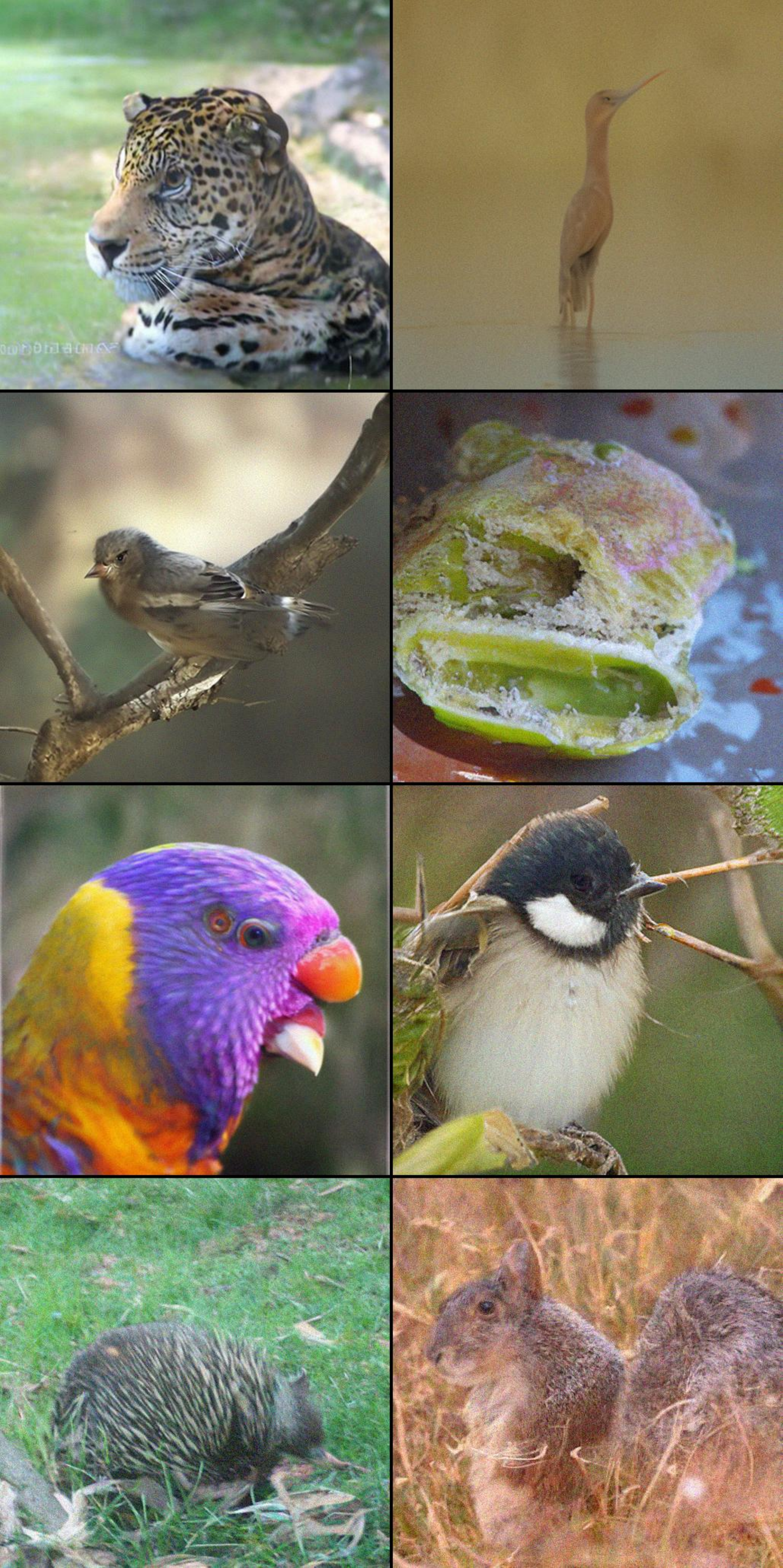}
\\
\multirow{-18.7}{*}{\parbox{0.8cm}{\centering DPM-Solver \cite{lu2022dpm}}}
& \includegraphics[width=0.221\textwidth]{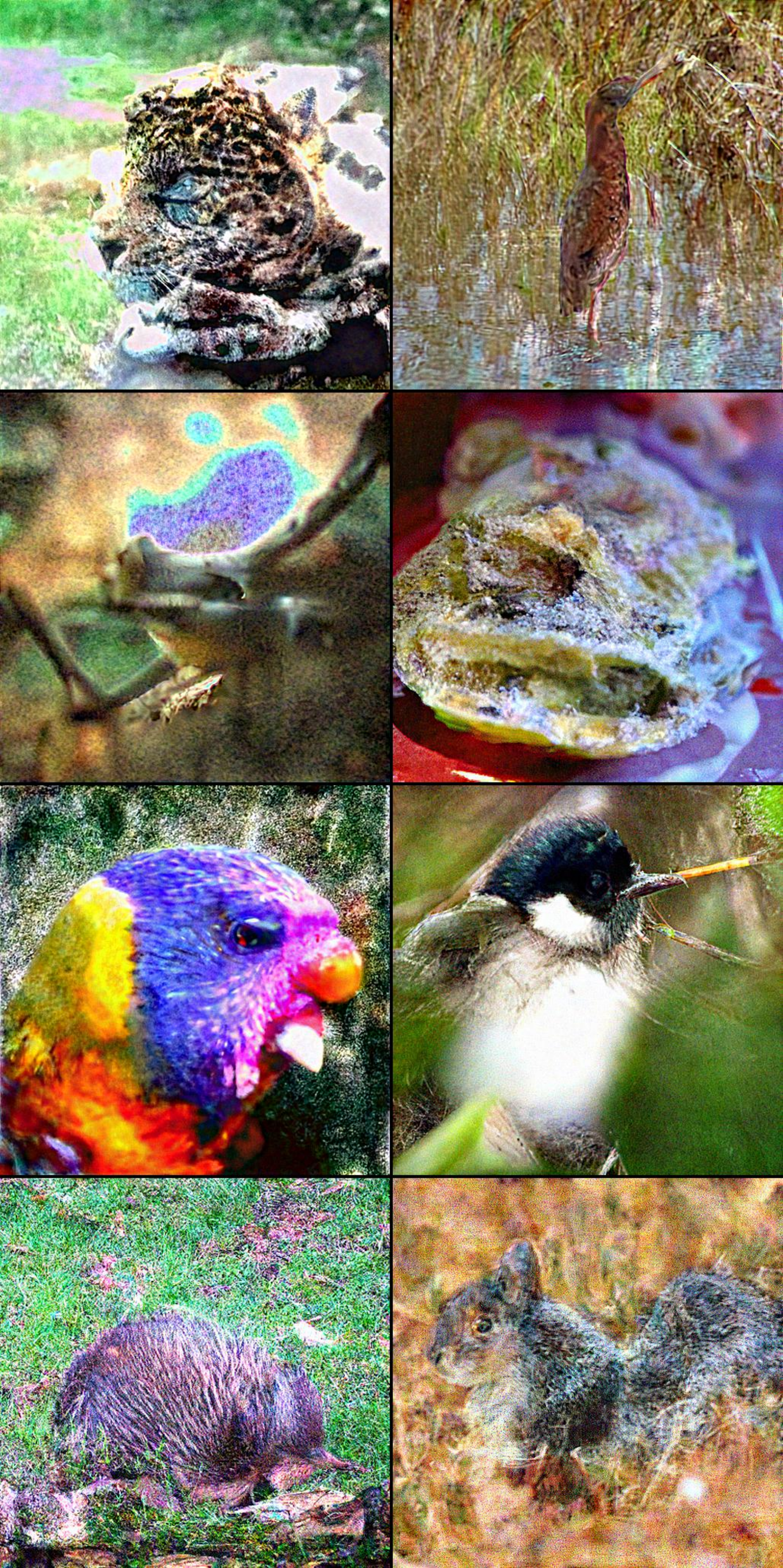} & \includegraphics[width=0.221\textwidth]{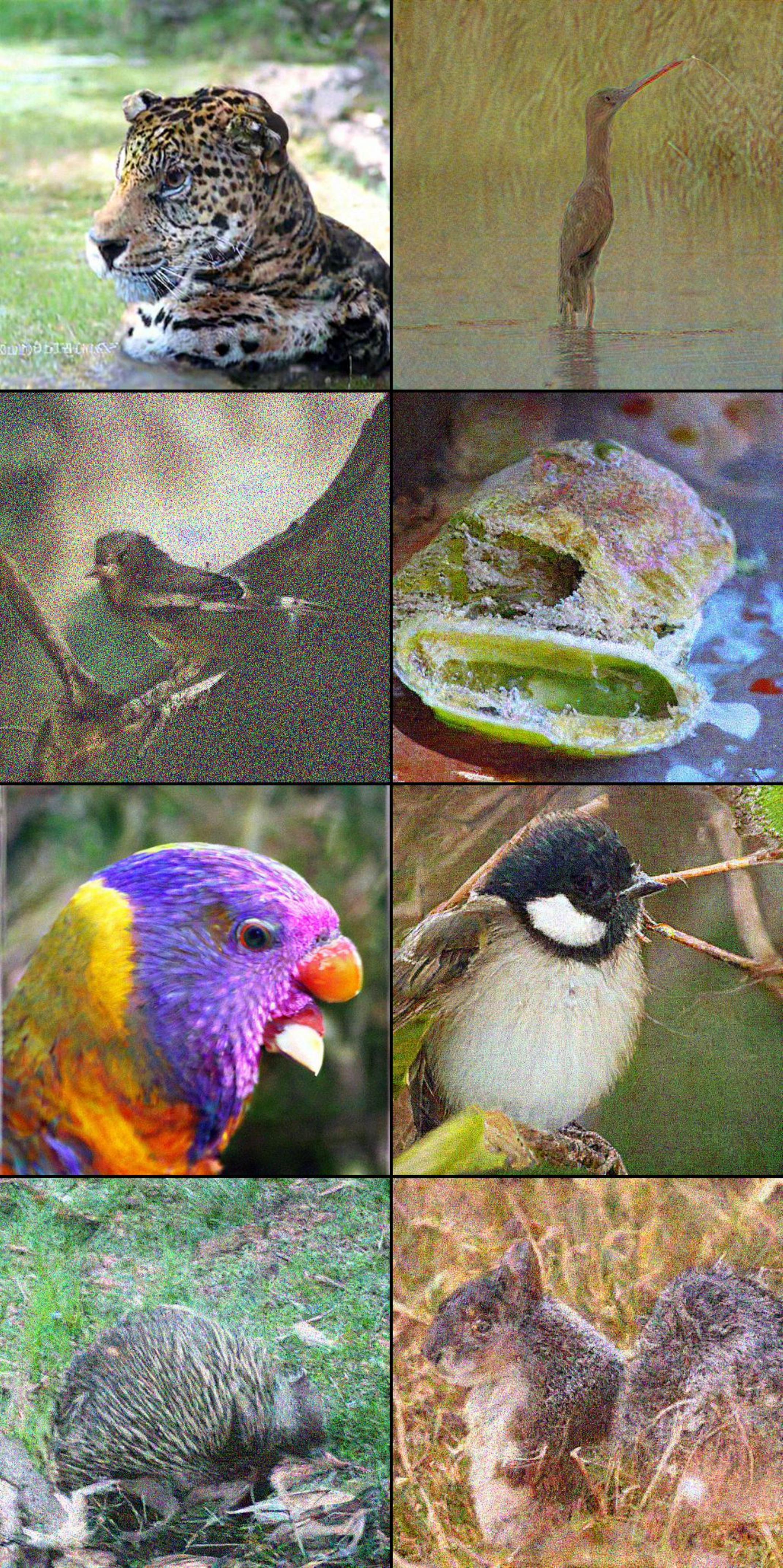} & \includegraphics[width=0.221\textwidth]{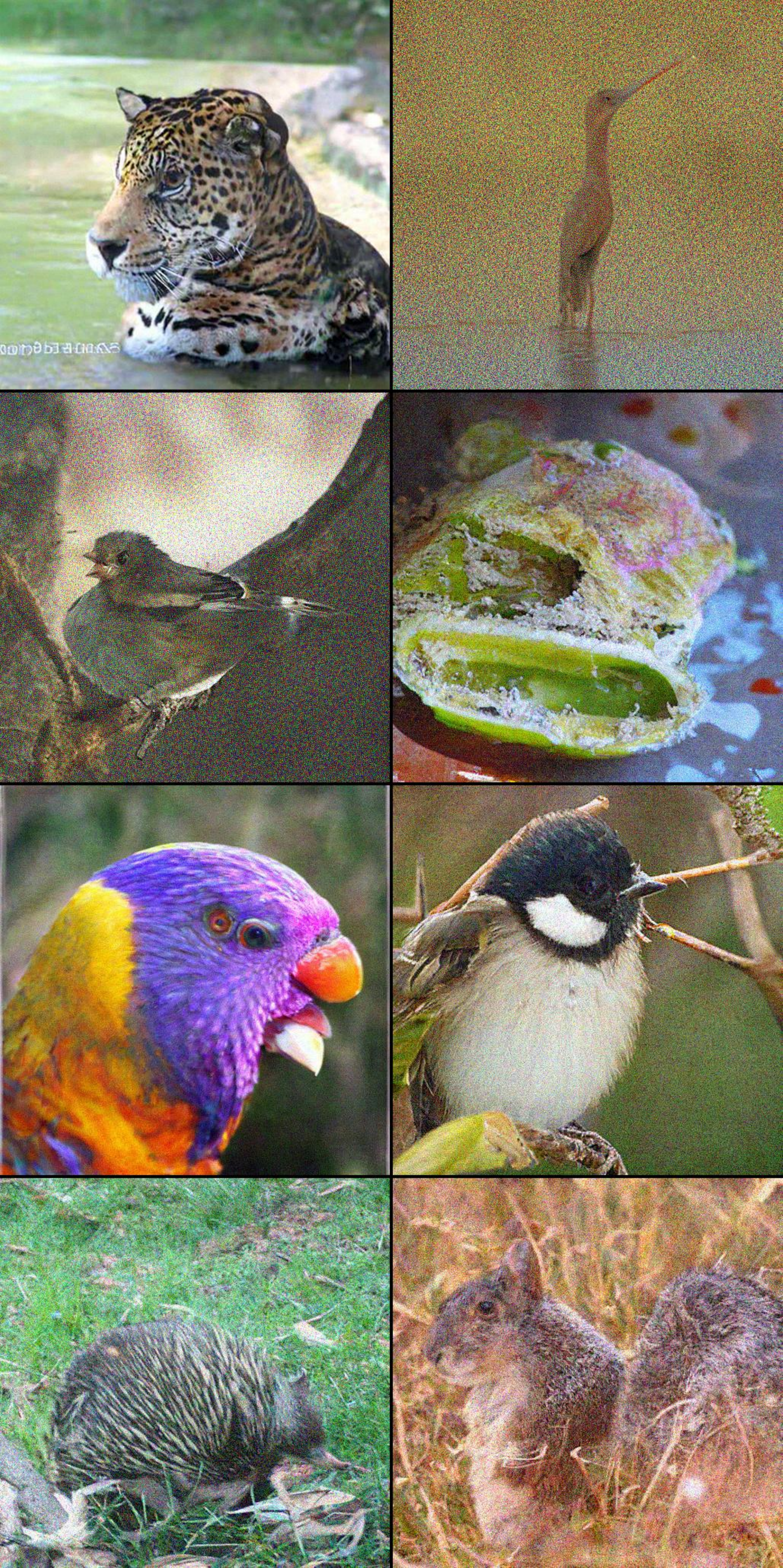} & \includegraphics[width=0.221\textwidth]{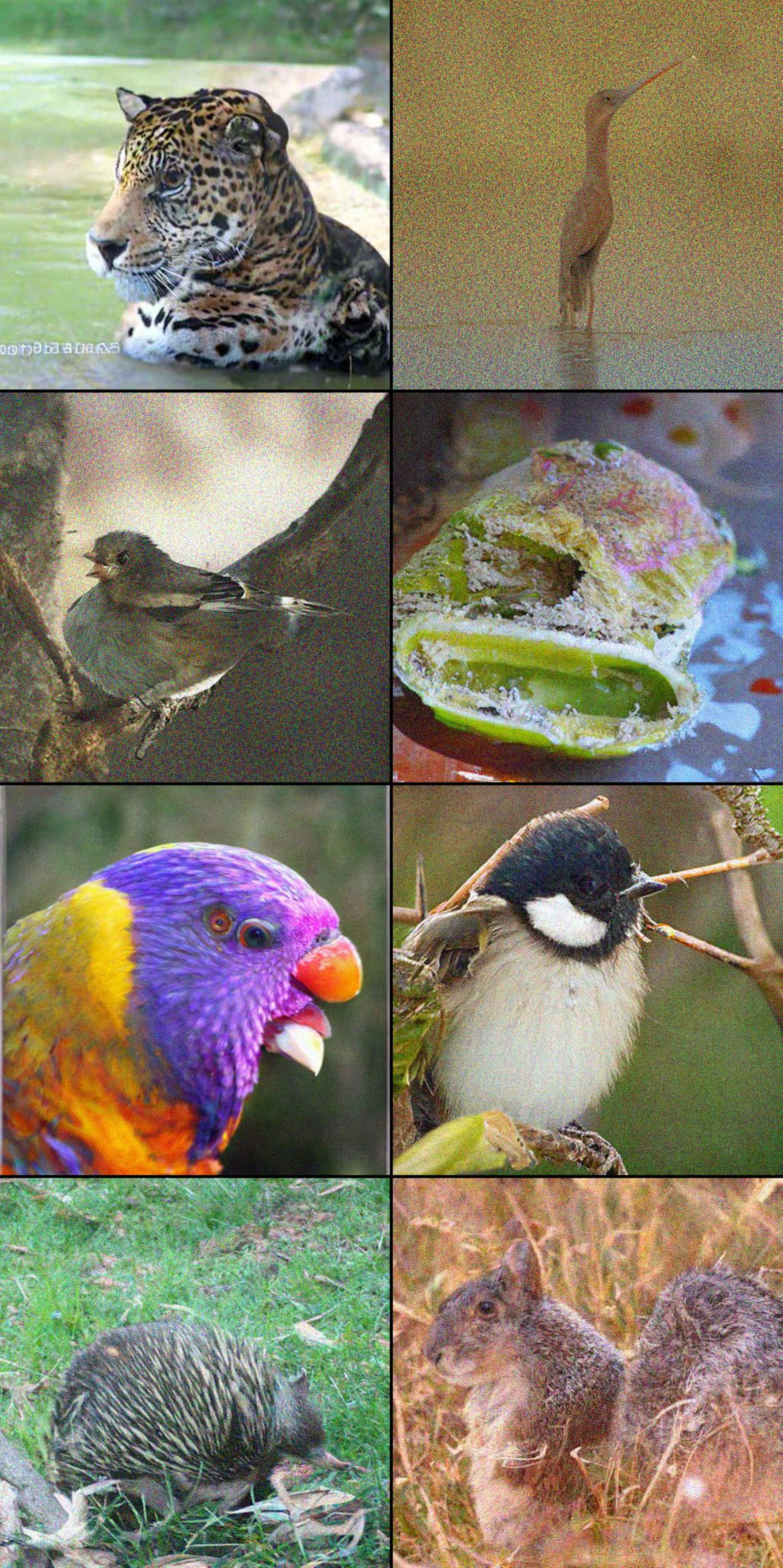}
\\
\multirow{-18.7}{*}{\parbox{0.8cm}{\centering SciRE-Solver (ours)}}
&\includegraphics[width=0.221\textwidth]{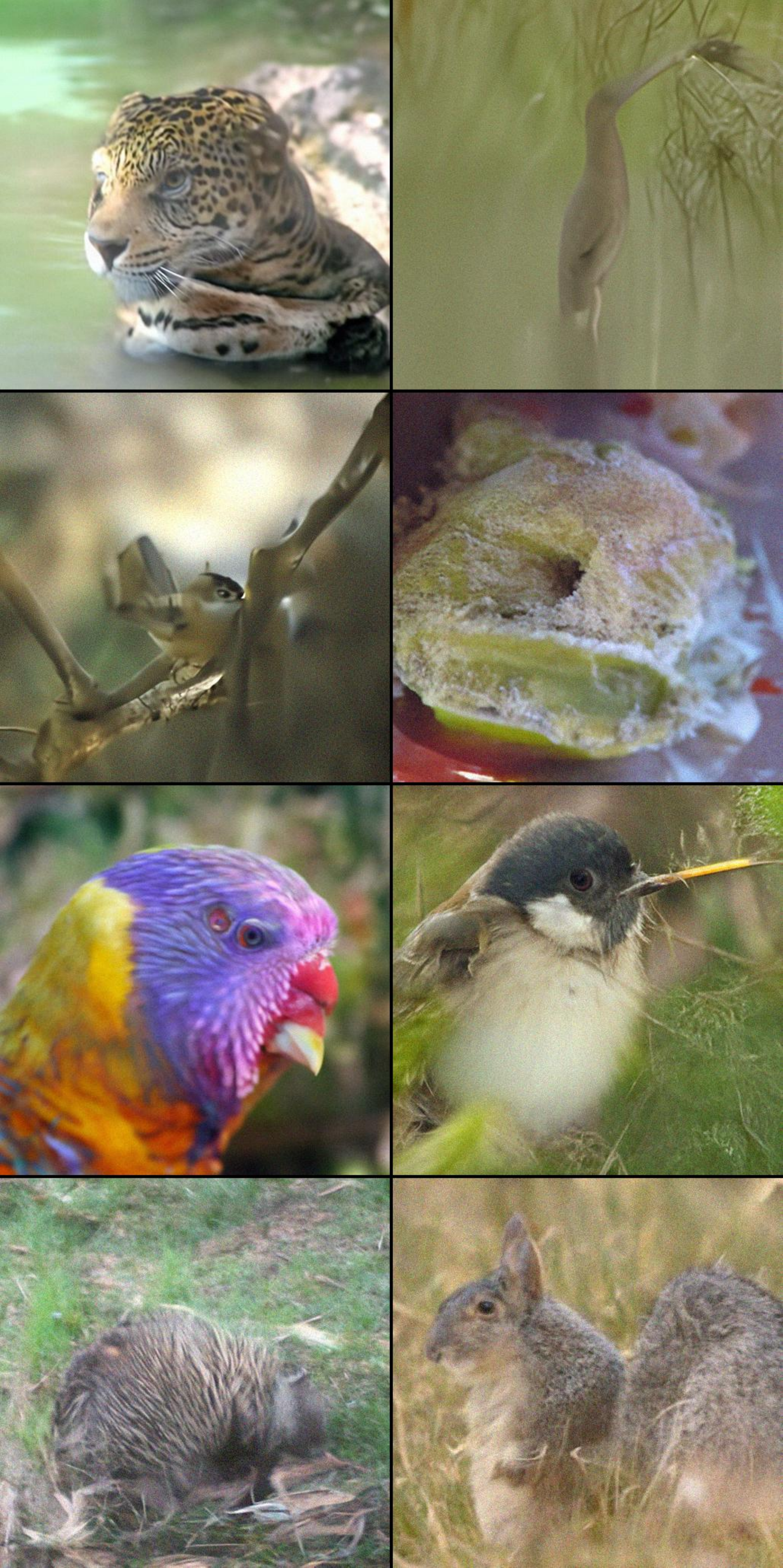} & \includegraphics[width=0.221\textwidth]{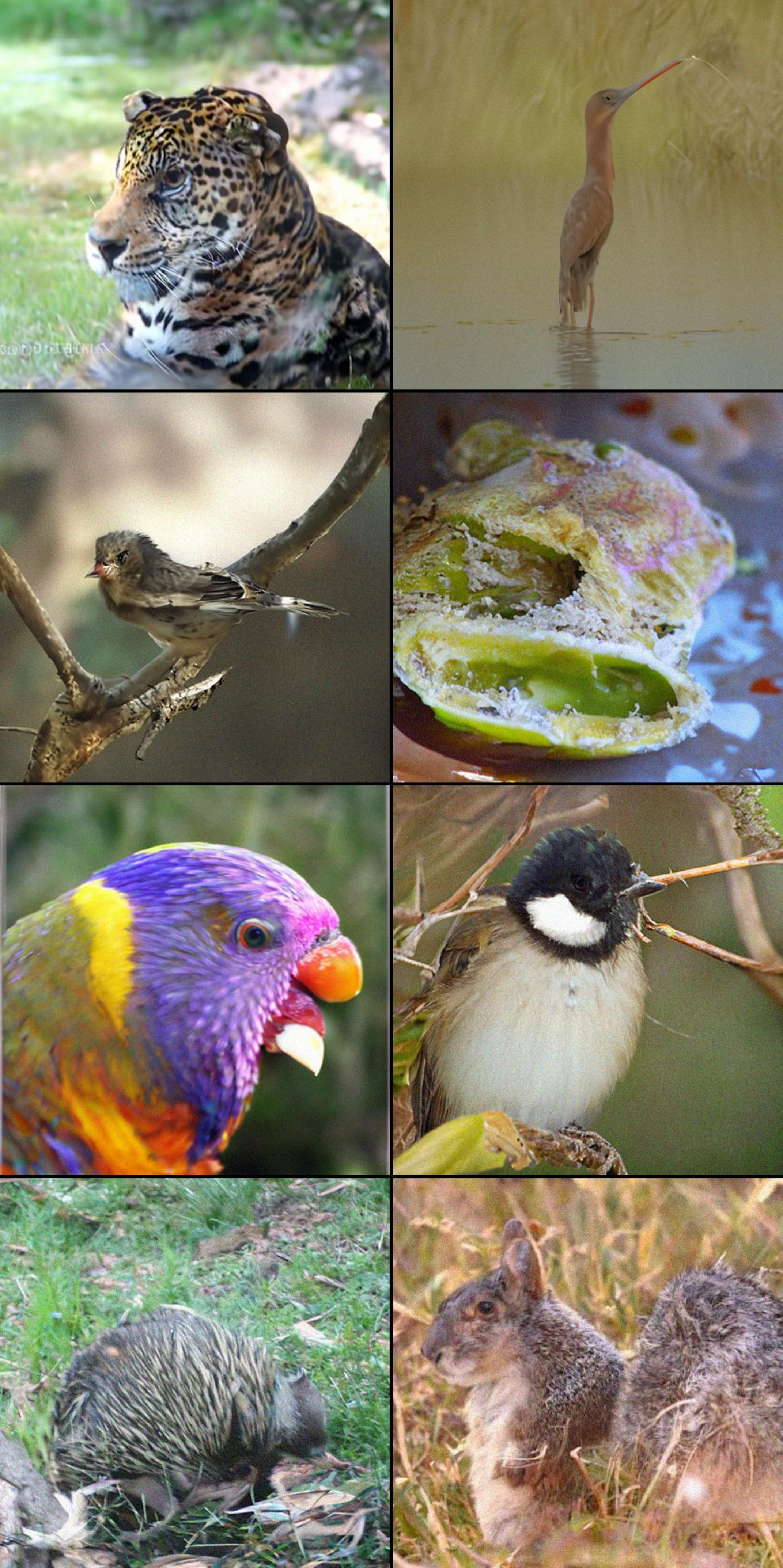} & \includegraphics[width=0.221\textwidth]{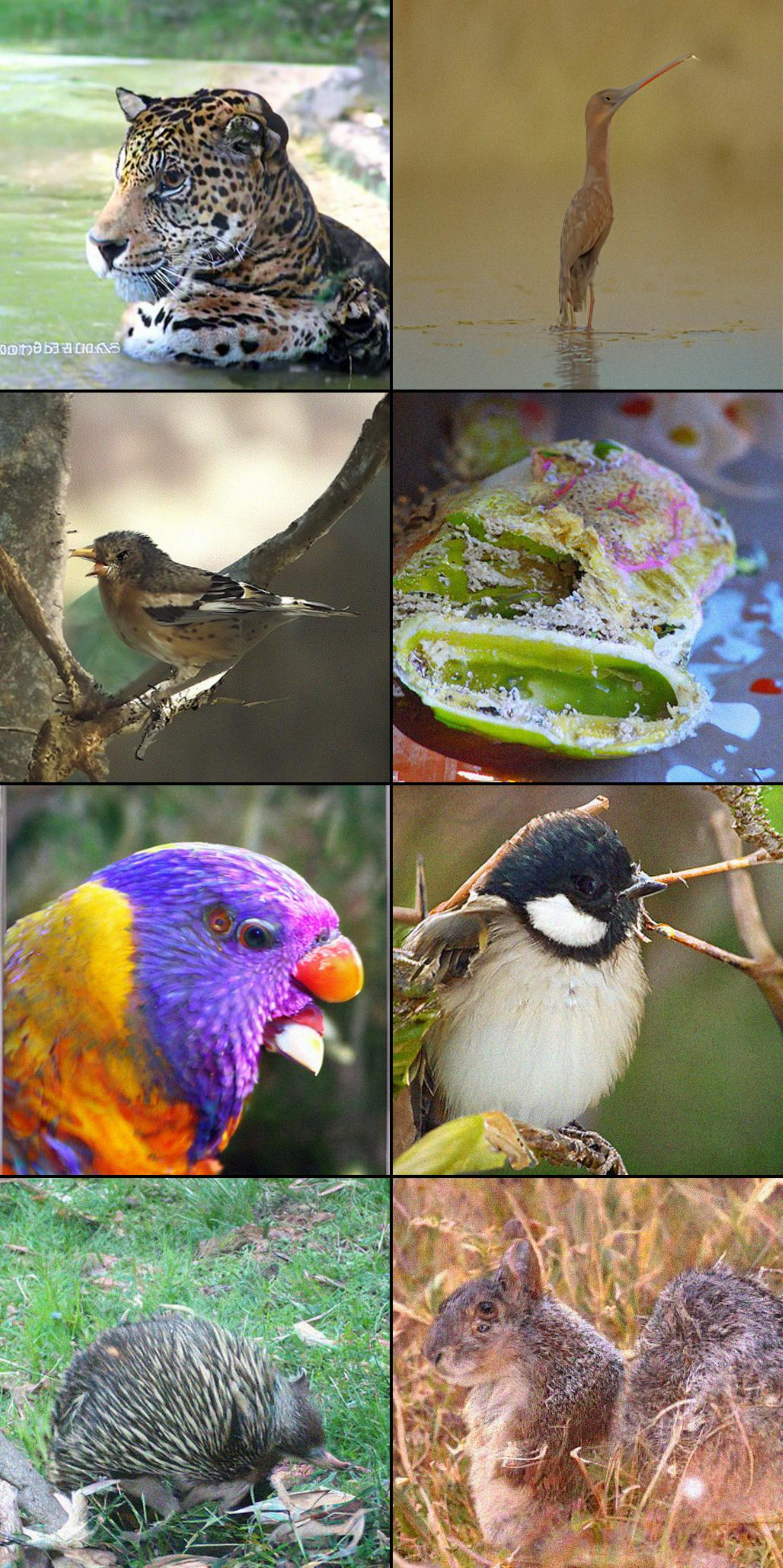} & \includegraphics[width=0.221\textwidth]{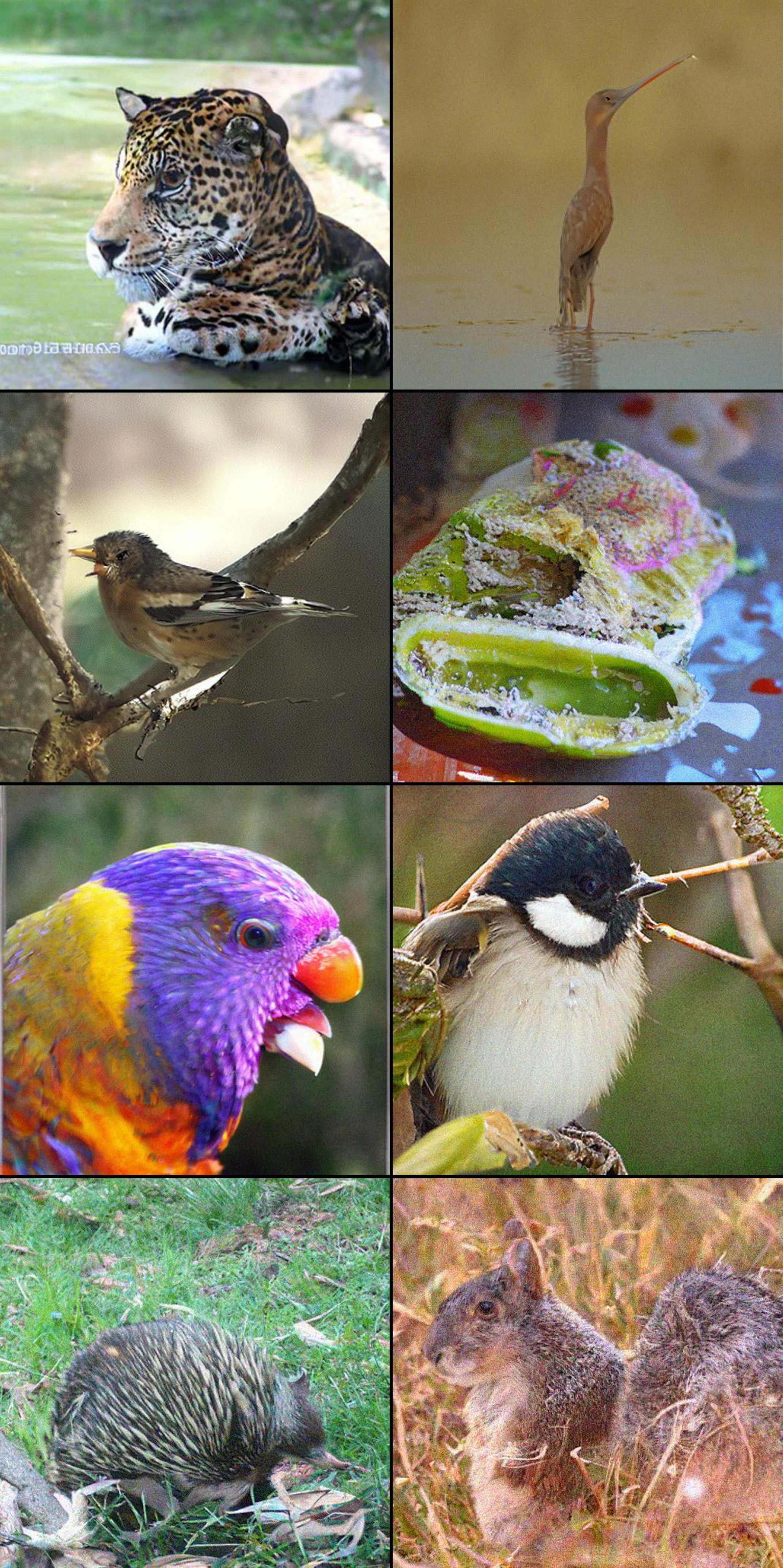} \\
\end{tabular}
 \caption{
Generated samples of the pre-trained DM \cite{dhariwal2021diffusion} on ImageNet 512$\times$512 (classifier scale: 1) using
6-30 sampling steps from different sampling methods with the same settings and codebase.
 }
\label{fig:imagenet512samecomparison}
\end{figure}

\begin{figure}[ht]
\centering
\begin{tabular}{m{0.8cm}p{2.7cm}p{2.7cm}p{2.7cm}p{2.7cm}}
   ~~ &~~~~~~~~~~NFE=$10$& ~~~~~~~~~~NFE=$15$  &~~~~~~~~~~NFE=$20$ &~~~~~~~~~~NFE=$50$ \\
\multirow{-8.3}{*}{\parbox{0.8cm}{\centering DDIM \cite{song2021denoising}}}
& \includegraphics[width=0.22\textwidth]{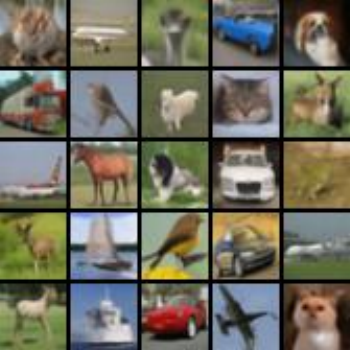} & \includegraphics[width=0.22\textwidth]{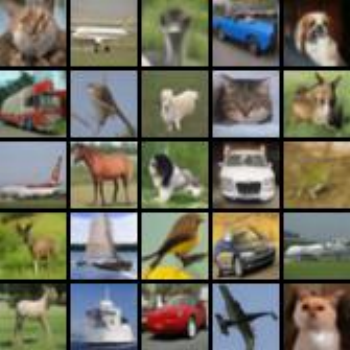} & \includegraphics[width=0.22\textwidth]{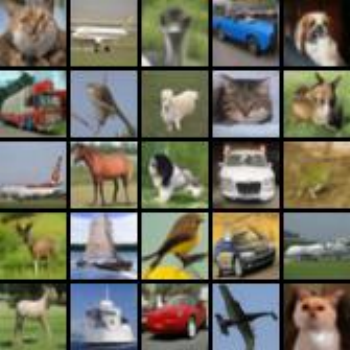} & \includegraphics[width=0.22\textwidth]{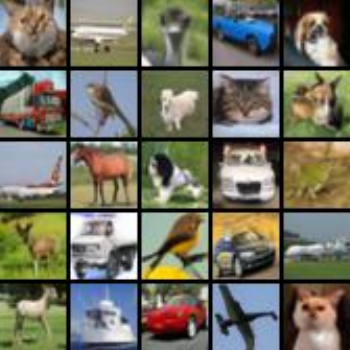}
\\
\multirow{-8.3}{*}{\parbox{0.8cm}{\centering DPM-Solver \cite{lu2022dpm}}}
& \includegraphics[width=0.22\textwidth]{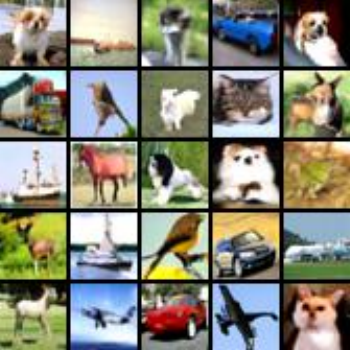} & \includegraphics[width=0.22\textwidth]{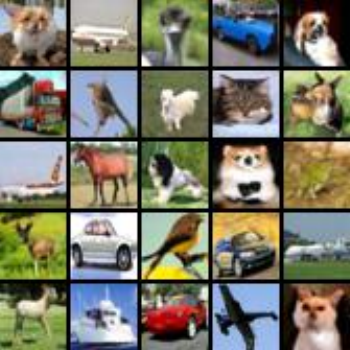} & \includegraphics[width=0.22\textwidth]{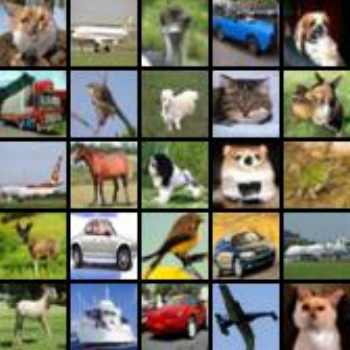} & \includegraphics[width=0.22\textwidth]{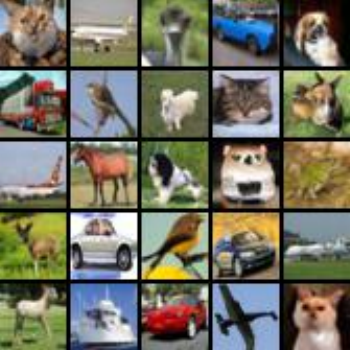}
\\
\multirow{-8.3}{*}{\parbox{0.8cm}{\centering SciRE-Solver (ours)}}
&\includegraphics[width=0.22\textwidth]{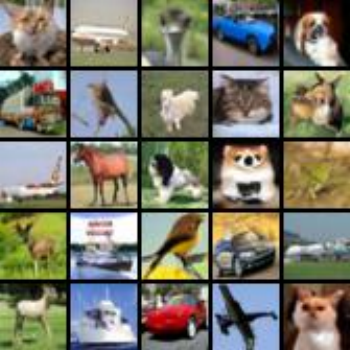} & \includegraphics[width=0.22\textwidth]{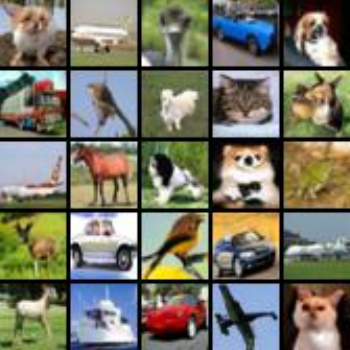} & \includegraphics[width=0.22\textwidth]{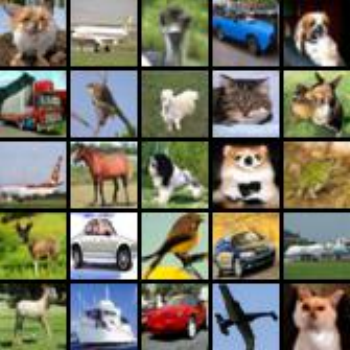} & \includegraphics[width=0.22\textwidth]{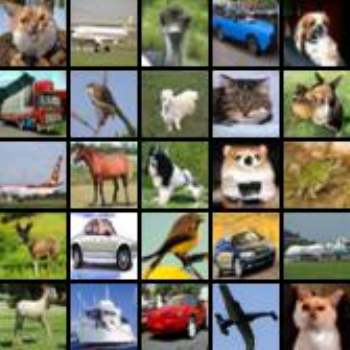} \\
\end{tabular}
 \caption{
 Random samples with the same random seed were generated by  DDIM \cite{song2021denoising} (uniform time steps), DPM-Solver \cite{lu2022dpm} (logSNR time steps), and SciRE-Solver (SNR time steps, $k=3.1$), employing the pre-trained
discrete-time DPM  \cite{ho2020denoising} on CIFAR-10.
 }
\label{fig:crfar10omparison}
\end{figure}

\begin{figure}[ht]
\centering
\begin{tabular}{m{0.8cm}p{2.7cm}p{2.7cm}p{2.7cm}p{2.7cm}}
   ~~ &~~~~~~~~~~NFE=$10$& ~~~~~~~~~~NFE=$15$  &~~~~~~~~~~NFE=$20$ &~~~~~~~~~~NFE=$50$ \\
\multirow{-8.3}{*}{\parbox{0.8cm}{\centering DDIM \cite{song2021denoising}}}
& \includegraphics[width=0.216\textwidth]{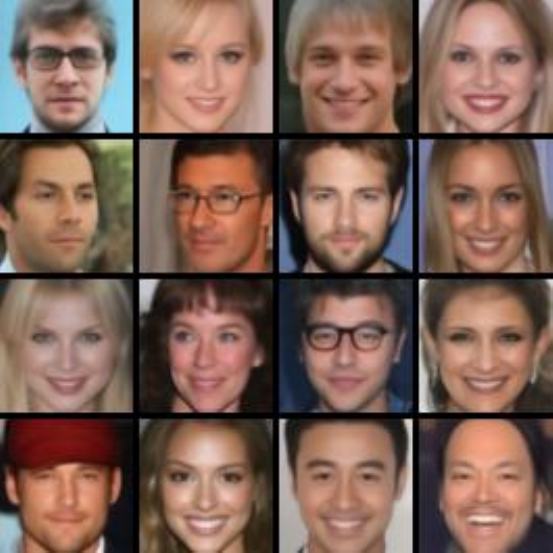} & \includegraphics[width=0.216\textwidth]{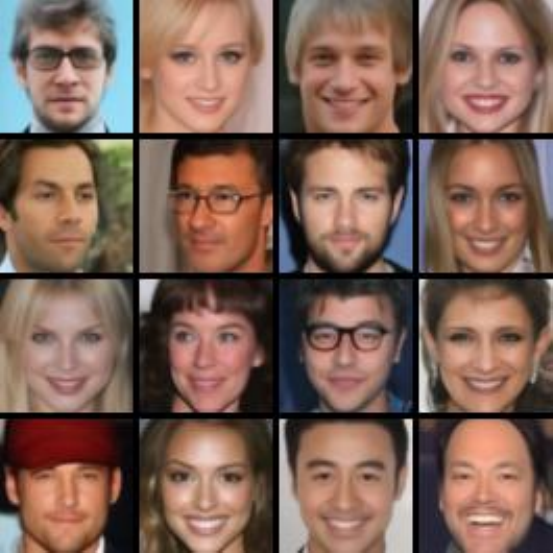} & \includegraphics[width=0.216\textwidth]{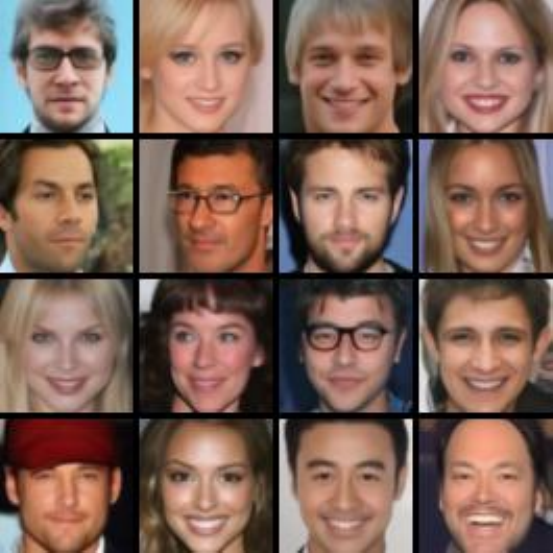} & \includegraphics[width=0.216\textwidth]{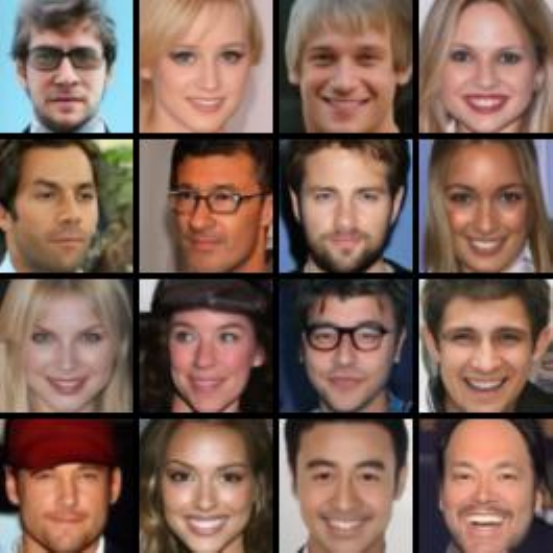}
\\
\multirow{-8.3}{*}{\parbox{0.8cm}{\centering DPM-Solver \cite{lu2022dpm}}}
& \includegraphics[width=0.216\textwidth]{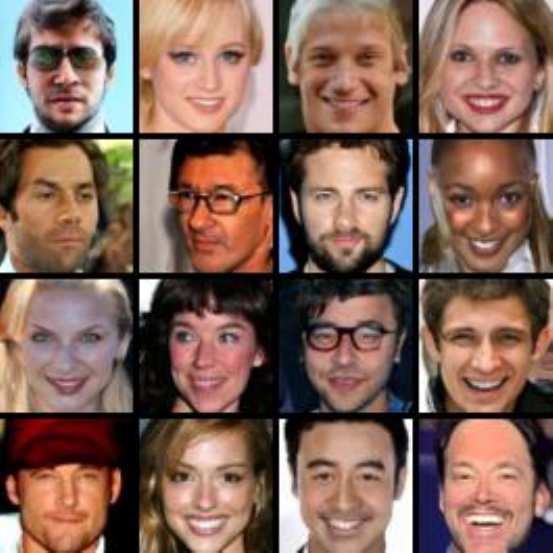} & \includegraphics[width=0.216\textwidth]{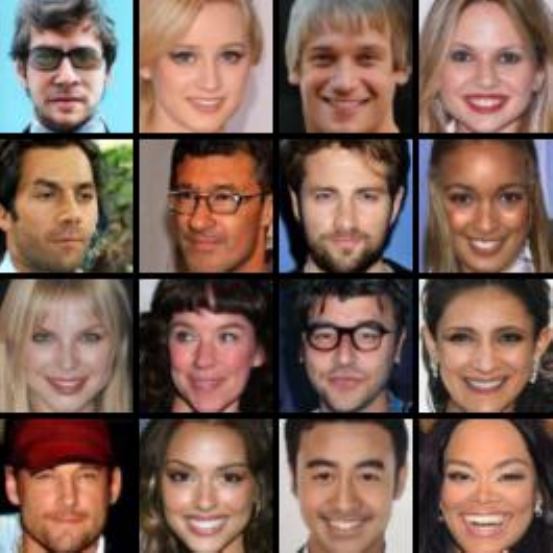} & \includegraphics[width=0.216\textwidth]{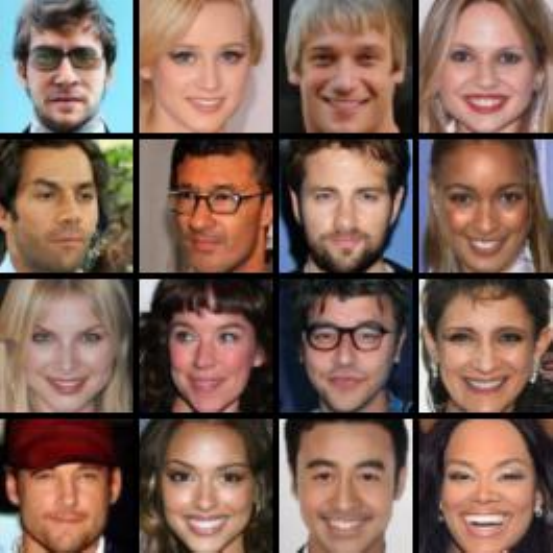} & \includegraphics[width=0.216\textwidth]{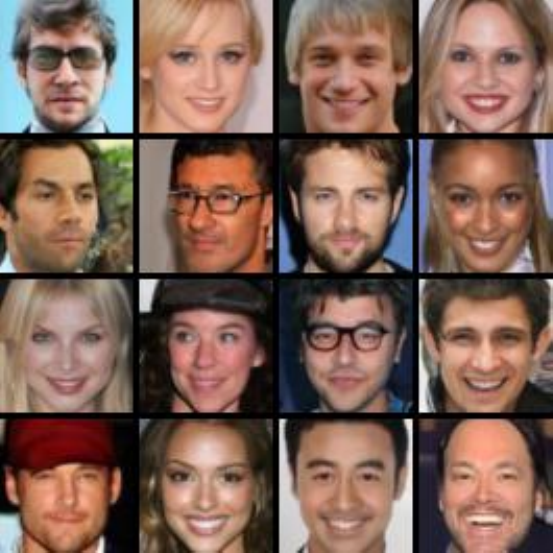}
\\
\multirow{-8.3}{*}{\parbox{0.8cm}{\centering SciRE-Solver (ours)}}
&\includegraphics[width=0.216\textwidth]{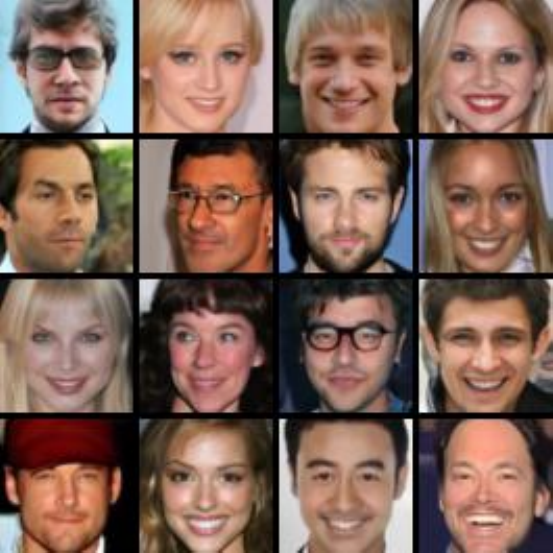} & \includegraphics[width=0.216\textwidth]{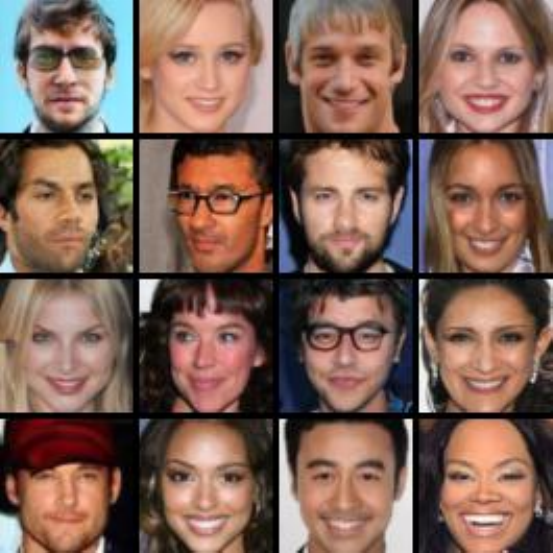} & \includegraphics[width=0.216\textwidth]{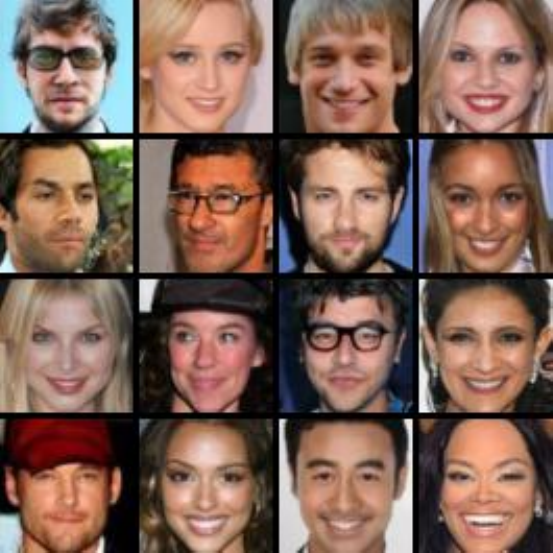} & \includegraphics[width=0.216\textwidth]{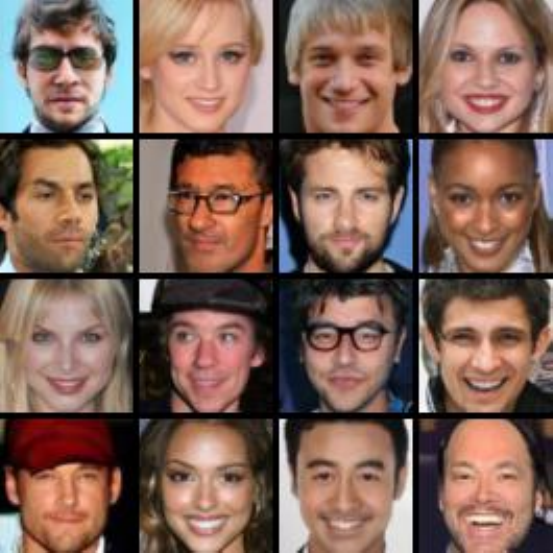} \\
\end{tabular}
 \caption{
 Random samples with the same random seed were generated by  DDIM \cite{song2021denoising} (uniform time steps), DPM-Solver \cite{lu2022dpm} (logSNR time steps), and SciRE-Solver (SNR time steps, $k=3.1$), employing the pre-trained
discrete-time DPM \cite{song2021denoising} on CelebA 64$\times$64.
 }
\label{fig:celeba64comparison}
\end{figure}

\begin{figure}[ht]
\centering
\begin{tabular}{m{0.8cm}p{2.7cm}p{2.7cm}p{2.7cm}p{2.7cm}}
   ~~ &~~~~~~~~~~NFE=$10$& ~~~~~~~~~~NFE=$15$  &~~~~~~~~~~NFE=$20$ &~~~~~~~~~~NFE=$50$ \\
\multirow{-8.3}{*}{\parbox{0.8cm}{\centering DDIM \cite{song2021denoising}}}
& \includegraphics[width=0.216\textwidth]{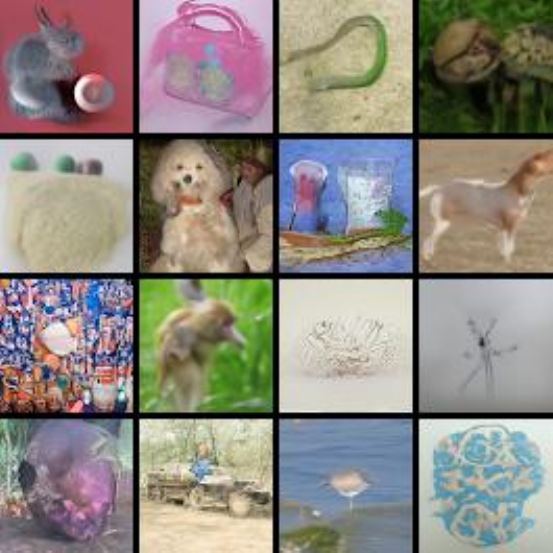} & \includegraphics[width=0.216\textwidth]{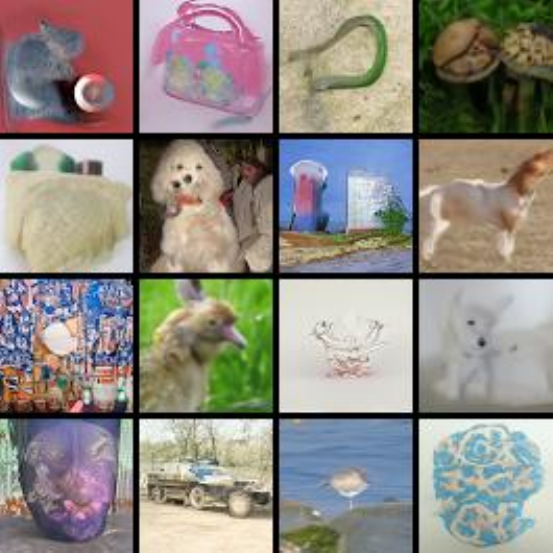} & \includegraphics[width=0.216\textwidth]{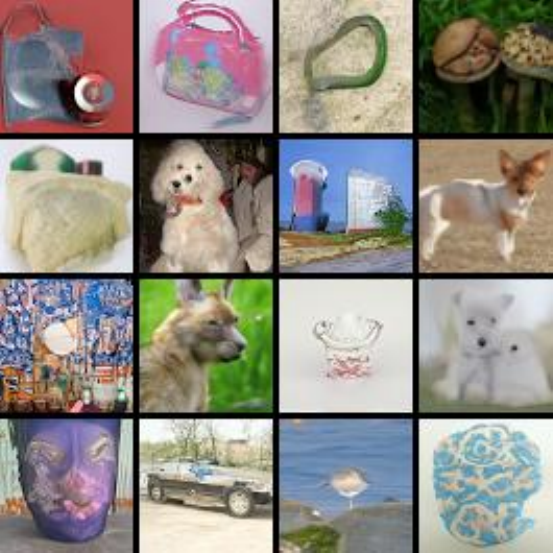} & \includegraphics[width=0.216\textwidth]{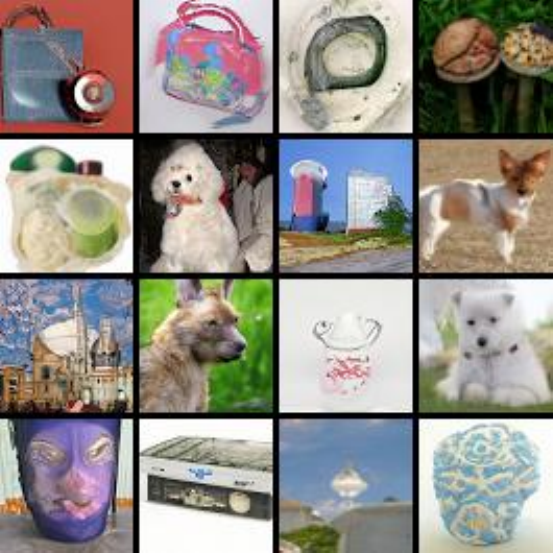}
\\
\multirow{-8.3}{*}{\parbox{0.8cm}{\centering DPM-Solver \cite{lu2022dpm}}}
& \includegraphics[width=0.216\textwidth]{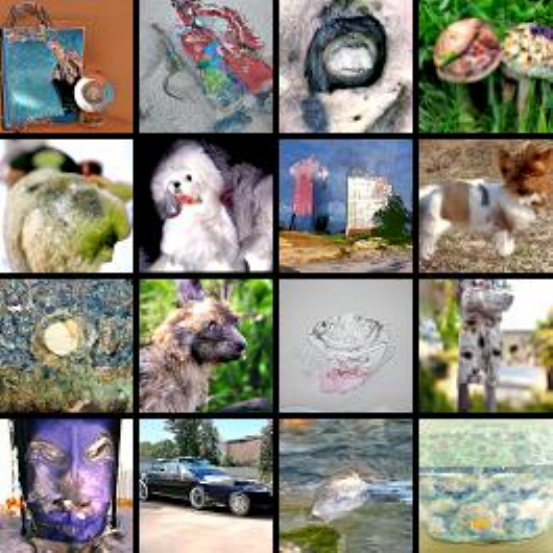} & \includegraphics[width=0.216\textwidth]{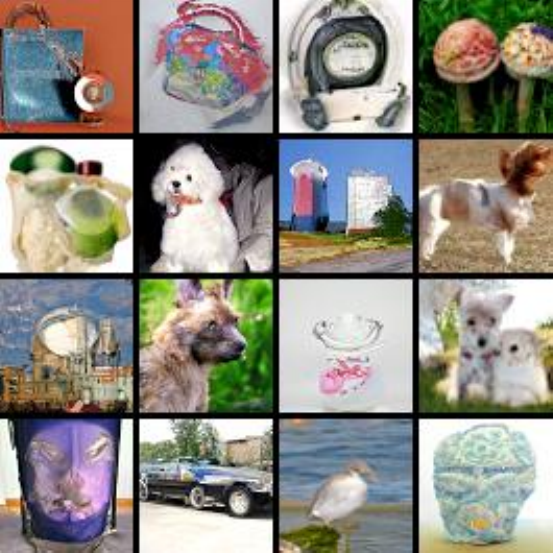} & \includegraphics[width=0.216\textwidth]{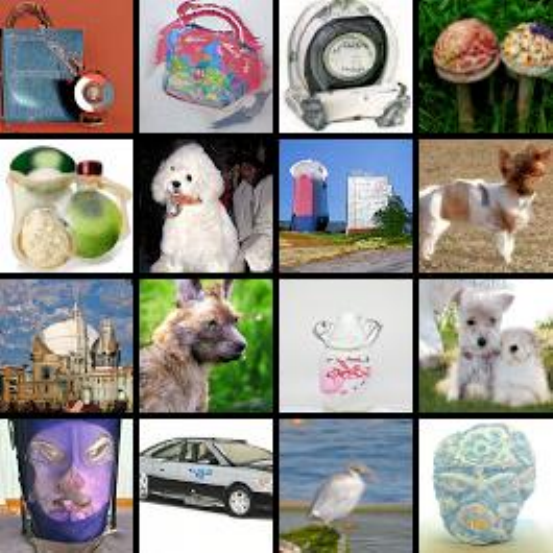} & \includegraphics[width=0.216\textwidth]{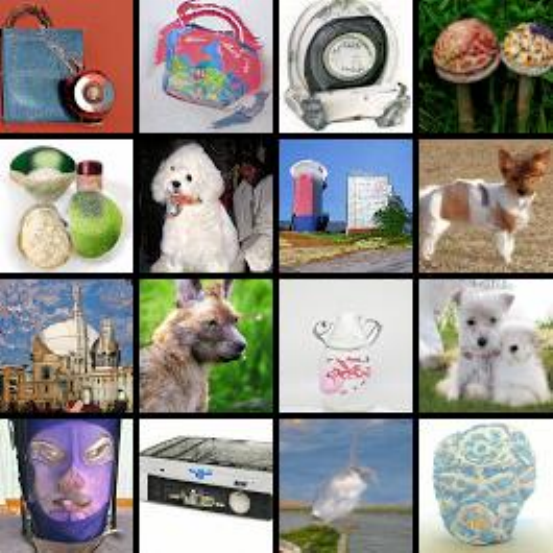}
\\
\multirow{-8.3}{*}{\parbox{0.8cm}{\centering SciRE-Solver (ours)}}
&\includegraphics[width=0.216\textwidth]{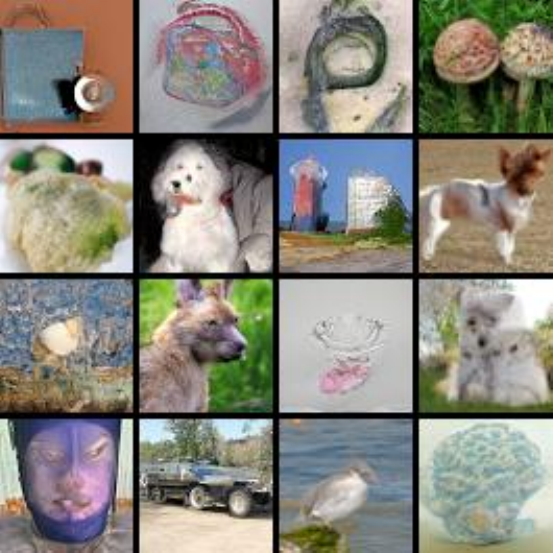} & \includegraphics[width=0.216\textwidth]{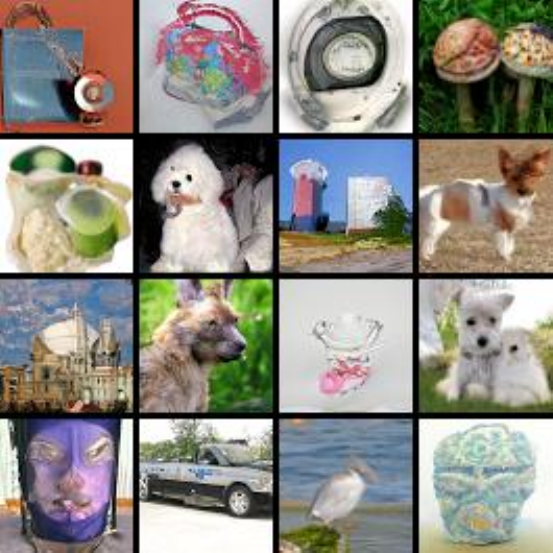} & \includegraphics[width=0.216\textwidth]{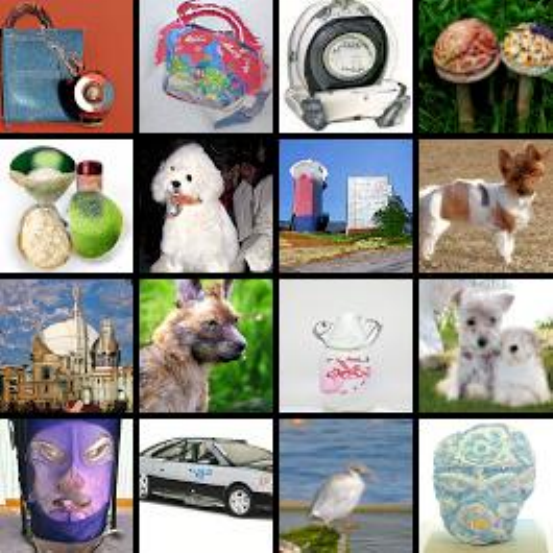} & \includegraphics[width=0.216\textwidth]{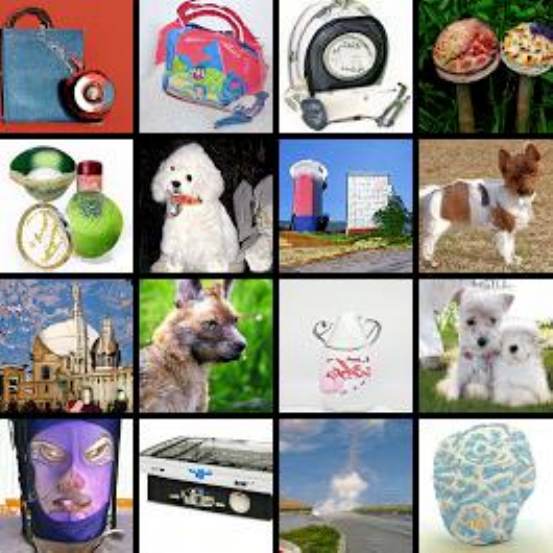} \\
\end{tabular}
 \caption{
Random samples with the same random seed were generated by  DDIM \cite{song2021denoising} (uniform time steps), DPM-Solver \cite{lu2022dpm} (logSNR time steps), and SciRE-Solver (SNR time steps, $k=3.1$), employing the pre-trained
discrete-time DPM \cite{nichol2021improved} on ImageNet 64$\times$64.
 }
\label{fig:imagenet64comparison}
\end{figure}

\begin{figure}[ht]
\centering
\begin{tabular}{m{0.8cm}p{2.7cm}p{2.7cm}p{2.7cm}p{2.7cm}}
   ~~ &~~~~~~~~~~NFE=$10$& ~~~~~~~~~~NFE=$15$  &~~~~~~~~~~NFE=$20$ &~~~~~~~~~~NFE=$50$ \\
\multirow{-8.3}{*}{\parbox{0.8cm}{\centering DDIM \cite{song2021denoising}}}
& \includegraphics[width=0.216\textwidth]{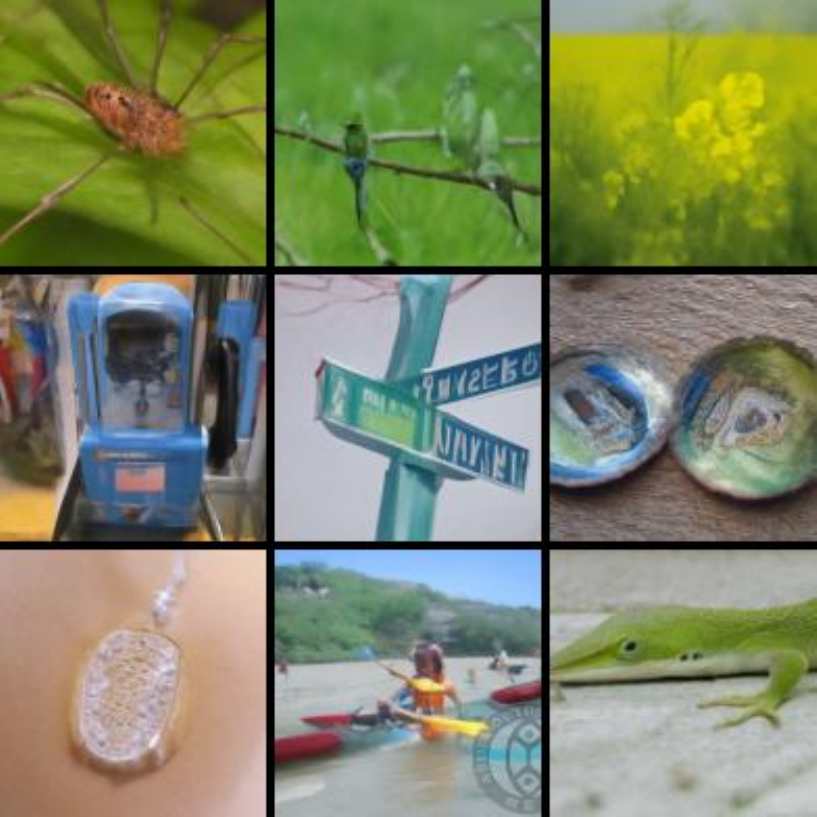} & \includegraphics[width=0.216\textwidth]{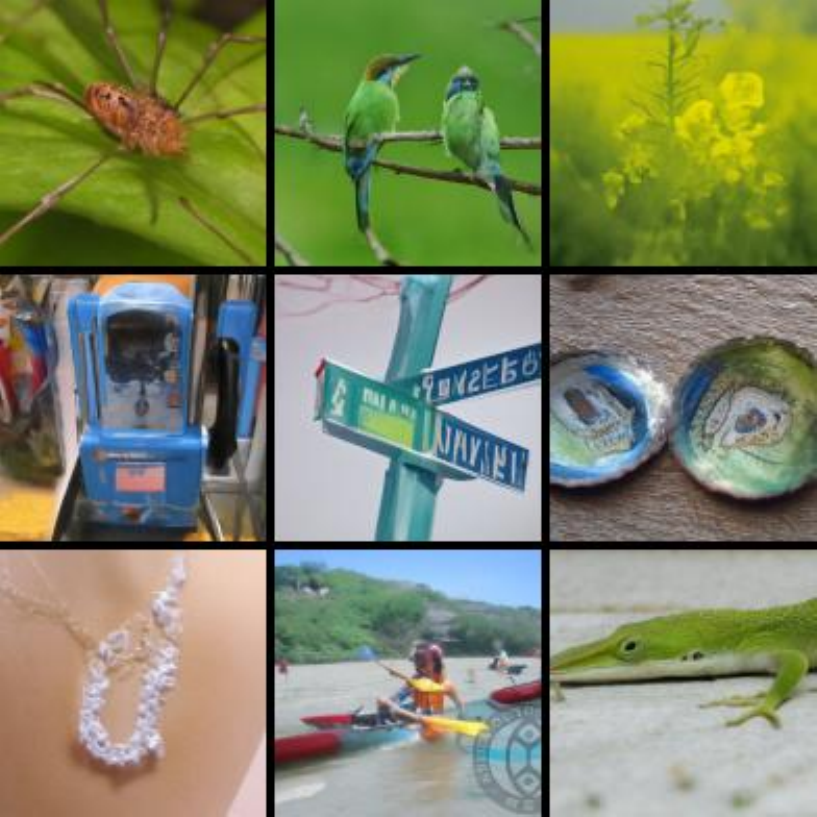} & \includegraphics[width=0.216\textwidth]{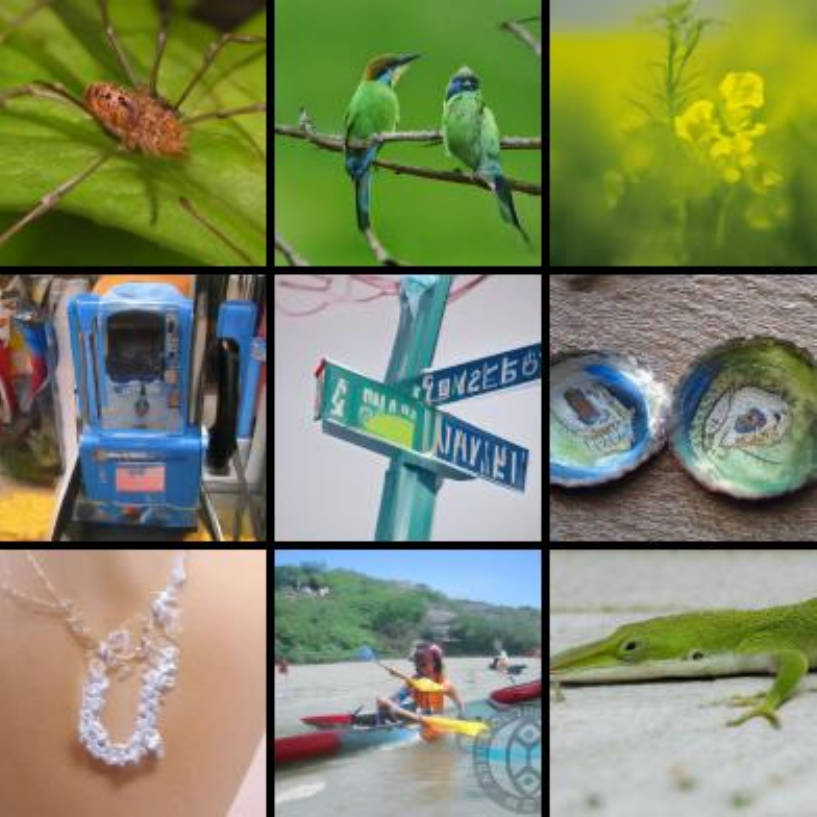} & \includegraphics[width=0.216\textwidth]{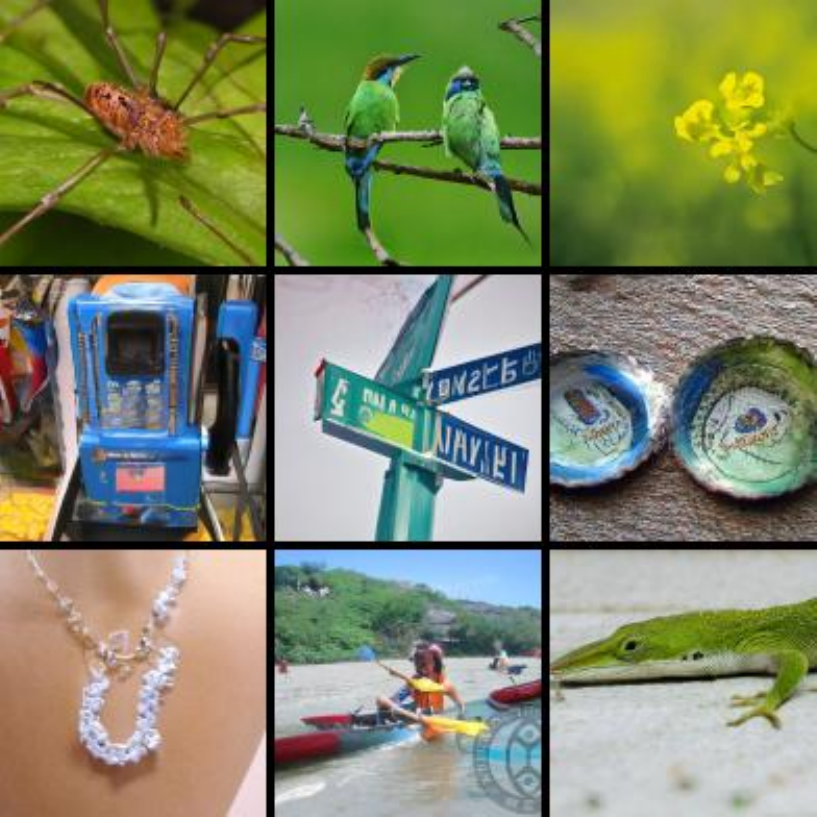}
\\
\multirow{-8.3}{*}{\parbox{0.8cm}{\centering DPM-Solver \cite{lu2022dpm}}}
& \includegraphics[width=0.216\textwidth]{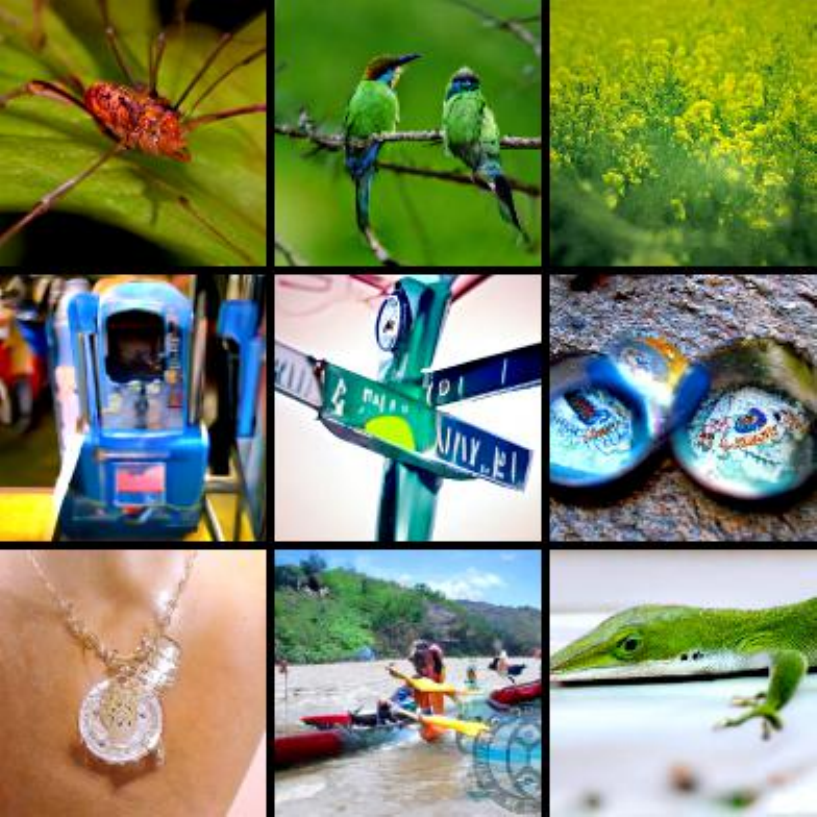} & \includegraphics[width=0.216\textwidth]{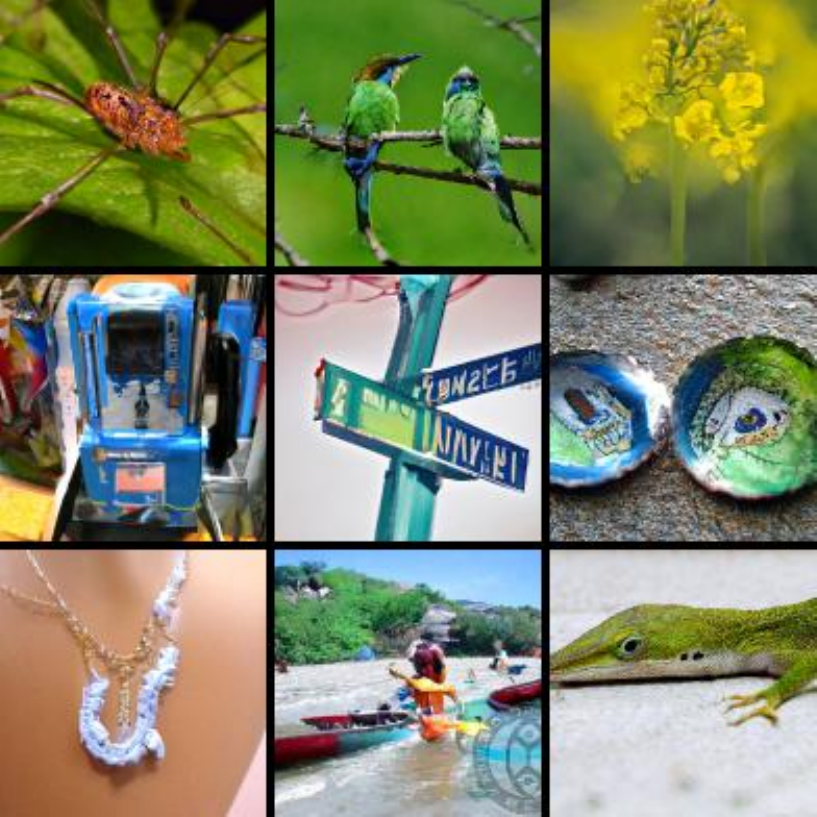} & \includegraphics[width=0.216\textwidth]{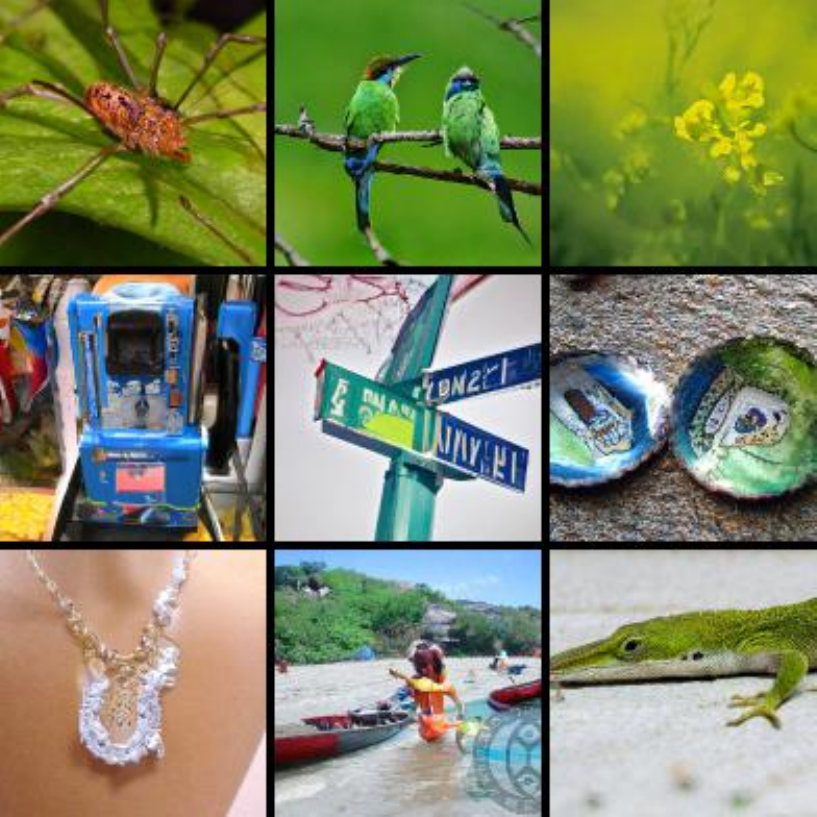} & \includegraphics[width=0.216\textwidth]{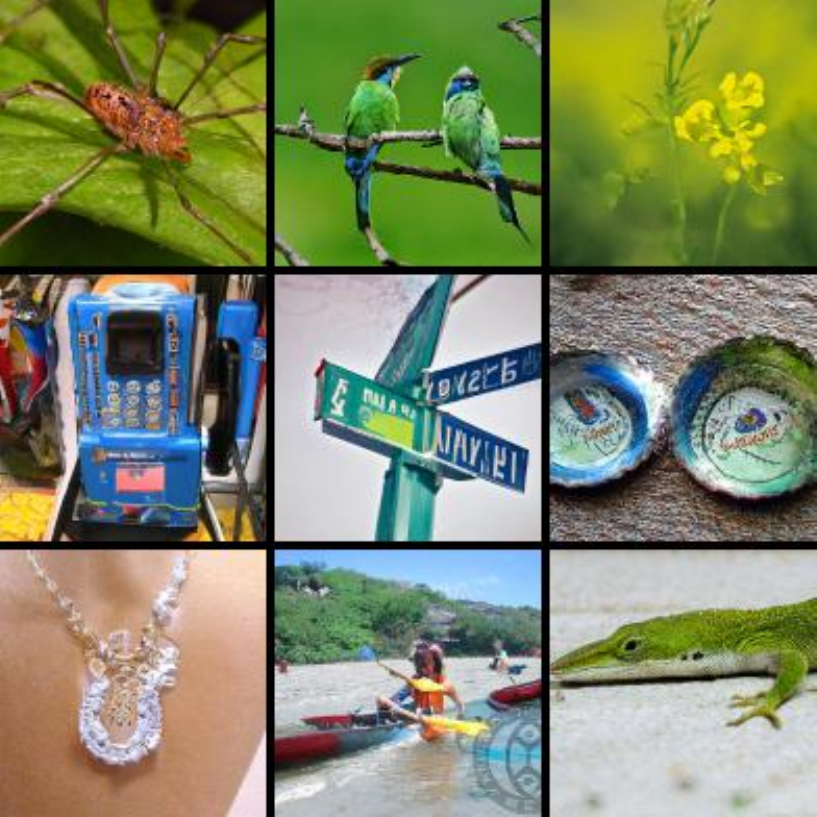}
\\
\multirow{-8.3}{*}{\parbox{0.8cm}{\centering SciRE-Solver (ours)}}
&\includegraphics[width=0.216\textwidth]{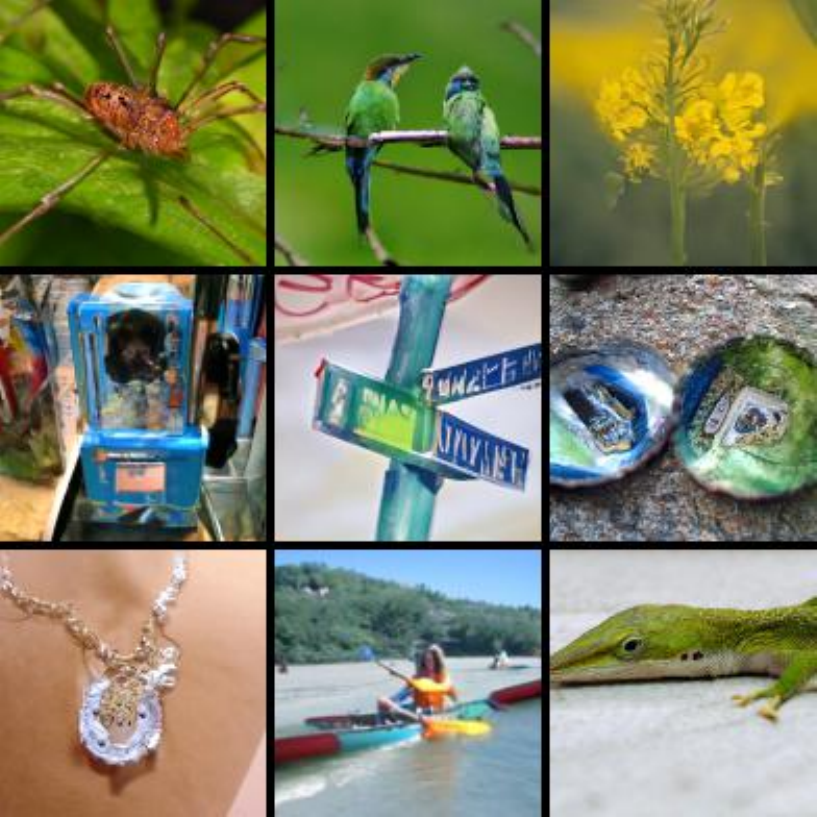} & \includegraphics[width=0.216\textwidth]{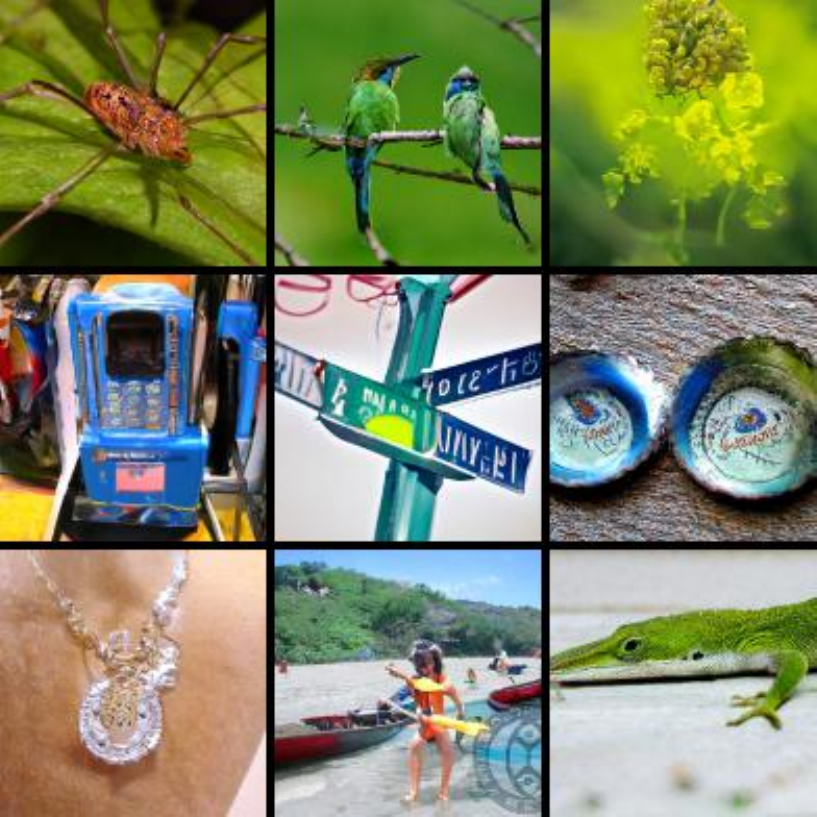} & \includegraphics[width=0.216\textwidth]{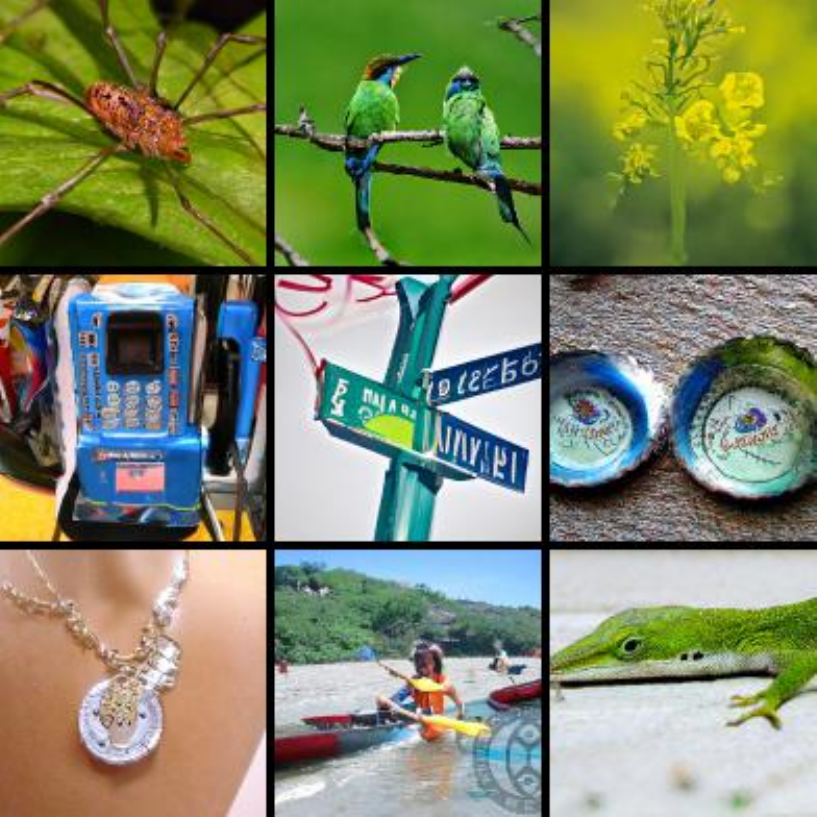} & \includegraphics[width=0.216\textwidth]{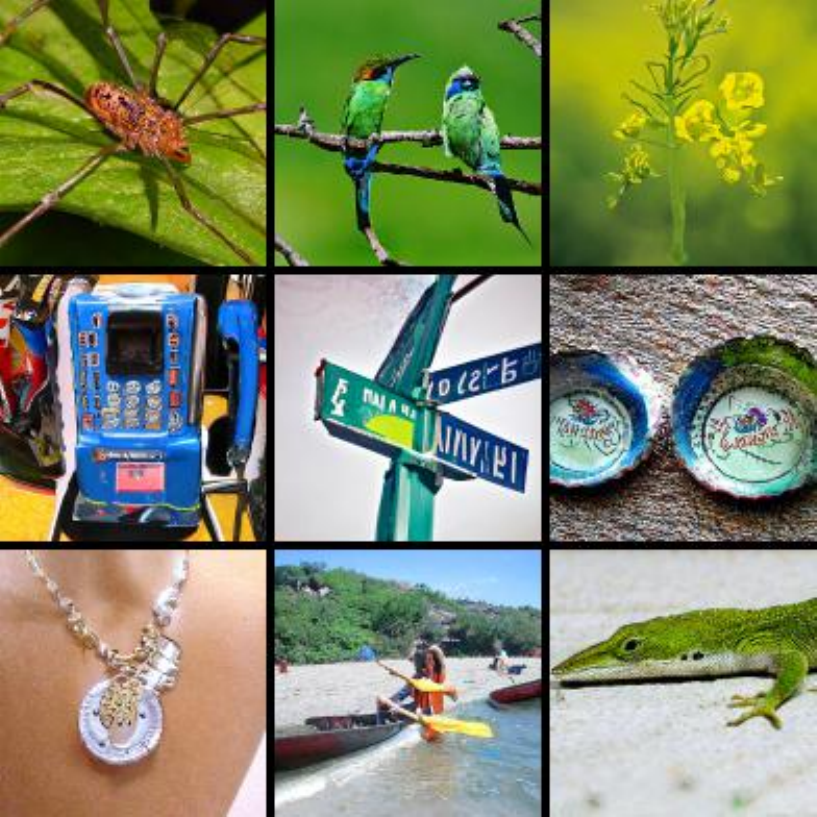} \\
\end{tabular}
 \caption{
Random samples with the same random seed were generated by  DDIM \cite{song2021denoising} (uniform time steps), DPM-Solver \cite{lu2022dpm} (logSNR time steps), and SciRE-Solver (SNR time steps, $k=3.1$), employing the pre-trained
discrete-time DPM \cite{dhariwal2021diffusion} on ImageNet 128$\times$128 (classifier scale: 1.25).
 }
\label{fig:imagenet128comparison}
\end{figure}

\begin{figure}[ht]
\centering
\begin{tabular}{m{0.8cm}p{2.7cm}p{2.7cm}p{2.7cm}p{2.7cm}}
   ~~ &~~~~~~~~~~NFE=$10$& ~~~~~~~~~~NFE=$15$  &~~~~~~~~~~NFE=$20$ &~~~~~~~~~~NFE=$50$ \\
\multirow{-8.3}{*}{\parbox{0.8cm}{\centering DDIM \cite{song2021denoising}}}
& \includegraphics[width=0.216\textwidth]{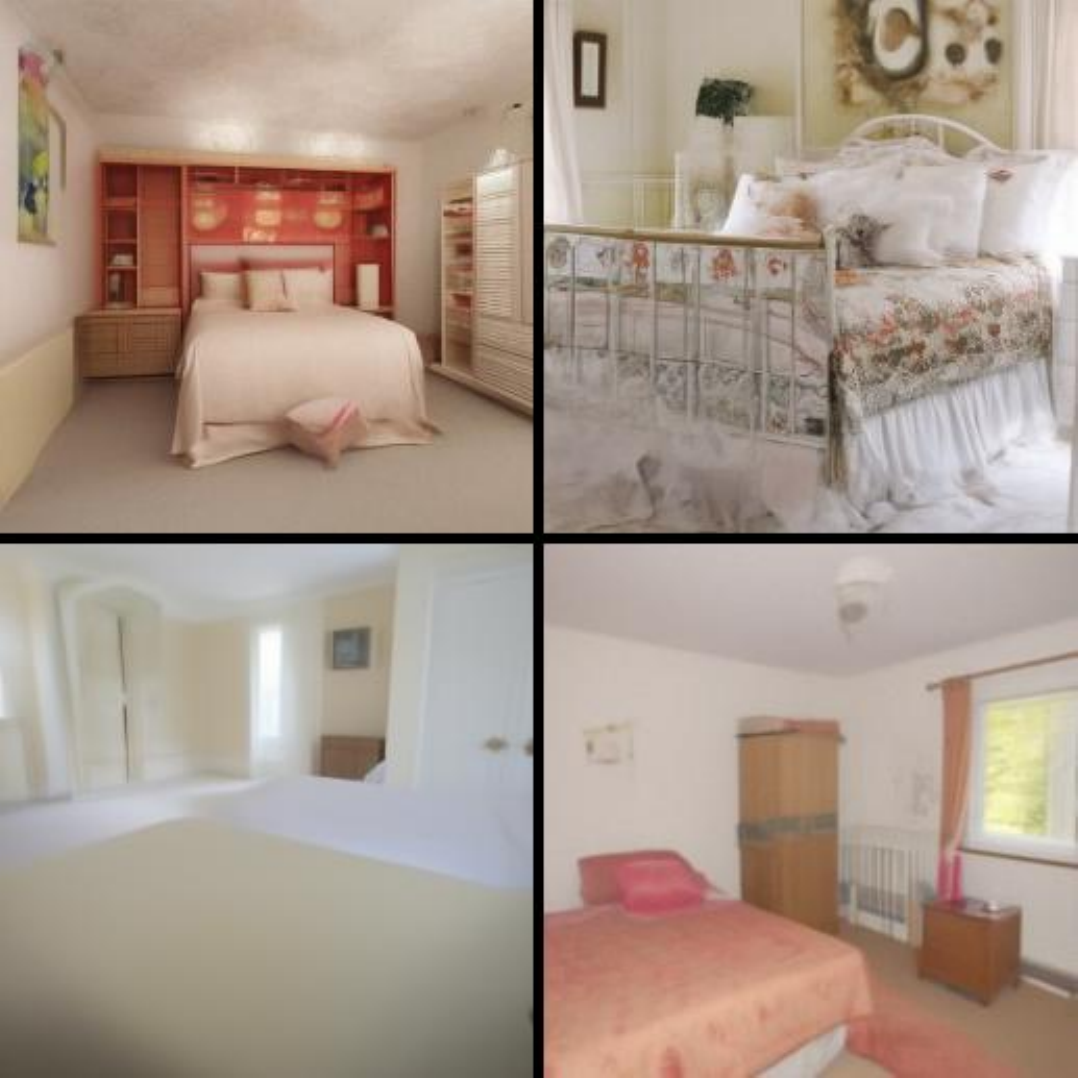} & \includegraphics[width=0.216\textwidth]{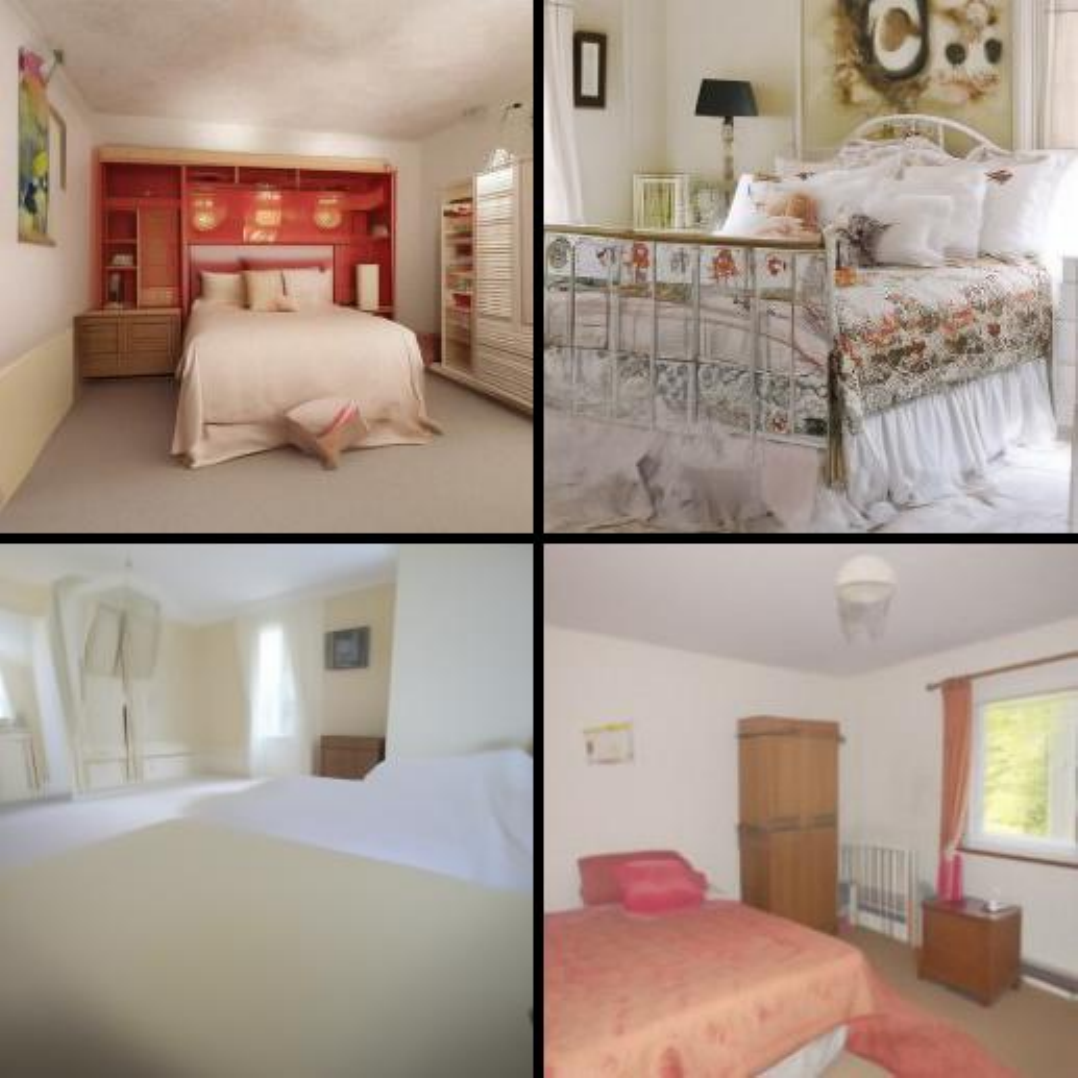} & \includegraphics[width=0.216\textwidth]{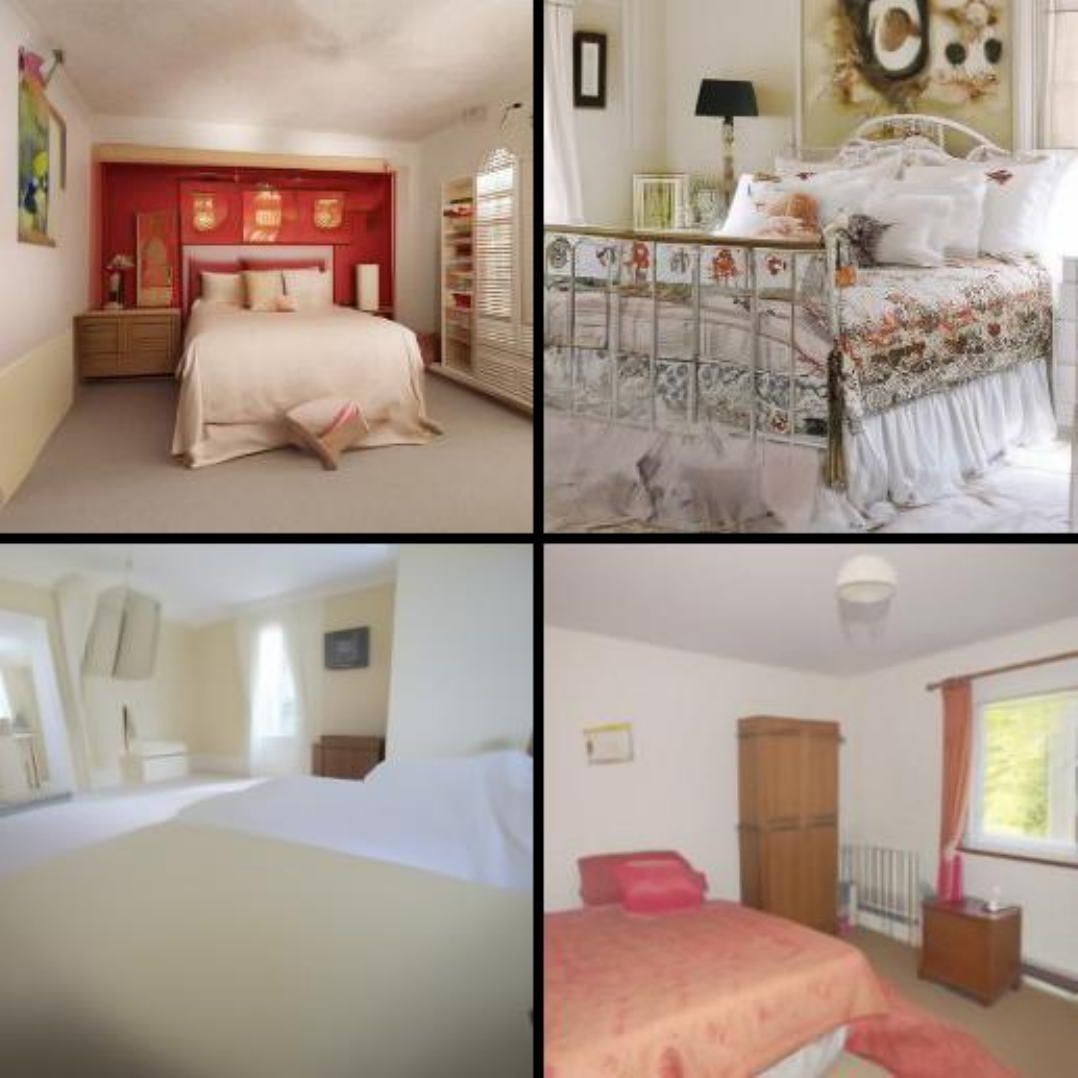} & \includegraphics[width=0.216\textwidth]{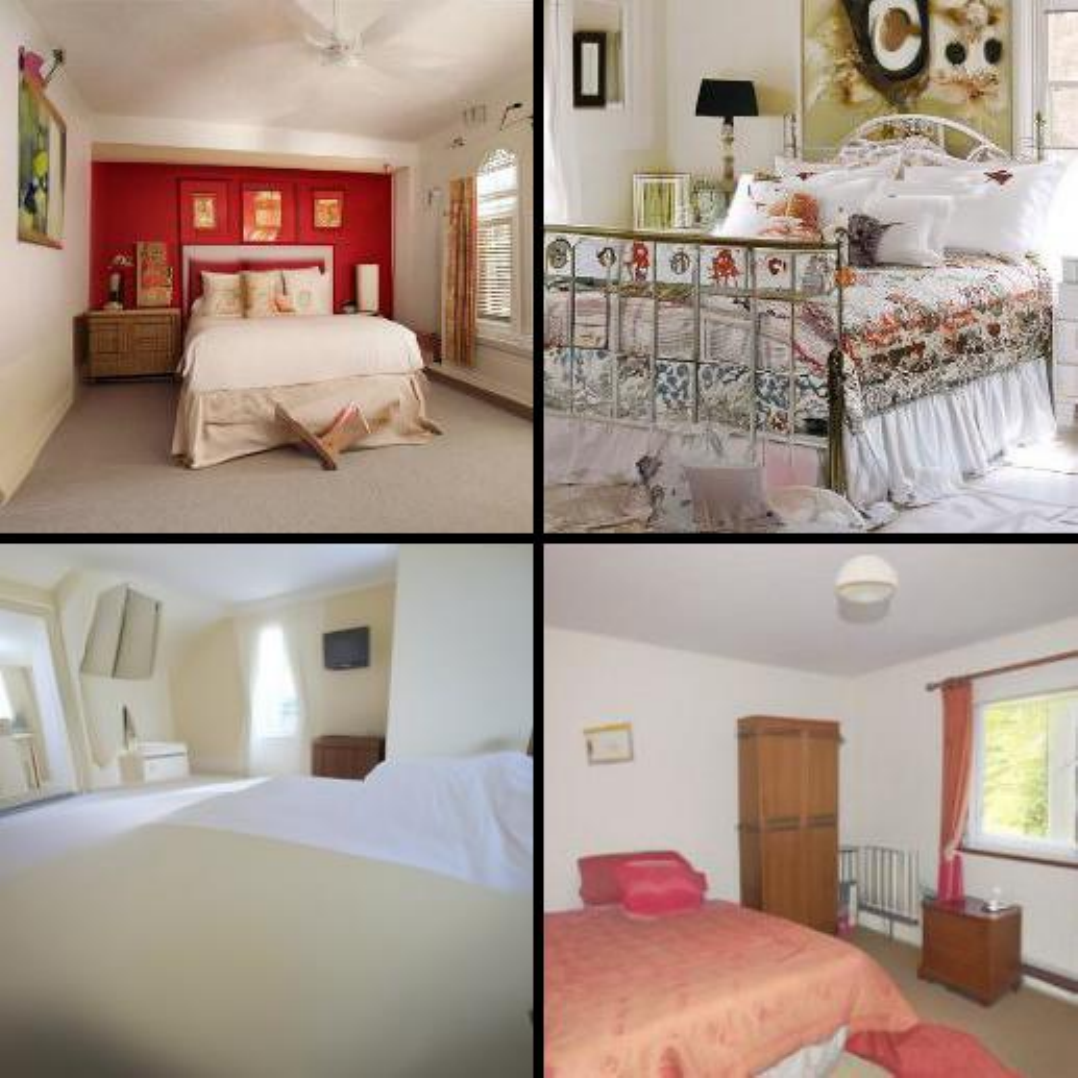}
\\
\multirow{-8.3}{*}{\parbox{0.8cm}{\centering DPM-Solver \cite{lu2022dpm}}}
& \includegraphics[width=0.216\textwidth]{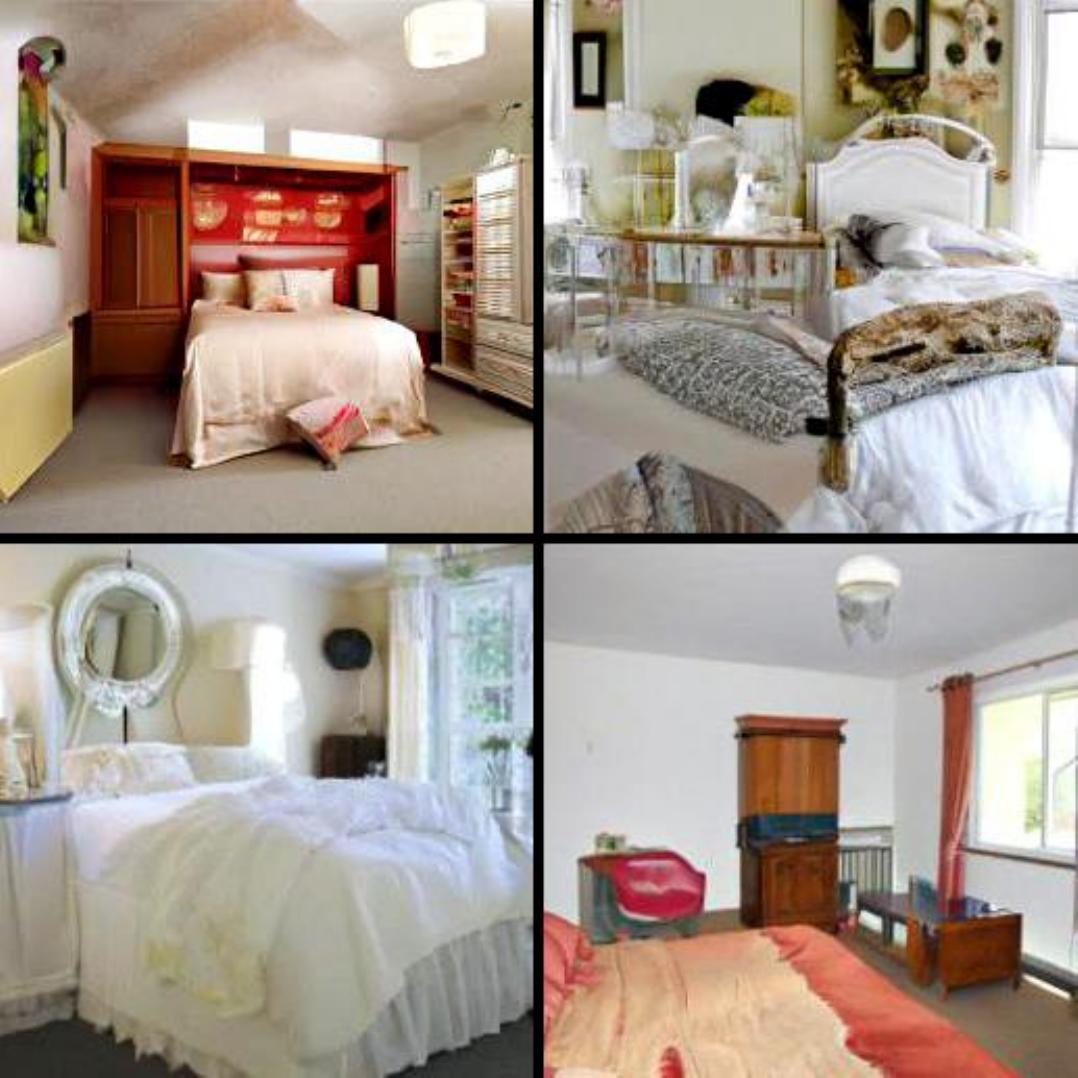} & \includegraphics[width=0.216\textwidth]{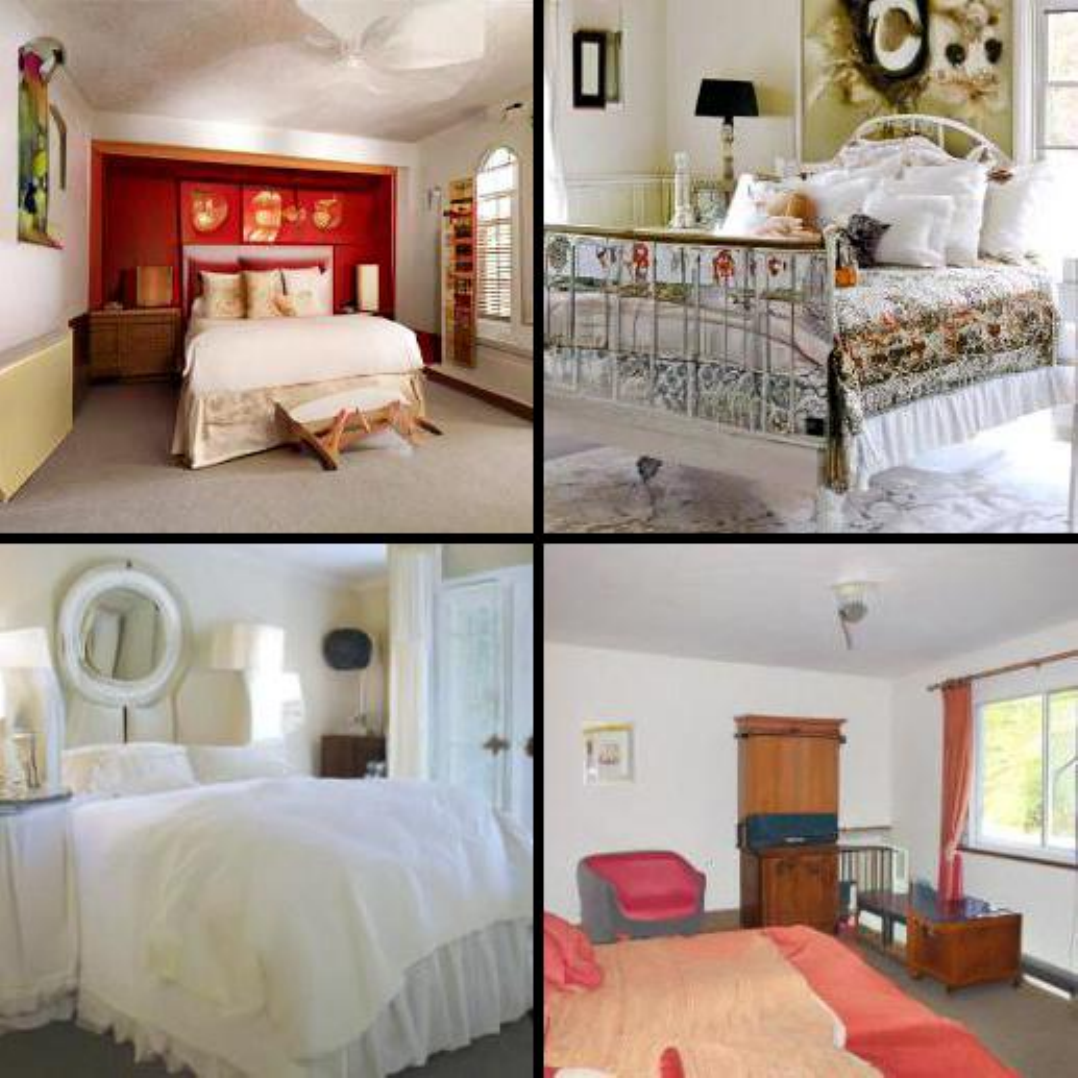} & \includegraphics[width=0.216\textwidth]{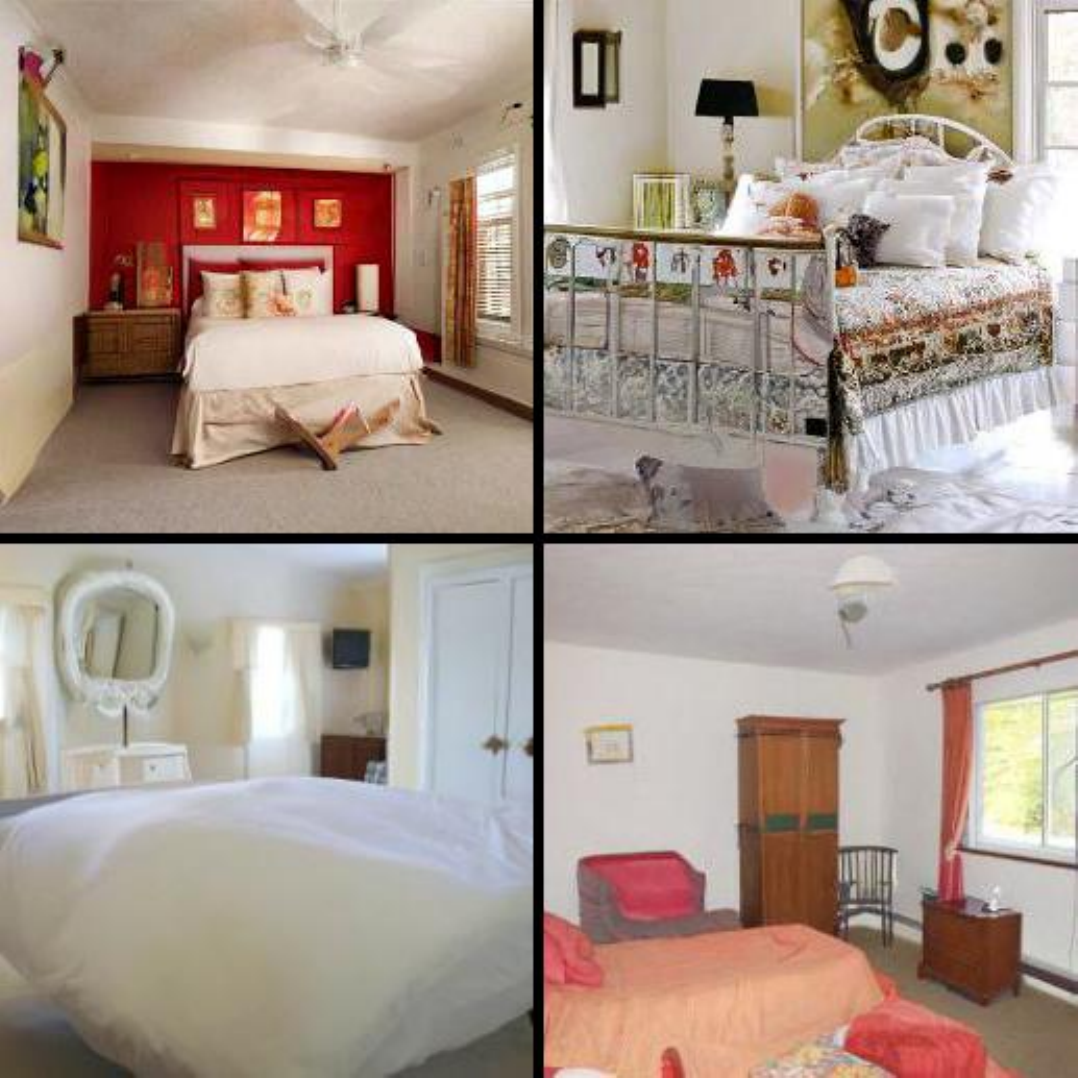} & \includegraphics[width=0.216\textwidth]{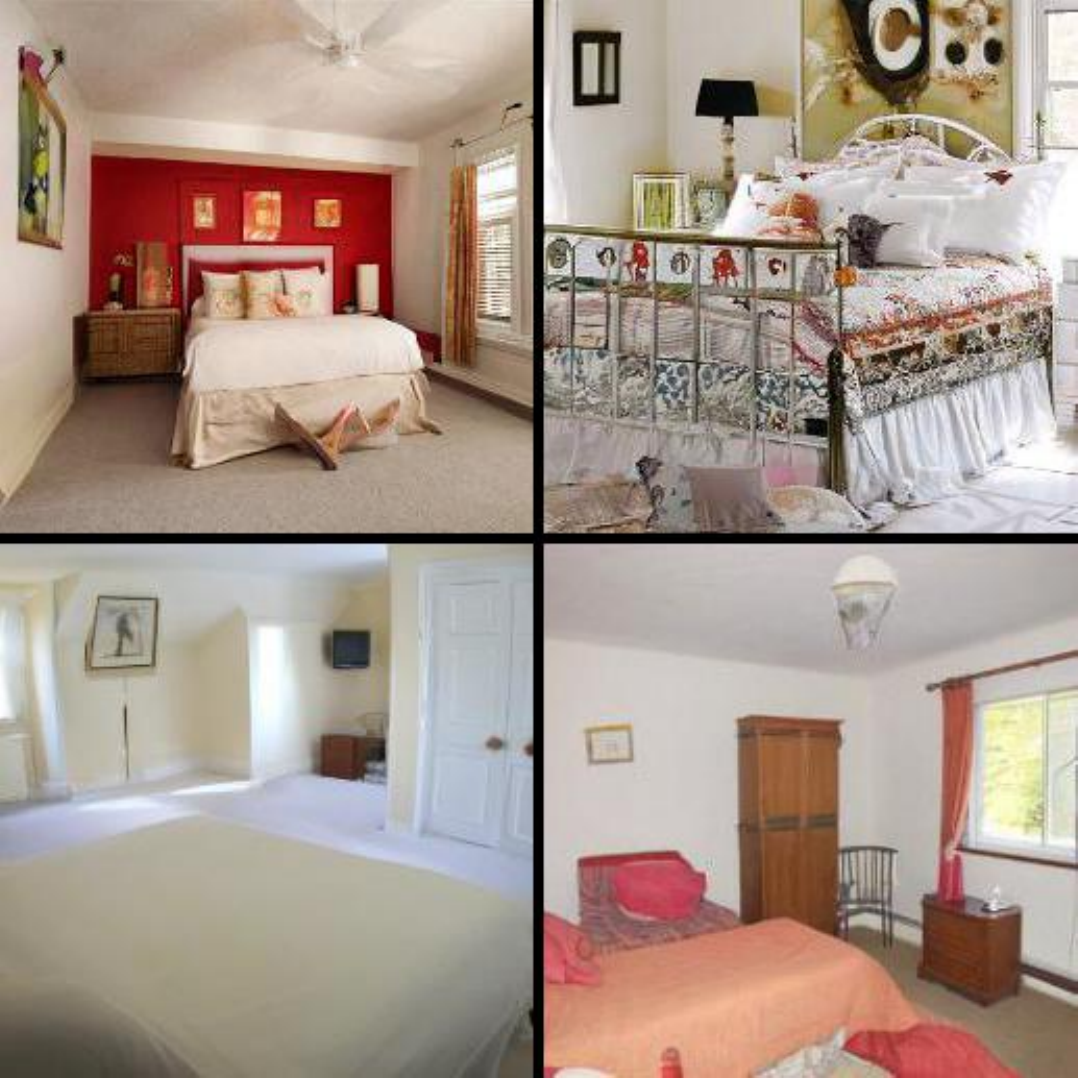}
\\
\multirow{-8.3}{*}{\parbox{0.8cm}{\centering SciRE-Solver (ours)}}
&\includegraphics[width=0.216\textwidth]{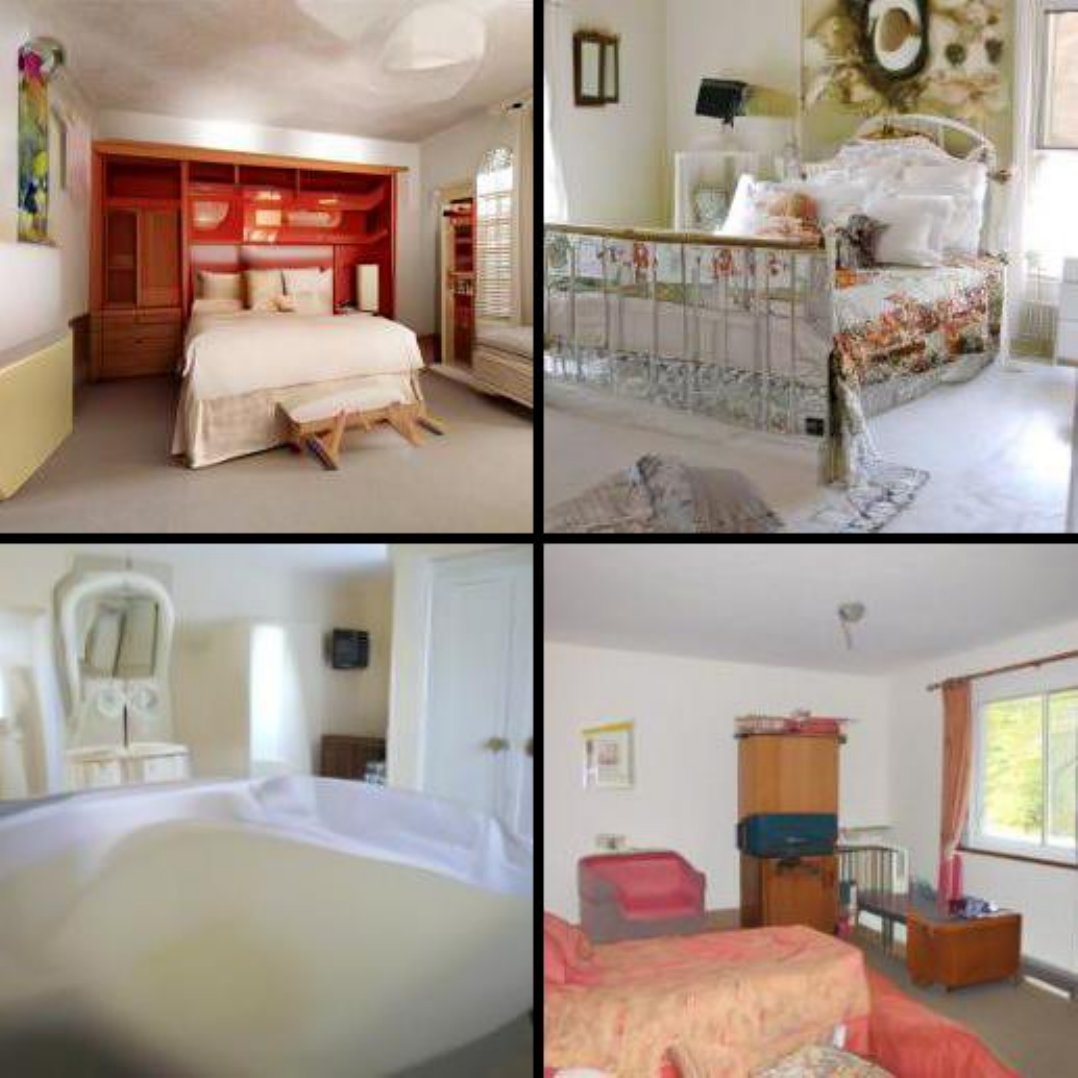} & \includegraphics[width=0.216\textwidth]{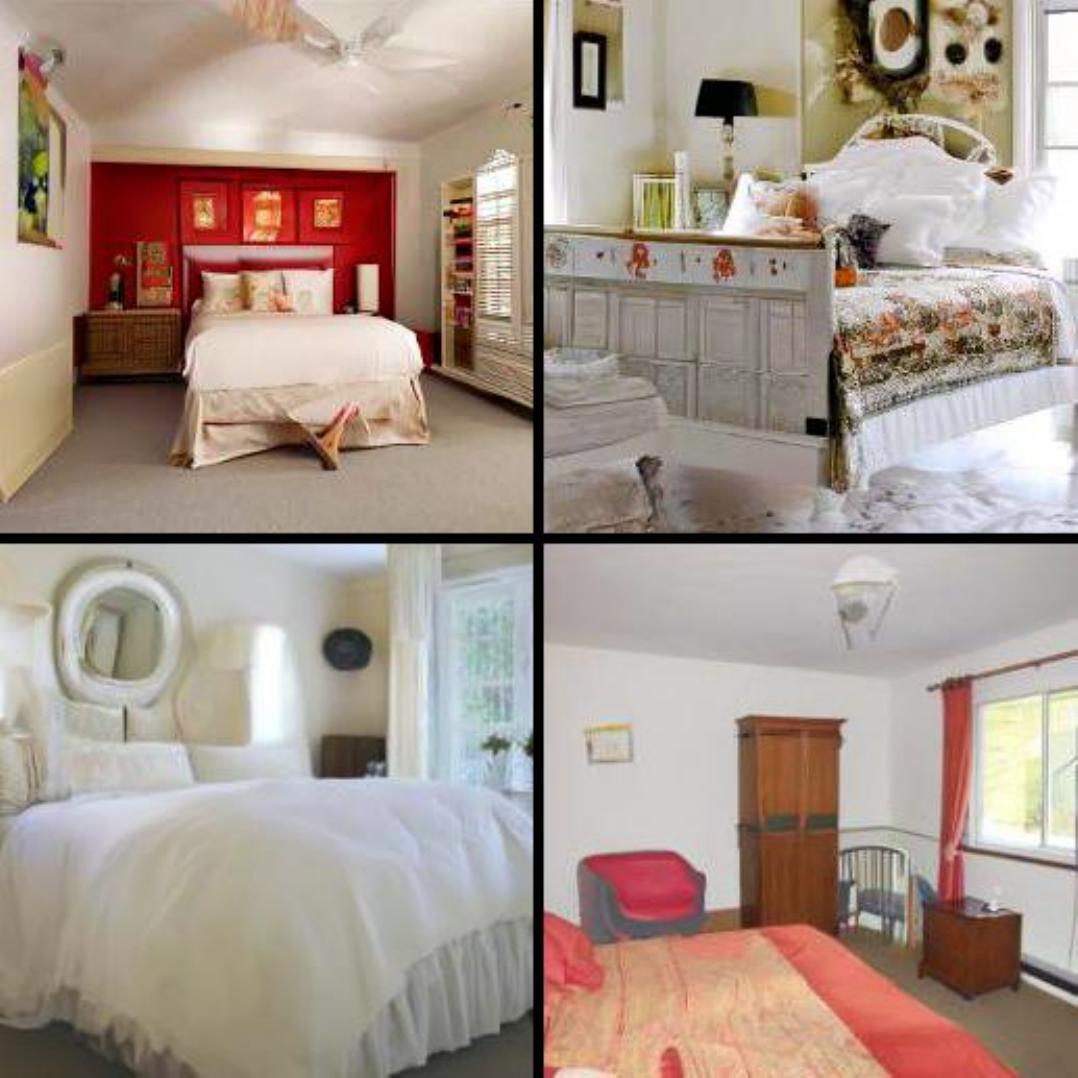} & \includegraphics[width=0.216\textwidth]{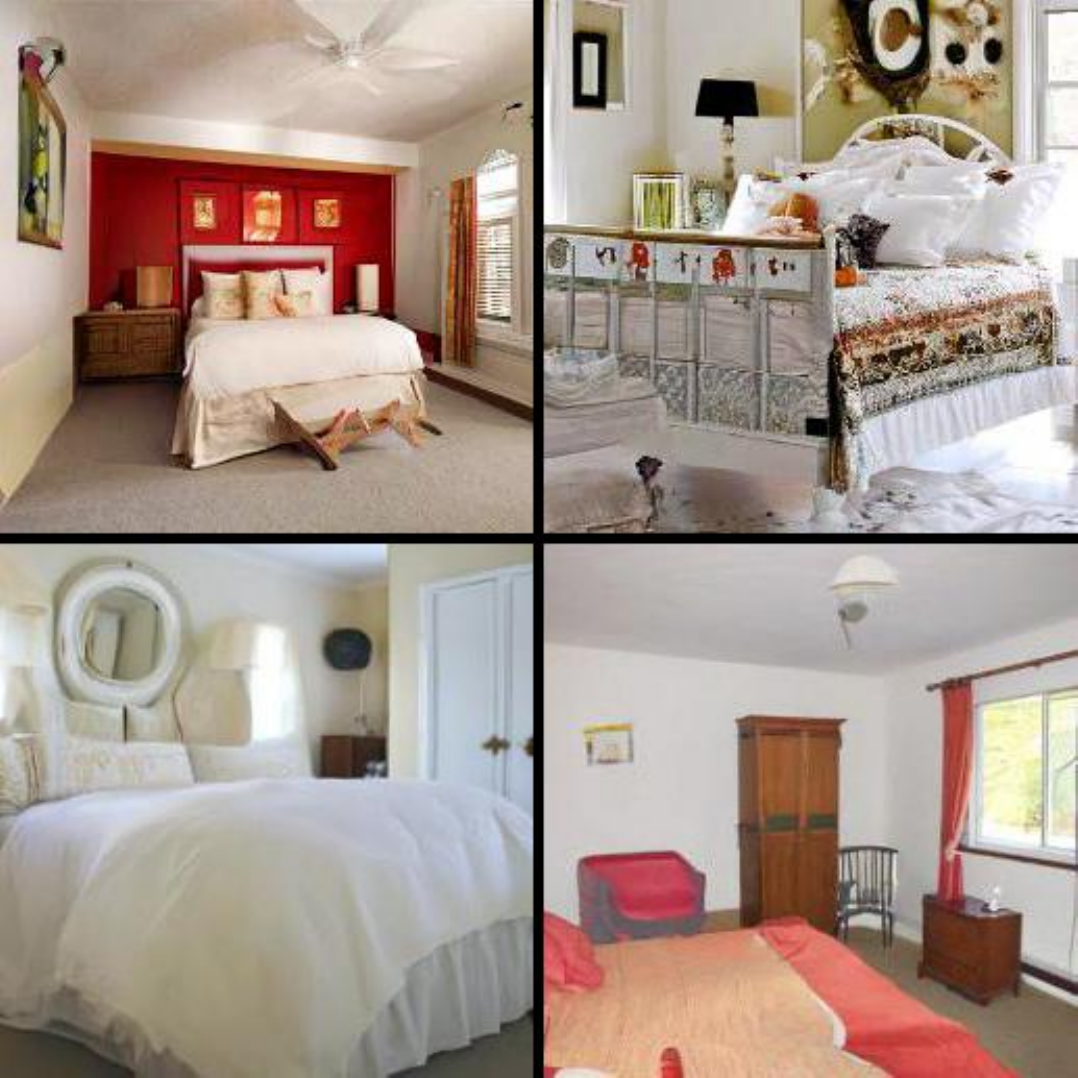} & \includegraphics[width=0.216\textwidth]{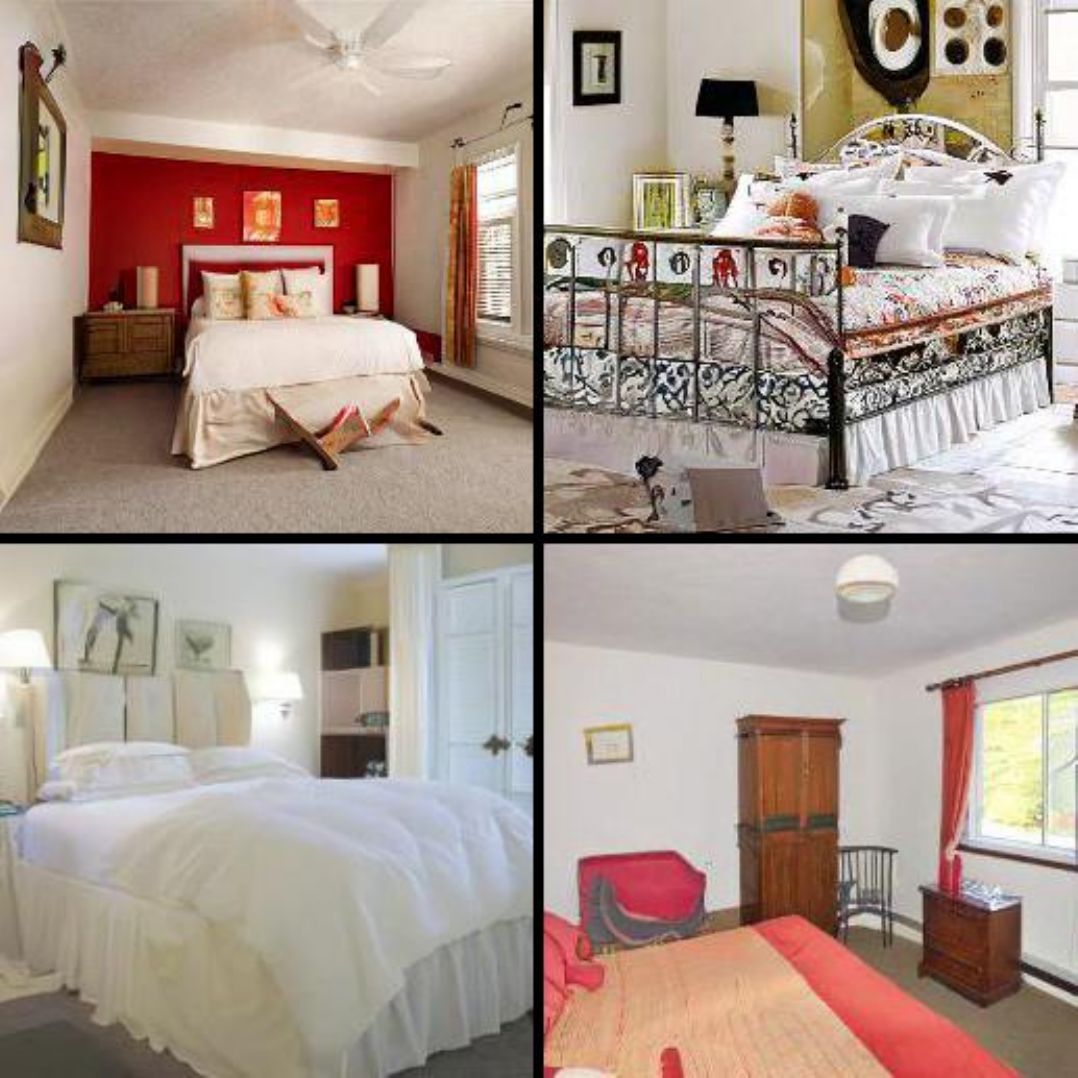} \\
\end{tabular}
 \caption{
 Random samples with the same random seed were generated by  DDIM \cite{song2021denoising} (uniform time steps), DPM-Solver \cite{lu2022dpm} (logSNR time steps), and SciRE-Solver (SNR time steps, $k=3.1$), employing the pre-trained discrete-time DPM \cite{dhariwal2021diffusion} on LSUN bedroom 256$\times$256.
 }
\label{fig:lsunbedroom256comparison}
\end{figure}

\begin{figure}[ht]
\centering
\begin{tabular}{m{0.8cm}p{2.7cm}p{2.7cm}p{2.7cm}p{2.7cm}}
   ~~ &~~~~~~~~~~NFE=$10$& ~~~~~~~~~~NFE=$15$  &~~~~~~~~~~NFE=$20$ &~~~~~~~~~~NFE=$50$ \\
\multirow{-8.3}{*}{\parbox{0.8cm}{\centering DDIM \cite{song2021denoising}}}
& \includegraphics[width=0.216\textwidth]{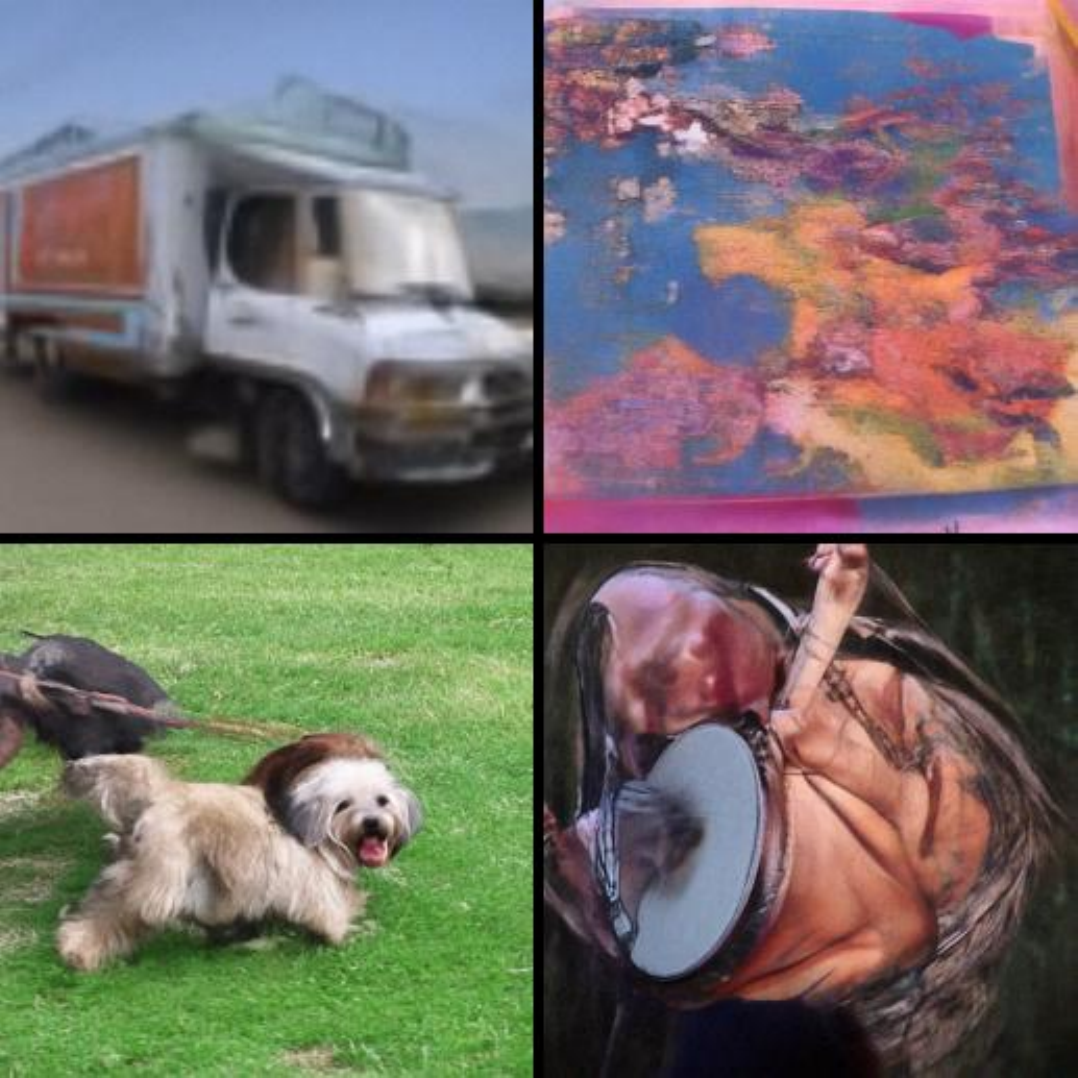} & \includegraphics[width=0.216\textwidth]{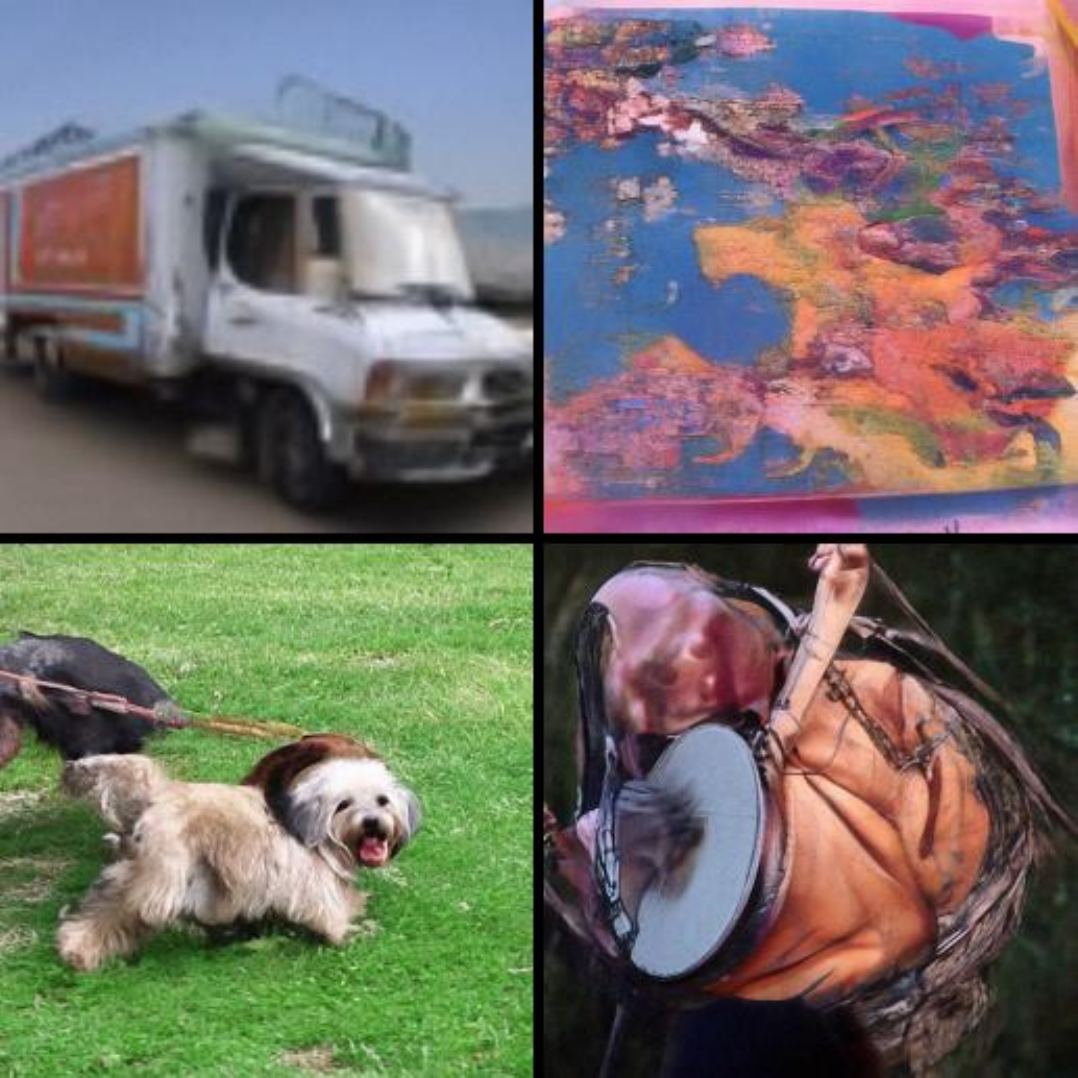} & \includegraphics[width=0.216\textwidth]{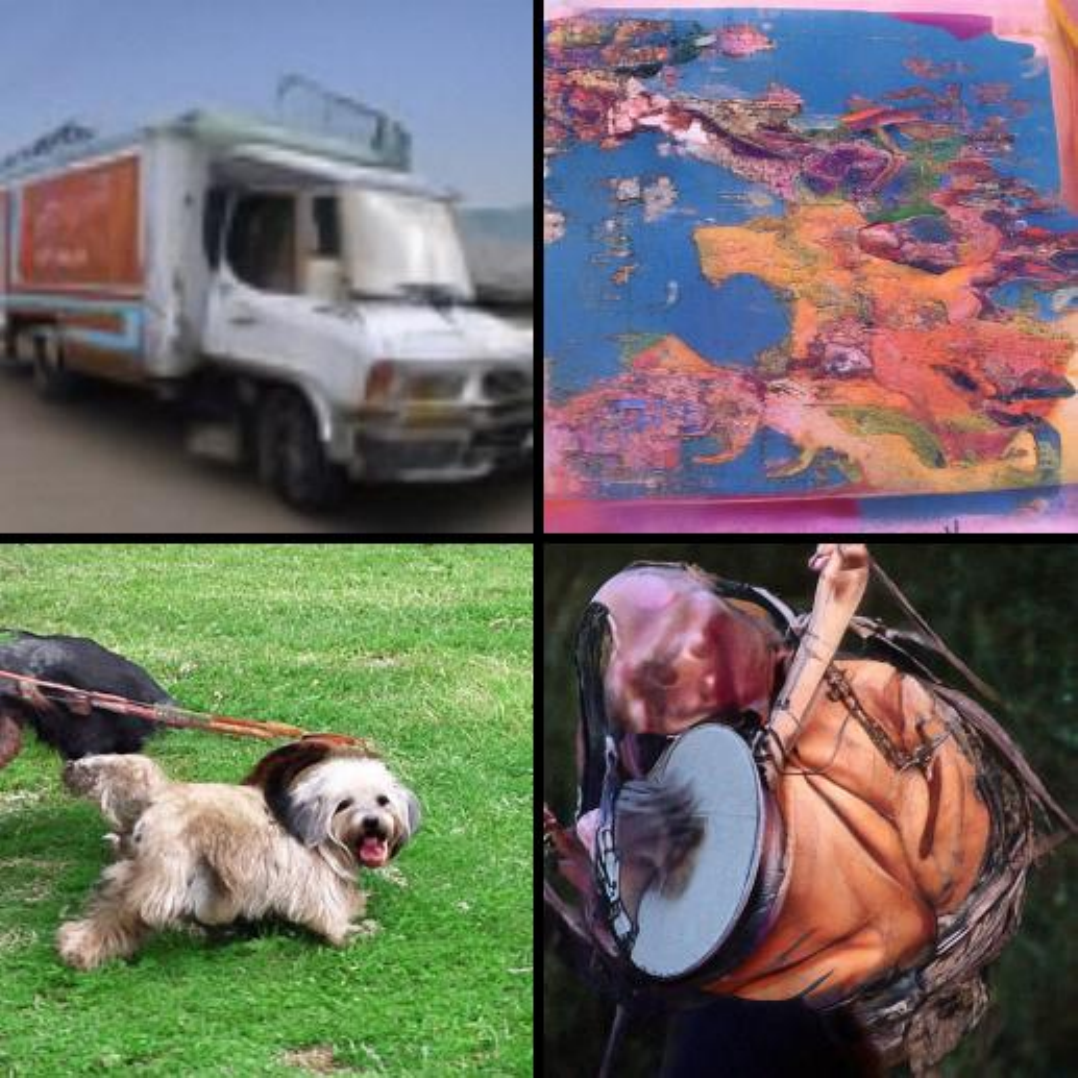} & \includegraphics[width=0.216\textwidth]{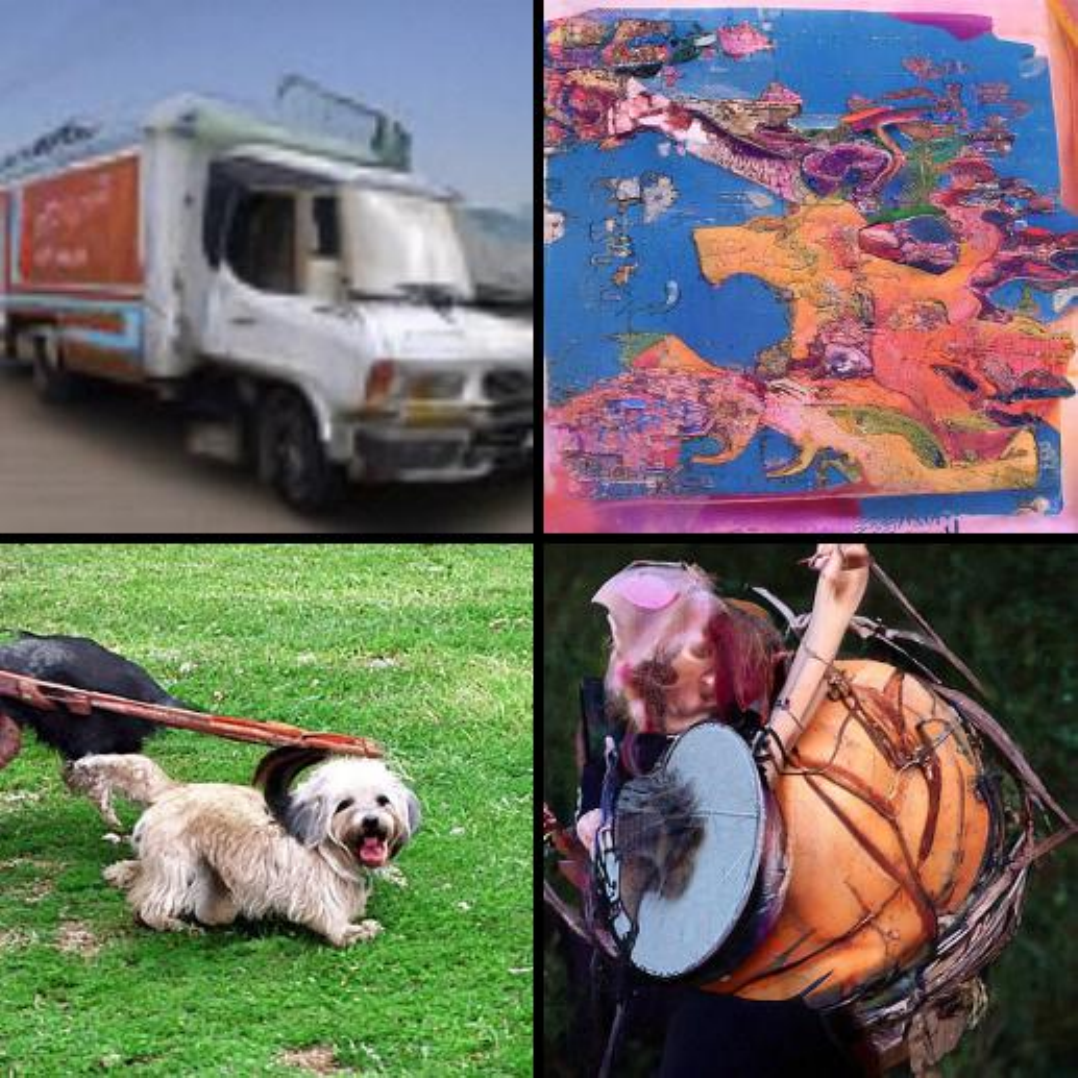}
\\
\multirow{-8.3}{*}{\parbox{0.8cm}{\centering DPM-Solver \cite{lu2022dpm}}}
& \includegraphics[width=0.216\textwidth]{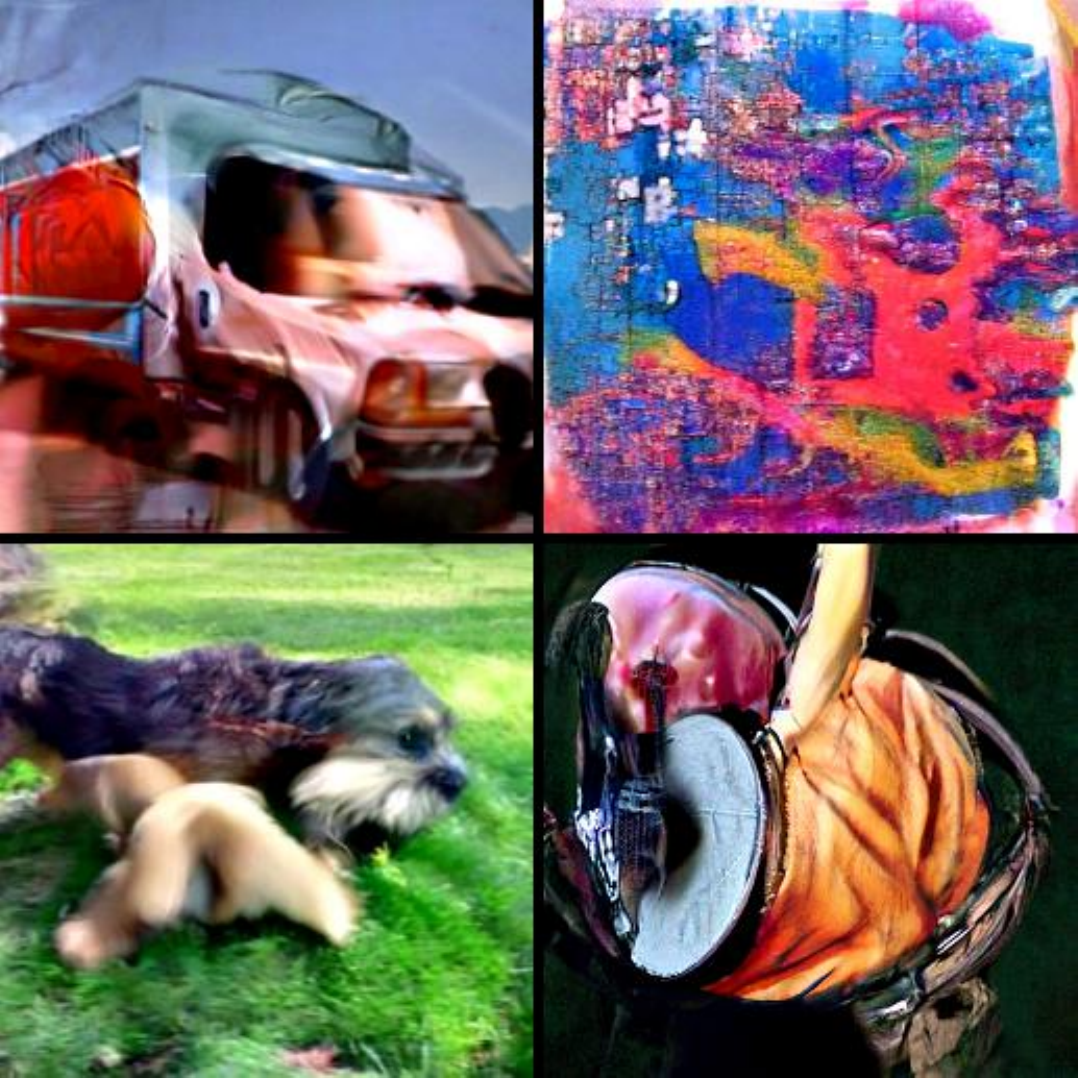} & \includegraphics[width=0.216\textwidth]{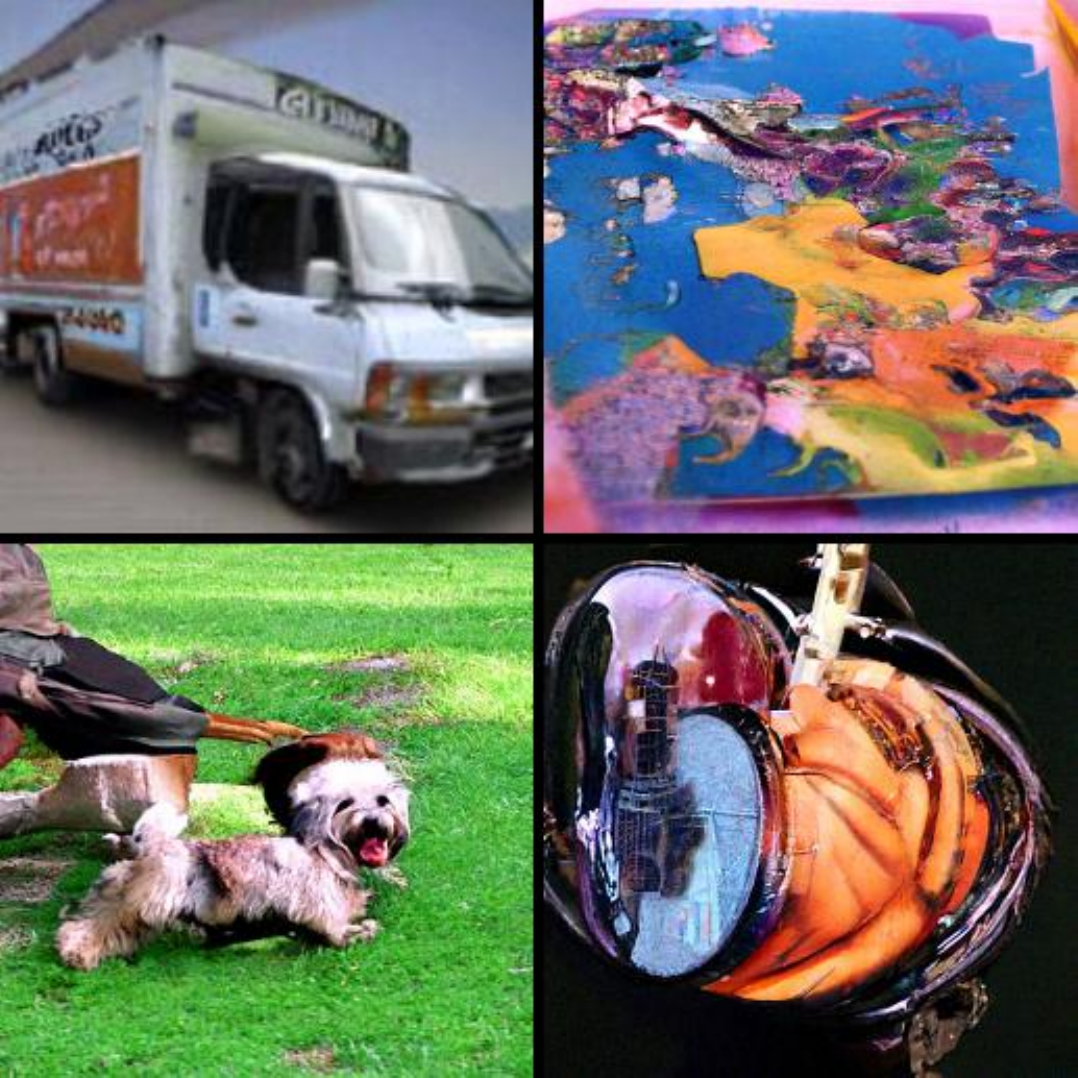} & \includegraphics[width=0.216\textwidth]{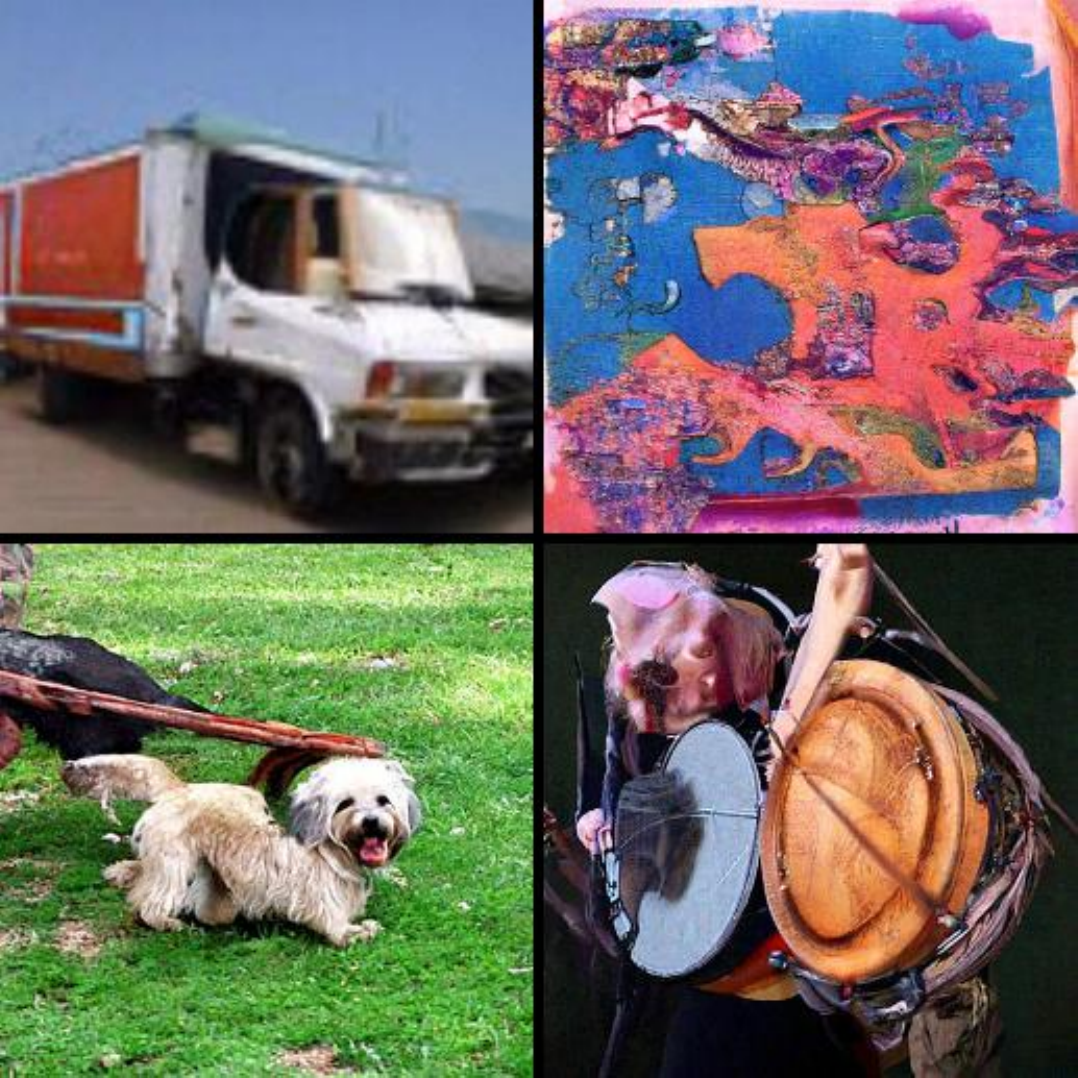} & \includegraphics[width=0.216\textwidth]{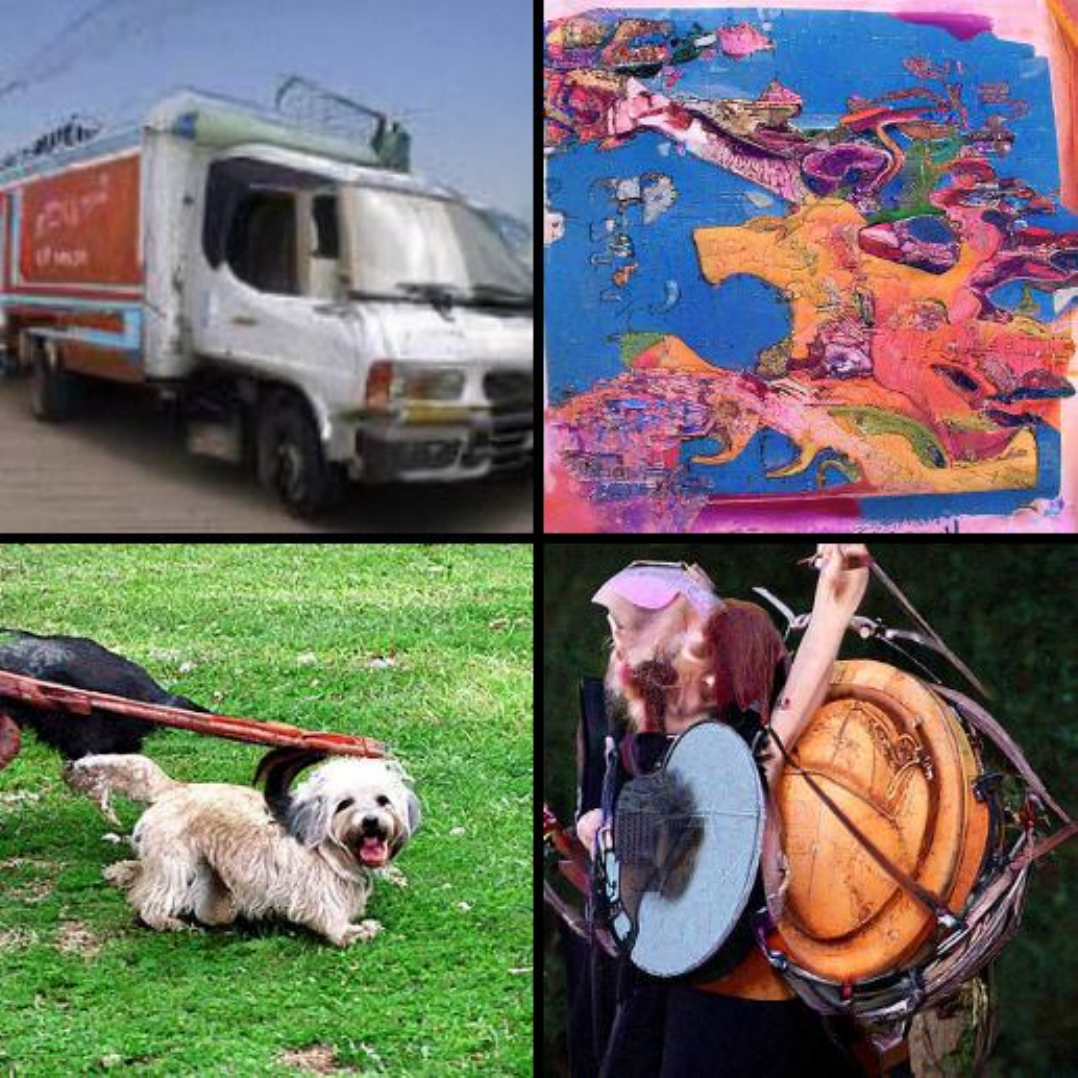}
\\
\multirow{-8.3}{*}{\parbox{0.8cm}{\centering SciRE-Solver (ours)}}
&\includegraphics[width=0.216\textwidth]{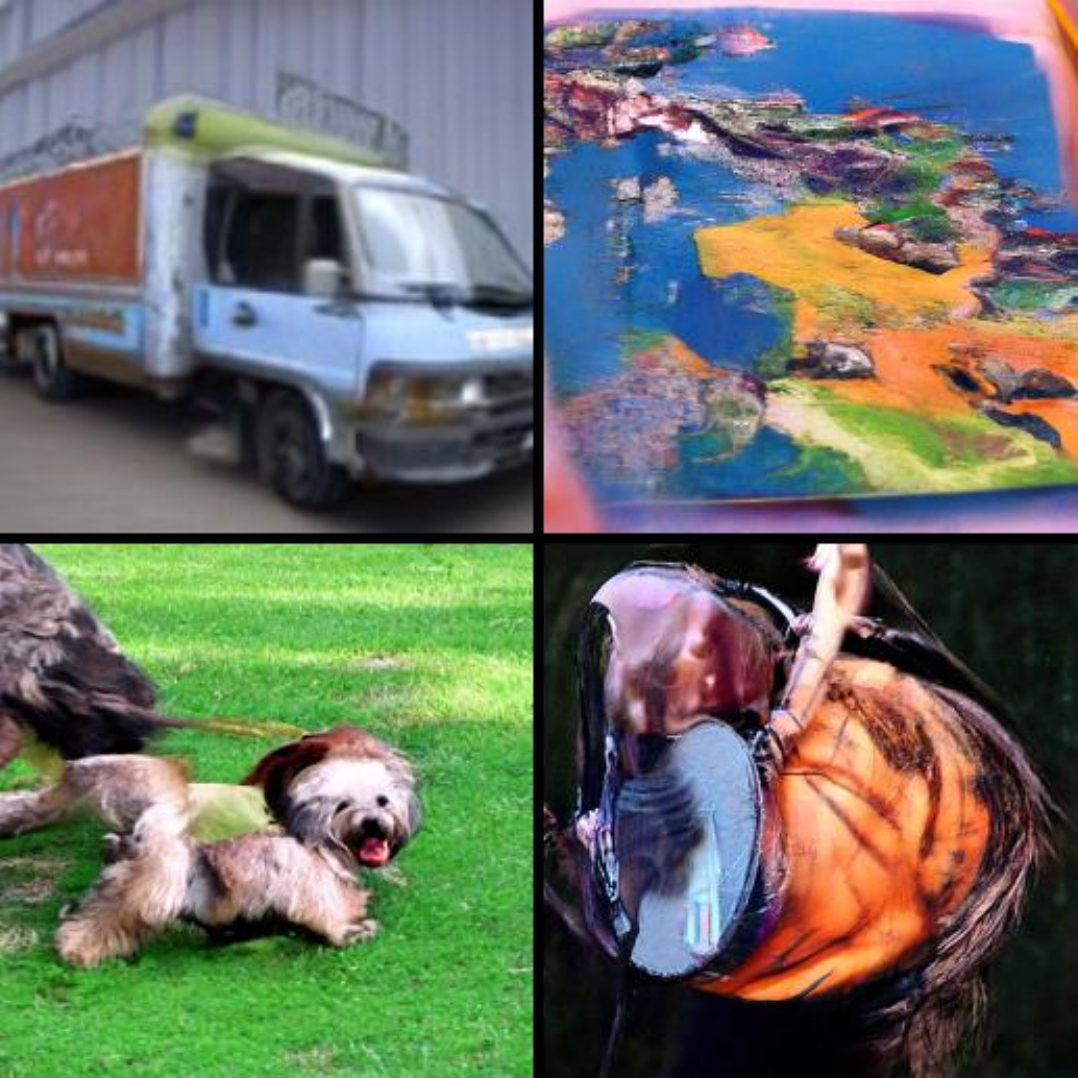} & \includegraphics[width=0.216\textwidth]{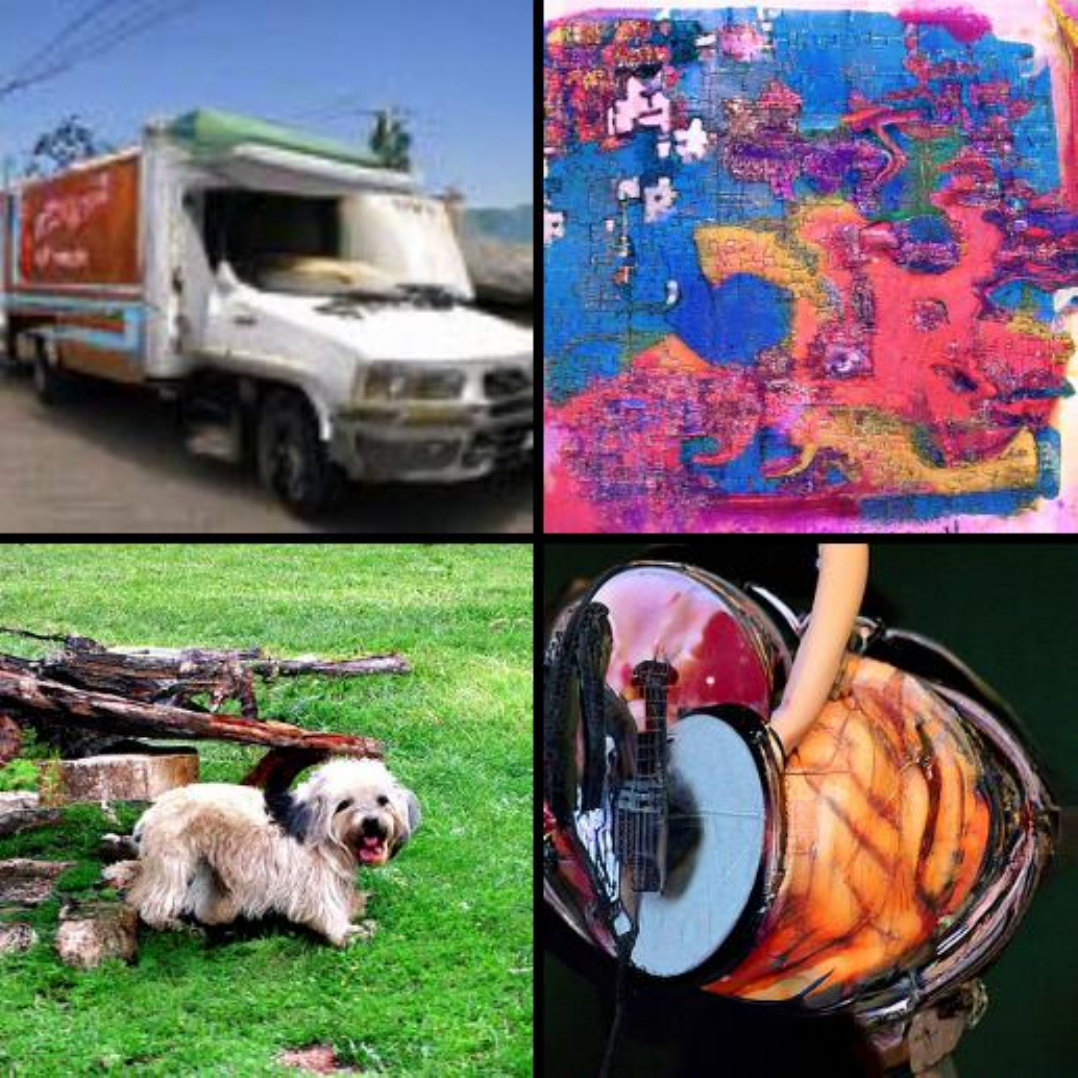} &
\includegraphics[width=0.216\textwidth]{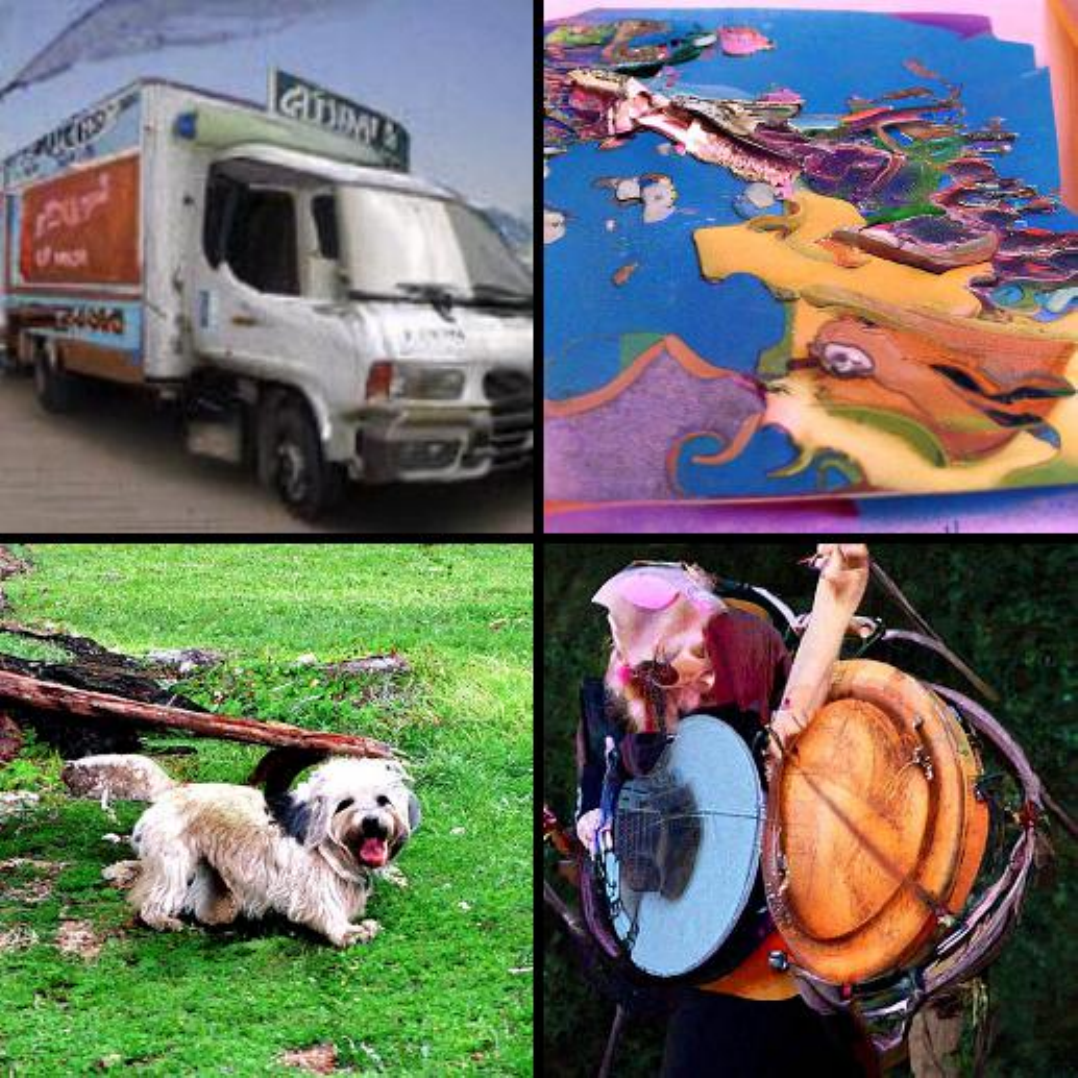} & \includegraphics[width=0.216\textwidth]{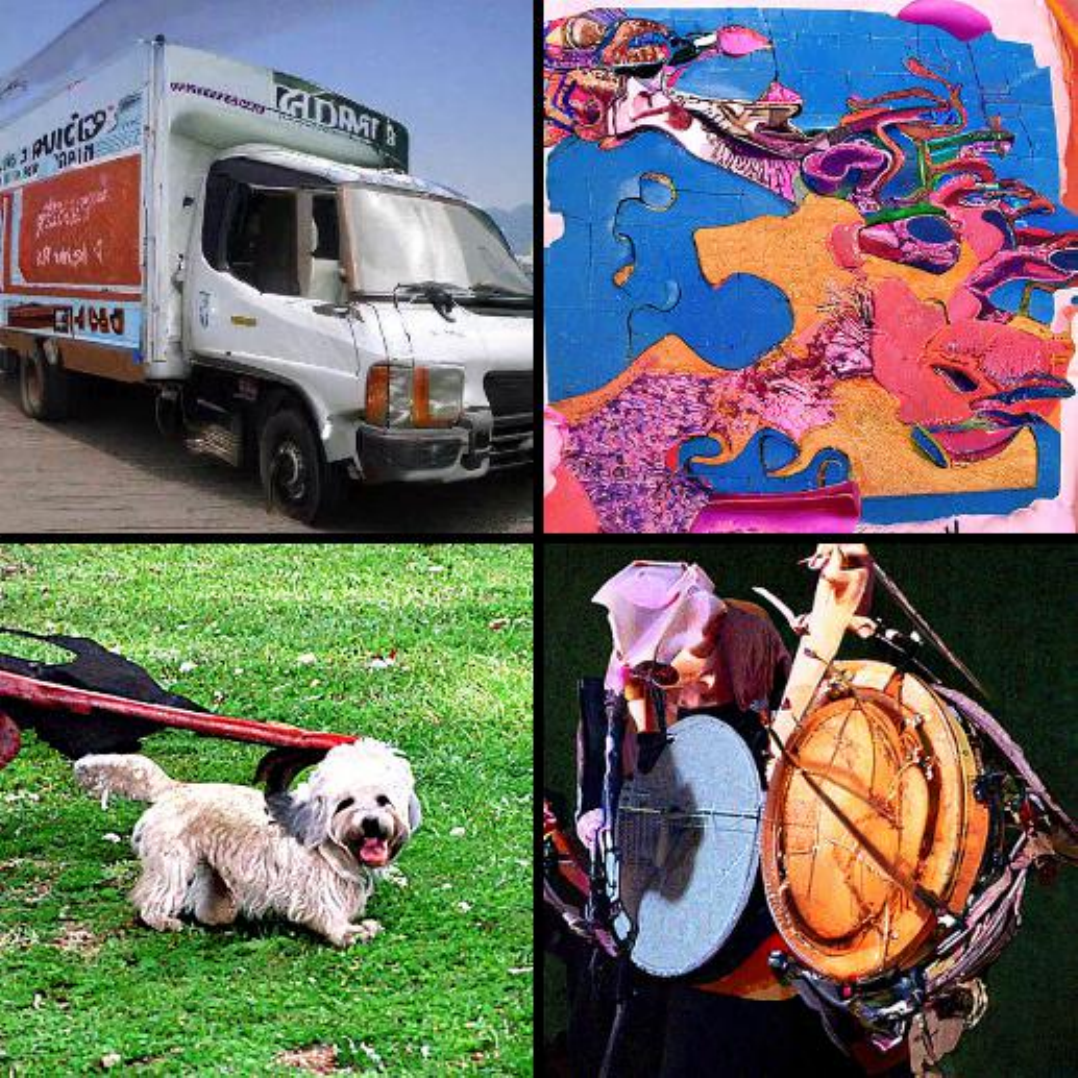} \\
\end{tabular}
 \caption{
 Random samples with the same random seed were generated by  DDIM \cite{song2021denoising} (uniform time steps), DPM-Solver \cite{lu2022dpm} (logSNR time steps), and SciRE-Solver (SNR time steps, $k=3.1$), employing the pre-trained
 DPM \cite{dhariwal2021diffusion} on ImageNet 256$\times$256 (classifier scale: 2.5).
 }
\label{fig:imagenet256comparison}
\end{figure}

\begin{figure}[ht]
\centering
\begin{tabular}{m{0.8cm}p{2.7cm}p{2.7cm}p{2.7cm}p{2.7cm}}
   ~~ &~~~~~~~~~~NFE=$10$& ~~~~~~~~~~NFE=$15$  &~~~~~~~~~~NFE=$20$ &~~~~~~~~~~NFE=$50$ \\
\multirow{-8.3}{*}{\parbox{0.8cm}{\centering DDIM \cite{song2021denoising}}}
& \includegraphics[width=0.216\textwidth]{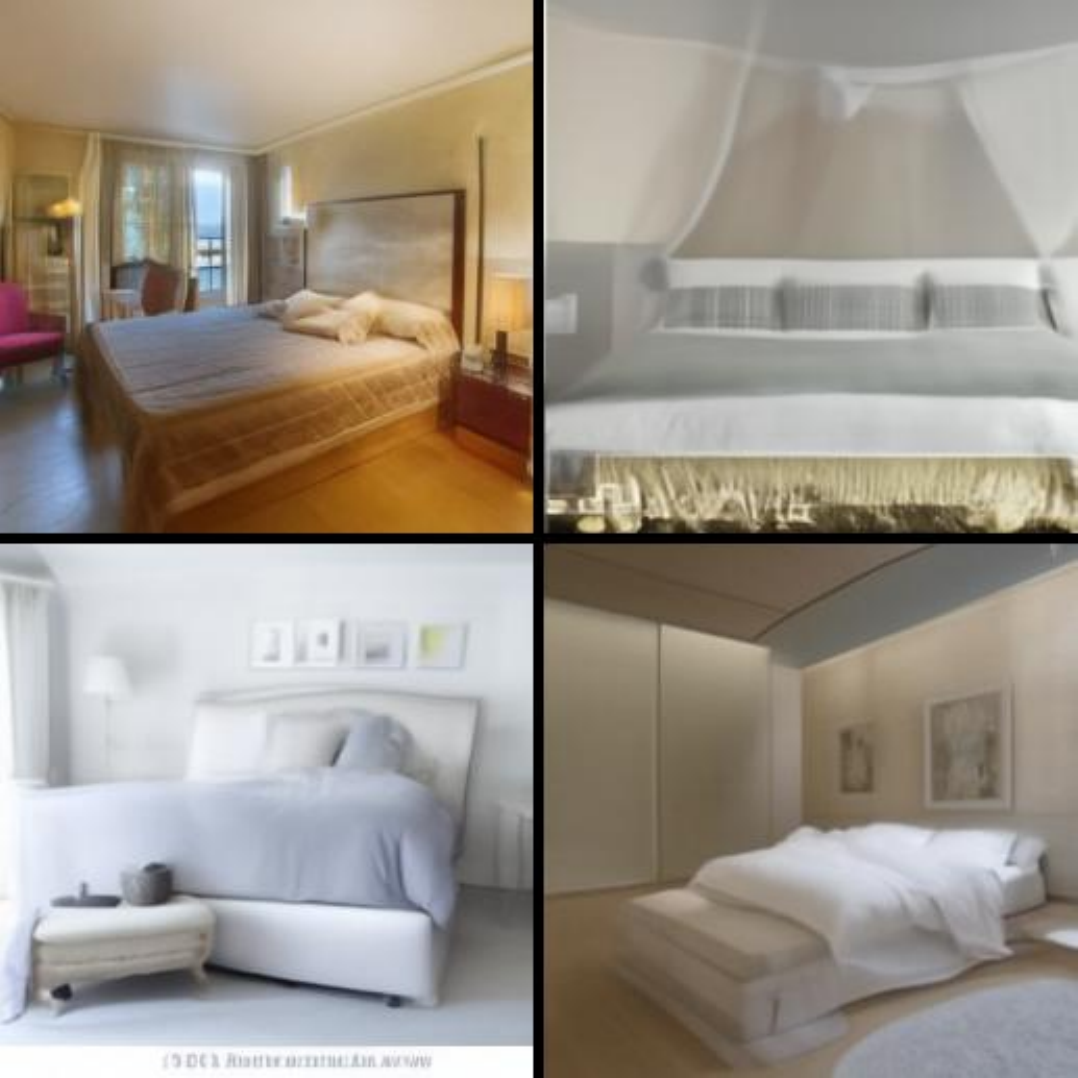} & \includegraphics[width=0.216\textwidth]{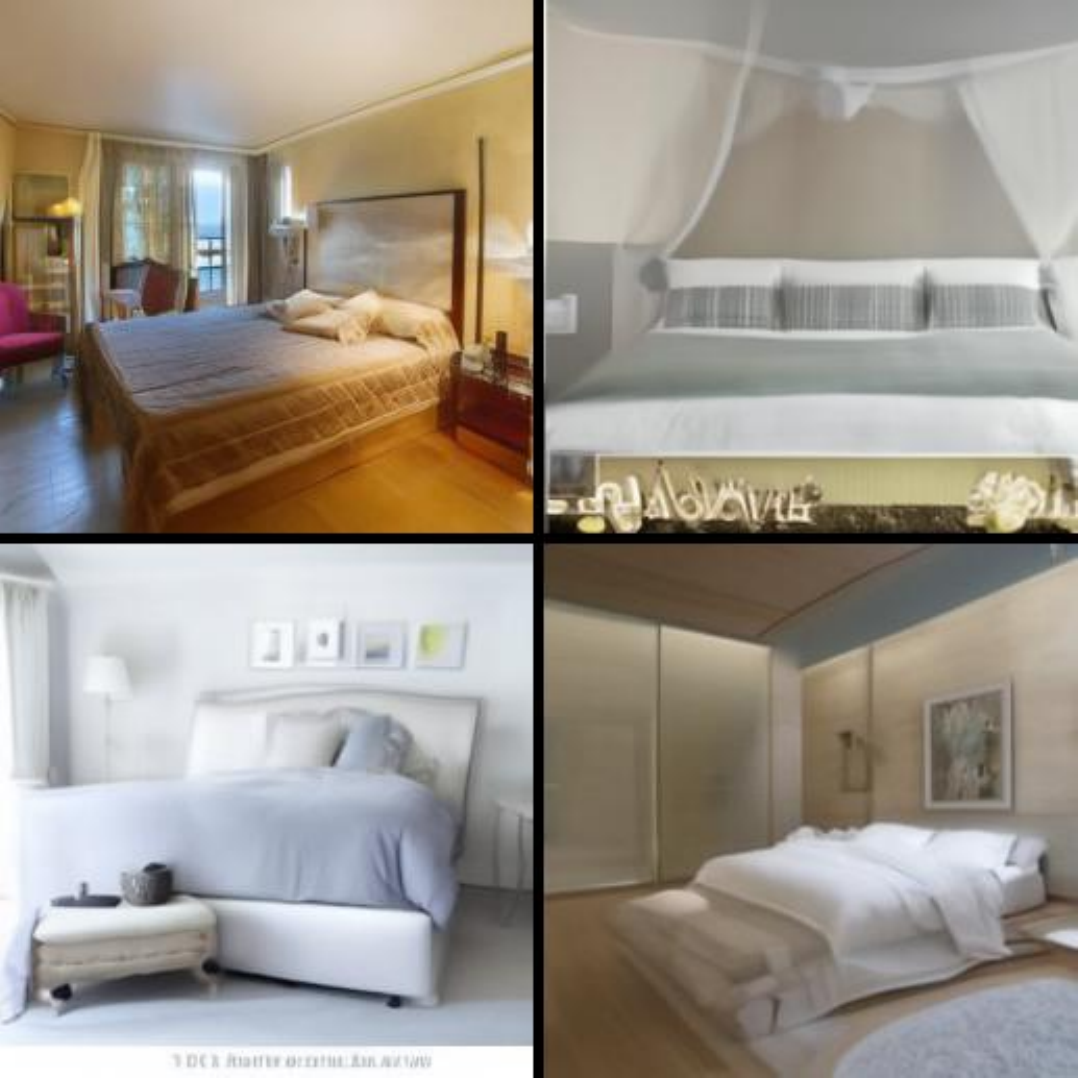} & \includegraphics[width=0.216\textwidth]{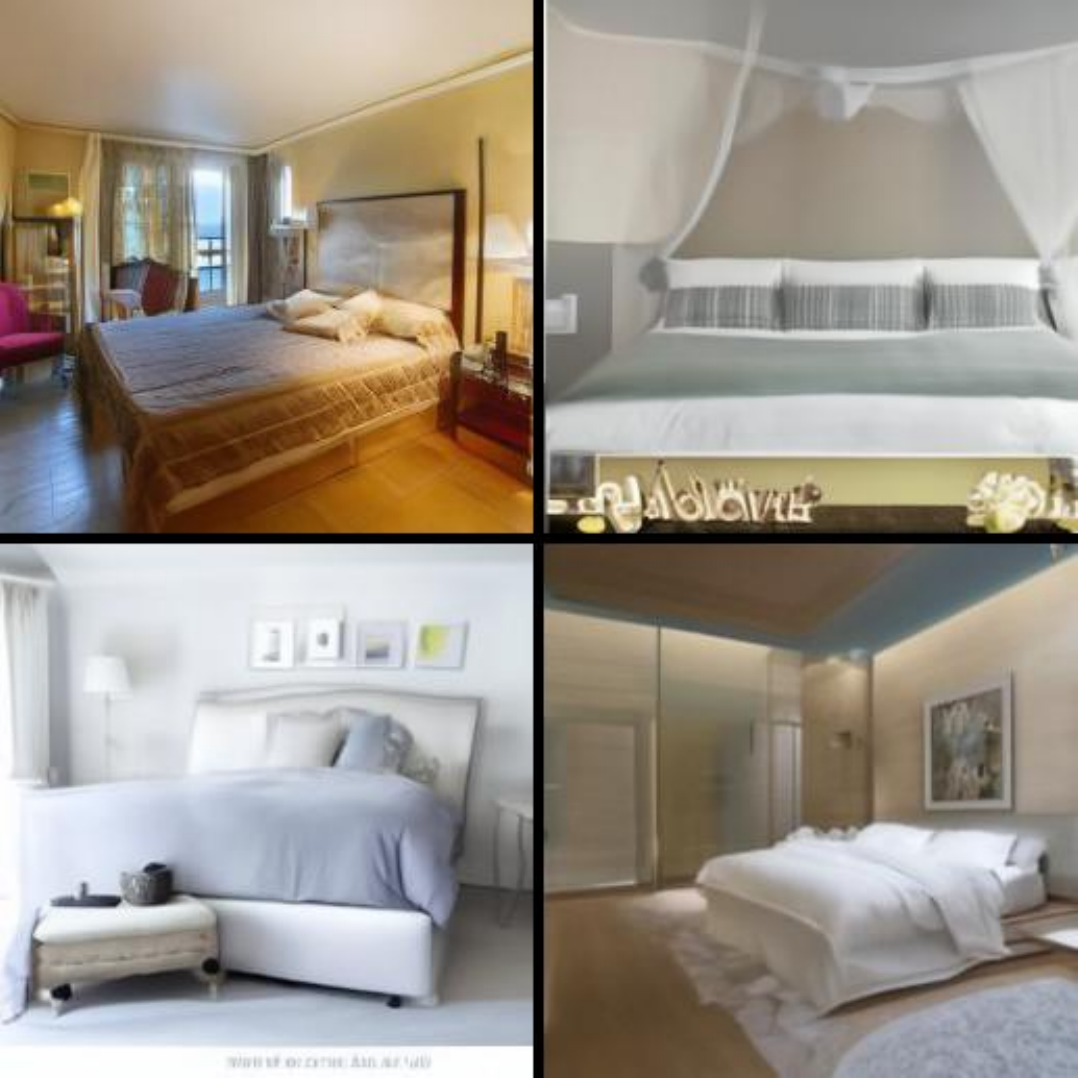} & \includegraphics[width=0.216\textwidth]{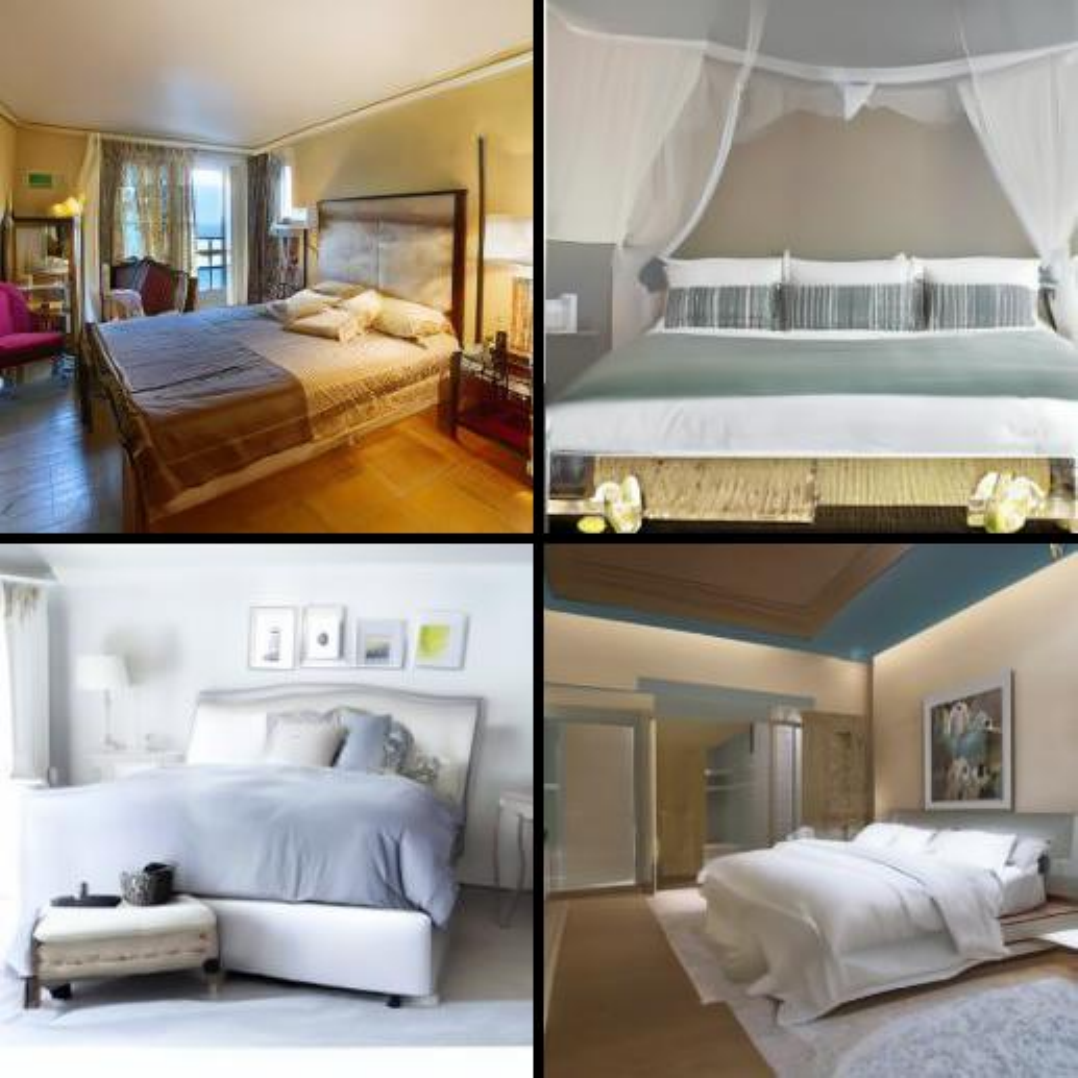}
\\
\multirow{-8.3}{*}{\parbox{0.8cm}{\centering DPM-Solver \cite{lu2022dpm}}}
& \includegraphics[width=0.216\textwidth]{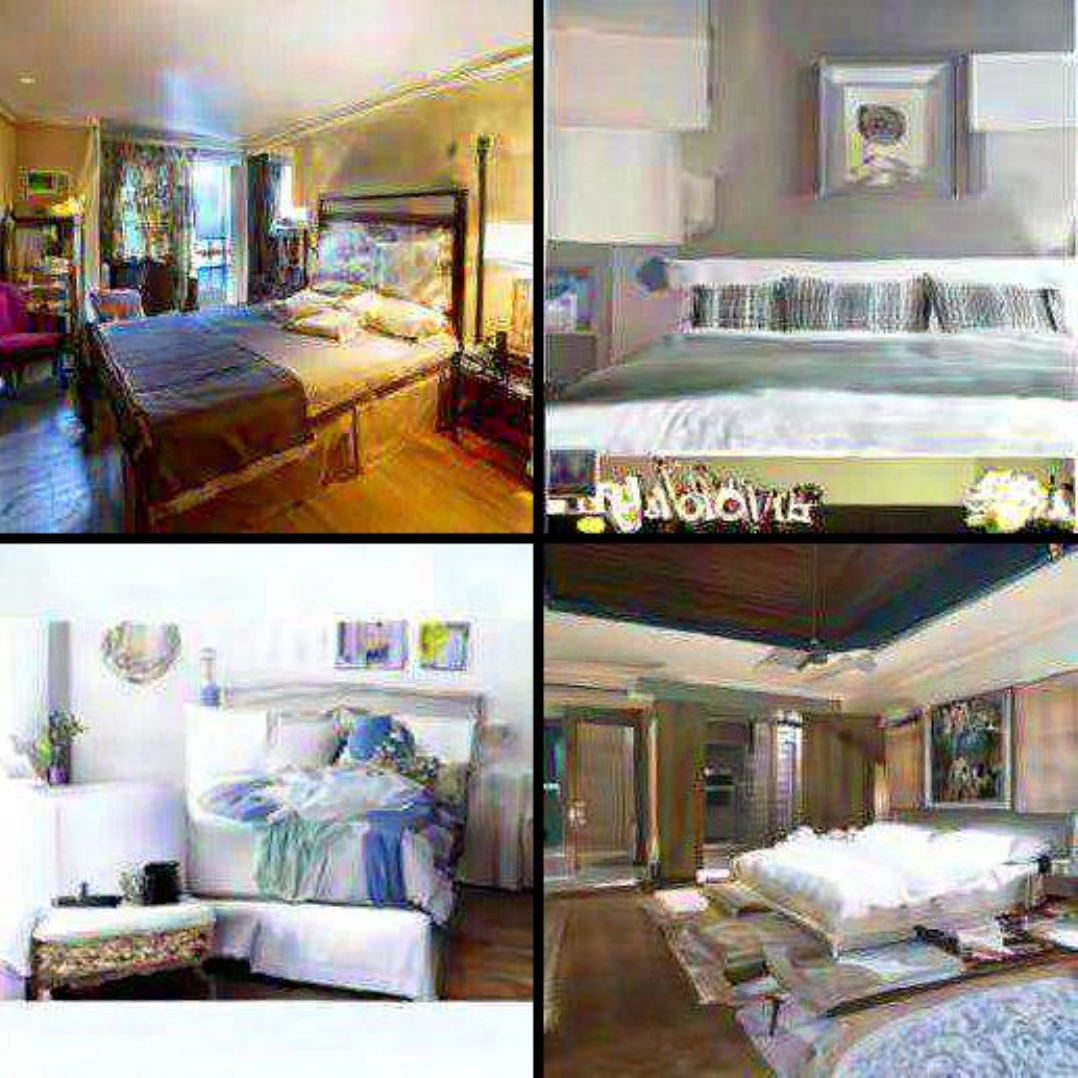} & \includegraphics[width=0.216\textwidth]{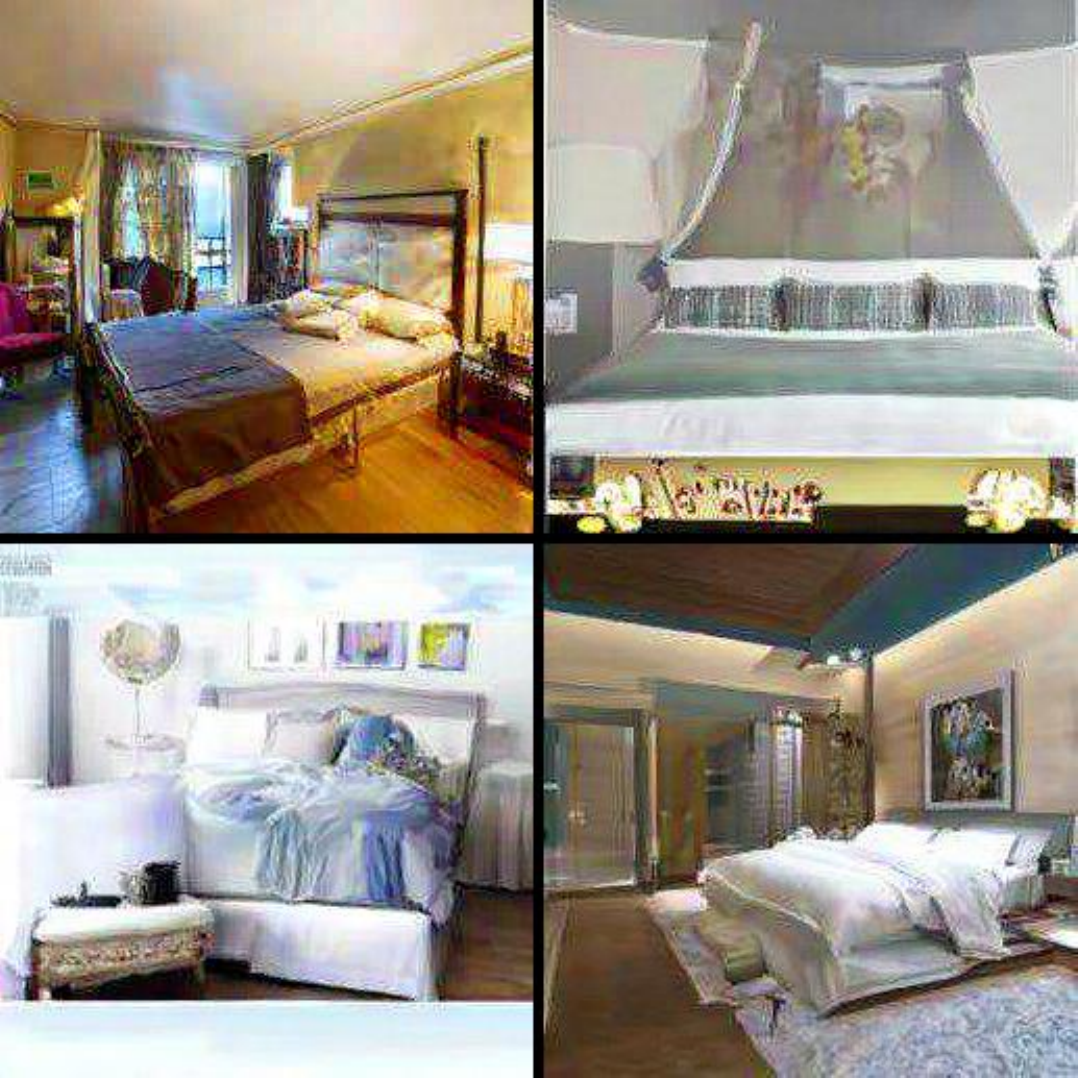} & \includegraphics[width=0.216\textwidth]{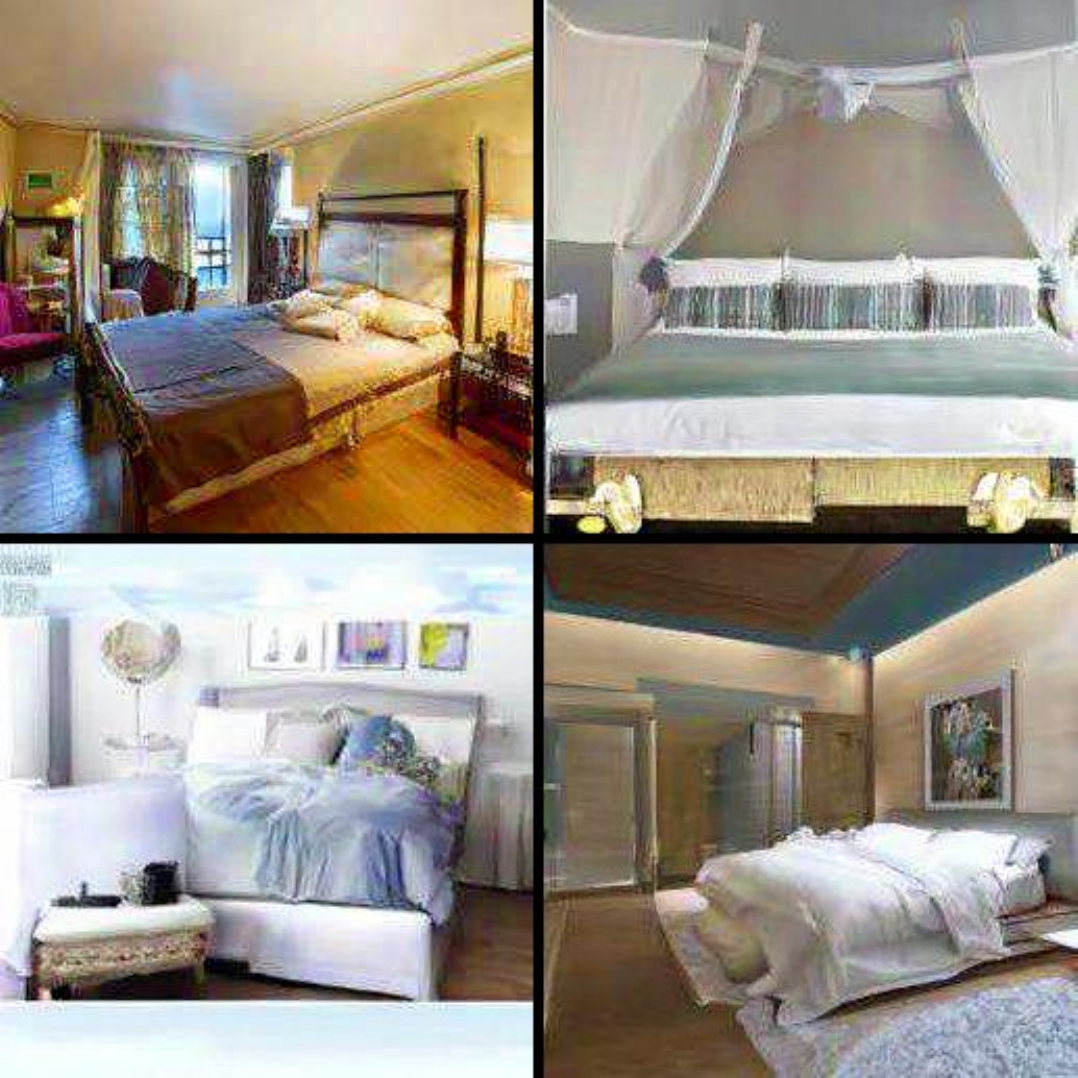} & \includegraphics[width=0.216\textwidth]{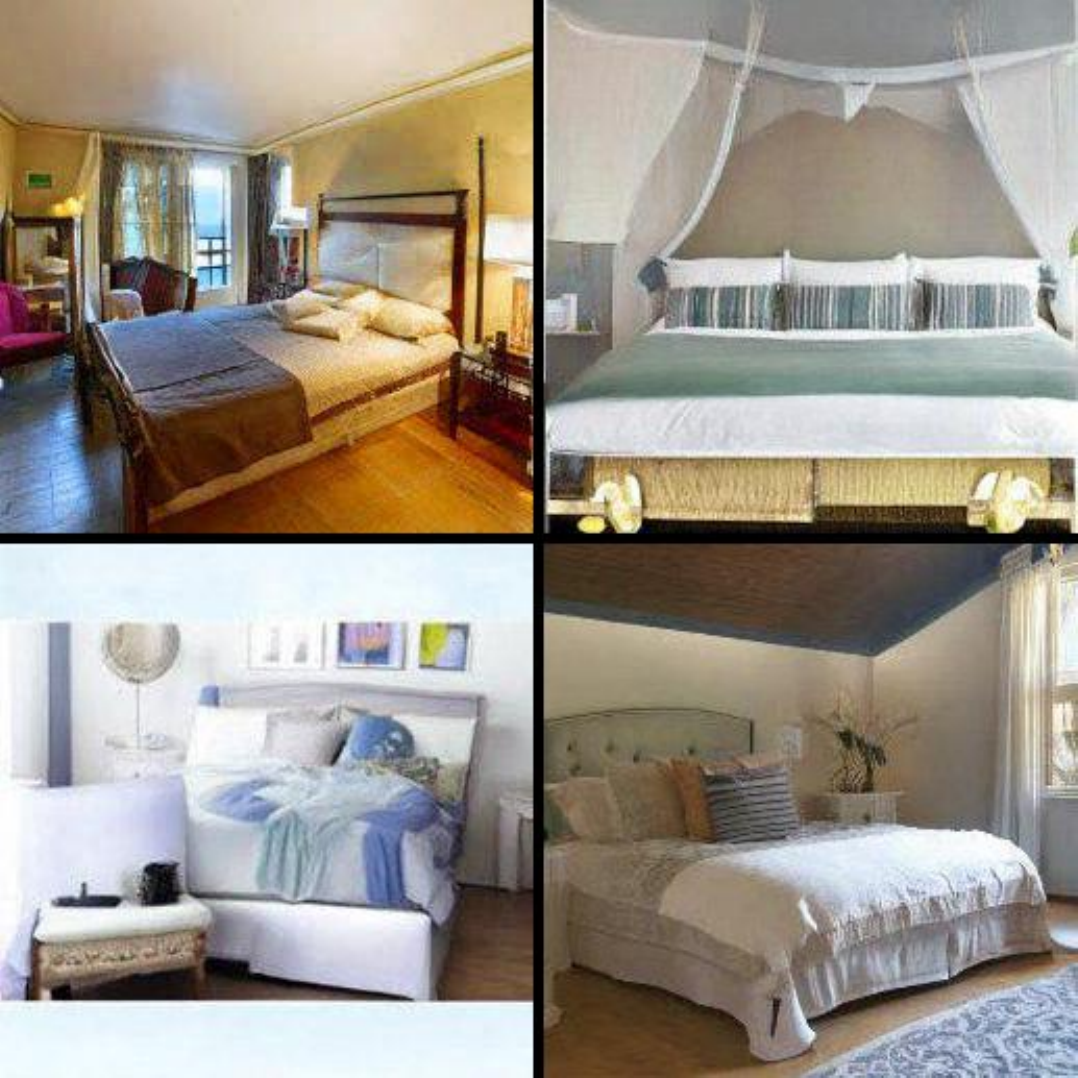}
\\
\multirow{-8.3}{*}{\parbox{0.8cm}{\centering SciRE-Solver (ours)}}
&\includegraphics[width=0.216\textwidth]{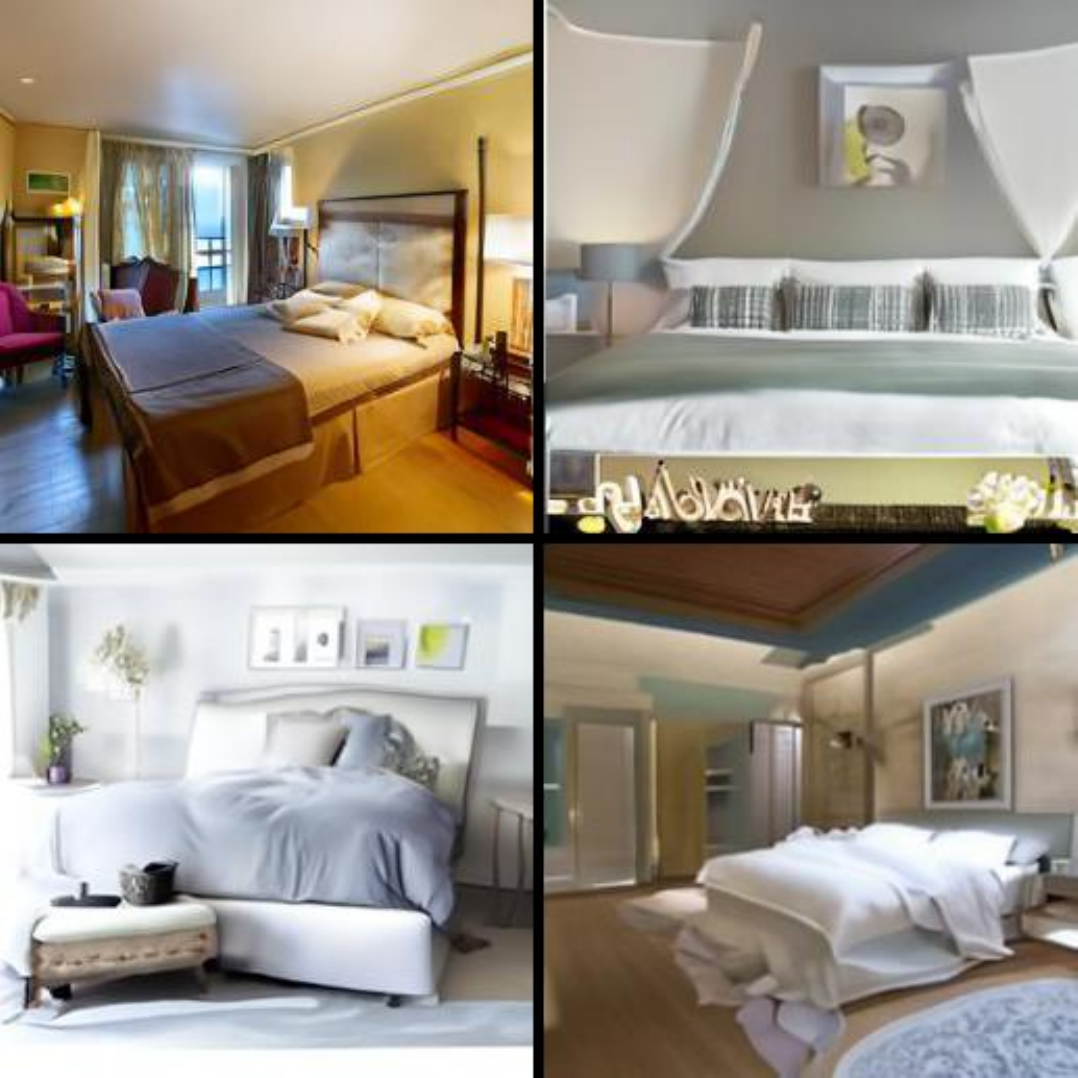} & \includegraphics[width=0.216\textwidth]{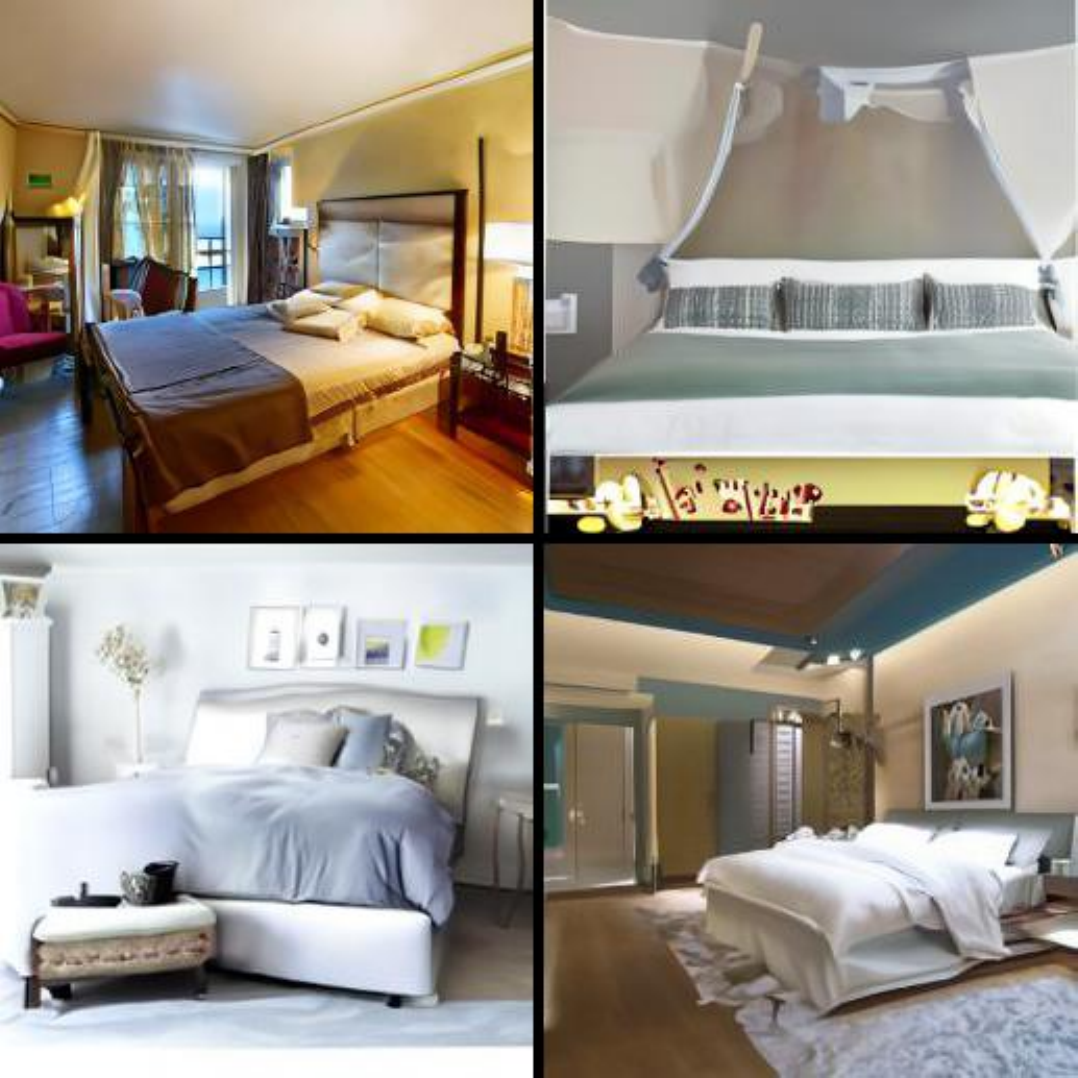} & \includegraphics[width=0.216\textwidth]{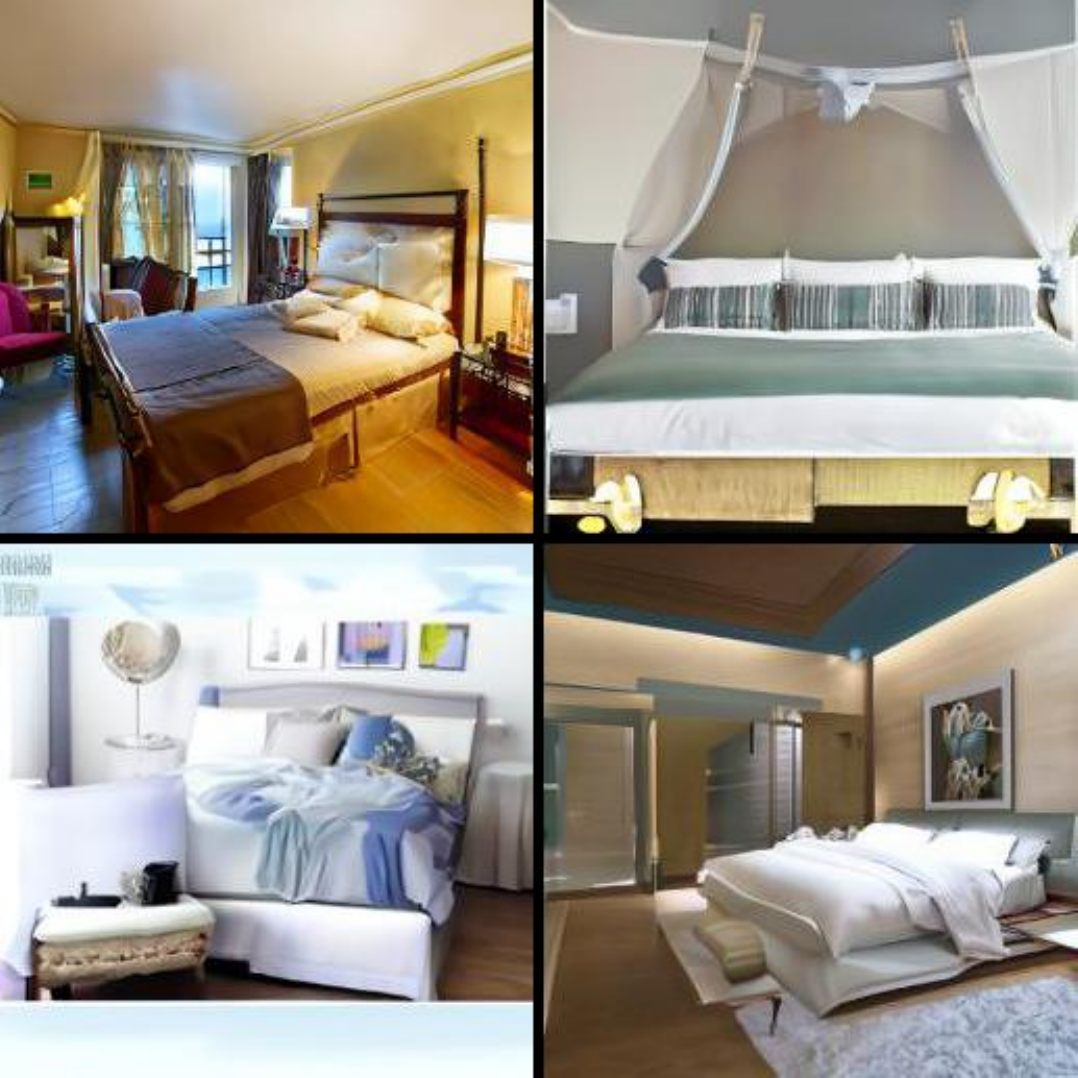} & \includegraphics[width=0.216\textwidth]{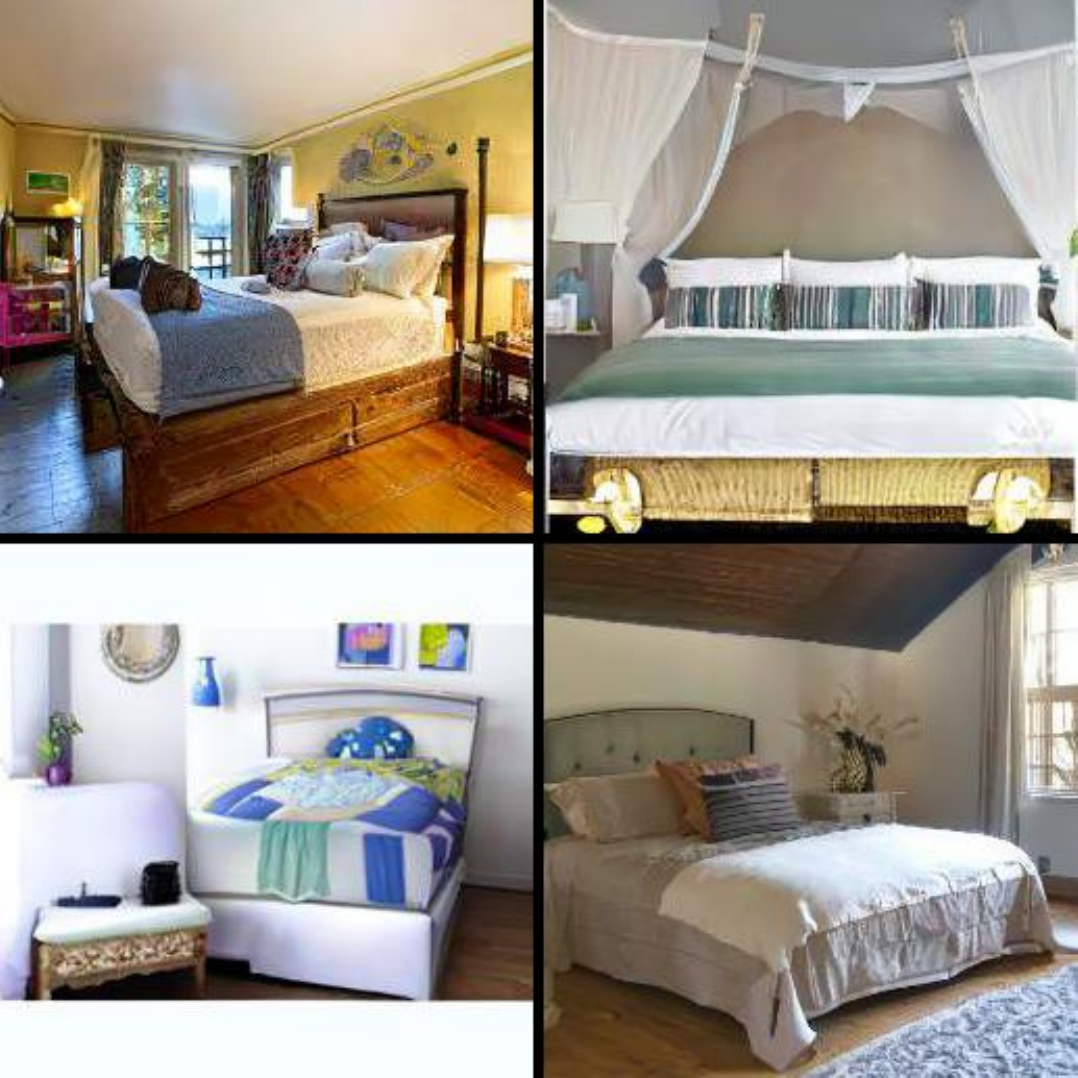} \\
\end{tabular}
 \caption{Random samples with the same random seed were generated by SciRE-Solver, DPM-Solver \cite{lu2022dpm}, and DDIM \cite{song2021denoising} with the consistently uniform time steps, employing the pre-trained discrete-time DPM \cite{dhariwal2021diffusion} on LSUN bedroom 256$\times$256.
}
\label{fig:lsunbedroomsamecomparison}
\end{figure}

\begin{figure}[ht]
    \centering
    \includegraphics[width=0.7\linewidth]{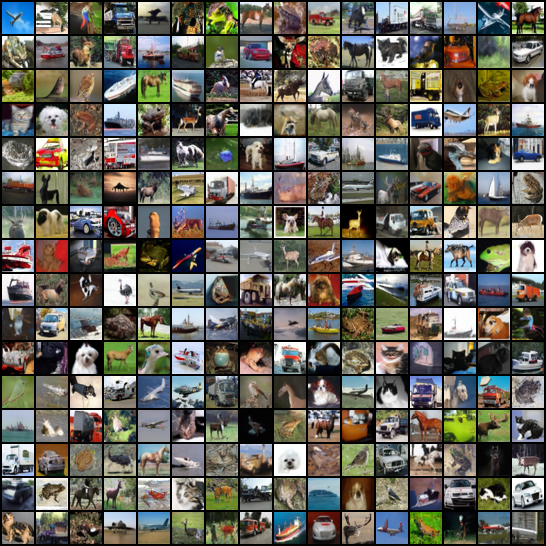}
    \caption{
    Random samples were generated by SciRE-Solver with $12$ NFE,  employing the pre-trained discrete-time DPM \cite{song2021scorebased} on continious-time CIFAR-10. We achieve an $3.48$ FID by using the \emph{Sigmoid}-type time trajectory with $k=0.65$, and setting the initial time as $\epsilon=10^{-3}$.
    }
    \label{fig:diagram-12NFE}
\end{figure}

\begin{figure}[ht]
\vspace{-0.6cm}
    \centering
    \includegraphics[width=0.7\linewidth]
    {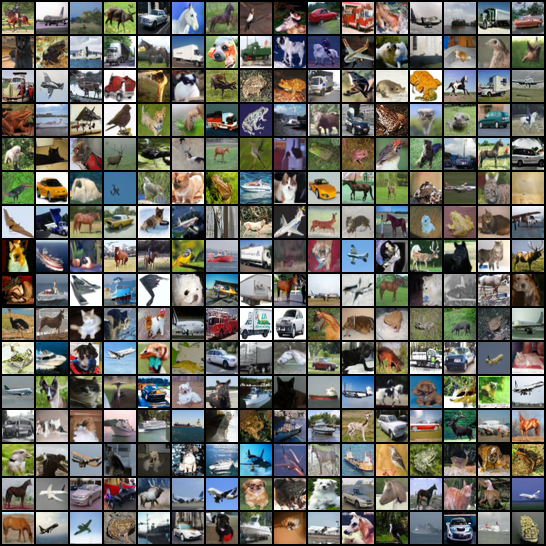}
    \caption{
    Random samples were generated by SciRE-Solver with $20$ NFE,  employing the pre-trained discrete-time DPM \cite{song2021scorebased} on continious-time CIFAR-10. We achieve an $2.42$ FID by using the \emph{NSR}-type time trajectory with $k=3.10$, and setting the initial time as $\epsilon=10^{-4}$.
    }
    \label{fig:diagram-20NFE}
\vspace{-0.4cm}
\end{figure}

\begin{figure}[ht]
    \centering
    \includegraphics[width=0.7
    \linewidth]
    {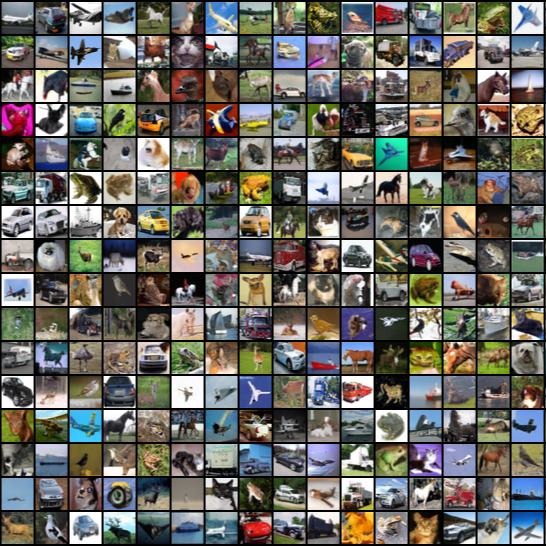}
    \caption{
    Random samples were generated by SciRE-Solver with $100$ NFE,  employing the pre-trained discrete-time DPM \cite{song2021scorebased} on continious-time CIFAR-10. We achieve an $2.40$ FID by using the \emph{NSR}-type time trajectory with $k=3.10$, and setting the initial time as $\epsilon=10^{-4}$.
    }
    \label{fig:diagram-20100NFE}
    \vspace{-0.4cm}
\end{figure}

\begin{figure}[ht]
    \centering
    \includegraphics[width=1\linewidth]{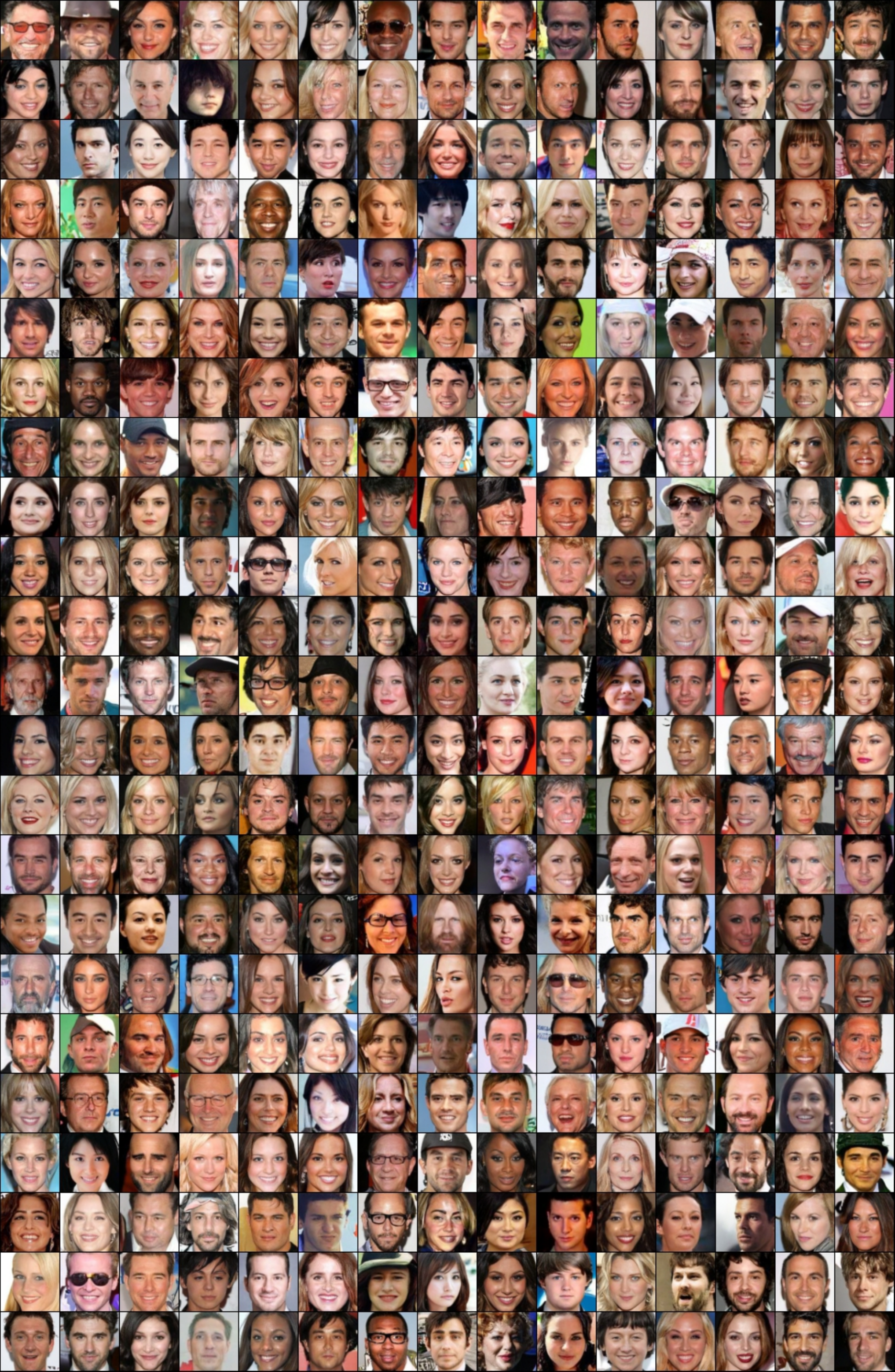}
    \caption{
    Random samples were generated by SciRE-Solver with $18$ NFE,  employing the pre-trained discrete-time DPM \cite{song2021denoising} on CelebA 64$\times$64. We achieve an $2.17$ FID by using the \emph{NSR}-type time trajectory with $k=3.10$, and setting the initial time as $\epsilon=10^{-4}$.
    }
    \label{fig:diagram-20NFEceleba}
\end{figure}

\begin{figure}[ht]
   \centering
   \includegraphics[width=1\linewidth]
    {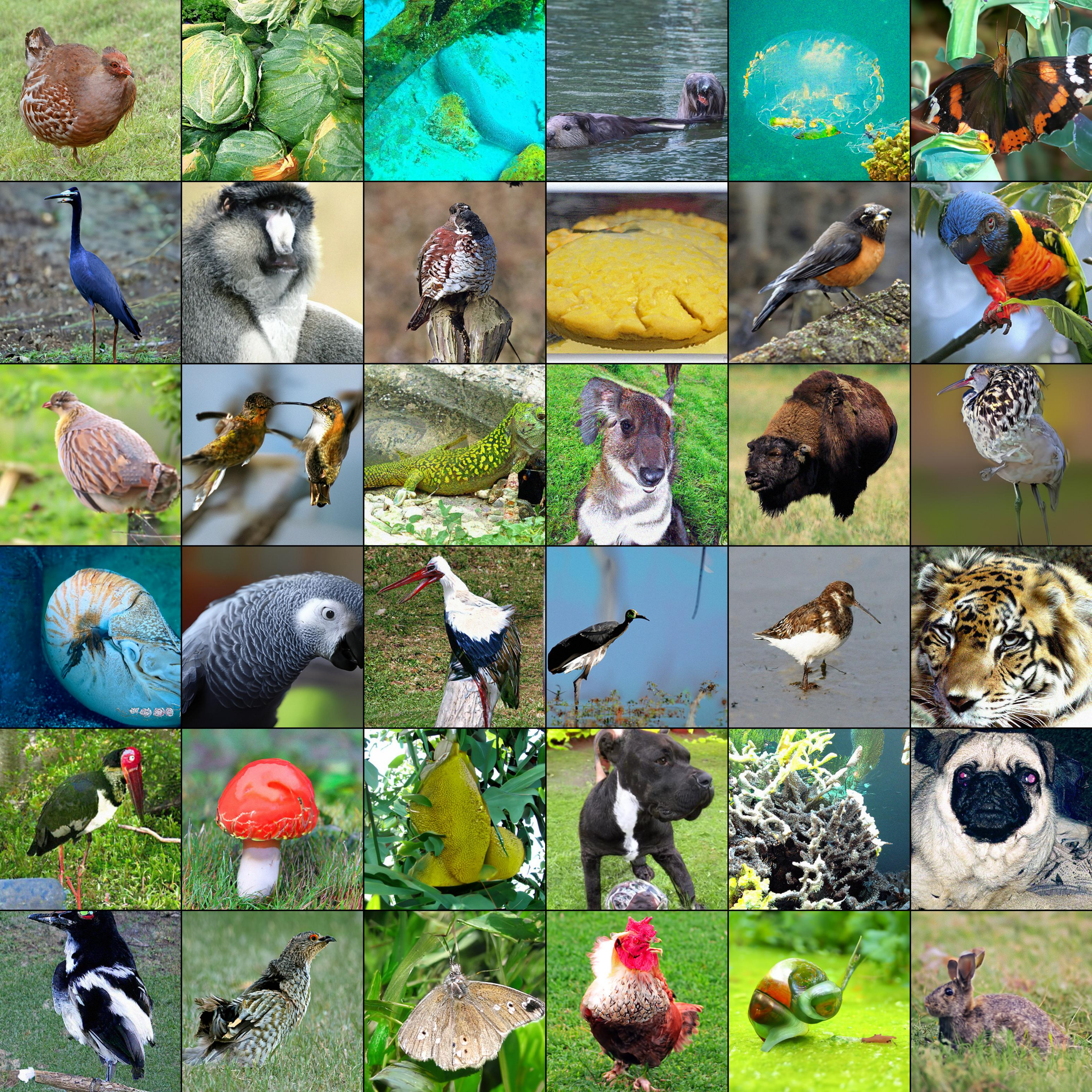}
   \caption{Random samples were generated by SciRE-Solver-2 with $20$ NFE and the uniform time steps, using the pre-trained discrete-time DPM \cite{dhariwal2021diffusion} on Imagenet 512$\times$512 (classifier scale: 4).}
   \label{fig:220imagenet512}
\end{figure}

\begin{figure}[ht]
   \centering
   \includegraphics[width=1\linewidth]
{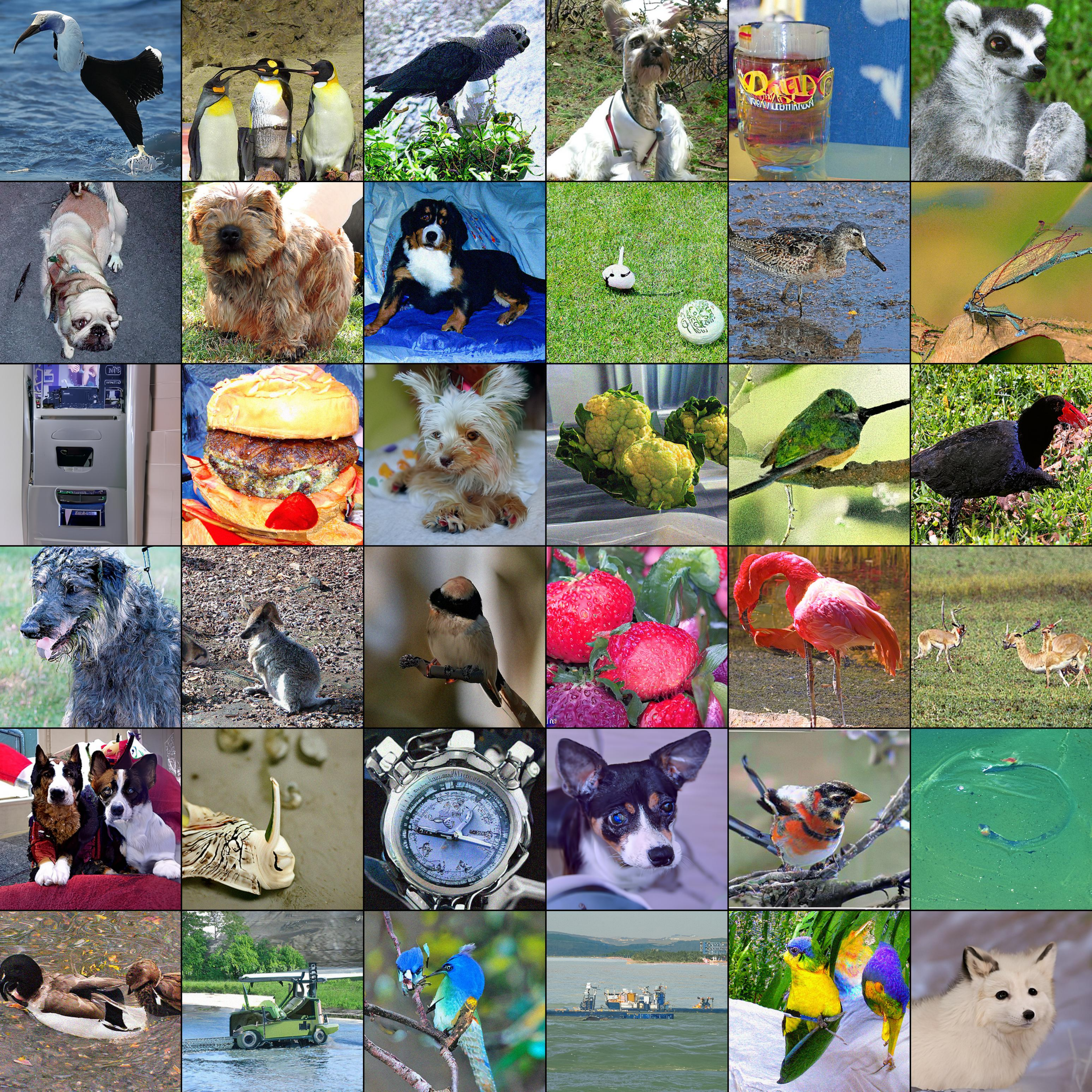}
   \caption{Random samples were generated by SciRE-Solver-3 with $18$ NFE and the uniform time steps, using the pre-trained discrete-time DPM \cite{dhariwal2021diffusion} on Imagenet 512$\times$512 (classifier scale: 1).}
   \label{fig:320imagenet512}
\end{figure}

\begin{figure}[ht]
   \centering
   \includegraphics[width=1\linewidth]
{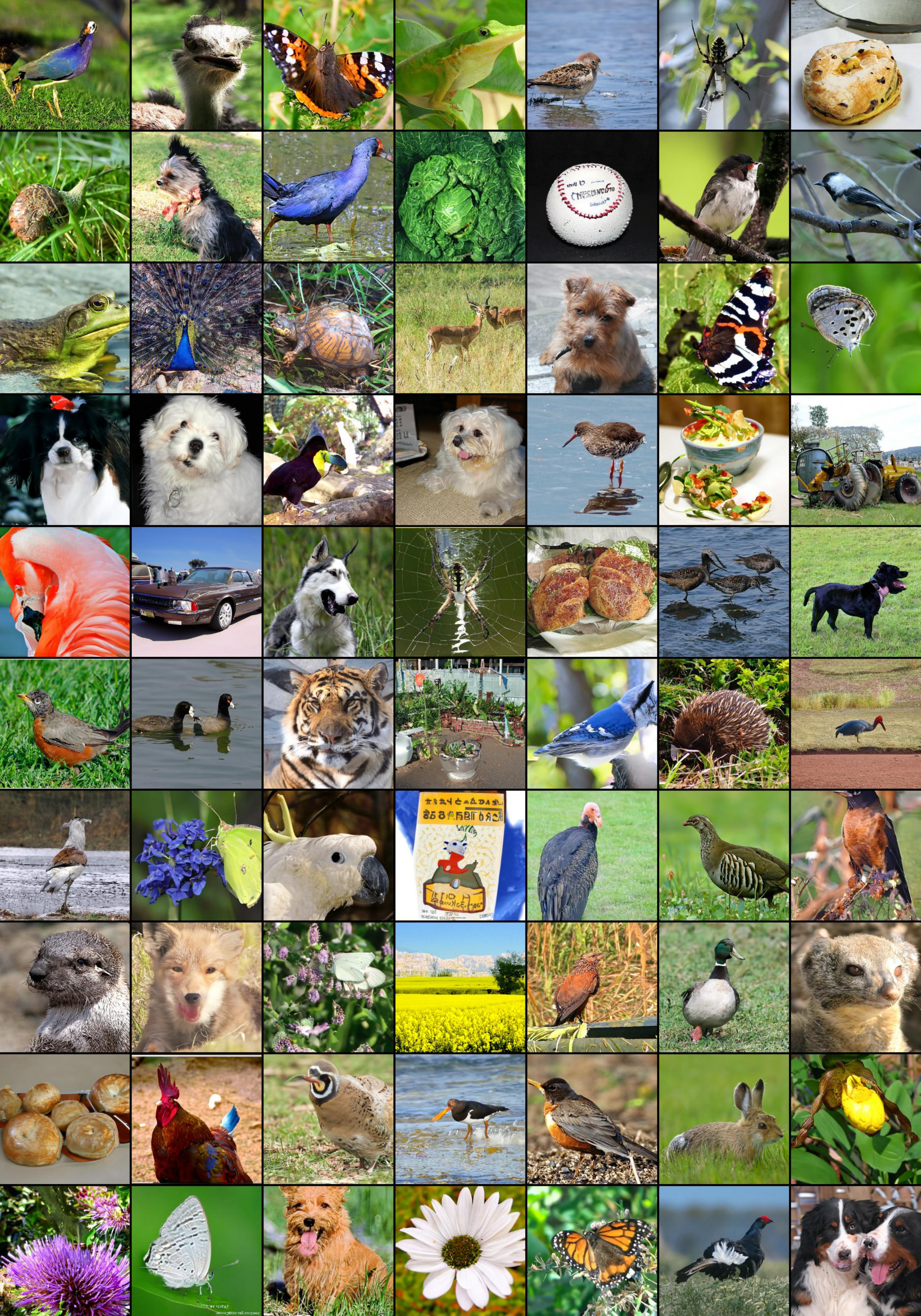}
   \caption{Random samples were generated by SciRE-Solver-2 with $20$ NFE and the uniform time steps, using the pre-trained discrete-time DPM \cite{dhariwal2021diffusion} on Imagenet 256$\times$256 (classifier scale: 1).}
   \label{fig:220imagenet256}
\end{figure}

\begin{figure}[ht]
   \centering
   \includegraphics[width=1\linewidth]
{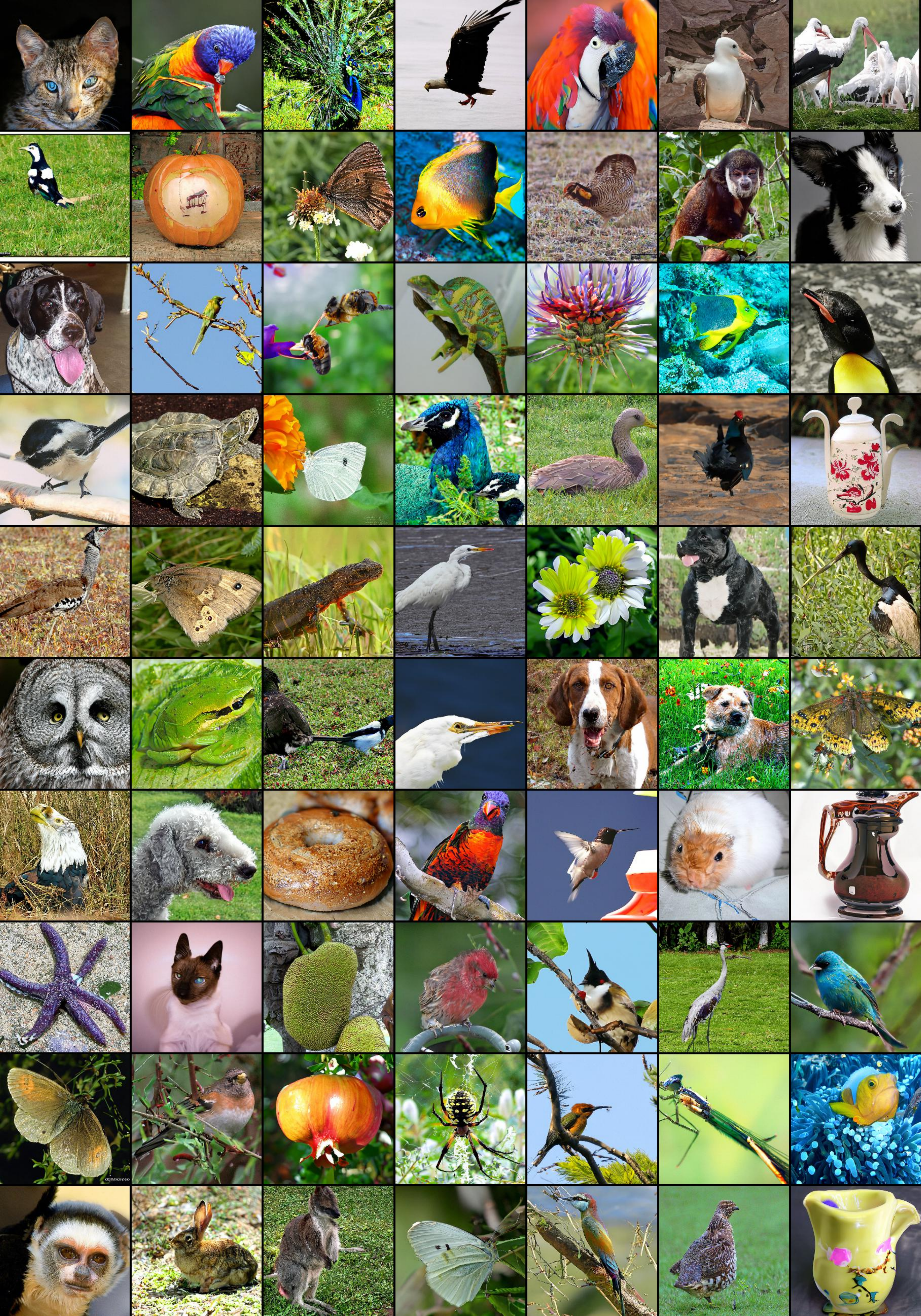}
   \caption{Random samples were generated by SciRE-Solver-3 with $18$ NFE and the uniform time steps, using the pre-trained discrete-time DPM \cite{dhariwal2021diffusion} on Imagenet 256$\times$256 (classifier scale: 1).}
   \label{fig:320imagenet256}
\end{figure}

In this section, we provide sample comparisons of random sampling using SciRE-Solver, DPM-Solver, and DDIM with the same codebase on different datasets, as depicted in Figures  \labelcref{fig:crfar10omparison,fig:celeba64comparison,fig:imagenet64comparison,fig:imagenet128comparison,fig:lsunbedroom256comparison,fig:imagenet256comparison,fig:lsunbedroomsamecomparison,fig:imagenet128samecomparison,fig:imagenet2128samecomparison,fig:imagenet512samecomparison}. Additionally, we present some generated samples on CIFAR-10, CelebA 64$\times$64, Imagenet 256$\times$256
 and Imagenet 512$\times$512, which reported in Figures  \labelcref{fig:diagram-12NFE,fig:diagram-20NFE,fig:diagram-20100NFE,fig:diagram-20NFEceleba,fig:220imagenet512,fig:320imagenet512,fig:220imagenet256,fig:320imagenet256}.


\newpage

\end{appendices}

\end{document}